\documentclass[preprint,3p,12pt]{elsarticle}
\usepackage{xfrac}  
\usepackage{lipsum}
\usepackage{amsfonts,dsfont}
\usepackage{graphicx}
\usepackage{epstopdf}
\usepackage{algorithmic}
\usepackage{tikz}
\usepackage{graphicx}
\usepackage[normalem]{ulem}

\usepackage{enumitem,comment}
\usepackage{subcaption}
\usepackage{graphicx}
\usepackage{hyperref}
\usepackage{multirow}
\usepackage{booktabs}
\usepackage{tabularray}
\usepackage{amssymb}
\usepackage[font=small,labelfont=bf]{caption}
\usepackage{amsmath}
\usepackage{amsthm}

\usepackage{cleveref}
\usepackage{float}
\usepackage[utf8]{inputenc}
\usepackage{pgfplots}
\DeclareUnicodeCharacter{2212}{−}
\usepgfplotslibrary{groupplots,dateplot}
\usetikzlibrary{patterns,shapes.arrows}
\pgfplotsset{compat=newest}

\ifpdf
\hypersetup{
    colorlinks=true,
    linkcolor=blue,
    filecolor=magenta,      
    urlcolor=red,
    pdftitle={Model-Constrained Deep Learning Approaches for Inverse Problems},
    pdfpagemode=FullScreen,
  }

\usepackage{multirow}
\usepackage{colortbl}
\fi
\usetikzlibrary{positioning}

\ifpdf
  \DeclareGraphicsExtensions{.eps,.pdf,.png,.jpg}
\else
  \DeclareGraphicsExtensions{.eps}
\fi
\usepackage{comment}

\usepackage{tikz}
\usetikzlibrary{arrows,shapes}
\usetikzlibrary{decorations.pathreplacing,calligraphy}
\usetikzlibrary{patterns}
\usetikzlibrary{arrows,shapes}
\usetikzlibrary{decorations.pathreplacing,calligraphy}

\usepackage{amsopn}
\DeclareMathOperator{\diag}{diag}

\makeatletter
\newcommand*{\addFileDependency}[1]{% argument=file name and extension
  \typeout{(#1)}% latexmk will find this if $recorder=0 (however, in that case, it will ignore #1 if it is a .aux or .pdf file etc and it exists! if it doesn't exist, it will appear in the list of dependents regardless)
  \@addtofilelist{#1}% if you want it to appear in \listfiles, not really necessary and latexmk doesn't use this
  \IfFileExists{#1}{}{\typeout{No file #1.}}% latexmk will find this message if #1 doesn't exist (yet)
}
\makeatother

\newcommand{\mynote}[3]{
	\textcolor{#2}{\fbox{\bfseries\sffamily\scriptsize#1}}
		{\textsf{\emph{#3}}}
}

\newcommand{\tanbui}[1]{\mynote{Tan}{magenta}{#1}}

\newcommand{\jauuei}[1]{\mynote{Jau-Uei}{blue}{#1}}

\newcommand{\nor}[1]{\left\| #1 \right\|} % norm
\newcommand{\snor}[1]{\left| #1 \right|} %semi-norm
\newcommand{\LRp}[1]{\left( #1 \right)} % adaptive left and right parentheses
\newcommand{\dLRs}[1]{\left[\!\!\left[ #1 \right]\!\!\right]} % adaptive left and right parentheses
\newcommand{\dLRss}[1]{\left[\hspace{-0.4ex}\left[ #1 \right]\hspace
{-0.4ex}\right]} % adaptive left and right parentheses
\newcommand{\LRs}[1]{\left[ #1 \right]} % adaptive left and right square brackets
\newcommand{\LRc}[1]{\left\{ #1 \right\}} % adaptive left and right curly brackets
\newcommand{\dLRc}[1]{\left\{\!\!\!\left\{ #1 \right\}\!\!\!\right\}} % adaptive left and right parentheses
\newcommand{\dLRcs}[1]{\left\{\hspace{-1ex}\left\{ #1 \right\}\hspace{-1ex}\right\}} % adaptive left and right parentheses
\newcommand{\pp}[2]{\frac{\partial #1}{\partial #2}} % adaptive partial derivatives

\newcommand{\mc}[1]{\mathcal{#1}} %mathcal
\newcommand{\mb}[1]{\mathbf{#1}} %math boldface
\newcommand{\mbb}[1]{\mathbb{#1}} %mathbb
\newcommand{\half}{\frac{1}{2}}

\newcommand{\bs}[1]{\boldsymbol{#1}}

\newcommand{\zb}{\bs{z}}
\newcommand{\zt}{\tilde{\bs{z}}}
\newcommand{\zbar}[1]{ \bar{\zb}^{{#1}}}
\newcommand{\pb}{\bs{u}}
\newcommand{\ub}{\pb}

\newcommand{\FW}{\mc{F}}

\newcommand{\figlab}[1]{\label{fig:#1}}
\newcommand{\eqnlab}[1]{\label{eq:#1}}
\newcommand{\theolab}[1]{\label{theo:#1}}

\newcommand{\tablab}[1]{\label{tab:#1}}

\newcommand{\theoref}[1]{\ref{theo:#1}}

\newcommand{\seclab}[1]{\label{sect:#1}}

\newcommand{\dt}{\Delta t}

\newcommand{\MSE}[1]{ \nor{#1}_{L^2\LRp{\Omega}}^2}

\newcommand{\ui}[1]{ {\ub}^{{#1}}}

\newcommand{\ut}[1]{ \tilde{\ub}^{{#1}}}
\newcommand{\ubar}[1]{ \bar{\ub}^{{#1}}}

\newcommand{\hai}[1]{\mynote{HAI}{green}{\bf #1}}

\newcommand{\eval}[2][\right]{\relax \ifx#1\right\relax \left.\fi#2#1\rvert}

\newcommand{\xb}{\bs{x}}
\newcommand{\nb}{\bs{n}}

\renewcommand{\u}{u}

\newcommand{\eg}[1]{ {\varepsilon}^{{#1}}}
\newcommand{\ev}[1]{ {\bs{e}_{\text{ML}}^{{#1}}}}
\renewcommand{\epsilon}{\varepsilon}
\newcommand{\vb}{\bs{v}}
\newcommand{\ubr}{\vb}
\newcommand{\epsb}{\bs{\epsilon}}

\newcommand{\DGGNN}{\Psi_\texttt{{DGNet}}}
\newcommand{\NNflux}{\Psi_\text{{flux}}}
\newcommand{\NNvol}{\Psi_\text{{vol}}}

\newcommand{\mcDGGNNa}{$\texttt{mcDGNet} \ $}
\newcommand{\nDGGNNa}{$\texttt{nDGNet} \ $}
\newcommand{\DGNet}{$\texttt{DGNet}$}
\newcommand{\oDGGNNa}{\DGNet{}$ \ $}

\newcommand{\mcDGNet}{\mcDGGNNa{}}
\newcommand{\nDGNet}{\nDGGNNa{}}
\newcommand{\oDGNet}{\oDGGNNa{}}

\newcommand{\fb}{\bs{f}}

\newcommand{\fbxaver}{\{\!\!\{ \fb_1 \}\!\!\}}
\newcommand{\fbyaver}{\{\!\!\{ \fb_2 \}\!\!\}}
\newcommand{\ubjump}{[\![ \ub ]\!]}

\newcommand{\ubhat}{\widehat{\ub}}
\newcommand{\uhat}{\widehat{u}}
\newcommand{\fbhat}{\widehat{\fb}}
\newcommand{\fhat}{\widehat{f}}

\newtheorem{theorem}{Theorem}
\newtheorem{remark}{Remark}
\journal{CMAME}

\begin{document}

\begin{frontmatter}

\title{A Model-Constrained Discontinuous Galerkin Network (\DGNet) for Compressible Euler Equations with Out-of-Distribution Generalization}

\author[label1]{Hai Van Nguyen} %% Author name
\author[label1]{Jau-Uei Chen}
% \author[label2]{William Cole Nockolds}
% \author[label2]{Wesley Lao}
\author[label1,label2]{Tan Bui Thanh} %% Author name

\affiliation[label1]{organization={Department of Aerospace Engineering and Engineering Mechanics, The University of Texas at Austin},%Department and Organization
            city={Austin},
            state={Texas},
            country={USA}}

\affiliation[label2]{organization={The Oden Institute for Computational Engineering and Sciences, The University of Texas at Austin},%Department and Organization
            city={Austin},
            state={Texas},
            country={USA}}

% % %%Graphical abstract
% % \begin{graphicalabstract}
% % %\includegraphics{grabs}
% % \end{graphicalabstract}

% % %%Research highlights
% % \begin{highlights}
% % \item Research highlight 1
% % \item Research highlight 2
% % \end{highlights}

% %% PACS codes here, in the form: \PACS code \sep code

% %% MSC codes here, in the form: \MSC code \sep code
% %% or \MSC[2008] code \sep code (2000 is the default)

% \end{keyword}
% REQUIRED
\begin{abstract}
Real-time accurate solutions of large-scale complex dynamical systems are critically needed for control, optimization, uncertainty quantification, and decision-making in practical engineering and science applications, particularly in digital twin contexts. Recent research on hybrid approaches combining numerical methods and machine learning in end-to-end training has shown significant improvements over either approach alone. However, using neural networks as surrogate models generally exhibits limitations in generalizability over different settings and in capturing the evolution of solution discontinuities. In this work, we develop a model-constrained discontinuous Galerkin Network (\oDGNet \hspace{-1ex}) approach, a significant extension to our previous work \cite{nguyen2022model}, for compressible Euler equations with out-of-distribution generalization. The core of \oDGNet is the synergy of several key strategies: (i) leveraging time integration schemes to capture temporal correlation and taking advantage of neural network speed for computation time reduction; (ii) employing a model-constrained approach to ensure the learned tangent slope satisfies governing equations; (iii) utilizing a GNN-inspired architecture where edges represent Riemann solver surrogate models and nodes represent volume integration correction surrogate models, enabling capturing discontinuity capability, aliasing error reduction, and mesh discretization generalizability; (iv) implementing the input normalization technique that allows surrogate models to generalize across different initial conditions, geometries, meshes, boundary conditions, and solution orders; and (v) incorporating a data randomization technique that not only implicitly promotes agreement between surrogate models and true numerical models up to second-order derivatives, ensuring long-term stability and prediction capacity, but also serves as a data generation engine during training, leading to enhanced generalization on unseen data. To validate the effectiveness, stability, and generalizability of our novel \oDGNet approach, we present comprehensive numerical results for 1D and 2D compressible Euler equation problems, including Sod Shock Tube, Lax Shock Tube, Isentropic Vortex, Forward Facing Step, Scramjet, Airfoil, Euler Benchmarks, Double Mach Reflection, and a Hypersonic Sphere Cone benchmark.
\end{abstract}

% a model-constrained discontinuous Galerkin Network (\DGNet) for compressible euler equations with out-of-distribution generalization

% REQUIRED
\begin{keyword}
hyperbolic system; Discontinuous Galerkin method; graph neural network; model-constrained machine learning; data randomization; generalization and stability.
\end{keyword}
 
\end{frontmatter}


\section{Introduction}
\seclab{introduction}

{ Modeling the evolution of dynamical systems is a fundamental challenge in various engineering disciplines, especially in computational fluid dynamics (CFD). In particular, the major challenges of simulating compressible flows lie in precisely capturing the propagation of different waves \cite{wagner_large-eddy_2007} and properly resolving the shock(s) induced by the nonlinearity \cite{tu_slope_2005, persson_sub-cell_2006, feistauer_robust_2007}. Preserving these physical features is crucial in a wide range of applications, including aerospace engineering \cite{donea_finite_2003,karniadakis_spectralhp_2005, hesthaven_nodal_2007, hartmann_discontinuous_2010,arndt_exadg_2020}, weather prediction \cite{tumolo_semi-implicit_2015, marras_review_2016}, and the study of supersonic \cite{hartmann_adaptive_2002, tu_slope_2005, persson_sub-cell_2006, feistauer_robust_2007, hesthaven_nodal_2007, wang_highorder_2013} and hypersonic flows \cite{hoskin_discontinuous_2024}  
\begin{comment}
{\tanbui{It is more convincing if we can provide a couple of references here}.} \hai{Jau-Uei will add a few soon}}
\end{comment}
{Discontinuous Galerkin (DG) methods have been proven to be successful in solving compressible flow \cite{cockburn_rungekutta_2001,hartmann_adaptive_2002,donea_finite_2003,wang_highorder_2013,dolejsi_discontinuous_2015}. Specifically, the high-order capability can bring tremendous accuracy to the approximation of waves. That is, less dissipation and less dispersion error in DG methods compared to other traditional methods such as finite volume (FV) methods or finite difference (FD) methods \cite{ainsworth_dispersive_2004,hesthaven_nodal_2007,wilcox_high-order_2010,wang_highorder_2013}. However, high-order approximations have a few shortcomings. First of all, stability issues need attention when explicit time integration schemes are employed. Explicit time integration schemes, while straightforward to implement and easy to compute, become inefficient due to the restrictive time step sizes required to satisfy the Courant-Friedrichs-Lewy (CFL) condition. In addition, higher order methods will amplify this restriction \cite{warburton_taming_2008,gassner_runge-kutta_2011}. The restriction can be relaxed by utilizing implicit schemes. Nonetheless, a nonlinear solver is required at each time step, which is computationally expensive and may pose challenges for parallelization. The other type of stability issue results from aliasing error incurred by the approximation of nonlinearity. Perhaps the simplest remedy is the so-called over-integration \cite{kirby_-aliasing_2003,spiegel_-aliasing_2015,kopriva_stability_2018} where high-order quadrature rules are applied to reduce the error produced by the nonlinear approximation and achieve a stabilization effect, this approach is more computational extensive.    
}

% \gencomment{This paragraph covers about accelerating numerical simulations with machine learning. PINN is popular one, but not generalizable}

In recent years, machine learning approaches for modeling dynamical systems have gained traction due to their potential to reduce computational cost and execution time compared to traditional numerical methods. Among these approaches, the Physics Informed Neural Network (PINN) \cite{raissi2017physics1, RaissiEtAl2019, RaissiKarniadakis2018, RaissiEtAl2017, YangPerdikaris2019, TripathyBilionis2018} has become particularly popular for modeling the dynamics of Partial Differential Equations (PDEs).
The PINN framework leverages both data and the governing physical laws represented by PDEs to learn surrogate solutions. 
For instance, in \cite{jin2021nsfnets}, authors utilize PINNs to solve the 2D incompressible Navier-Stokes equations, demonstrating the efficacy of this approach in fluid dynamics. 
Similarly, the authors in \cite{jagtap2020conservative} introduced a conservative PINN for solving conservation law equations. In this method, the computational domain is decomposed into subdomains, with flux continuity enforced at the interfaces by adding an extra flux loss term to the standard PINN loss. Different neural networks are applied to each subdomain, facilitating parallel computing and improving computational efficiency. 
For handling shock problems, PINNs have shown promising results. 
The study by \cite{mao2020physics} explores the application of the original PINN to high-speed flow problems. The authors emphasize the importance of clustering training points near shock positions to accurately capture shock dynamics. 
Additionally, in \cite{liu2024discontinuity}, authors introduced a weighted PINN approach specifically designed for capturing discontinuities. In this method, the PINN loss is weighted down at collocation points near shocks, and a Rankine–Hugoniot loss term is incorporated. This approach is tested on 1D and 2D Euler equation problems, showing improved performance in handling discontinuities. The approaches in \cite{mao2020physics,liu2004discontinuous}, however, require the location of the shocks {\em a priori}, which is unlikely practical for applications with time-dependent complex shock interactions. 
In \cite{chen2021deep}, authors learned a neural network that maps the coordinates of the center of a DG element and time variable to element solutions at all quadrature points on the corresponding element. This method uses a DG residual loss training function and captures discontinuous solutions using numerical fluxes like the Lax-Friedrichs flux. 
\begin{comment}
\tanbui{does this require knowledge about the shock location a priori?} \hai{this is the paper that Thomas has been assigned to implement as a test project, it is a variant of PINN. After training, it solves one instance for linear nD transport equation/1D burger equations. NO shock location is needed}. 
\end{comment}
% Boundary conditions are enforced through penalty loss terms in \cite{mcfall2009artificial}. In order to handle mixed boundary conditions on arbitrarily shaped domains, the authors have developed length factor functions which enable the network to determine whether Dirichlet or Neumann boundary conditions should be satisfied at a given boundary point. \textcolor{blue}{Comment: In \cite{mcfall2009artificial}, the penalty on the loss term is the main way in which the boundary conditions are enforced. The "trial function approach," which are called "length factor functions" in the paper, are only used to extend their method to arbitrarily shaped domains with mixed boundaries.} 
Another notable method is presented in \cite{huang2022neural}, where a neural network is trained to map solutions at two coarse mesh grid levels to finer mesh grid solutions. This technique allows the neural network to predict high-resolution solutions accurately, given inexpensive numerical solutions on a coarse grid. However, while PINNs and related approaches offer powerful tools for solving PDEs, they are not generalizable to new scenarios, such as different boundary conditions, initial conditions, geometry, or new parameter values. Furthermore, PINN typically requires a retrain for each new unseen scenario.
% Indeed, PINN only solves a specified instance of the problem that is used for training.  
% Neural ODE (NODE) is another approach for simulating the evolution of dynamical systems. This method defines the ODE as a neural network to speed up the computation of the right-hand side and uses traditional numerical integration techniques to generate solutions. The NODE architecture has been combined with physics constraints and autoencoders to construct reduced order models for high dimensional PDEs \cite{Shol}.
% \textcolor{blue}{Comment: Continued literature review for more specific topics (conservation laws and CFD}

% \gencomment{This paragraph covers direct prediction + its drawbacks , mcTangent learning + its drawbacks on spatial causality, GNN is  flexible:}

In contrast to the PINN approach, naive data-driven deep learning methods learn a surrogate neural network model from a large dataset of training samples. Once trained, the neural network can predict future solutions given the current (and some past) states for unseen scenarios. In various works, solutions of a discrete ODE or PDE system are predicted directly as the output of the neural network architecture \cite{SMAOUI200427, LIN2003611, Pan_2018}. However, this direct prediction approach is typically incapable of capturing, simultaneously, both spatial and temporal correlations in the data. Complex neural network architectures can be deployed to improve the result, however, optimizers may struggle to achieve optimal neural network parameters that generalize well. Additionally, test predictions obtained from the learned model are often limited to uniform time points corresponding to the training data step size, as there is no parameter to control the time step size in the method.
To overcome these limitations, alternative approaches have been developed to impart time-invariance to neural networks by learning the tangent slope of the dynamical system. In \cite{661124}, it was demonstrated that a neural network could learn the tangent slope within a Runge-Kutta scheme to produce accurate numerical solutions. Similarly, the authors of \cite{Zhuang_2021} computed the tangent for the Runge-Kutta method in a reduced subspace of the original system. Although these approaches are time-invariant, they do not respect the governing equations of the dynamical system. 
This gap motivated our previous work \cite{nguyen2022model} in which we developed a model-constrained approach, called \texttt{mcTangent}, for learning the tangent slope of the dynamical system. In this approach, two consecutive solutions are constrained to satisfy the discretized governing dynamic equations. However, in \cite{nguyen2022model}, a simple fully connected neural network is employed, which neglects the spatial correlation that is crucial in scientific problems. To better capture local interactions in space, the graph neural network (GNN) architecture is particularly flexible, as it can handle unstructured mesh data from complex geometries \cite{iakovlev2020learning, belbute2020combining, zhao2022learning}.

% This approach was employed in \cite{um2021solverintheloop} in which a network was trained to correct low resolution numerical simulations to their corresponding high-resolution solutions.

% \subsection{shock, viscosity, slope limiter related}
% While recent developments in machine learning for dynamical systems and CFD problem have produced significant positive results, developing algorithms that can tackle more complex physics remains an open question. Specifically shock-type problems in CFD remain a unique challenge for machine learning methods due to \textcolor{blue}{something about why shock is difficult to capture}. The following paragraph is a review of current literature on machine learning methods for capturing shock phenomena in CFD simulations.

% \gencomment{This paragraph covers learning the computationally demanding part of numerical simulation.}

Significant effort has been devoted to developing hybrid learning approaches for simulating dynamical systems. Hybrid learning procedures primarily replace computationally expensive components of numerical methods with machine learning algorithms. Instead of learning the composition of complex maps, it can be more efficient to focus on computationally demanding segments of numerical simulations. 
% In the context of Computational Fluid Dynamics (CFD), simulations are computationally intensive because the equations describing compressible flows are not analytically solvable. Therefore, they need to be numerically integrated, and the grid quality predominantly dictates the solution accuracy.
An example of such an approach is presented in \cite{ZANDSALIMY2024113063}, where the authors propose an automated mesh refinement algorithm based on Principal Component Analysis (PCA). This approach helps stabilize CFD simulations while increasing the speed of the mesh refinement process.
Another successful strategy is augmenting numerical method solutions with additional information from a neural network. For instance, the authors in \cite{um2020solver} learned the correction for upsampling from low-resolution data to high-resolution solutions within a differentiable numerical simulation framework. Similarly, the authors in \cite{kochkov2021machine} propose approaches to correct the velocity field via interpolation and correction networks learned within a differentiable CFD solver. Additionally, a differentiable computational fluid mechanics package using the FV method, as developed in \cite{bezgin2023jax}, facilitates end-to-end learning approaches and can be used for optimization design. These hybrid approaches combine the strengths of traditional numerical methods with the flexibility and efficiency of machine learning, enabling more accurate and efficient simulations of complex dynamical systems.
{ For nonlinear conservation laws, such as a compressible flow, shocks can naturally be developed, even with smooth initial/boundary conditions. Like other high-order methods, the DG methods suffer from oscillations around discontinuities unless equipped with special treatments. The phenomenon is also referred to as Gibb's phenomenon \cite{gottlieb_gibbs_1997}. Approaches used to eliminate or mitigate the oscillations are usually called shock-capturing techniques and there are two popular ones in the DG community: limiting (or slope limiter) \cite{cockburn_runge-kutta_1990,cockburn_rungekutta_1998,tu_slope_2005} and artificial viscosity \cite{persson_sub-cell_2006,feistauer_robust_2007,barter_shock_2010,klockner_viscous_2011,yu_study_2020}. As suggested by its name, the limiter approach limits the slopes of the local approximation in a way that the slopes monotonically increase or decrease across "trouble" elements. The method of artificial viscosity, on the other hand, introduces an additional diffusion term to the governing equations, where the amount of viscosity is local and depends on the oscillatory nature of the solution. Moreover, these techniques can be improved by leveraging the power of machine learning. A surrogate model for the slope limiter designated for the DG method can be built by a neural network and was proposed in \cite{ray2018artificial} for 1D problems. Later, it was extended to 2D problems in \cite{ray2019detecting}. In \cite{caldana2024discovering}, the authors proposed an end-to-end viscosity surrogate model mapping element solutions to the viscosity mode for the DG method. In this approach, the neural network is embedded within a numerical DG solver during training and then employed similarly for solving test cases. Another unique approach, presented in \cite{morgan2020machine}, involves training a neural network to detect shocks in 2D problems. This network is then integrated into the high-order method residual distribution Lagrangian hydrodynamic method to simulate multidimensional shock-driven flows. Besides limiter and artificial viscosity, it turns out that a smoothness indicator can also be learned by machine learning. This indicator plays an essential role in shock-capturing techniques in the sense that we do not want loose accuracy brought by the high-order approximation and would like to apply the limiter or artificial viscosity to the limited number of the elements indicator as "trouble." In \cite{kossaczka2021enhanced}, smoothness indicators are proposed using neural network surrogates within the WENO-Z approach. Similarly, the authors of \cite{yu2022data} use neural networks to predict indicators from element solutions, which are then used to determine the required decay rate for artificial viscosity in the DG method.}

% \jauuei
{The other critical ingredient needed by an upwind numerical method is a numerical flux. The ambiguity introduced by the discontinuous approximation space or discontinuous solution allows us to resort to the numerical fluxes, which is a relatively mature field of research. Most of the numerical fluxes (i.e., Godunov-type methods \cite{godunov_finite_1959,toro_riemann_2013}) require locally solving Riemann problems either exactly or approximately in order to preserve conversation laws and capture the propagation of waves. The task is usually computationally demanding but deep learning approaches could overcome the obstacle. The authors of \cite{magiera2020constraint} proposed a deep learning constraint-aware approach to learn a surrogate model for the Riemann solver. This neural network takes the state variables on both sides of the interface and outputs the Riemann problem solutions, including trace states and wave speeds. Similarly, \cite{wang2023fluxnet} proposes four different Riemann solver surrogate neural network schemes to predict numerical fluxes at the interface. The loss function for optimization includes constraints on mass, momentum, and energy. However, for each problem, a large number of Riemann solver solutions need to be generated in advance to train the surrogate model. Such large training data sets might not be available in practical engineering problems. 
}

% \jauuei{For Hai: Note that the Riemann solver is introduced to design a numerical flux. It is NOT a shock-capturing technique! The idea behind it is that we can regard each interface of adjacent elements as a local Riemann problem provided the approximate solution is discontinuous across the interface. This is true for both DG and FV methods. Once we solve this local Riemann problem approximating or exactly, we then can decide how to define (or compute) the numerical flux.}

% \gencomment{This paragraph covers: adding gradient/Jacobian w.r.t. input data to loss function improves accuracy, we can achieve by data randomization technique.}

In the scientific machine learning field, improving the generalization and long-term stability of the learned surrogate models is crucially important. The authors in \cite{pan2018long, yu2022gradient, drucker1992improving, ross2018improving, finlay2021scaleable} proposed methods to enhance the accuracy and generalization of neural networks by incorporating the network gradient with respect to input into the loss function. However, explicitly forming a Jacobian matrix via back-propagation and then optimizing the loss function is computationally expensive, especially for high-dimensional input problems, as it requires double back-propagation. In \cite{o2024derivative}, the low-rank properties of the Jacobian matrix can be exploited to reduce the cost of evaluating the Jacobian matrix in the loss function. However, as demonstrated in our previous work \cite{nguyen2022model}, we can achieve similar regularization effects through a data randomization technique. This technique is as simple as adding an appropriate amount of noise to the data inputs of the neural network. Adding noise to training data effectively enhances long-term predictive stability \cite{sanchez2020learning, poggio1990networks}. In \cite{reed1992regularization, bishop1995training, matsuoka1992noise} the authors proved that adding noise to inputs of the neural network is equivalent to Tikhonov regularization which smooths the neural network with respect to its inputs.

% \gencomment{This paragraph covers, what is done in our work, appealing features}

Building on the insights from our previous approach \cite{nguyen2022model}, this work presents an end-to-end learning framework called the model-constrained Discontinuous Galerkin network (with the abbreviation \DGNet) for compressible Euler equations with out of distribution generalization. \oDGNet aims to learn a surrogate model for the DG spatial discretization via learning the tangent slope of the DG-discretized conservation law systems. The surrogate model is inspired by the graph neural network architecture and is constructed with the high-order DG method framework in a hybrid manner. As depicted in \cref{fig:DG_and_graph_dual_mesh}, the graph neural network is the dual of the  DG mesh. The nodal values of the DG method are treated as graph neural network node attributes. Meanwhile, the graph neural network edges, equivalent to the role of a numerical flux in the DG method, model the interaction between pairs of points on the shared edges of adjacent elements. The \oDGNet approach is a synergy of the \texttt{mcTangent} approach, the DG method, and graph neural networks, leading to several appealing features over existing approaches. 
First, it learns the underlying spatial discretization\textemdash the DG in this case. Thus, once trained, the neural networks are readily employed with either explicit or implicit time integration schemes: the approach thus temporal-discretization-invariant, but generally spatial-discretization-variant.
Second, it enforces the surrogate model to satisfy the discretized governed equations during training, ensuring adherence to conservation laws. As such, it automatically respects the underlying governing equations.
Third, it is equipped with the data randomization technique and thus promotes the similarity of derivatives of the learned tangent slope of the surrogate model and numerical discretized solver, without explicitly penalizing the differences. Therefore, the prediction error is bounded, leading to stability for long-term prediction far beyond the training temporal horizon.
Fourth, the inputs to the neural networks are normalized to a unified range $\LRs{-1,1}$ regardless of data sets. Thus, our approach has great generalizability in solving unseen scenarios or even completely different problems on different complex geometry.
Fifth, it could preserve the high-order convergence rate of the DG approach, ensuring numerical accuracy is maintained, as we shall see.

\begin{figure}[htb!]
    \centering
    \raisebox{-0.5\height}{\resizebox{1\textwidth}{!}{\input{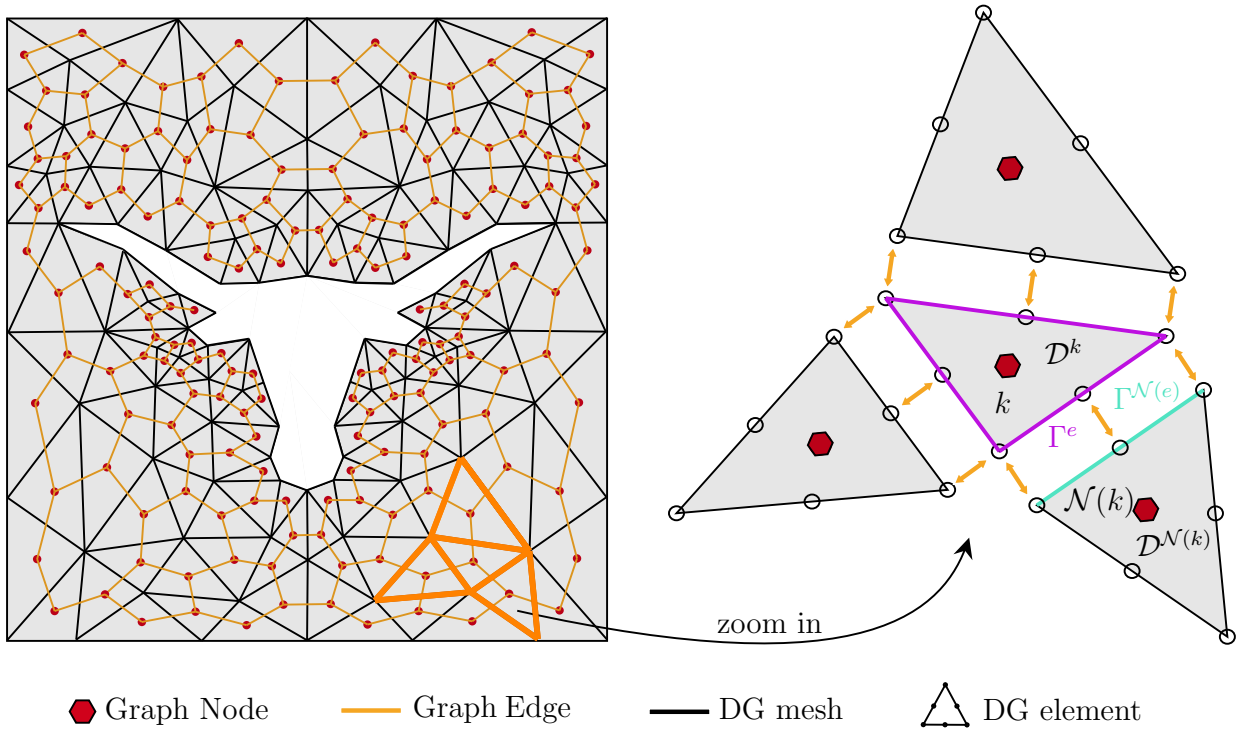}}} 
    % \begin{tabular*}{\textwidth}{c@{\hskip -0.0cm} c@{\hskip -0.0cm} }
    %     \centering
    %     % Comment out to have the original mesh
    %     % \raisebox{-0.5\height}{\resizebox{0.49\textwidth}{!}{\input{Figs/data_noise/image_dual_mesh.tex}}}
    %     \raisebox{-0.5\height}{\resizebox{0.49\textwidth}{!}{\input{Figs/data_noise/image_dual_mesh_UT.tex}}} &
    %     \raisebox{-0.5\height}{\resizebox{0.48\textwidth}{!}{\input{Figs/data_noise/Graph_edge_DG.tex}}}
    % \end{tabular*}
    \caption{{\bf Left figure} shows the duality of a Discontinuous Galerkin (DG) mesh (black) and graph neural network (GNN) architecture (red nodes and orange edges).  {\bf Right figure} illustrates the $k$-th  element  $\mathcal{D}^k$ (triangle with purple boundary) and its set of faces $\Gamma^e \subset \partial \mathcal{D}^k$; $\mathcal{D}^{\mathcal{N}\LRp{k}}$ (bottom right) represents an immediate neighboring element of the $k$-th element; $\Gamma^{\mathcal{N}\LRp{e}} \subset \mathcal{D}^{\mathcal{N}\LRp{k}}$ denotes a neighboring face of $\Gamma^e$.  The node attribute of the GNN $k$-th vertex is a collection of all nodal values of the $k$-th DG element.  The edge feature for the GNN edge between the $k$-th vertex and its neighboring vertex $\mc{N}(k)$  represents the numerical flux  between the DG elements $\mathcal{D}^k$ and $\mathcal{D}^{\mathcal{N}\LRp{k}}$;
    %through the face pair of $\Gamma^e$ and $\Gamma^{\mathcal{N}\LRp{e}}$.
    For illustration, circles represent the nodal value of $2^{nd}-$order DG elements. 
    } 
    \figlab{DG_and_graph_dual_mesh}
\end{figure}

\begin{comment}
    \tanbui{Hai: can you pick two element and then put $\mc{D}^k$ and $\mc{D}^{k+1}$ inside the elements. Then label the corresponding two dual vertices as $\mc{N}^k$ and $\mc{N}^{k+1}$. Also make the edge connecting these two vertices stand-out so that the discussion the relationship between DG and GNN at the end of section 2.1 can be clear.} \hai{Tan: We do not have $\mc{N}^{k+1}$ notation, I added comments below. Also, please add comments if you like more or clean comments if it looks good to you}
\end{comment}

% \gencomment{This paragraph covers, the organization of the paper, }

The paper is structured as follows. In the \cref{sect:methodology}, we develop the \oDGNet approach in detail for general conservation laws. We begin by introducing the motivation behind the \oDGNet approach, demonstrating the challenges of the DG method in solving shock-type problems and the limitations of our original machine learning \texttt{mcTangent} approach \cite{nguyen2022model} in \cref{sect:motivation}. In \cref{sect:DGGNNTagent_framework}, we discuss the \oDGNet architecture and normalization technique which enable effective generalization across various initial conditions, boundary conditions, geometries, mesh discretizations, and solution orders. Data randomization enhances the stability and generalizability of our approach, as detailed in \cref{sect:Data_rand}. Notably, data randomization promotes agreement between the tangent slope of learned surrogate models and the Discontinuous Galerkin solver up to the second-order derivative, resulting in improved stability for long-term prediction capacity. Furthermore, the data randomization technique functions as a data generation engine during training, thus substantially enriching training data information and leading to greater generalization. Error estimation of predictions for unseen test cases is analyzed in \cref{sect:Error_estimation}. Numerical results for 1D and 2D compressible Euler equation problems, including Sod Shock Tube, Lax Shock Tube, Isentropic Vortex, Forward Facing Step, Scramjet, Airfoil, Euler Benchmarks, Double Mach Reflection, and Hypersonic Sphere Cone are presented in \cref{sect:numerics}. Additionally, we provide the general training settings and hyperparameters in  \cref{sect:general_training}, convergence rate study in \cref{sect:isentropic_vortex}, the effect of data randomization in \cref{sect:P6_2D_Noise_corruption}, robustness to randomness using different neural network initializers in \cref{sect:Random_initializers_and_noise_seeds}, and computational time considerations in \cref{sect:Train_validation_test_computation_time}. Finally, we conclude the paper and discuss future research directions in \cref{sect:conclusions}.

% \clearpage
\section{Methodology}
\seclab{methodology}
\subsection{A brief review of the nodal discontinuous Galerkin Approach}
\seclab{motivation}
% \jauuei{Assume that we already have the discussion about DGM in the introduction. (i.e.,Discontinuous Galerkin (DG) methods have been proven to be successful in solving shock-type problems [CITE].)}
% \hai{collocation-type integration OR collocation-type integration?}
% \jauuei{collocation-type integration}
% \hai{Let me change everywhere again one more time! Thank you sir.}
The \oDGNet approach is an extension of \texttt{mcTangent} approach \cite{nguyen2022model} for hyperbolic conservation laws equations. The main idea of the \oDGNet approach is to design a neural network framework to capture the spatial causality, while the temporal causality is taken into consideration with traditional temporal discretization.  {\em The goal of \oDGNet is to learn the DG spatial discretization. Though we limit ourselves to DG discretization, the approach has a natural and straightforward extension to upwind finite difference and finite volume methods.} The principle design for the \oDGNet approach stems from our realization of the similarity between the Discontinuous Galerkin (DG) method and the graph neural network. In this section, we briefly review the general concept of DG methods to pave the way for elaborating the philosophy behind \oDGNet in \cref{sect:DGGNNTagent_framework}. Afterward, we discuss drawbacks in \texttt{mcTangent} approach together with the inspiration from the graph network and DG dual meshes in capturing spatial causality/correlation.

To begin with, let $\Omega\subset\mbb{R}^d$, $d=1,2,3$ be bounded domain and $T>0$. $\partial\Omega$ denotes the boundary of $\Omega$.
Considering $\ub\LRp{\xb,t}=\LRp{\u_1\LRp{\xb,t},\dots,\u_m\LRp{\xb,t}}^T$ where $\xb=\LRp{x_1,\dots,x_d}^T$, $m$ denotes the dimension of the conservative variables $\ub$ with components $\u_q:\mbb{R}^d\times\mbb{R}^+\mapsto\mbb{R}$ for $q=1,\dots,m$. An abstract system of conservation law for $\ub\LRp{\xb,t}$ can be written in the following form
\begin{equation}
    \eqnlab{conservation_law}
    \begin{aligned}
        \pp{\ub}{t}\LRp{\xb,t} + \nabla \cdot \fb \LRp{\ub\LRp{\xb,t}} & = 0, & & \quad \xb \in \Omega, \quad t \in \LRs{0,T},\\
        \ub\LRp{\xb,t} & = \ub_{bc}\LRp{\xb,t}, & & \quad \xb \in \partial \Omega, \quad t \in \LRs{0,T},\\
        \ub\LRp{\xb,0} & = \ub_0\LRp{\xb}, & & \quad \xb \in \Omega,
    \end{aligned}
\end{equation}
in which $\fb\LRp{\ub}
=\LRp{\fb_1\LRp{\ub},\dots,\fb_d\LRp{\ub}}$ where $\fb_i\LRp{\ub}=\LRp{f_{i1}\LRp{\ub},\dots,f_{im}\LRp{\ub}}^T$ with $f_{iq}:\mbb{R}^m\mapsto\mbb{R}$ for $i=1,\dots,d$ and $q=1,\dots,m$, and $\nabla \cdot \fb := \sum_{i=1}^d \pp{\fb_{i}}{x_i}$. In addition, $\ub_{bc}\LRp{\xb,t}$ is the boundary conditions and $\ub_0\LRp{\xb}$ is the initial condition. In DG methods, the domain $\Omega$ is partitioned into a set of non-overlapping elements $\mc{T}_h = \LRc{\mc{D}^k}_{k=1}^{K}$, where $K$ is the number of elements. The global solution $\ub$ is approximated by a piecewise $N-$order polynomial function $\ub_h\LRp{\xb,t}=\LRp{\u_{h,1}\LRp{\xb,t},\dots,\u_{h,m}\LRp{\xb,t}}$ with components $\u_{h,q}:\mbb{R}^d\times\mbb{R}^+\mapsto\mbb{R}$ for $q=1,\dots,m$. To reduce notational complexity and confusion, we shall remove the subscript $h$ for all approximate quantities. For example, we shall use $\ub\LRp{\xb,t}$, instead of $\ub_h\LRp{\xb,t}$, for the approximate solution. The approximate solution is given by,
\begin{comment}
\tanbui{Throughout the manuscript, for all equations/expressions that we do not refer to or cite later, we need to remove the equation number with equation*.}\hai{Hai will take care of it and back to comment out once done}
\end{comment}
\begin{equation*}
    %\ub\LRp{\xb,t} \approx
    \ub\LRp{\xb,t} = \bigoplus_{k=1}^K \ub^k\LRp{\xb,t},
\end{equation*}
\begin{comment}
\tanbui{We should remove the subscript $h$ to avoid too many subscripts later which is very difficult to follow. You can do that by saying "To reduce notational complexity and confusion, we shall remove the subscript $h$ for all approximate quantities. For example, we shall use $\ub\LRp{\xb,t}$, instead of $\ub_h\LRp{\xb,t}$, for the approximate solution."}  \hai{Jau-Uei, please take care DG-related notations, I will follow up in other sections based on Jau-Uei's new format}\jauuei{done}
\end{comment}
which is a direct sum of local discretized solutions $\ub^k\LRp{\xb,t}=\LRp{\u_{1}^k\LRp{\xb,t},\dots,\u_{m}^k\LRp{\xb,t}}^T$ on each element $\mc{D}^k$. Here, the superscript $k$ indicates the restriction on an element $\mc{D}^k$ (i.e., $\u_{q}^k\LRp{\xb,t}=\eval{\u_{q}\LRp{\xb,t}}{_{\xb\subseteq \mc{D}^k}}$ for $q=1,\dots,m$). However, it should be pointed out that the superscript $k$ will not always represent the restriction. Sometimes it simply suggests that the quantity is somehow associated with the domain $\mc{D}^k$ (i.e., the nodal data located at the interpolating node defined within the domain $\mc{D}^k$).
% \hai{which quantities associated with k? sounds confusing to me}\jauuei{for example, the $k$ in the hat terms is simply referred to as the data point is on $\xb_i\in\mc{D}^k$. Note that those hat terms are functions of time and hence it make no sense to say that the restriction on the "spatial" domain.}. 
To further breakdown the expression of $\ub^k\LRp{\xb,t}$, let $\ubhat^{k}\LRp{t}:=\LRp{\ubhat^{k}_{1}\LRp{t},\dots,\ubhat^{k}_{m}\LRp{t}  }$ where $\ubhat^{k}_{q}\LRp{t}=\LRp{ \uhat^{k}_{q,1}\LRp{t},\dots,\uhat^{k}_{q,{N_p}}\LRp{t} }^T$ with $\uhat^{k}_{q,l}\LRp{t}:=\u^{k}_{q}\LRp{\xb_l,t}$ for $q=1,\dots,m$ and $l=1,\dots,Np$. $N_p$ is the number of nodal values of the element
%\footnote{For simplicity, we use the same order, and hence the same number of nodal values, for all elements.} 
and $\xb_l$ for $l=1,\dots,N_p$ denotes a set of interpolating points defined in an element $D^k$. Throughout the paper, hatted quantities denote the nodal values. In each element, the approximate solution $\u_{q}^k\LRp{\xb,t}$ is given by 
\begin{equation*}
    \u_{q}^k\LRp{\xb,t} = \sum_{l=1}^{N_p} \uhat_{q,l}^k(t) \phi_l^k\LRp{\xb},
\end{equation*}
where $\phi_l^k$ is the Lagrange polynomial associated with an interpolating point $\xb_l\in\mc{D}^k$. Moreover, the flux tensor $\fb$ can be approximated in a similar way.\footnote{There are several ways to approximate the flux. In this work, we simply approximate it by interpolating the flux to the polynomial function with degree $N$. We refer the readers to \cite{hesthaven_nodal_2007} for a comprehensive discussion.} In particular, we set $\fb^{k}\LRp{\ub^k}=\LRp{\fb_{1}^k\LRp{\ub^k},\dots,\fb_{d}^k\LRp{\ub^k}}$ and $\fb_{i}^k\LRp{\ub^k}=\LRp{f_{i,1}^k\LRp{\ub^k},\dots,f_{i,m}^k\LRp{\ub^k}}^T$. 
Denoting $\fbhat^k_{i}\LRp{t}=\LRp{ \fbhat^k_{i,1}\LRp{t},\dots,\fbhat^k_{i,m}\LRp{t} }$ where $\fbhat^k_{i,q}\LRp{t}=\LRp{ \fhat^k_{i,q,1}\LRp{t},\dots,\fhat^k_{i,q,{N_p}}\LRp{t} }^T$ with $\fhat^k_{i,q,l}\LRp{t}:=f_{i,q}\LRp{ \ub^{k}\LRp{\xb_l,t} }$ for $i=1,\dots,d$, $q=1,\dots,m$, and $l=1,\dots,N_p$, we approximate the $i$th component flux corresponding to the $q$th equation  as
\begin{equation*}
    f_{i,q}^k\LRp{\xb,t}:=\sum_{l=1}^{N_p}\fhat^k_{i,q,l}(t)\phi_l^k\LRp{\xb}.
\end{equation*}
For the clarity of the presentation, from now on we remove the explicit dependence on $\xb$ and $ t$ for all unknowns/variables: for example, we write $f_{i,q}^k$ instead of $f_{i,q}^k\LRp{\xb,t}$.

The set of $\LRc{\phi_l^k}_{l=1}^{N_p}$ forms a basis of the $N$-th order polynomial space defined on the element $\mc{D}^k$. Such a space is also referred to as the trial function space and will be denoted as $\mc{V}^k$ in this paper. In DG methods, the test space is chosen the same as the trial space. Therefore, a DG discretization of \cref{eq:conservation_law} reads: for $l = 1, \ldots, N_p,\,\text{and} \quad k = 1, \ldots, K,$
\begin{equation}
    \label{eq:DG_weak_form}
    \begin{aligned}
        \int_{\mc{D}^k} \pp{\ub^k}{t} \phi_l^k \, d\xb & =  \int_{\mc{D}^k}  \fb^k \cdot \nabla \phi_l^k \, d\xb - \int_{\partial \mc{D}^k} \nb \cdot \fb^* \phi_l^k d\xb,\,\text{for},
    \end{aligned}
\end{equation}
where $\fb^k \cdot \nabla \phi_l^k:=\sum_{i=1}^d\fb_{i}^k\pp{\phi_l^k}{x_i}$, $\nb=\LRp{n_1,\dots,n_d}$ is the unit outward normal vector on the element boundaries $\partial\mc{D}^k$, $\fb^*=\LRp{\fb^*_1,\dots,\fb^*_d}$ is the numerical flux, and $\nb \cdot \fb^*:=\sum_{i=1}^dn_i\fb^*_i$. 

The numerical flux plays a key role in the stabilization and can be computed by solving the Riemann problem either exactly or approximately on each interface between two adjacent elements. Perhaps the simplest approach  is the Lax-Friedrichs method in which the speeds of waves are assumed to be the same \cite{lax_weak_1954,toro_riemann_2013} on the common interface for both adjacent elements. In particular, the wave speed is selected to be at least as large as the fastest wave speed appearing in the Riemann solution. Given that the definition of a numerical flux is usually associated with adjacent elements, for an element $k$, we will use $\mc{N}(k)$ to index an immediate neighbor element that share an edge with element $k$. It is popular in the community of DG methods to use superscripts "$-$" and "$+$" to describe the interior and exterior information respectively. However, in this work, we adopt the notation that is widely used in graph neural network theory to facilitate readability since we will soon discuss the connection between the DG method and GNN in \cref{sect:DGGNNTagent_framework}. For example, an element $\mc{D}^{\mc{N}(k)}$ is said to be a neighbor of the element $\mc{D}^k$ when the intersection of elemental boundaries $\partial \mc{D}^k\cap \partial\mc{D}^{\mc{N}(k)}$ has a positive $d-1$ Lebesgue measure. 

We also define a face of the element $\mc{D}^k$ by $\Gamma^e\subset\partial\mc{D}^k$ where $e$ is the local index of elemental faces (i.e., for a simplex element, $e=1,\dots,d+1$). Similarly, $\Gamma^{\mc{N}(e)}\subset\partial\mc{D}^{\mc{N}(k)}$.  
%On the other hand, let $\circledcirc$ be some generic function associated with $\circledast$ which is an index of a domain (i.e., $\circledast$ can be set to $k$ or $e$.), then the average operator $\dLRc{ \circledcirc^{\circledast} }$ is defined as $\half\LRp{ \circledcirc^{\circledast}+\circledcirc^{\mc{N}(\circledast)} }$ and the jump operator $\dLRs{ \circledcirc^{\circledast} }$ as $\circledcirc^{\circledast}-\circledcirc^{\mc{N}(\circledast)}$.
On the common face of an element  $\mc{D}^k$ and one of its neighbor $\mc{D}^{\mc{N}(k)}$, we define the average of the approximate solution $\ub^k$ as $\dLRcs{\ub^k} := \half\LRp{ \ub^{k}+\ub^{\mc{N}(k)}}$ and the jump (looking from element $\mc{D}^k$) as $\dLRs{\ub^k}:= \ub^{k}-\ub^{\mc{N}(k)}$. The average and the jump of any other quantity can defined analogously. 
%Moreover, if $\mc{N}(\circledast)$ is nonexistent and the intersection of the boundary $\partial\Omega$ and a domain with am index $\circledast$ has a positive $d-1$ Lebesgue measure, then $\dLRc{ \circledcirc^{\circledast} }:=\circledcirc^{\circledast}$ and $\dLRs{ \circledcirc^{\circledast} }:=\circledcirc^{\circledast}$.
%\hai{how we can define boundary condition? if average and jump from element itself}
%\jauuei{In our code, we simply replace those nodal data with Dirichlet data. Although a better way to do it is to enforce it weakly. (i.e., $\int_{\Gamma^e\cap\partial\Omega}u_h-u_{bc}\,d\xb=0$.)}
%\hai{what if I have reflective wall conditions or inflow as well? we always need ghost adjacent edges right?}
In this work, we consider the local Lax-Friedrichs flux  given as
\begin{equation*}
    \fb^*_i{\LRp{\ub^{k}, \ub^{\mc{N}(k)}}} = \dLRc{\fb^k_{i}} + \frac{\lambda}{2} \dLRs{\ub^k},\,\text{for}\,i=1,\dots,d.
\end{equation*}
The constant $\lambda$ is the local\footnote{In contrast to the local Lax-Friedrichs flux, we will have global Lax-Friedrichs flux if the global maximum eigenvalue is chosen as the wave speed. However, the computation of the global Lax-Friedrichs flux is not parallelization-friendly.} maximum eigenvalue of the directional flux Jacobian, computed by
\begin{equation*}
%    \lambda = \max_{\ub^k_h\in \LRc{\ub^k_h,\ub^{\mc{N}(k)}_h}}\eval{\snor{\nb \cdot \pp{ \fb}{\ub}}}_{\ub=\ub^{k}_h}.
\lambda = \max_{\ub^k\in \LRc{\ub^k,\ub^{\mc{N}(k)}}} \max_{\theta \in \eval{{\text{eig}\LRp{\nb \cdot \pp{ \fb}{\ub}}}}_{\ub=\ub^{k}}}\snor{\theta}
\end{equation*}
where $\nb \cdot \pp{ \fb}{\ub}:=\sum_{i=1}^dn_i\pp{ \fb_{i}}{\ub}$ and $\text{eig}\LRp{\nb \cdot \pp{ \fb}{\ub}}$ denotes the set of absolute values for eigenvalues of $\nb \cdot \pp{ \fb}{\ub}$. 
%To facilitate the discussion in the latter sections, we also define $\widehat{\,}$ quantity for $\fb^*_i$ in a similar way of $\ubhat_{h}$ and $\fbhat_{h,i}$. 
Considering a subset of interpolating nodes 
\begin{comment}
    \tanbui{Please use $N_e$ in place of $N_{ep}$ to avoid cumbersome notations and misunderstanding.}\jauuei{done}
\end{comment}
% \hai{Jau-Uei, do we have any reason to use $N_{ep}$? otherwise we should change as Tan suggested}
$\LRc{\xb^e_l}_{l=1}^{N_{e}}\subset\LRc{\xb_l}_{l=1}^{N_{p}}$ where $\xb_l\in\mc{D}^k$ and $N_{e}$ is the number of interpolating nodes residing on the face $e$. %We use superscript $e$ to indicate that the quantity under consideration is evaluated on face $e$.  
For $i=1,\dots,d$, let us denote the $i$th flux tensor on face $e$ as 
\begin{comment}
\tanbui{See my previous comment on removing the subscript $h$. The subscripts for the flux are too cumbersome and hard to follow. Best is to use commas to separate subscripts. For example, $\fhat^e_{i,q,l}$ is more readable than $\fhat^e_{h,iql}$. Need to use commas for more than 1 subscripts/superscripts.}\jauuei{done}
\end{comment}
$\fbhat^e_{i}\LRp{t}$ is defined as $\LRp{ \fbhat^e_{i,1}\LRp{t},\dots,\fbhat^e_{i,m}\LRp{t} }$ where $\fbhat^e_{i,q}\LRp{t}:=\LRp{ \fhat^e_{i,q,1}\LRp{t},\dots,\fhat^e_{i,q,{N_{e}}}\LRp{t} }^T$ with $\fhat^e_{i,q,l}\LRp{t}:=f_{i,q}\LRp{\ub^k(\xb^e_l,t)}$  $q=1,\dots,m$, and $l=1,\dots,N_{e}$. In addition, $\ubhat^e\LRp{t}$ is defined as $\LRp{ \ubhat^e_{1}\LRp{t},\dots,\ubhat^e_{m}\LRp{t} }$ where $\ubhat^e_{q}\LRp{t}:=\LRp{ \uhat^e_{q,1}\LRp{t},\dots,\uhat^e_{q,N_{e}}\LRp{t} }^T$ with $\uhat^e_{q,l}\LRp{t}:=u^k_{q}\LRp{\xb^e_l,t}$ for $q=1,\dots,m$ and $l=1,\dots,N_{e}$. 
%Similar to the usage of the superscript $k$, the superscript $e$ is also used to specify the restriction or to simply indicate that the quantity is somehow related to $e$. 
We can now rewrite the Lax-Friedrichs flux  $\fbhat^*:=\LRp{\fbhat^*_1,\dots,\fbhat^*_d}$ on a face $e$ of an element $\mc{D}^k$ in terms of nodal values as
\begin{equation*}
    \fbhat^*_i = \dLRc{\fbhat^e_{i}} + \frac{\lambda}{2} \dLRss{\ubhat^e},\,\text{for}\,i=1,\dots,d.    
\end{equation*}

With the natations in place, we now rewrite DG weak form \cref{eq:DG_weak_form}  in a matrix form. To that end, we define the following matrices
\begin{equation}
\eqnlab{VK_formula}
    \begin{aligned}
    &V_i^k:={(M^k)^{-1}S_i},
    &&\text{where }M^k_{rs}:=\int_{\mc{D}^k}\phi^k_r\phi^k_s\,d\xb,\,\text{and }
    S_{i,rs}:=\int_{\mc{D}^k}\pp{\phi^k_r}{x_i}\phi^k_s\,d\xb,\\
    &
    &&\text{for }r,s=1,\dots,N_p,
    \end{aligned}
\end{equation}
and $E^{k,e}:\underbrace{\mbb{R}^{N_{ep}}\times\dots\times\mbb{R}^{N_{ep}}}_{m\text{ times}}\mapsto\underbrace{\mbb{R}^{N_{p}}\times\dots\times\mbb{R}^{N_{p}}}_{m\text{ times}}$ is defined as
\begin{equation}
\eqnlab{EK_formula}
    \begin{aligned}
    % & E^{k,e}:\underbrace{\mc{R}^{N_{ep}}\times\dots\times\mc{R}^{N_{ep}}}_{m\text{ times}}\mapsto\underbrace{\mc{R}^{N_{p}}\times\dots\times\mc{R}^{N_{p}}}_{m\text{ times}},
    % &&\\
    & E^{k,e}(\cdot):=\LRp{M^k}^{-1}\mc{E}\LRp{M^e\LRp{\cdot}}
    &&\text{where }M^e_{sr}:=\int_{\Gamma^e}\phi^k_r\phi^{k,e}_s\,d\xb,\\
    &
    &&\text{for }r=1,\dots,N_p,\,s=1,\dots,N_{ep},
\end{aligned}
\end{equation}
where 
%
%$\LRc{\phi^{k,e}_s}_{s=1}^{N_{e}}$ are Lagrange polynomials in $\LRc{\phi^k_l}_{l=1}^{N_{p}}$ %i.e., the tuple $\LRp{k,e,s}$ can be paired with $\LRp{k,l}$) and is a set of 
whose restriction on face $e$ are non-trival.
%$\LRc{\xb^e_s}_{s=1}^{N_{ep}}$. 
The $E^{k,e}(\cdot)$ is a lift operator that lifts nodal values defined on the face $e$ to nodal values defined on the corresponding element $\mc{D}^k$.
The operator $\mc{E}\LRp{\cdot}$ is defined in the similar way described on page 188 in \cite{hesthaven_nodal_2007}. The major difference is that in this work $\mc{E}\LRp{\cdot}$ only takes nodal values of $e$ while in \cite{hesthaven_nodal_2007} it takes nodal values of $\partial\mc{D}^k$.  
\begin{comment}
\tanbui{The preceding sentence is too vague for me to understand how $\mc{E}\LRp{\cdot}$ is defined. If it is similar to some operator in \cite{hesthaven_nodal_2007}, then say it and point out where the reader can find the definition.}\jauuei{edited}
\end{comment}
%Furthermore, the output matrix \tanbui{which matrix?} is formed by simply inserting the input $\LRp{\cdot}$ in the positions corresponding to the face nodes in the element and stuffing the other positions with zeros (see also Section 6.2 and Section 10.2 in \cite{hesthaven_nodal_2007}) \tanbui{This is a very unclear definition/statement that you made}. 
%With all notations defined above, we are now in the position of rewriting 
We now can write \cref{eq:DG_weak_form} in a matrix form
\begin{equation}
    \eqnlab{DG_matrix_form_origin}
    \begin{aligned}
        \frac{d}{dt}\ubhat^k  & = \sum_{i=1}^d V^k_i\fbhat_{i}^k -\sum_{\Gamma^{e}\subset\partial\mc{D}^k} {E}^{k,e}\LRp{ \nb\cdot\fbhat^* },\quad k = 1, \ldots, K,
    \end{aligned}
\end{equation}
where $\frac{d}{dt}\ubhat^k=\LRp{\frac{d}{dt}\ubhat^{k}_{1},\dots,\frac{d}{dt}\ubhat^{k}_{m}}$, $\nb\cdot\fbhat^*:=\sum_{i=1}^d\LRp{n_i\fbhat^*_i} $, $V^k_i\fbhat_{i}^k=\LRp{ V^k_i\fbhat^k_{i,1}\LRp{t},\dots,V^k_i\fbhat^k_{i,m}\LRp{t} }$, $M^e\fbhat_{i}^k=\LRp{ M^e\fbhat^e_{i,1}\LRp{t},\dots,M^e\fbhat^e_{i,m}\LRp{t} }$, and $M^e\ubhat_h^e=\LRp{M^e\ubhat^{e}_{1},\dots,M^e\ubhat^{e}_{m}}$. 
% \hai{why we need two last equalities?}\jauuei{For average term and jump term in the $\fbhat^*_{h,i}$}
How to evaluate the integrals in \cref{eq:DG_weak_form} is critical in a DG method. In particular, we need to choose appropriate quadrature rules to balance computational resources and the stability of the DG scheme. There are two popular approaches. The first is the so-called discontinuous Galerkin collocation spectral element method (DGSEM) \cite{hesthaven_nodal_2007,kopriva_stability_2018} where Gauss–Lobatto points are the typical choices for both quadrature and the interpolation nodes. This approach is convenient but may suffer from instability induced by aliasing errors which could be easily observed in nonlinear conservation laws. To fix it, additional tricks might be required to ensure stability (i.e., discrete entropy stability or discrete energy stability). The second approach is over-integration where high-order quadrature nodes (i.e., Gauss quadrature or cubature rules) are deployed for de-aliasing \cite{kirby_-aliasing_2003,spiegel_-aliasing_2015,kopriva_stability_2018}. However, the latter is more computationally demanding than the former. In \cref{sect:DGGNNTagent_framework}, we shall develop an end-to-end learning framework to concurrently learn surrogate models for the Riemann solver and rectify the mismatch arising from the collocation-type integration and over-integration. 

\subsection{Discontinuous Galerkin graph neural network (\texttt{DGNet}) framework} 
\seclab{DGGNNTagent_framework}

In our previous work \cite{nguyen2022model}, we introduced the model-constrained tangent slope learning approach, $\texttt{mcTangent}$, for learning spatial discretizations of dynamical systems. This method aims to learn a surrogate model for the dynamical system's evolution operator so that the solutions of the learned dynamical system resemble solutions of the discretized governing equations. Consequently, the learned dynamical system can produce accurate and stable solutions over long time horizons with less computational time. Moreover, the learned dynamical model, once trained, is time-invariant and can be deployed with explicit or implicit time integration schemes. However, this approach has certain limitations due to simple neural network architecture. First, a dense fully connected neural network 
%is used to directly map the current states to the dynamical system's tangent slope. This approach 
overlooks the causality of the spatial correlation inherent in physics, where the evolution at a certain point is typically influenced by its neighbors. %but not all points as the way the dense network works.
Second, the neural network architecture fails to ensure important physics properties, such as conservation laws throughout the physics domain, making it incapable of solving shock-type problems. Third, the approach is not applicable when the domain and/or the mesh change in the prediction stage. Fourth, from the optimization and computation perspective, the dense fully connected neural network can be seen as overparameterized with many unimportant, or even redundant, parameters connecting states of distant points to the point under consideration. These unnecessary parameters complicate training and thus challenge the optimizer to achieve optimal solutions. Additionally, 
%the generalization capacity of the learned surrogate model is limited since the neural network input is fixed and high dimensional. For example, 
the learned dynamical system cannot generalize for different meshes and geometries.

%Observing the pros and cons of our previous
Aiming at overcoming the aforementioned limitations of the $\texttt{mcTangent}$ approach, this work presents an end-to-end learning  \oDGNet framework for learning DG discretizations\footnote{The approach can be easily adapted to other numerical methods such as upwind finite difference and finite volume methods.} for
%. The new approach also aims to learn a surrogate model to approximate the tangent slope of 
systems of nonlinear conservation laws \cref{eq:conservation_law} that could have discontinuous (shock) solutions. The tangent slope of the DG semi-discrete system \cref{eq:DG_matrix_form_origin} is the DG-spatial discretization operator, i.e. the right hand side, denoted as $\mc{F}\LRp{\ubhat^k}$. We can thus write \cref{eq:DG_matrix_form_origin} succinctly as
%of hyperbolic equations aligned with the DG method is defined as
\begin{equation}
    \mc{F}\LRp{\ubhat^k} = \frac{d}{dt}\ubhat^k.
    \eqnlab{DG_matrix_form}
\end{equation}
%where the right hand side is given in \cref{eq:DG_matrix_form_origin}. 
In this paper, our proposed approach for learning the  tangent slope is inspired by the similarity between the graph neural network (GNN) update  and the DG method.
%the high-order Discontinuous Galerkin (DG) method. 
As depicted in \cref{fig:DG_and_graph_dual_mesh}, the graph neural network is the dual of the DG mesh. 
The DG nodal values of an element $\mc{D}^k$ are treated as node attributes of the corresponding GNN vertex $\mc{N}^k$. On the other hand, a GNN edge connecting two vertices $\mc{N}^k$ and $\mc{N}^{\mc{N}(k)}$ models the flux of information exchanged between the corresponding two DG elements $\mc{D}^k$ and $\mc{D}^{\mc{N}(k)}$ .
\begin{comment}
\hai{Tan: we use $k$ for considering element and $\mathcal{N}(k)$ for neighboring elements. I guess your $\mc{N}(k)$ is $k$ and $\mc{N}^{k+1}$ would be $\mc{N}(k)$}. 
\end{comment}
The attribute for this edge can be used to represent the numerical flux between the two elements. 
%In our approach, this numerical flux surrogate part is the key component. 
From a data-driven scientific machine learning (SciML) perspective, GNN edges in   \oDGNet capture local interactions/correlation between subdomains on an unstructured mesh. 
%From the numerical method perspective, the numerical flux map can be seen as an approximation of a Riemann solver, which is typically computationally expensive for capturing discontinuities. 
In summary, {\em the \oDGNet approach combines the strengths of the \texttt{mcTangent} approach, the Discontinuous Galerkin method, and graph neural networks,  to deliver a SciML approach that could learn the way DG solves hyperbolic conservation laws. Since \oDGNet is a surrogate to the DG solver, it can generalize to challenging unseen scenarios, including new mesh and new geometry, that may be otherwise infeasible for existing methods.}

In the following two subsections, we discuss a model-constrained neural network block $\DGGNN\LRp{\ub;\bs{\theta}}$, where vector $\bs{\theta}$ is the collection of all network parameters, for learning the tangent slope $\FW \LRp{\ubhat^k}$ for an element $\mc{D}^k$. Since the block is applicable for all elements, we shall drop subscript, superscipt, and `` $\hat{}$ '' on $\ub$ for clarity. 
Unless otherwise specified, $\ub$, $\fb$, and $\fb^*$ will be referred to as nodal data/vectors for the rest of the paper. We also drop the explicit dependence of the network block $\DGGNN$ on $\bs{\theta}$ for the clarity of the exposition. We first discuss how to train the neural network block $\DGGNN\LRp{\ub}$\textemdash that is, finding the best network parameters $\bs{\theta}$\textemdash of the \oDGNet\hspace{-1ex}
in \cref{sect:trainingDGNetBlock}, and then discuss the architecture of the $\DGGNN\LRp{\ub}$ block in \cref{sect:designDGNetBlock}.

\subsection{How to train \oDGNet block $\DGGNN\LRp{\ub}$?}
\label{sect:trainingDGNetBlock}
The $\DGGNN\LRp{\ub}$ block is trained using both DG solutions $\ub$ and DG spatial discretization operator $\mc{F}(\ub)$. For temporal discretization, any explicit time integration scheme can be used for training $\DGGNN\LRp{\ub}$ \cite{nguyen2022model}.  An implicit approach can also be used to train \oDGNet at the expense of high computational cost. Once trained, the \oDGNet can evolve using any implicit or explicit time integration. In this paper, we deploy one-step $2^{\text{nd}}$-order strong stability-preserving Runge-Kutta (2nd-SSP-RK) time integration scheme \cite{hesthaven2007nodal} to solve semi-discretized equations \cref{eq:DG_matrix_form} for training $\DGGNN\LRp{\ub}$.  Our experience shows that a one-step temporal discretization for training is more efficient and practical for handling large-scale problems.
%in order to preserve the convergence rate of DG method. 
Specifically, given the snapshot of the solution $\ui{i}$ at time $t^i$, the DG solution at time $t^{i+1}$ is obtained by the following steps:
\begin{equation}
    \eqnlab{DG_2nd_scheme}
    \begin{aligned}
        \ui{i,1} & = S\LRp{\ui{i} + \dt \FW\LRp{\ui{i}}}, \\
        \ui{i+1} = \ui{i,2} & = S \LRp{\frac{1}{2}\LRp{\ui{i,1} + \ui{i} + \dt \FW\LRp{\ui{i,1}}}},
    \end{aligned}
\end{equation}
where the second superscripts ($1 $ and $ 2$) denote the first and second step of the 2nd SSP RK scheme, $\FW$ is DG spatial discretization in  \cref{eq:DG_matrix_form},
%a differentiable DG numerical solver evaluating the tangent slope of PDEs system,
and $S$ is a smooth slope limiter operator \cite{tu2005slope}.\footnote{\oDGNet can be used for other shock-capturing methods including artificial viscosity.} In the same fashion, we define the \oDGNet prediction $\ut{i+1}$ as follows:
\begin{equation}
\eqnlab{NN_2nd_scheme}
    \begin{aligned}
        \ut{i,1} & = S \LRp{\ui{i} + \dt \DGGNN\LRp{\ui{i}}} \\
        \ut{i+1} = \ut{i,2} & =  S \LRp{\frac{1}{2}\LRp{\ut{i,1} + \ui{i} + \dt \DGGNN\LRp{\ut{i,1}}}}.
    \end{aligned}
\end{equation}
For training the network block $\DGGNN{}$, we generate the training data $\LRc{\ui{i}}_{i=1}^{n_t}$ of $n_t$ snapshots from \cref{eq:DG_2nd_scheme}. Given a data set $\LRc{\ui{i}}_{i=1}^{n_t}$, {\em we consider  two strategies for learning the DG spatial discretization surrogate $\DGGNN$}: i) a naive data-driven tangent slope learning approach \texttt{nDGNet}, and ii) a model-constrained tangent slope learning approach \mcDGNet \hspace{-1ex}. 
% The key difference is that the former requires only the training data, while the latter requires a DG solver to impose the constraint that the (induced) neural network solution $\LRc{\ut{i}}_{i=1}^{n_t}$ satisfy the fully discrete DG scheme \cref{eq:DG_2nd_scheme} to a certain desirable level. We now present the two approaches in more detail.
%on machine learning predictions to ensure discretized governed equations \cite{nguyen2022model}. 

In the first approach, \nDGNet  with the $L^2-$ loss 
% \tanbui{is this L2 loss based on the nodal value, and thus the Euclidean L2 norm, or the L2-norm with integral? We need to define it here.}\tanbui{When I get to the randomization section 2.5, I think it is the big $L^2\LRp{\Omega}$: so write it explicitly everywhere for clarity: you need to define it of course.} \hai{it is L2-norm integral} 
(defined as $\nor{\ub}_{L^2\LRp{\Omega}} = \LRp{\int_\Omega \snor{\ub}^2 d\xb }^\half$) between \oDGNet solutions and DG solutions at all Runge-Kutta stages:
%. The loss function for this approach reads
\begin{equation}
    \label{eq:nDGGNNTagent_loss}
    \mc{L}_n = \sum_{i=2}^{n_t} \LRp{\nor{\ui{i,1} - \ut{i,1}}_{L^2\LRp{\Omega}}^2 + \nor{\ui{i,2} - \ut{i,2}}_{L^2\LRp{\Omega}}^2},
\end{equation}
% The flow of how to compute this loss function from \cref{eq:DG_2nd_scheme} and \cref{eq:NN_2nd_scheme} is given in \cref{fig:NN_architecture} with $\epsilon = 0$. In \cref{fig:NN_architecture},
where $\ut{i,1}$ and $\ut{i,2}$ are obtained from \cref{eq:NN_2nd_scheme}, and $\ui{i,1}$ and $\ui{i,2}$, the pre-computed Runge-Kutta stages data obtained from \cref{eq:DG_2nd_scheme}, are loaded during training.
% \hai{$\ui{i,1}$ and $\ui{i,2}$ are training data generated in advance, please see my answer below}

In the second approach, \mcDGNet \hspace{-1ex}, 
the loss function is given as
\begin{equation}
    \label{eq:mcDGGNNTagent_loss}
    \mc{L} = \mc{L}_{n} + \alpha \mc{L}_{mc},
\end{equation}
where $\alpha$ is the loss balance parameter and
\begin{equation*}
    \mc{L}_{mc} = \sum_{i=2}^{n_t} \LRp{\nor{\ubar{i,1} - \ut{i,1}}_{L^2\LRp{\Omega}}^2 + \nor{\ubar{i,2} - \ut{i,2}}_{L^2\LRp{\Omega}}^2}
\end{equation*}
with $\ubar{}$ and $\ut{}$ computed on-the-fly. How? For each time step, we randomize the pre-computed DG data $\ui{}$ ({\em not the Runge-Kutta stages}), which are then fed to the right-hand sides of \cref{eq:DG_2nd_scheme} to compute $\ubar{}$ and \cref{eq:NN_2nd_scheme} to compute $\ut{}$. 
\cref{fig:NN_architecture} illustrates the computation of $\ubar{i,1},\ubar{i,2}, \ut{i,1}$, and $\ut{i,2}$: \oDGNet and DG branches are corresponding to the on-the-fly computation of $\ut{}$ and $\ubar{}$, respectively. Meanwhile, the left side presents the procedure for computing the loss term $\mc{L}_n$. The key difference between the first and the second approaches is that the former requires only pre-computed training data, while the latter requires both pre-computed training data and DG spatial discretization $\mc{F}$ to compute the Runge-Kutta stages. As we shall show in  \cref{sect:Data_rand}, these two loss terms $\mc{L}_n$ and $\mc{L}_{mc}$ in the second approach induces implicit regularization that leads to a more accurate and stable $\DGGNN$ network. 
% the $\DGGNN$ network is obtained by minimizing the $L^2-$ loss between the direct differentiable DG numerical solver solution \tanbui{what is the direct differentiable DG numerical solver solution?} and neural network solutions. The schematic of the model-constrained \mcDGNet approach is presented in the \cref{fig:NN_architecture}. In this approach, we integrate the data randomization technique to introduce regularization, achieving a more accurate and stable $\DGGNN$ network than the first approach, which we will discuss next in \cref{sect:Data_rand}. The loss function for this approach reads
% \begin{equation}
%     \eqnlab{mcDGGNNTagent_loss}
%     \mc{L}_{mc} = \sum_{i=2}^{n_t} \nor{\ubar{i,1} - \ut{i,1}}_{L^2\LRp{\Omega}}^2 + \nor{\ubar{i,2} - \ut{i,2}}_{L^2\LRp{\Omega}}^2,
% \end{equation}
% where $\ubar{i}$ is the model-constrained solutions computed by \cref{eq:DG_2nd_scheme} with randomized inputs. It is worth noting that if there is no noise added to the training data, the \mcDGNet approach is equivalent to the \nDGNet approach.

% \input{Figs/Architecture.tex}
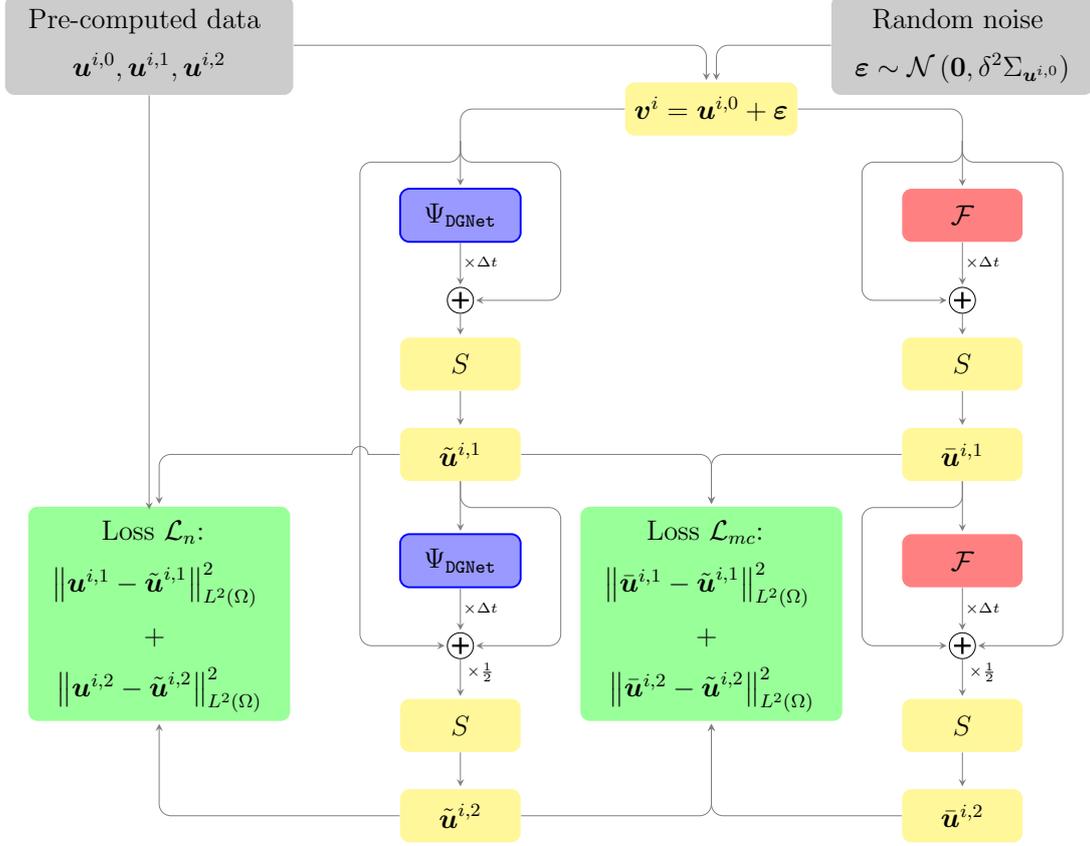
\begin{figure}[htb!]
    \centering
    \resizebox{.9\textwidth}{!}{\def\layersep{1.5cm}
\def\nodeinlayersep{.8cm}
\usetikzlibrary{calc}

\begin{tikzpicture}[
    node distance=\layersep,
    edge/.style={-stealth,shorten >=1pt, draw=black!50,thin, rounded corners=5pt},
    edge2/.style={draw=black!50,thin},
    neuron/.style={circle,fill=black!25,minimum size=10pt,inner sep=0pt},
    operator/.style={rectangle,fill=green!,minimum height= \nodeinlayersep, minimum width= 1.2 * \layersep, inner sep=0pt, rounded corners},
    input neuron/.style={neuron, fill=green!50,minimum size=12pt},
    output neuron/.style={neuron, fill=green!50,minimum size=12pt},
    hidden neuron/.style={neuron,,draw=blue,thick fill=blue!5},
    Forward map/.style={operator, fill=red!50},
    annot/.style={text width=4em, text centered},
    every node/.style={scale=1.0},
    node1/.style={scale=2.0},
    cross/.style={path picture={ 
        \draw[black, shorten <=2pt, shorten >=2pt, line width=1pt] 
            (path picture bounding box.south) -- (path picture bounding box.north); 
        \draw[black, shorten <=2pt, shorten >=2pt, line width=1pt] 
            (path picture bounding box.west) -- (path picture bounding box.east);
    }}
]

\node[rectangle,fill=yellow!50,minimum height= \nodeinlayersep, minimum width= 1.2 * \layersep, rounded corners] (u_in) at (3.5*\layersep, -.00*\nodeinlayersep) {$\ubr^{i} = \ui{{i,0}} + \epsb$};

% DG GNN RHS tikz figure
\node[rectangle,draw=blue,thick,fill=blue!40,minimum height= \nodeinlayersep, minimum width= 1.2 * \layersep, rounded corners] (DGGNN-1) at (1*\layersep, -2*\nodeinlayersep) {$\DGGNN$};
\node[rectangle,fill=yellow!50,minimum height= \nodeinlayersep, minimum width= 1.2 * \layersep, rounded corners] (SL-DGGNN-1) at (1*\layersep, -4.8*\nodeinlayersep) {$S$};
\node[rectangle,fill=yellow!50,minimum height= \nodeinlayersep, minimum width= 1.2 * \layersep, rounded corners] (DGGNN-RK-1) at (1*\layersep, -6.5*\nodeinlayersep) {$\ut{{i},1}$};
\node [draw,circle,cross,minimum width=.02*\layersep,line width=.3pt](cross_1) at (1*\layersep, -3.6*\nodeinlayersep){};
\node[rectangle,minimum height= \nodeinlayersep, minimum width= 1.2 * \layersep, rounded corners] (dt) at (1.2*\layersep, -2.9*\nodeinlayersep) {\tiny $ \times \dt$};

\draw[edge,thin] (u_in) -- (1*\layersep, -.00 *\nodeinlayersep) -- (DGGNN-1);
\draw[edge,thin] (1*\layersep, -0.6 *\nodeinlayersep) -- (1*\layersep, -1 *\nodeinlayersep) -- (2*\layersep, -1 *\nodeinlayersep) -- (2*\layersep, -3.6 *\nodeinlayersep) -- (cross_1);
\draw[edge,thin] (DGGNN-1) -- (cross_1);
\draw[edge,thin] (cross_1) -- (SL-DGGNN-1);
\draw[edge,thin] (SL-DGGNN-1) -- (DGGNN-RK-1);

\node[rectangle,draw=blue,thick,fill=blue!40,minimum height= \nodeinlayersep, minimum width= 1.2 * \layersep, rounded corners] (DGGNN-2) at (1*\layersep, -8.5*\nodeinlayersep) {$\DGGNN$};
\node[rectangle,fill=yellow!50,minimum height= \nodeinlayersep, minimum width= 1.2 * \layersep, rounded corners] (SL-DGGNN-2) at (1*\layersep, -11.6*\nodeinlayersep) {$S$};
\node[rectangle,fill=yellow!50,minimum height= \nodeinlayersep, minimum width= 1.2 * \layersep, rounded corners] (DGGNN-RK-2) at (1*\layersep, -13.3*\nodeinlayersep) {$\ut{{i},2}$};
\node [draw,circle,cross,minimum width=.02*\layersep,line width=.3pt](cross_2) at (1*\layersep, -10.1*\nodeinlayersep){};
\node[rectangle,minimum height= \nodeinlayersep, minimum width= 1.2 * \layersep, rounded corners] (dt) at (1.2*\layersep, -9.4*\nodeinlayersep) {\tiny $ \times \dt$};
\node[rectangle,minimum height= \nodeinlayersep, minimum width= 1.2 * \layersep, rounded corners] (dt) at (1.2*\layersep, -10.6*\nodeinlayersep) {\tiny $ \times \frac{1}{2}$};

\draw[edge,thin] (DGGNN-RK-1) -- (DGGNN-2);
\draw[edge,thin] (1*\layersep, -7.*\nodeinlayersep) -- (1*\layersep, -7.5*\nodeinlayersep) -- (2*\layersep, -7.5 *\nodeinlayersep) -- (2*\layersep, -10.1 *\nodeinlayersep) -- (cross_2);
\draw[edge,thin] (DGGNN-2) -- (cross_2);
\draw[edge,thin] (cross_2) -- (SL-DGGNN-2);
\draw[edge,thin] (SL-DGGNN-2) -- (DGGNN-RK-2);

\draw[edge,thin] (1*\layersep, -.6 *\nodeinlayersep) -- (1*\layersep, -1 *\nodeinlayersep) -- (0.*\layersep, -1 *\nodeinlayersep) -- (0.*\layersep, -10.1 *\nodeinlayersep) -- (cross_2);

% DG solver RHS tikz figure
\node[rectangle,fill=red!50,minimum height= \nodeinlayersep, minimum width= 1.2 * \layersep, rounded corners] (DGsol-1) at (6*\layersep, -2*\nodeinlayersep) {$\mc{F}$};
\node[rectangle,fill=yellow!50,minimum height= \nodeinlayersep, minimum width= 1.2 * \layersep, rounded corners] (SL-DGsol-1) at (6*\layersep, -4.8*\nodeinlayersep) {$S$};
\node[rectangle,fill=yellow!50,minimum height= \nodeinlayersep, minimum width= 1.2 * \layersep, rounded corners] (DGsol-RK-1) at (6*\layersep, -6.5*\nodeinlayersep) {$\ubar{{i},1}$};
\node [draw,circle,cross,minimum width=.02*\layersep,line width=.3pt](cross_3) at (6*\layersep, -3.6*\nodeinlayersep){};
\node[rectangle,minimum height= \nodeinlayersep, minimum width= 1.2 * \layersep, rounded corners] (dt) at (6.2*\layersep, -2.9*\nodeinlayersep) {\tiny $ \times \dt$};

\draw[edge,thin] (u_in) -- (6*\layersep, -.00 *\nodeinlayersep) -- (DGsol-1);
\draw[edge,thin] (6*\layersep, -.6 *\nodeinlayersep)  -- (6*\layersep, -1 *\nodeinlayersep) -- (5*\layersep, -1 *\nodeinlayersep) -- (5*\layersep, -3.6 *\nodeinlayersep) -- (cross_3);
\draw[edge,thin] (DGsol-1) -- (cross_3);
\draw[edge,thin] (cross_3) -- (SL-DGsol-1);
\draw[edge,thin] (SL-DGsol-1) -- (DGsol-RK-1);

\node[rectangle,fill=red!50,minimum height= \nodeinlayersep, minimum width= 1.2 * \layersep, rounded corners] (DGsol-2) at (6*\layersep, -8.5*\nodeinlayersep) {$\mc{F}$};
\node[rectangle,fill=yellow!50,minimum height= \nodeinlayersep, minimum width= 1.2 * \layersep, rounded corners] (SL-DGsol-2) at (6*\layersep, -11.6*\nodeinlayersep) {$S$};
\node[rectangle,fill=yellow!50,minimum height= \nodeinlayersep, minimum width= 1.2 * \layersep, rounded corners] (DGsol-RK-2) at (6*\layersep, -13.3*\nodeinlayersep) {$\ubar{{i},2}$};
\node [draw,circle,cross,minimum width=.02*\layersep,line width=.3pt](cross_4) at (6*\layersep, -10.1*\nodeinlayersep){};
\node[rectangle,minimum height= \nodeinlayersep, minimum width= 1.2 * \layersep, rounded corners] (dt) at (6.2*\layersep, -9.4*\nodeinlayersep) {\tiny $ \times \dt$};
\node[rectangle,minimum height= \nodeinlayersep, minimum width= 1.2 * \layersep, rounded corners] (dt) at (6.2*\layersep, -10.6*\nodeinlayersep) {\tiny $ \times \frac{1}{2}$};

\draw[edge,thin] (DGsol-RK-1) -- (DGsol-2);
\draw[edge,thin] (6*\layersep, -7.*\nodeinlayersep) -- (6*\layersep, -7.5*\nodeinlayersep) -- (5*\layersep, -7.5 *\nodeinlayersep) -- (5*\layersep, -10.1 *\nodeinlayersep) -- (cross_4);
\draw[edge,thin] (DGsol-2) -- (cross_4);
\draw[edge,thin] (cross_4) -- (SL-DGsol-2);
\draw[edge,thin] (SL-DGsol-2) -- (DGsol-RK-2);

\draw[edge,thin] (6*\layersep, -.6 *\nodeinlayersep) -- (6*\layersep, -1 *\nodeinlayersep) -- (7.*\layersep, -1 *\nodeinlayersep) -- (7.*\layersep, -10.1 *\nodeinlayersep) -- (cross_4);

\node[rectangle,fill=green!40,minimum height= \nodeinlayersep, minimum width= 1.6 * \layersep, rounded corners] (loss_mc) at (3.5*\layersep, -9.5*\nodeinlayersep) {\begin{tabular}{c} Loss  $\mc{L}_{mc}$: \vspace*{.2cm} \\  $\MSE{\ubar{{i},1} - \ut{{i},1}}$ \vspace*{.2cm} \\ $ + $ \vspace*{.2cm} \\ $ \MSE{\ubar{{i},2} - \ut{{i},2}}$ \end{tabular}};

\draw[edge,thin] (DGGNN-RK-1) -- (3.5*\layersep, -6.5 *\nodeinlayersep) -- (loss_mc);
\draw[edge,thin] (DGGNN-RK-2) -- (3.5*\layersep, -13.3 *\nodeinlayersep) -- (loss_mc);

\draw[edge,thin] (DGsol-RK-1) -- (3.5*\layersep, -6.5 *\nodeinlayersep) -- (loss_mc);
\draw[edge,thin] (DGsol-RK-2) -- (3.5*\layersep, -13.3 *\nodeinlayersep) -- (loss_mc);

% ML loss  term 
\node[rectangle,fill=green!40,minimum height= \nodeinlayersep, minimum width= 1.6 * \layersep, rounded corners] (loss_ml) at (-2.*\layersep, -9.5*\nodeinlayersep) {\begin{tabular}{c} Loss $\mc{L}_n$: \vspace*{.2cm} \\  $\MSE{\ui{{i},1} - \ut{{i},1}}$ \vspace*{.2cm} \\ $ + $ \vspace*{.2cm} \\ $ \MSE{\ui{{i},2} - \ut{{i},2}}$ \end{tabular}};

\draw[edge2,thin] (DGGNN-RK-1) -- (.08*\layersep, -6.5 *\nodeinlayersep)
    arc [start angle=0, end angle=180, radius=0.08*\layersep] 
    -- (-1.*\layersep, -6.5 *\nodeinlayersep);
\draw[edge,thin] (-1.*\layersep, -6.5 *\nodeinlayersep) -- (-2.*\layersep, -6.5 *\nodeinlayersep) -- (loss_ml);
\draw[edge,thin] (DGGNN-RK-2) -- (-2.*\layersep, -13.3 *\nodeinlayersep) -- (loss_ml);

% Pre-computed data
\node[rectangle,fill=gray!40,minimum height= \nodeinlayersep, minimum width= 1.6 * \layersep, rounded corners] (data_set) at (-2.1*\layersep, 1.2*\nodeinlayersep) {\begin{tabular}{c} Pre-computed data \vspace*{.2cm} \\  $\ui{i,0}, \ui{i,1}, \ui{i,2}$ \end{tabular}};
\draw[edge,thin] (data_set) -- (-2.1*\layersep, -7.58 *\nodeinlayersep);

\draw[edge,thin] (data_set) -- (3.45*\layersep, 1.2 *\nodeinlayersep) -- (3.45*\layersep, .5 *\nodeinlayersep);

% Randomize machine engine
\node[rectangle,fill=gray!40,minimum height= \nodeinlayersep, minimum width= 1.6 * \layersep, rounded corners] (random_noise_engine) at (6*\layersep, 1.2*\nodeinlayersep) {\begin{tabular}{c} Random noise \vspace*{.2cm} \\  $\epsb \sim \mc{N} \LRp{\mathbf{0}, \delta^2 \Sigma_{\ui{i,0}}}$ \end{tabular}};
\draw[edge,thin] (random_noise_engine) -- (3.55*\layersep, 1.2 *\nodeinlayersep) -- (3.55*\layersep, .5 *\nodeinlayersep);

\end{tikzpicture}}
    \caption{The schematic of constructing the \mcDGNet loss \cref{eq:mcDGGNNTagent_loss} using $2^\text{nd}$ order strong stability-preserving Runge-Kutta (2nd-SSP-RK) time integration scheme. To embed the data randomization technique, the random noise vector $\epsb  \sim \mc{N}\LRp{\bs{0}, \delta^2 \Sigma_{\ui{i,0}}}$ is added to the input of the neural network, where $\Sigma_{\ui{i,0}}=\diag\LRp{\LRp{\ui{i,0}}^2}$.
    % (\tanbui{each time step or each epoch? Do you imply that each epoch is only for one time step. In the general setting, each epoch could involve multiple time steps. Also, you didn't mention any epoch in \cref{sect:trainingDGNetBlock}, and this could be a confusion for the reader. Either you mention epoch in both places or remove it entirely.} \hai{I removed the "at each epoch" and we discuss more in the general setting section \cref{sect:general_training}, I meant adding noise to all samples (all time steps that fed to neural network) in the batch at each epoch})
    $\mc{F}$ is the DG spatial discretization operator. $S$ is the slope limiter operator \cite{tu2005slope} applied at the end of each stage of 2nd-SSP-RK scheme.}
    \figlab{NN_architecture}
\end{figure}

\subsection{The architecture design of the $\DGGNN$ block}
\label{sect:designDGNetBlock}
The $\DGGNN$ block is designed following a similar structure to graph neural networks, as illustrated in \cref{fig:NN_block_architecture} for a representative element $\mc{D}^k$. At the node level, the element nodal values $\ub^k$ serve as \oDGNet node attributes. 
%The volume flux vector ${\fb}^k$ is evaluated from $\ub^k$ and is then normalized. The normalization is applied to each component of the flux vector separately. To be more specific, 
The volume flux vector $\fb^k$  can be rearranged in the third order tensor form $f_{i,q,l}^k$ for $i = 1\hdots d, q = 1\hdots m, l = 1\hdots N_p$. We define the tensor $\bs{\eta}^k$ where $\eta_{i,q}^k = \max_{l} \LRp{\snor{f_{i,q,l}^k}, \beta}$ with $\beta = 10^{-16}$ to avoid zero division. The normalized flux $\overline{\fb}^k$ is computed and rearranged as
\begin{equation}
    \overline{f}_{i,q,l}^k = \frac{f_{i,q,l}^k}{\eta_{i,q}^k}
    \eqnlab{normalized_vol}
\end{equation}
which normalizes the value of $\overline{f}_{i,q,l}^k $ to be in $ \LRs{-1,1}$.  The normalized flux is passed through the volume neural network $\NNvol$ which modifies the normalized flux to learn the correction to collocation-type integration \cite{hesthaven_nodal_2007,kopriva_stability_2018}. This correction is necessary since the collocation-type integration method may suffer from instability induced by aliasing errors. 
% \tanbui{In order to correct something, you need to tell the reason why it needs a correction. We cannot expect the readers to understand this over-integration learning to avoid aliasing. Also, it is a good idea to make a connection with super-resolution with citations, which is essentially what it is trying to do here.} \hai{I added some sentences, cited again from discussion in the motivation section \cref{sect:motivation}}
The output of $\NNvol: \mbb{R}^{N_p} \mapsto \mbb{R}^{N_p}$ is denormalized by multiplying $\eta_{i,q}^k$ as follows
\begin{equation*}
    \tilde{\fb}_{i,q}^k = \eta_{i,q}^k \NNvol \LRp{\overline{\fb}_{i,q}^k}.
\end{equation*}
where $\tilde{\fb}_{i,q}^k$ is the corrected volume flux. Finally, we apply a collocation-type integration to compute the volume term. 

At the edge level, $\NNflux: \mbb{R}^{d + 1} \mapsto \mbb{R}^{1}$ is the numerical flux neural network representing the edges of \oDGNet. It plays the role of the Riemann solver in computing the numerical fluxes at the boundary points. The input of $\NNflux$ comprises the normalized average flux terms and normalized state jump terms, which are computed directly from flux and state at each shared face 
% \tanbui{Do we use edge or face? Choose one and uniformly use it everywhere to help the readers.} 
% \hai{I am gonna use face uniformly}
point between the current element $k$ and its corresponding neighboring element $\mc{N}\LRp{k}$
% \tanbui{Do you need all elements or you only need the corresponding neighboring element?}. 
% \hai{only need the corresponding neighboring element}
The normalization procedure is similar to the one at the node level. The average flux and state jump can be rearranged as 
$\dLRc{f_{i,q,j}^e}$ 
and 
$\dLRs{u_{q,j}^e}$, respectively, for $i = 1\hdots d, q = 1\hdots m, j = 1\hdots N_{e}$. 
We define the tensor $\bs{\psi}^e$ where $ \psi_{q,j}^e= \max \LRp{\snor{n_1 \dLRc{f_{1,q,j}^e}}, \hdots, \snor{n_d \dLRc{f_{d,q,j}^e}}, \snor{\dLRs{u_{q,j}^e}}, \beta}$ with $\beta = 10^{-16}$ to avoid dividing by zero.
% The operator $\odot$ denotes the elementwise multiplication operator, e.g. $\LRp{a_1, a_i, \hdots, a_d} \odot \LRp{b_1, b_i, \hdots, b_d} = \LRp{a_1 b_1, a_i b_i, \hdots, a_d b_d}$.
The normalized average and jump are computed as
\begin{equation}
    % \begin{aligned}
        \overline{\dLRs{u_{q,j}^e}}= \frac{\dLRs{u_{q,j}^e}}{\psi_{q,j}^e} 
        \quad \text{and} \quad 
        \overline{n_i \dLRc{f_{i,q,j}^e}}= \frac{n_i \dLRc{f_{i,q,j}^e}}{\psi_{q,j}^e},
    % \end{aligned}
    \eqnlab{normalized_flux}
\end{equation}
which normalize both $\overline{\dLRs{u_{q,j}^e}} $ and $  \overline{n_i \dLRc{f_{i,q,j}^e}} $ to be in $ \LRs{-1,1}$. The output of the numerical flux network $\NNflux$ is denormalized by multiplying ${\psi}_{q,j}^e$ to obtain the approximated numerical fluxes $\LRp{\nb \cdot \fb^*}$ as follows
\begin{equation*}
    \LRp{\nb \cdot \fb^*}_{q,j} = \NNflux \LRp{\overline{n_1 \dLRc{f_{1,q,j}^e}}, \hdots, \overline{n_d \dLRc{f_{d,q,j}^e}},  \overline{\dLRs{u_{q,j}^e}}}.
\end{equation*}
Subsequently, we compute boundary integrals (the second term on the right-hand side of \cref{eq:DG_weak_form}) with collocation-type integration. It is worth noting that the matrices $V^k$ and $E^{k,e}$ (derived from the element's geometry) are precomputed with \cref{eq:VK_formula} and \cref{eq:EK_formula} for evaluating the volume and flux terms, respectively.
\begin{remark}
{\em Note that we use the same flux network and volume network for all $p$ conservative components. We have found that the normalization step is key to the generalization of the \oDGNet framework for new geometries and/or meshes}. As we shall see in \cref{sect:numerics}, \oDGNet can be trained with some limited training data sets and can solve problems with completely different (out-of-distribution) initial conditions, boundary conditions, geometries, meshes. Moreover, since the inputs to the neural network are normalized in the $\LRs{-1,1}$ range, as recommended in \cite{lecun2002efficient}, we adopt the hyperbolic tangent activation function for faster convergence in training.    
\end{remark}

\begin{figure}[htb!]
    \centering
    \def\layersep{2.4cm}
    \def\nodeinlayersep{0.8cm}
    \usetikzlibrary{calc}
    \resizebox{\textwidth}{!}{\begin{tikzpicture}[
        node distance=\layersep,
        edge/.style={-stealth,shorten >=1pt, draw=black!50,thin},
        neuron/.style={circle,fill=black!25,minimum size=10pt,inner sep=0pt},
        operator/.style={rectangle,fill=green!,minimum height= \nodeinlayersep, minimum width= 0.8 * \layersep, inner sep=0pt, rounded corners},
        input neuron/.style={neuron, fill=green!50,minimum size=12pt},
        output neuron/.style={neuron, fill=green!50,minimum size=12pt},
        hidden neuron/.style={neuron, fill=blue!50},
        Forward map/.style={operator, fill=red!50},
        annot/.style={text width=4em, text centered},
        every node/.style={scale=1.0},
        node1/.style={scale=2.0},
        cross/.style={path picture={ \draw[black, shorten <=2pt, shorten >=2pt, line width=1pt] (path picture bounding box.south) -- (path picture bounding box.north); \draw[black, shorten <=2pt, shorten >=2pt, line width=1pt] (path picture bounding box.west) -- (path picture bounding box.east);}}
    ]
    % \draw[draw=blue, thick, fill=blue!5, rounded corners] (-2.2*\layersep, -12.*\nodeinlayersep) rectangle (6.8*\layersep, .2*\nodeinlayersep); 
    
    \draw[draw=black, thick, fill=blue!10, rounded corners] (-1.*\layersep, -9.4*\nodeinlayersep) rectangle (6.3*\layersep, -0.2*\nodeinlayersep);

    \draw[draw=black!50,thick, dashed] (-2.*\layersep, -3.4*\nodeinlayersep) -- (6.5*\layersep, -3.4*\nodeinlayersep);
    \node [align=left](Node Level) at (-1.6*\layersep, -3.0*\nodeinlayersep) {Node Level };
    \node [align=left](Edge Level) at (-1.6*\layersep, -3.8*\nodeinlayersep) {Edge Level};

    \node[rectangle, draw = black,fill=yellow!50,minimum height= \nodeinlayersep, minimum width= 0.8 * \layersep, rounded corners] (u_in) at (-1.6*\layersep, -2*\nodeinlayersep) {$\ub^{{k}}$};
    
    \node[rectangle, draw = black,fill=white!50,minimum height= \nodeinlayersep, minimum width= 0.8 * \layersep, rounded corners] (fu) at (2.*\layersep, -2*\nodeinlayersep) {$\fb^{k}$};

    % \node[rectangle,fill=yellow!50,minimum height= \nodeinlayersep, minimum width= 0.8 * \layersep, rounded corners] (volintegration) at (4.*\layersep, -2*\nodeinlayersep) {$\LRp{M^k}^{-1} \int_{\Omega^k} \NNvol \LRp{\fb^k} \pp{\phi\LRp{\xb}}{\xb} d\xb$};

    % \node[rectangle,fill=yellow!50,minimum height= \nodeinlayersep, minimum width= 0.8 * \layersep, rounded corners] (fluxintegration) at (4.*\layersep, -7*\nodeinlayersep) {$-\LRp{M^k}^{-1} \int_{\partial\Omega^k}  \nb \cdot \fb^* \phi\LRp{\xb} d\xb$};

    \node[rectangle, draw = black,fill=white!50,minimum height= \nodeinlayersep, minimum width= 0.8 * \layersep, rounded corners] (corrected_flux) at (3.9*\layersep, -2*\nodeinlayersep) {$\tilde{\fb}_{i,q}^k = \eta_{i,q}^k \NNvol \LRp{\overline{\fb}_{i,q}^k}$};

    \node[rectangle, draw = black,fill=white!50,minimum height= \nodeinlayersep, minimum width= 0.8 * \layersep, rounded corners] (volintegration) at (5.3*\layersep, -2*\nodeinlayersep) {$\sum_{i}^d V_i^k \tilde{\fb}_i^k $};

    \node[rectangle, draw = black,fill=white!50,minimum height= \nodeinlayersep, minimum width= 0.8 * \layersep, rounded corners] (fluxintegration) at (4.7*\layersep, -8.4*\nodeinlayersep) {$ - \sum_{e} E^{k,e} \LRp{\nb \cdot \fb^*}$};

    \node[rectangle, draw = black,fill=white!50,minimum height= \nodeinlayersep, minimum width= 0.8 * \layersep, rounded corners, align=center] (flux_ML) at (1.3*\layersep, -8.4*\nodeinlayersep)
    {$\LRp{\nb \cdot \fb^*}_{q,j}  = {\psi}_{q,j}^e \NNflux \LRp{\overline{n_1 \dLRc{f_{1,q,j}^e}}, \hdots, \overline{n_d \dLRc{f_{d,q,j}^e}}, \overline{\dLRs{u_{q,j}^e}}}$};

    \node[rectangle, draw = black,fill=white!100,minimum height= \nodeinlayersep, minimum width= 0.8 * \layersep, rounded corners] (u_jump) at (0.6*\layersep, -4.8*\nodeinlayersep) {$\dLRs{\ub^e}$};
    \node[rectangle, draw = black,fill=white!100,minimum height= \nodeinlayersep, minimum width= 0.8 * \layersep, rounded corners] at (2.*\layersep, -4.8*\nodeinlayersep) (f_average) {$\left\{\!\!\left\{ {\fb^e} \right\}\!\!\right\}$};

    \draw[edge,thin] (u_in) -- (fu);
    \draw[edge,thin] (fu) -- (corrected_flux);
    \draw[edge,thin] (fu) -- (f_average);
    \draw[edge,thin] (f_average) -- (2.*\layersep, -7.7*\nodeinlayersep);
    \draw[edge,thin] (0.6*\layersep, -2.*\nodeinlayersep) -- (u_jump);
    \draw[edge,thin] (u_jump) -- (0.6*\layersep, -7.7*\nodeinlayersep);

    \draw[edge,thin] (flux_ML) -- (fluxintegration);

    \node[rectangle, draw = black,fill=white!100,minimum height= \nodeinlayersep, minimum width= 1.8 * \layersep, rounded corners] at (1.25*\layersep, -6.45*\nodeinlayersep) {normalization};

    % \node[rectangle, draw = black,fill=white!100,minimum height= \nodeinlayersep, minimum width= 0.8 * \layersep, rounded corners] at (2.*\layersep, -4.8*\nodeinlayersep) {$\dLRc{\fb^e}$};

    \node [draw,circle,cross,minimum width=.02*\layersep,line width=.3pt](cross_1) at (6.*\layersep, -2*\nodeinlayersep){};

    \node[rectangle, draw = black, rotate = 90,fill=white!100,minimum height= \nodeinlayersep, minimum width= 0.8 * \layersep, rounded corners] at (2.8*\layersep, -2*\nodeinlayersep) {normalization};

    \draw[edge,thin] (fluxintegration) -- (6.*\layersep, -8.4*\nodeinlayersep) -- (cross_1);
    \draw[edge,thin] (corrected_flux) -- (volintegration);
    \draw[edge,thin] (volintegration) -- (cross_1);
    \draw[edge,thin] (cross_1) -- (6.5*\layersep, -2*\nodeinlayersep);

    % Neighboring elements
    % \draw[draw=black, thick, fill=blue!0, rounded corners] (-1.*\layersep, -11.6*\nodeinlayersep) rectangle (6.3*\layersep, -9.8*\nodeinlayersep);

    % \draw[draw=black!50,thick, dashed] (-2.*\layersep, -13.6*\nodeinlayersep) -- (6.5*\layersep, -13.4*\nodeinlayersep);
    % \node [align=left](Node Level 2) at (-1.6*\layersep, -12.0*\nodeinlayersep) {Node Level};
    % \node [align=left](Edge Level 2) at (-1.6*\layersep, -14.5*\nodeinlayersep) {Edge Level};

    \node[rectangle, draw = black,fill=yellow!50,minimum height= \nodeinlayersep, minimum width= 0.8 * \layersep, rounded corners] (u_in_neighbor) at (-1.6*\layersep, -4.80*\nodeinlayersep) {$\ub^{\mathcal{N}\LRp{k}}$};
    % \node[rectangle, draw = white,fill=white!0,minimum height= \nodeinlayersep, minimum width= 0.8 * \layersep, rounded corners] (fu) at (2.5*\layersep, -10.80*\nodeinlayersep) {The same procedure for neighboring element $\mc{N}(k)$};
    
    \draw[edge,thin] (u_in_neighbor) -- (u_jump);
    % \draw[edge,thin] (u_in_neighbor) -- (-.6*\layersep, -10.80*\nodeinlayersep);
    
    % \node[rectangle, draw = black,fill=yellow!50,minimum height= \nodeinlayersep, minimum width= 0.4 * \layersep, rounded corners] (u_total) at (-2.5*\layersep, -6.50*\nodeinlayersep) {$\ub$};

    % \draw[edge,thin] (u_total) -- (-2.1*\layersep, -6.5*\nodeinlayersep) -- (-2.1*\layersep, -2*\nodeinlayersep) -- (u_in);
    % \draw[edge,thin] (u_total) -- (-2.1*\layersep, -6.5*\nodeinlayersep) -- (-2.1*\layersep, -10.80*\nodeinlayersep) -- (u_in_neighbor);
    % \draw[edge,thin] (5.6*\layersep, -10.80*\nodeinlayersep) -- (6.6*\layersep, -10.80*\nodeinlayersep) -- (6.6*\layersep, -6.50*\nodeinlayersep) -- (7.0*\layersep, -6.50*\nodeinlayersep);
    
\end{tikzpicture}}
    \caption{{\bf Description of $\DGGNN$ architecture block}. For the $k$-th element $\mc{D}^k$, at the node level, the element nodal values $\ub^k$ serve as node attributes in the \oDGNet\hspace{-1ex}. The volume flux vector ${\fb}^k$ is evaluated %algebraically 
    from $\ub^k$ and then normalized to $\overline{\fb}^k$ as in \cref{eq:normalized_vol}. The volume neural network $\NNvol:\mbb{R}^{N_p} \mapsto \mbb{R}^{N_p}$ modifies the normalized flux to learn the corrections. The output is denormalized by multiplying factor ${\eta}_{i,q}^k$ to get the corrected volume flux, $\tilde{\fb}_{i,q}^k$. Finally, we apply the collocation-type integration rule to compute the volume term. 
    At the edge level, the edge network $\NNflux$ represents the graph neural network edges and surrogates the Riemann solver. It computes the numerical fluxes at the shared face points of the element $k$ and its corresponding element $\mc{N}\LRp{k}$. The input for $\NNflux : \mbb{R}^{d+1} \mapsto \mbb{R}^{1} $ comprises the normalized average flux terms
    % $\overline{n_i \dLRc{f_{i,q,j}^e}}$ 
    and normalized state jump, 
    % $\overline{\dLRs{u_{q,j}^e}}$ 
    as described in \cref{eq:normalized_flux}. The output of the numerical flux network is denormalized by multiplying the factor ${\psi}_{q,j}^e$ to evaluate the numerical flux. Subsequently, the approximated numerical fluxes undergo a collocation-type integration rule for evaluating the flux term. The flux and volume terms are added up, returning the tangent slope for the $k$-th element.}
    % The same procedure is applied to all elements. Finally, the \oDGNet tangent slope is obtained by concatenating the elemental tangent slopes from all elements.
    %\tanbui{we can remove the block for $\mc{D}^{\mc{N}(k)}$ to really mean the block is for every element. I will talk to you about that.} \hai{I have modified as you suggested} } 
    \figlab{NN_block_architecture}
\end{figure}
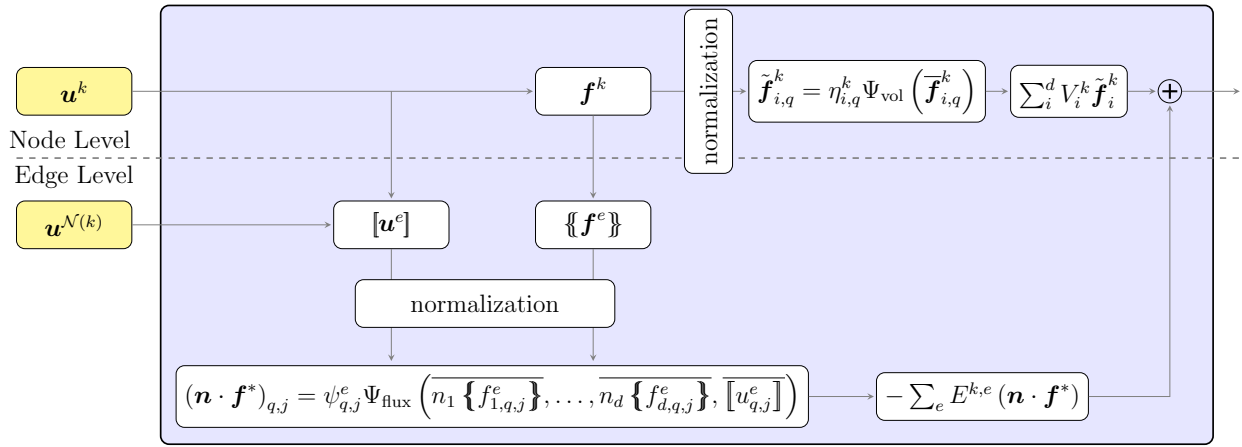

\begin{remark}
We would like to point out that the proposed \oDGNet can handle new boundary conditions weakly, just like DG methods, by appropriately specifying $\ub^{\mc{N}\LRp{k}}$ and $\fb^{\mc{N}\LRp{k}}$ for all elements on the domain boundaries. 
\end{remark}

\subsection{Data randomization}
\seclab{Data_rand}

Adding a small amount of noise to the training data, as noted in \cite{sanchez2020learning}, enhances generalization on unseen data and reduces accumulated errors in long-term predictions. Specifically, corrupted training data simulates the accumulated error, which could be amplified when fed into neural networks for subsequent predictions. Additionally, randomization is known to induce regularization of the gradient of the loss function with respect to the inputs \cite{reed1992regularization}. In \cite{bishop1995training}, it was shown that adding noise to data is analogous to introducing Tikhonov regularization to the loss function, with noise variance acting as the regularization parameter, thereby improving model generalization.  When an appropriate noise level is applied, the neural network is encouraged to learn a smooth function of the input data
%, in the sense of sensitivity to input variations 
\cite{matsuoka1992noise}, and thus enhancing the stability of long-term predictions \cite{poggio1990networks}.

In our previous work \cite{nguyen2022model}, we demonstrated that data randomization induces regularization, which is crucial for the long-term prediction capability of trained networks. In this work, we carry out a similar analysis to the \oDGNet approach, particularly for \mcDGNet \hspace{-1ex}. As we shall illustrate, randomization promotes similarity
not only between $\DGGNN\LRp{\ub}$ and $\mc{F}(\ub)$ but also their Jacobian with respect to $\ub$. 
%between the derivative of the DG numerical solver $\FW$ and the derivative of the \oDGNet neural network,
This enhances the stability (if the DG method does) and accuracy of the \oDGNet method.\footnote{Suppose, in a hypothetical scenario, $\DGGNN\LRp{\ub}$ and $\mc{F}(\ub)$ match their values and derivatives at multiple values of $\ub$. Then,  $\DGGNN{\LRp\ub}$ is a Hermite interpolation of  $\mc{F}(\ub)$ with second-order accuracy.} To start, we randomize the input $\ub$ by adding a small amount of noise
% \tanbui{refer to my concern about epoch versus time step above}
. The randomized input $\ubr$ is defined as
\begin{equation}
    \label{eq:add_noise}
    \ubr = \ub + \epsb,
\end{equation}
where $\epsb$ is a standard normal random vector $\epsb \sim \mc{N}\LRp{\bs{0}, \delta^2 \Sigma_{\ub}}$ with a factor $\delta^2 > 0$ and diagonal matrix $\Sigma_{\ub} = \diag(\ub^2)$. The details on how to implement in practical code is presented in \cref{sect:general_training}.
%\hai{changing the Gaussian distribution as we do in numerical results}
% \hai{we need to discuss noise is the multiplication between random noise and $\ub$ magnitude there! that is how we implement it in the code}
It is important to note that the following arguments apply to any random vector with independent components, each being a random variable with zero mean and variances $\delta^2$. Let $\mathbb{E}\LRs{\cdot}$ denote the expectation with respect to $\epsb$. Following \cite{an1996effects}, for a generic loss function $\mc{L}\left(\ub\right)$, we have:
\begin{equation}
\label{eq:Expect_form}
\begin{aligned}
    \mathbb{E}\LRs{\mc{L}\LRp{\ubr}} = & \mc{L}\LRp{\ub} + \mathbb{E}\LRs{\eval{\pp{\mc{L}}{\ub}}_{\ub} \epsb} +  \half \mathbb{E}\LRs{ \epsb^T \eval{\pp{^2 \mc{L}}{\ub^2}}_{\ub} \epsb} 
    +  \mathbb{E}\LRs{o\LRp{\nor{\epsb}^2}}
    \\
    \approx & \mc{L}\LRp{\ub} +  \half \mathbb{E}\LRs{ \epsb^T \eval{\pp{^2 \mc{L}}{\ub^2}}_{\ub} \epsb},
    \end{aligned}
\end{equation}
We assume the noise $\epsilon$ is sufficiently small (relative to $\ub$) such that the
with standard ``small oh" notation for the
higher-order term $o\left(|\epsb|^2\right)$. For clarity, we define several new variables $\zbar{1}, \zbar{2}$, $\zt^{1}, \zt^{2}$, $F^1, F^2$ and $\Psi^1, \Psi^2$ 
%are Runge-Kutta stage 
as functions of input $\vb$ for DG  and oDGNet solutions as follows:
% \oDGNet updated solutions (before applying the slope limiter operator), re, $F^1, F^2$ and $\Psi^1, \Psi^2$ are functions of input $\vb$ for DG  and \oDGNet solutions after imposing the slope limiter.
% \tanbui{Please list them here. The readers cannot know what are new and what are old.}
% {I added two sentences}  \hai 
\begin{equation*}
    \eqnlab{operators}
    \begin{aligned}
        \zbar{1}(\ubr^i) & = \ubr^i + \dt \FW(\ubr^i), & \ubar{i,1}\LRp{\ubr^i} = F^1(\ubr^i) & = S(\zbar{1}(\ubr^i)), \\
        \zt^1(\ubr^i) & = \ubr^i + \dt \DGGNN(\ubr^i), & \ut{i,1}\LRp{\ubr^i} = \Psi^1(\ubr^i) & = S(\zt^1(\ubr^i)), \\
        \zbar{2}(\ubr^i) & = \half\LRp{\ubar{i,1} + \ubr^i + \dt \FW(\ubar{i,1})}, & \ubar{i,2}\LRp{\ubr^i} = F^2(\ubr^i) & = S(\zbar{2}(\ubr^i)), \\
        \zt^2(\ubr^i) & = \half\LRp{\ut{i,1} + \ubr^i + \dt \DGGNN(\ut{i,1})}, & \ut{i,2}\LRp{\ubr^i} = \Psi^2(\ubr^i) & = S(\zt^2(\ubr^i)).
    \end{aligned}
\end{equation*}

% \hai{[TODO: modify proofs]}

It is important to note that the slope limiter operator $S\LRp{\cdot}$ is nonlinear and applied to the conservative variables at the end of each stage in the 2nd-SSP-RK scheme. Given the randomized inputs $\ubr$ from \cref{eq:add_noise} and the defined operators from \cref{eq:operators}, the randomized loss function version of the original loss function \cref{eq:mcDGGNNTagent_loss} reads:
% \begin{equation}
%     \label{eq:mcDGGNNTagent_loss_rand}
%     \begin{aligned}
%         \mc{L}_\text{rand} = & \sum_{i=1}^{n_t-1} \underbrace{ \nor{\ui{i+1,1} - \Psi^1\LRp{\ubr^{i}}}_{L^2\LRp{\Omega}}^2 + \nor{\ui{i+1,2} - \Psi^2\LRp{\ubr^{i}}}_{L^2\LRp{\Omega}}^2}_{\mc{L}_{n}\LRp{\ubr^{i}}} \\
%         + \alpha & \sum_{i=1}^{n_t-1} \underbrace{ \nor{F^1\LRp{\ubr^{i}} - \Psi^1\LRp{\ubr^{i}}}_{L^2\LRp{\Omega}}^2 + \nor{F^2\LRp{\ubr^{i}} - \Psi^2\LRp{\ubr^{i}}}_{L^2\LRp{\Omega}}^2}_{\mc{L}_{mc}\LRp{\ubr^{i}}}.
%     \end{aligned}
% \end{equation}

\begin{equation}
    \label{eq:mcDGGNNTagent_loss_rand}
    \begin{aligned}
        \mc{L}_\text{rand} = & \sum_{i=1}^{n_t-1}  \LRp{\nor{\ui{i+1,1} - \Psi^1\LRp{\ubr^{i}}}_{L^2\LRp{\Omega}}^2 + \nor{\ui{i+1,2} - \Psi^2\LRp{\ubr^{i}}}_{L^2\LRp{\Omega}}^2} \\
        + \alpha & \sum_{i=1}^{n_t-1}  \LRp{\nor{F^1\LRp{\ubr^{i}} - \Psi^1\LRp{\ubr^{i}}}_{L^2\LRp{\Omega}}^2 + \nor{F^2\LRp{\ubr^{i}} - \Psi^2\LRp{\ubr^{i}}}_{L^2\LRp{\Omega}}^2}.
    \end{aligned}
\end{equation}

Let us define
\begin{equation}
    \eqnlab{P1}
    \mc{P}_1 \LRp{\ui{i}} = \mb{Tr} \LRs{\LRp{\eval{\nabla_{\ub}\Psi^1}_{\ui{i}}}^T M \Sigma_{\ui{i}} \LRp{\eval{\nabla_{\ub}\Psi^1}_{\ui{i}}}} + \mb{Tr} \LRs{\LRp{\eval{\nabla^2_{\ub}\Psi^1}_{\ui{i}}} \odot M \Sigma_{\ui{i}} \LRp{\ui{i+1,1} - \eval{\Psi^1}_{\ui{i}}}}, 
\end{equation}
\begin{equation}
    \eqnlab{P2}
    \mc{P}_2 \LRp{\ui{i}} = \mb{Tr} \LRs{\LRp{\eval{\nabla_{\ub}\Psi^2}_{\ui{i}}}^T M \Sigma_{\ui{i}} \LRp{\eval{\nabla_{\ub}\Psi^2}_{\ui{i}}}}
    + \mb{Tr} \LRs{\LRp{\eval{\nabla^2_{\ub}\Psi^2}_{\ui{i}}} \odot M \Sigma_{\ui{i}} \LRp{\ui{i+1,2} - \eval{\Psi^2}_{\ui{i}}}},
\end{equation}
\begin{equation}
    \eqnlab{R1}
    \begin{aligned}
        \mc{R}_1\LRp{\ui{i}} = & \underbrace{\mb{Tr} \LRs{\LRp{\eval{\nabla_{\ub}F^1}_{\ui{i}} - \eval{\nabla_{\ub}\Psi^1}_{\ui{i}}}^T M \Sigma_{\ui{i}} \LRp{\eval{\nabla_{\ub}F^1}_{\ui{i}} - \eval{\nabla_{\ub}\Psi^1}_{\ui{i}}}}}_{\mc{R}_{1A}} \\
        & + \underbrace{\mb{Tr} \LRs{\LRp{\eval{\nabla^2_{\ub}F^1}_{\ui{i}} - \eval{\nabla^2_{\ub}\Psi^1}_{\ui{i}}} \odot M \Sigma_{\ui{i}} \LRp{\eval{F^1}_{\ui{i}} - \eval{\Psi^1}_{\ui{i}}}}}_{\mc{R}_{1B}},
    \end{aligned}
\end{equation}
and 
\begin{equation}
    \eqnlab{R2}
    \begin{aligned}
        \mc{R}_2\LRp{\ui{i}} = & \mb{Tr} \LRs{\LRp{\eval{\nabla_{\ub}F^2}_{\ui{i}} - \eval{\nabla_{\ub}\Psi^2}_{\ui{i}}}^T M \Sigma_{\ui{i}} \LRp{\eval{\nabla_{\ub}F^2}_{\ui{i}} - \eval{\nabla_{\ub}\Psi^2}_{\ui{i}}}} \\
        & + \mb{Tr} \LRs{\LRp{\eval{\nabla^2_{\ub}F^2}_{\ui{i}} - \eval{\nabla^2_{\ub}\Psi^2}_{\ui{i}}} \odot M \Sigma_{\ui{i}} \LRp{\eval{F^2}_{\ui{i}} - \eval{\Psi^2}_{\ui{i}}}},
    \end{aligned}
\end{equation}
where $\odot$ denotes the product of a third-order tensor and a vector, $M$ is the global mass matrix assembled from local element mass matrix $M^k$, $\Sigma_{\ui{i}}$ is the covariance matrix with respect to the sample $\ui{i}$. %Note that $M$ is used for notation purposes in the analysis; this large 
Note that the global mass matrix is not explicitly computed using the DG method. Replacing $\mc{L}$ in \cref{eq:Expect_form} with $\mc{L}_\text{rand}$, we have the following result.
% \begin{equation}
%     \eqnlab{Expect_form_rand}
%     \begin{aligned}
%         \mathbb{E}\LRs{\mc{L}_\text{rand}} \approx & \mc{L}_\text{mc} + \delta^2 \sum_{i=1}^{n_t-1} \LRp{\mc{R}_1\LRp{\ui{i}} + \mc{R}_2\LRp{\ui{i}}},
%     \end{aligned}
% \end{equation}
\begin{theorem}
    %Given a training snapshot data set $\LRc{\ub^i}_1^{n_t}$, a new randomized version is generated per training epoch by \cref{eq:add_noise}. Then,
    Let the data $\LRc{\ub^i}_{i=1}^{n_t}$ be randomized as in \cref{eq:add_noise} for every epoch.    There holds:
    \begin{equation}
        \mathbb{E}\LRs{\mc{L}_\text{rand}} = \mc{L} + \delta^2 \sum_{i=1}^{n_t-1} \LRs{ \mc{P}_1\LRp{\ui{i}} + \mc{P}_2\LRp{\ui{i}} + \alpha \LRp{\mc{R}_1\LRp{\ui{i}} + \mc{R}_2\LRp{\ui{i}}}} +  \mathbb{E}\LRs{o\LRp{\nor{\epsb}^2}}.
        \label{eq:Expect_form_rand}
    \end{equation}
    \label{theo:data_rand}
\end{theorem}

Assuming that the order of minimization and expectation can be
interchanged\footnote{The conditions under which the interchange is
valid can be consulted in \cite[Theorem 14.60]{RockafellarWetts98}.} and  that $\mathbb{E}\LRs{o\LRp{\nor{\epsb}^2}}$ in \cref{eq:Expect_form} is negligible, we have
\[
\mathbb{E}\LRs{\min_{\bs{\theta}} \mc{L}_\text{rand}} = \min_{\bs{\theta}} \mathbb{E}\LRs{\mc{L}_\text{rand}} = \min_{\bs{\theta}} \mc{L}_\text{reg},  
\]
where
\begin{equation}
    \mc{L}_\text{reg} := \mc{L} + \delta^2 \sum_{i=1}^{n_t-1} \LRs{ \mc{P}_1\LRp{\ui{i}} + \mc{P}_2\LRp{\ui{i}} + \alpha \LRp{\mc{R}_1\LRp{\ui{i}} + \mc{R}_2\LRp{\ui{i}}} }.
    \label{eq:mcDGGNNTagent_loss_rand_reg}
\end{equation}

{\em Thus, on average, training the randomized loss function $\mc{L}_\text{rand}$ in \cref{eq:mcDGGNNTagent_loss_rand} is (approximately) equivalent to training regularized loss $\mc{L}_\text{reg}$ in \cref{eq:mcDGGNNTagent_loss_rand_reg}, which consists of the original loss function $\mc{L}$ and regularization terms $\mc{P}_1, \mc{P}_2, \mc{R}_1$ and $\mc{R}_2$ with the noise variance $\delta^2$ as the regularization parameter.}
\begin{remark}
    In practice, we do not randomize the whole training data at each epoch, as stated in \cref{theo:data_rand}, but its random minibatch. Our approach is thus a minibatch stochastic gradient descent (SGD) for solving the optimization problem: $ \min_{\bs{\theta}} \mathbb{E}\LRs{\mc{L}_\text{rand}}$. The convergence of our minibatch SGD can be analyzed using the standard settings, which can be consulted from \cite{BottouEtAl2018}.
    \label{rem:SGD}
\end{remark}

Now, suppose that $\mc{L}_\text{rand}$ is at its minimum (on average). Since the minimal value of $\mc{L}_\text{rand}$ is non-negative, so is the minimal value of $\mc{L}_\text{reg}$. 
% In the first induced regularization term $\mc{P}_1$ \cref{eq:P1}, the first term is non-negative and thus minimized the quadratic term $\eval{\nabla_{\ub}\Psi^1}_{\ui{i}}$ during training. 
We opt to analyze the regularization term $\mc{R}_1$ \cref{eq:R1} which provides the same procedure for gaining insights for the other three terms $\mc{P}_1, \mc{P}_2$, and $\mc{R}_2$. The first term $\mc{R}_{1A}$ in $\mc{R}_1\LRp{\ui{i}}$ is non-negative, and thus cannot be large. Let us discuss the second term $\mc{R}_{1B}$ in $\mc{R}_1\LRp{\ui{i}}$ as it could be negative. Since $\LRp{\eval{\nabla_{\ub}F^1}_{\ui{i}} - \eval{\nabla_{\ub}\Psi^1}_{\ui{i}}}$, the second factor in $\mc{R}_{1B}$, appears quadratically in $\mc{R}_{1A}$, it cannot be large at the minimum of $\mc{L}_\text{reg}$. The first factor of $\mc{R}_{1B}$ cannot be large either as it makes $\mc{R}_{1B}$ either too positive, which is not possible due to minimization, or too negative, which is not possible due to the non-negativeness of $\mc{L}_\text{reg}$. Note that the case in which the first factor (second derivative) is of the same magnitude as the second factor (the first derivative), but has the opposite sign, is unlikely due to the Poincar\'e-Friedrichs inequality (see, e.g., \cite{Beckner89,Chen88,Chen87}) which says that the former is (possibly much) greater than the latter. A similar argument applies to $\mc{P}_1\LRp{\ui{i}}, \mc{P}_2\LRp{\ui{i}}$, and $\mc{R}_2\LRp{\ui{i}}$. In summary, at the minimum of  $\mc{L}_\text{reg}$, all the following quantities cannot be large (if not small):
% \[
% \begin{aligned}
%     &  &&  &&\LRp{\eval{F^1}_{\ui{i}} - \eval{\Psi^1}_{\ui{i}}}, &&  \LRp{\eval{F^2}_{\ui{i}} - \eval{\Psi^2}_{\ui{i}}},\\
%     & \eval{\nabla_{\ub}\Psi^1}_{\ui{i}}, && \eval{\nabla_{\ub}\Psi^2}_{\ui{i}}, &&\LRp{\eval{\nabla_{\ub}F^1}_{\ui{i}} - \eval{\nabla_{\ub}\Psi^1}_{\ui{i}}}, && \LRp{\eval{\nabla_{\ub}F^2}_{\ui{i}} - \eval{\nabla_{\ub}\Psi^2}_{\ui{i}}},\\
%     & \eval{\nabla^2_{\ub}\Psi^1}_{\ui{i}}, && \eval{\nabla^2_{\ub}\Psi^2}_{\ui{i}}, &&\LRp{\eval{\nabla^2_{\ub}F^1}_{\ui{i}} - \eval{\nabla^2_{\ub}\Psi^1}_{\ui{i}}}, && \LRp{\eval{\nabla^2_{\ub}F^2}_{\ui{i}} - \eval{\nabla^2_{\ub}\Psi^2}_{\ui{i}}}.
% \end{aligned}
% \]
\[
\begin{aligned}
    & \LRp{\eval{F^1}_{\ui{i}} - \eval{\Psi^1}_{\ui{i}}}, && \eval{\nabla_{\ub}\Psi^1}_{\ui{i}}, && \LRp{\eval{\nabla_{\ub}F^1}_{\ui{i}} - \eval{\nabla_{\ub}\Psi^1}_{\ui{i}}}, && \eval{\nabla^2_{\ub}\Psi^1}_{\ui{i}}, && \LRp{\eval{\nabla^2_{\ub}F^1}_{\ui{i}} - \eval{\nabla^2_{\ub}\Psi^1}_{\ui{i}}},\\
    & \LRp{\eval{F^2}_{\ui{i}} - \eval{\Psi^2}_{\ui{i}}}, && \eval{\nabla_{\ub}\Psi^2}_{\ui{i}}, && \LRp{\eval{\nabla_{\ub}F^2}_{\ui{i}} - \eval{\nabla_{\ub}\Psi^2}_{\ui{i}}}, && \eval{\nabla^2_{\ub}\Psi^2}_{\ui{i}}, && \LRp{\eval{\nabla^2_{\ub}F^2}_{\ui{i}} - \eval{\nabla^2_{\ub}\Psi^2}_{\ui{i}}}.
\end{aligned}
\]
% Another key point is that gradients always exist as we are able to evaluate them with the auto-differentiation functionality of machine learning platforms. 
To gain further insight into the positive effects of the regularization terms, we further expand the first terms in both \cref{eq:R1} and \cref{eq:R2} as
% \tanbui{A quick look does not tell me how you get the following by chain rule. I need to talk to you in person on the following two expressions.} \hai{I made zzz\_derivation.tex for derivation, we have discussed about this, need to check the order}
\begin{equation}
    \label{eq:R1_expand}
    \begin{aligned}
        \eval{\nabla_{\ub}F^1}_{\ui{i}} - \eval{\nabla_{\ub}\Psi^1}_{\ui{i}} 
        & = \eval{\nabla_{\zb}S}_{\zbar{i,1}\LRp{\ub^i}} - \eval{\nabla_{\zb}S}_{\zt^{i,1}\LRp{\ub^i}} \\
        &+  \dt \LRs{\eval{\nabla_{\zb}S}_{\zbar{i,1}\LRp{\ub^i}}\eval{\nabla_{\ub}\FW}_{\ui{i}} - \eval{\nabla_{\zb}S}_{\zt^{i,1}\LRp{\ub^i}} \eval{\nabla_{\ub}\DGGNN}_{\ui{i}}}
    \end{aligned}
\end{equation}
and 
\begin{multline}
    \label{eq:R2_expand}
    \begin{aligned}
        &\eval{\nabla_{\ub}F^2}_{\ui{i}} - \eval{\nabla_{\ub}\Psi^2}_{\ui{i}} 
         = \frac{1}{2}\LRp{\eval{\nabla_{\zb}S}_{\zbar{i,2}\LRp{\ub^i}} - \eval{\nabla_{\zb}S}_{\zt^{i,2}\LRp{\ub^i}}} \\
        & + \frac{1}{2} 
        \LRp{
            \eval{\nabla_{\zb}S}_{\zbar{i,2}\LRp{\ub^i}} \eval{\nabla_{\ub}F^1}_{\ui{i}}
            - \eval{\nabla_{\zb}S}_{\zt^{i,2}\LRp{\ub^i}} \eval{\nabla_{\ub}\Psi^1}_{\ui{i}}
        }\\
        & + \frac{\dt}{2} 
        \LRp{
            \eval{\nabla_{\zb}S}_{\zbar{i,2}\LRp{\ub^i}} \eval{\nabla_{\ub}\FW}_{\ubar{i,1}\LRp{\ub^i}} \eval{\nabla_{\ub}F^1}_{\ui{i}}
            - \eval{\nabla_{\zb}S}_{\zt^{i,2}\LRp{\ub^i}} \eval{\nabla_{\ub}\DGGNN}_{\ut{i,1}\LRp{\ub^i}} \eval{\nabla_{\ub}\Psi^1}_{\ui{i}}
        },
    \end{aligned}
    \end{multline}
respectively.  From \cref{eq:mcDGGNNTagent_loss_rand_reg} and \cref{rem:SGD}, we know that the discrepancies $\ubar{i,1}\LRp{\ui{i}}-\ut{i,1}\LRp{\ui{i}}$ and $\ubar{i,2}\LRp{\ui{i}}-\ut{i,2}\LRp{\ui{i}}$ are small (especially when $\delta^2$ is small) at the optimal solution (supposed that the optimization problem is solved exactly). Since all the operators involved, including $\DGGNN$ with hyperbolic tangent activation functions and the slope limiter, are continuous, it is reasonable\footnote{This is for certain if the slope limiter $S$ is the identity, that is, we do not apply limiter. Otherwise, the limiter \cite{tu2005slope}, though smooth, is not injective over certain regions of its input. However, upon convergence (e.g. with vanishingly small learning rate) of the minibatch SGD, the input for the limiter $S$ changes slightly, and thus locally the inverse of $S$ is differentiable. Consequently, when  discrepancies $\ubar{i,1}-\ut{i,1}$ and $\ubar{i,2}-\ut{i,2}$ are small, $\zbar{i,1}$ and $\zbar{i,2}$ are close to $\zt^{i,1}$ and $\zt^{i,2}$, respectively.} to then conclude that $\zbar{i,1}\LRp{\ui{i}}$ and $\zbar{i,2}\LRp{\ui{i}}$ are close to $\zt^{i,1}\LRp{\ui{i}}$ and $\zt^{i,2}\LRp{\ui{i}}$, respectively. Consequently, the differences $\eval{\nabla_{\zb}S}_{\zbar{i,1}\LRp{\ui{i}}} - \eval{\nabla_{\zb}S}_{\zt^{i,1}\LRp{\ui{i}}}$ and $\eval{\nabla_{\zb}S}_{\zbar{i,2}\LRp{\ui{i}}} - \eval{\nabla_{\zb}S}_{\zt^{i,2}\LRp{\ui{i}}}$ is small. Since the left-hand side of \cref{eq:R1_expand} is smaller (as argued above), it follows that the difference $\eval{\nabla_{\ub}\FW}_{\ui{i}} - \eval{\nabla_{\ub}\DGGNN}_{\ui{i}}$ is small. That is, {\em randomization implicitly encourages not only $\DGGNN$ and $\FW$ to match, but also their Jacobians for the first stage of the Runge-Kutta approach!} A similar argument for \cref{eq:R2_expand} shows that {\em randomization also promotes the matching between the Jacobians of $\DGGNN$ and $\FW$ for the second stage.} Similar arguments for the second terms in both \cref{eq:R1} and \cref{eq:R2} show that the matching for the Hessians at both stages is also promoted with randomization. Following the same procedure, we can show that regularization terms $\mc{P}_1, \mc{P}_2$ enhance the smoothness and curvature of $\DGGNN$ w.r.t. the input.

%\begin{remark}
{\em  In summary, our randomization approach implicitly ensures that the error of the \oDGNet block $\DGGNN$ in approximating the DG spatial discretization $\mc{F}$ is under control at the training data. When trained well (see \cref{sect:general_training} for our empirical settings), $\DGGNN$ can match $\mc{F}$ up to second order derivatives with small errors at least at the training data points. 
%hai{we might need to talk more about smoothness and curvature in this paragraph here if needed.}
}      
%\end{remark}

% In summary, the discrepancies between the \oDGNet block $\DGGNN$ and its DG spatial discretization counterpart $\mc{F}$ at both Runge-Kutta stages, and the discrepancies in their derivatives up to second order)

\begin{remark}
    We can turn the above arguments into rigorous statements\footnote{Such a rigorous argument is straightforward, but cumbersome and we omit the details here to keep the paper at the reasonable length.} by expressing the errors in matching the Jacobian and Hessian of $\DGGNN$ and $\FW$ in terms of the discrepancies $\ubar{i,1}\LRp{\ui{i}}-\ut{i,1}\LRp{\ui{i}}$ and $\ubar{i,2}\LRp{\ui{i}}-\ut{i,2}\LRp{\ui{i}}$ (and Lipschitz constants of the inverse of $S$, $\nabla_{\zb}S$ etc), but the above semi-rigorous arguments are sufficient for explaining and providing insights into why our randomization approach helps obtain accurate and long-time stable \mcDGNet network. 
    \label{rem:randomization}
\end{remark}

It is important to note that an appropriate noise level is crucial and varies depending on the problem. An excessively large noise level, while resulting in strong regularization for derivative matching, can cause training data samples to become random noise samples,
% \tanbui{why? I do not why "become indistinguishable"?} \hai{my observation is from intuition. An extreme scenario would be all training data samples become random noise or dominated by noise, which does not provide any information. From optimal noise level to an extreme case, the randomization benefit decreases.} 
% \hai{it makes me think that mixing noise level for data randomization would be an interesting test! besides Euler-GPT, one scenario training data.} 
thereby losing valuable information. In our work, we manually tune the noise level for each problem. By carefully controlling the noise level, the model-constrained approach \mcDGNet learns a surrogate model that is at least as effective as the \nDGNet approach trained without data randomization.
% \tanbui{Not clear on this sentence}. 
Notably, when the noise level is set to zero, the model-constrained approach \mcDGNet simplifies to the pure data-driven approach \nDGNet. 
% \tanbui{I do not understand the last few sentences either} \hai{there are cases where the optimal noise level is very small, 0.1 \%, from numerical results, \mcDGNet and \nDGNet are almost same accuracy. I explain the reason in the next paragraph. Because we can/have to tune the noise level to find the optimal value, and in the worst/extreme case, the optimal noise level is negligible, i.e., 0\%. With 0\% noise, \mcDGNet becomes \nDGNet{}.}

From \cref{theo:data_rand}, it can be seen that the effectiveness of data randomization depends on the availability of training data and the complexity of the dynamical system's tangent slope. For instance, if the dynamical system is simple, i.e., a linear system,
% \tanbui{define what simple is here} \hai{the simplest system is linear where we can exactly learn the map}
the naive data-driven approach \nDGNet with a sufficient amount of training data can present exactly the tangent slope for the linear dynamical system as proved in \cite{nguyen2022model}.
% \hai{I thought about a linear sytem derivation as mcTangent, but it seems to distract us from the shock-type problems}
On the other hand, data randomization is particularly valuable for more complex dynamical systems in low-data regime scenarios. Indeed, data randomization can be interpreted as a synthetic data engine. Specifically, given a clean data point, we can obtain many random data points within its neighborhood ball with a radius of $\delta$, as visualized in \cref{fig:rand_data_enrichment}. Therefore, when the training data points are spread uniformly over the entire potential data space, data randomization generates new training data points that could fill in the gaps between the clean training data points in the data space. In \cref{sect:numerics}, we quantitatively demonstrate that data randomization significantly enriches the training data when the training data is insufficiently informative (uniform). Conversely, when the training data is ample, the data randomization technique is less effective, and only the regularization effect is active.

\begin{figure}[htb!]
    \centering
    \begin{tikzpicture}
    \begin{axis}[
        xlabel={X-axis label},
        ylabel={Y-axis label},
        width=10cm,
        height=8cm,
        grid=major,
        legend style={at={(.8,.3)},anchor= west, nodes={right}},
        xmin=-3, xmax=3,
        ymin=-3, ymax=3,
        hide axis, 
    ]
        \addplot[only marks, mark=*, mark options={scale=7, fill=blue!20, draw=blue!20}, forget plot] coordinates {
            (0,.7)
            (1,1.5)
            (0,0)
            (0.5,0.5)
            (-0.5,-0.7)
            (-1.5,-1.5)
        };
        \addplot[only marks, mark=*, mark options={scale=3, fill=blue!20, draw=blue!20}] coordinates {
        (0,0)
        };
        \addlegendentry{Data enrichment area}
        \addplot[only marks, color=red, mark=*] table {
            x y
            0 .7
            1 1.5
            0. 0
            .5 .5
            -.5 -.7
            -1.5 -1.5
            1.8 1.8
        };
        \addlegendentry{Training data points}
        \addplot[only marks, color=black, mark=*] table {
            x y
            0.2 .4
            -1 -1
        };
        \addlegendentry{Test data points}
        \addplot [
            domain=0:360,
            samples=100,
            smooth,
            variable=\t,
            fill=gray!20,
            forget plot,
        ] (
            {3.2 * cos(\t) * cos(45) - 1 * sin(\t) * sin(45)},
            {3.2 * cos(\t) * sin(45) + 1 * sin(\t) * cos(45)}
        );
        % Add custom legend for the elliptical area
        % \addlegendentry{}
        \addlegendimage{area legend, fill=gray!20, draw=black}
        \addlegendentry{Data Space}
        \begin{scope}
            \draw[thick, fill=blue!20, color = blue!20] (axis cs:1.8, 1.8) circle (.5cm);
            \draw[->, thick, black] (axis cs:1.8,1.8) -- (1.8,1.3) node[midway, right] {$\delta$};
        \end{scope}
    \end{axis}
\end{tikzpicture}
    \caption{Visualization of data enrichment effect from the data randomization strategy. Data randomization expands a single training data point to the area of the ball with a radius of $\delta$. As a result, test samples are more likely to be discovered during the training process.} 
    \figlab{rand_data_enrichment}
\end{figure}
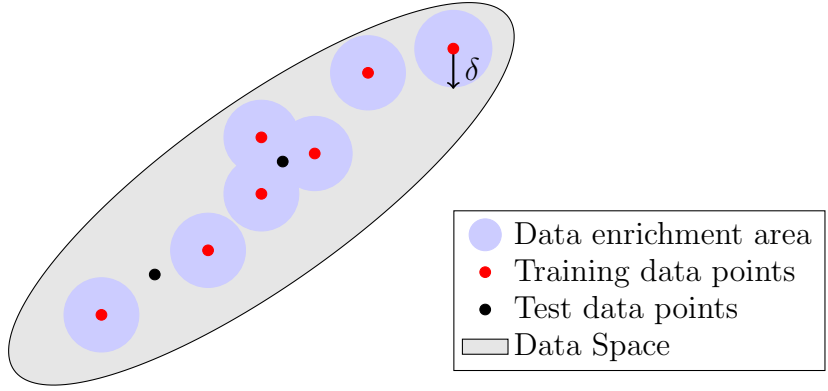

\subsection{Error estimation}
\seclab{Error_estimation}

In this section, we present the error estimation of the \oDGNet predictions for the unseen cases. To begin with, the prediction error is defined as %\tanbui{Hai: this error estimation is not useful as it does not provide a computable way to estimate the error. The better approach is to expand around $\ut{i}$ as you did in the original mcTangent paper.}
\begin{equation}
    \eqnlab{err_v}
    \ev{}\LRp{\ut{i}} = \ui{i+1} - \ut{i+1} = F^2\LRp{\ui{i}} - \Psi^2\LRp{\ut{i}}, \quad  \eg{i+1} = \nor{\ev{}\LRp{\ut{i}}}_{L^2\LRp{\Omega}}.
\end{equation}
By applying the Taylor expansion for $\Psi^2\LRp{\ut{i}} = \Psi^2\LRp{\ui{i} - \ev{}\LRp{\ut{i-1}}}$, we have
\begin{equation}
    \eqnlab{err_taylor}
    \begin{aligned}
        \ev{}\LRp{\ut{i}} & = F^2\LRp{\ui{i}} - \Psi^2\LRp{\ui{i}}  + \eval{\nabla_{\ub} \Psi^2}_{\ui{i}} \ev{}\LRp{\ut{i-1}} + o\LRp{\eg{i}} \\
        & = \LRs{F^2\LRp{\ui{i}} - \Psi^2\LRp{\ui{i}}}  +\LRs{ \eval{\nabla_{\ub} \Psi^2}_{\ui{i}} \ev{}\LRp{\ut{i-1}} - \eval{\nabla_{\ub} F^2}_{\ui{i}} \ev{}\LRp{\ut{i-1}}} \\
        &  + \eval{\nabla_{\ub} F^2}_{\ui{i}} \ev{}\LRp{\ut{i-1}}  + o\LRp{\eg{i}}.
    \end{aligned}
\end{equation}
Applying triangle inequality and then Cauchy-Schwarz inequality for \cref{eq:err_taylor} and using definitions in \cref{eq:err_v} yields
\begin{equation}
    \eqnlab{err_estimation}
    \begin{aligned}
        \eg{i+1} & \leq \nor{F^2\LRp{\ui{i}} - \Psi^2\LRp{\ui{i}}}_{L^2\LRp{\Omega}} 
        \\ & + \nor{\eval{\nabla_{\ub} \Psi^2}_{\ui{i}} - \eval{\nabla_{\ub} F^2}_{\ui{i}}}_{L^2\LRp{\Omega}} \eg{i} + \nor{\eval{\nabla_{\ub} F^2}_{\ui{i}} }_{L^2\LRp{\Omega}} \eg{i} + o\LRp{\eg{i}}
    \end{aligned}
\end{equation}
We can see that the first term in \cref{eq:err_estimation} is similar to the loss term, but evaluated at the unseen solution $\ui{i}$. Thus, if the unseen solution $\ui{i}$ belongs to one of the balls in \cref{fig:rand_data_enrichment} and both $F$ and $\Psi$ are smooth in the ball (which is exactly what randomization is trying to enforce as discussed in \cref{rem:randomization}), then the first term in \cref{eq:err_estimation} is  bounded and/or small after training. A similar conclusion can be drawn for the second term.
%\tanbui{Hai: This approach is not as nice as the original mcTangent approach where you add the purely data driven term in the loss. Note that we do not quite have the control over $\nor{\eval{\nabla_{\ub} F^2}_{\ui{i}} }_{L^2\LRp{\Omega}}$. Please look at how S computed in the yellow book. We need to expand S here. I can talk to you more. The term $\nor{\eval{\nabla_{\ub} F^2}_{\ui{i}} }_{L^2\LRp{\Omega}}$ is a bad term.}
The second term is the induced regularization term in \cref{eq:R2} which also is implicitly minimized during training. The third term, the Jacobian of $\nor{\eval{\nabla_{\ub} F^2}_{\ui{i}} }_{L^2\LRp{\Omega}}$, depends on the stiffness of the system of equations and the underlying discretization. For well-posed problems and well-designed discretization, it is reasonable to assume that $\nor{\eval{\nabla_{\ub} F^2}_{\ui{i}} }_{L^2\LRp{\Omega}}$ is bounded. It can be seen that $\eg{0} = 0$, $\eg{1}$ is bounded, and thus $\eg{i}$ is also bounded for all $\geq 0$ by induction. %This leads us to conclude the prediction error estimation with the following theorem.

\begin{theorem}
    \theolab{mainTheo}
    Assume that the second derivative of $F^2\LRp{\ub}$ with respect to $\ub$ is uniformly bounded.
    Let
    \[
    f^{i+1} := \nor{F^2\LRp{\ui{i}} - \Psi^2\LRp{\ui{i}}}_{L^2\LRp{\Omega}},
    \]
    and
    \[
    g^{i+1} :=  \nor{\eval{\nabla_{\ub} \Psi^2}_{\ui{i}} - \eval{\nabla_{\ub} F^2}_{\ui{i}}}_{L^2\LRp{\Omega}} + \nor{\eval{\nabla_{\ub} F^2}_{\ui{i}} }_{L^2\LRp{\Omega}} + c^i,
    \]
    where $c^i = \mc{O}\LRp{\eg{i}}$. Then,
    the prediction error $\eg{n}$ at time $t_n$ satisfies
    \[
    \eg{n} \le \sum_{j = 1}^{n}\LRp{\Pi_{i = j+1}^{n}g^i} f^j.
    \]
    \end{theorem}
\begin{proof}
The proof is a simple application of a discrete Gronwall's inequality on \cref{eq:err_estimation}.
\end{proof}
% \begin{itemize}
%     \item \textcolor{red}{JAU-UEI, please take a look at this, (1) make more sense/correct argument on remark 2.3. (2) Deriving the convergence rate analysis as the general DG scheme with element order N, then forming a theorem like theorem 2.2. We can form some theorem like theorem 2 and 3 in this \href{https://link.springer.com/article/10.1007/s00366-024-01955-7.}{paper + LINK} }
%     \item \href{https://arxiv.org/pdf/2208.04995}{original mcTangent, remark 2.9}
%     \item reading subsection 2.1 and the beginning of subsection 2.2 and make them more sense.
%     \item Adding DG method for shock-type problem in 1. introduction 
%     \item numerical subsection 3.9
% \end{itemize}
\begin{remark}
    The boundedness of the first and second derivatives of $F^2\LRp{\ub}$ with respect to $\ub$ holds for problem \cref{eq:DG_matrix_form} when the tangent slope is smooth. If the prediction distribution is closed to the training data distribution, the boundedness of $f^i$ and $g^i$ is not necessarily restrictive. As discussed in \cref{sect:Data_rand}, data randomization ensures small values for $h^i$ and $g^i$ at the training points. Furthermore, due to the smoothness of $\Psi^2 \LRp{\ub}$ and $F^2\LRp{\ub}$ and their similarity in both values and derivatives (also a result of randomization), the continuity ensures that $h^i$ and $g^i$ remain small (at least bounded) during inference.
\end{remark}

% \hai{TODO: add a theorem there: in practice, we do not have the true solution, a computable error bound for machine learning prediction can be achieved by performing Taylor expansion around $\ut{i}$ instead}
Nevertheless, expensive numerical solutions mostly are not available in practice, especially when we would like to estimate the error of $\DGGNN$, the computable error estimation on machine learning predictions can be obtained by performing the Taylor expansion for \cref{eq:err_taylor} around $\ut{i}$ instead. The following theorem states the error estimation at each time step in the inference phase.

\begin{theorem}
    \label{thm:mainTheo_computable}
    Assume that the second derivative of $F^2\LRp{\ub}$ with respect to $\ub$ is uniformly bounded.
    Let
    \[
    f^{i+1} := \nor{F^2\LRp{\ut{i}} - \Psi^2\LRp{\ut{i}}}_{L^2\LRp{\Omega}},
    \]
    and
    \[
    g^{i+1} :=  \nor{\eval{\nabla_{\ub} F^2}_{\ut{i}} - \eval{\nabla_{\ub} \Psi^2}_{\ut{i}}}_{L^2\LRp{\Omega}} 
    + \nor{\eval{\nabla_{\ub} \Psi^2}_{\ut{i}}}_{L^2\LRp{\Omega}} + c^i,
    \]
    where $c^i = \mc{O}\LRp{\eg{i}}$. Then,
    the prediction error $\eg{n}$ at time $t_n$ satisfies
    \[
    \eg{n} \le \sum_{j = 1}^{n}\LRp{\Pi_{i = j+1}^{n}g^i} f^j.
    \]
\end{theorem}

%\hai{this theorem is exactly mcTangent bound with a factor of $\nabla_z S$, if we further expand using chain rule.}

\cref{theo:mainTheo} allows us to bound the error between the neural network prediction and the exact solution of the original system, provided that we have an error estimation for the solution of the discretized equation \cref{eq:DG_matrix_form}.  Compared to 
\theoref{mainTheo}, $g^i$ are bounded and can be controlled by randomization as the Jacobian of the neural network $\nor{\eval{\nabla_{\ub} \Psi^2}_{\ut{i}}}_{L^2\LRp{\Omega}}$ is implicitly made small by the randomization as we have discussed in \cref{sect:Data_rand}.

Suppose the error between the discretized solution $\ub^n$ and the exact solution $\ub\LRp{t_n}$ at time $t_n$ is bounded\footnote{Since the 2nd-SSP-RK method is applied, the error contributed from temporal discretization is approximately $O(\dt^2)$. On the other hand, the error from the Discontinuous Galerkin (DG) spatial discretization is approximately $O(h^{N+\half})$, assuming that the exact solution is sufficiently smooth. It is important to note that 
though the error estimate $O(h^{N+\half})$ is only provable for certain linear and nonlinear hyperbolic conservation laws, it is the rate that one typically observes numerically.} by $\mc{O}\LRp{\dt^2 + h^{N+\half}}$
%there is no general result for error estimation of DG discretization for nonlinear problems.}
%The $O(h^{N+\half})$ result is obtained by considering a scalar linear advection problem \cite{hesthaven_nodal_2007}.}, 
where $h$ is the mesh size and $N$ is the order of accuracy of the underlying spatial DG discretization. Then, by applying the triangle inequality, we can bound the error between the neural network prediction and the exact solution.

\[
{\ut{n} - \ub\LRp{t_n}} = \mc{O}\LRp{\dt^2 + h^{N+\half} + \sum_{j = 1}^{n}\LRp{\Pi_{i = j+1}^{n}g^i} f^j},
\]
which shows that in order to achieve optimal accuracy and computational efficiency, we must balance not only the errors from temporal and spatial discretization but also the errors introduced by the neural network. 
Using the universal approximation theorems (see, e.g., \cite{BuiUniversal23}, and the references therein), we can show that there exists a block $\DGGNN$ such that 
\[
\sum_{j = 1}^{n}\LRp{\Pi_{i = j+1}^{n}g^i} f^j = \mc{O}\LRp{\dt^2 + h^{N+\half}},
\]
and thus 
\begin{equation}
    {\ut{n} - \ub\LRp{t_n}} = \mc{O}\LRp{\dt^2 + h^{N+\half}}.
    \label{eq:errorDGNet}
\end{equation}

We note that while the temporal and spatial discretization errors are attainable with sufficiently small $\dt$, $h$, and suffiently smooth solutions, the error incurred from $\DGGNN$ may not be  achievable as it depends on the specific training process and the inherent randomness involved. In \cref{sect:isentropic_vortex} we numerically verify this error estimation for a problem with smooth solutions.
%In the ideal case that training data is sufficiently rich, e.g. the whole data space in \cref{fig:rand_data_enrichment}, the machine learning errors can be mitigated significantly.

% \clearpage

\section{Numerical results}
\seclab{numerics}

In this section, we present numerical results to demonstrate the effectiveness of the proposed \oDGNet approach in approximate solutions of hyperbolic conservation laws. We shall show that the \oDGNet approach can achieve high-order accuracy similar to the underlying DG methods for both smooth and non-smooth solutions. We validate the generalization capabilities of the trained neural networks across various geometrical configurations in different problems including 1D Sod and Lax shock tube problems (\cref{sect:1dsodtube}), 2D Isentropic Vortex (\cref{sect:isentropic_vortex}), 2D Forward Facing Step (\cref{sect:forward_facing_conner}), 2D Scramjet (\cref{sect:sramjet}), 2D Airfoil (\cref{sect:airfoil}), 2D Euler Benchmarks (\cref{sect:2d_euler_benchmarks}), 2D Double Mach Reflection (\cref{sect:2d_euler_double_mach}), and 2D Hypersonic flow over a sphere-cone (\cref{sect:Hypersonic_Flow}). For the isentropic vortex problem with closed form solution in (\cref{sect:isentropic_vortex}), we numerically verify  the theoretical convergence rate of the \oDGNet approach in \cref{eq:errorDGNet} . Training data enrichment with data randomization will be discussed in \cref{sect:P6_2D_Noise_corruption}. A quantitative analysis of the generalizability of the \oDGNet approach across different problems is provided in \cref{sect:sramjet_generalization}. In \cref{sect:forward_facing_conner_Roe_flux}, we explore the flexibility of the \oDGNet approach when trained with Harten-Lax-van Leer numerical flux data, and emphasize the robustness of \oDGNet approach against various random neural network initializers in \cref{sect:Random_initializers_and_noise_seeds}. Finally, the computational training time and the speed improvements over the traditional Discontinuous Galerkin (DG) methods are summarized \cref{sect:Train_validation_test_computation_time}. It is important to note that while our investigated numerical problems are 1D and 2D Euler equations, the \oDGNet approach can be applied to other types of equations and 3D problems as well. 

\subsection{General settings and learning hyperparameters}
\label{sect:general_training}
To keep the following sections manageable, we now discuss common features shared across all problems. These include data generation, training settings, neural network selection strategy, noise levels, and the slope limiter operator. A comprehensive summary of training information for all numerical examples is provided in \cref{tab:summary_table}.

\begin{table}[htb!]
	\centering
	\caption{Summary of training settings for all problems}
	\tablab{summary_table}
	\resizebox{\textwidth}{!}{
    \begin{tblr}{
  column{3} = {c},
  column{4} = {c},
  column{5} = {c},
  column{6} = {c},
  column{7} = {c},
  column{8} = {c},
  column{10} = {c},
  cell{1}{1} = {c=2}{},
  cell{1}{9} = {c},
  cell{2}{1} = {r=2}{c},
  cell{2}{9} = {c},
  cell{3}{4} = {c=7}{},
  cell{4}{1} = {r=2}{c},
  cell{4}{9} = {c},
  cell{5}{4} = {c=7}{},
  cell{6}{1} = {r=2}{c},
  cell{6}{9} = {c},
  cell{7}{4} = {c=7}{},
  cell{8}{1} = {c=2}{},
  cell{8}{9} = {c},
  cell{9}{1} = {c=2}{},
  cell{9}{9} = {c},
  cell{10}{1} = {c=2}{},
  cell{10}{3} = {c=8}{},
  cell{11}{1} = {c=2}{},
  cell{11}{3} = {c=8}{},
  cell{12}{1} = {c=2}{},
  cell{12}{9} = {c},
  cell{13}{1} = {c=2}{},
  cell{13}{3} = {c=8}{},
  cell{14}{1} = {c=2}{},
  cell{14}{5} = {c=6}{},
  cell{15}{1} = {c=2}{},
  cell{15}{3} = {c=8}{},
  cell{16}{1} = {c=2}{},
  cell{16}{3} = {c=8}{},
  cell{17}{1} = {c=2}{},
  cell{17}{3} = {c=6,r=2}{},
  cell{17}{9} = {c=2,r=2}{},
  cell{18}{1} = {c=2}{},
  cell{19}{1} = {c=2}{},
  cell{19}{3} = {c=8}{},
  cell{20}{1} = {c=2}{},
  cell{20}{5} = {c=6}{},
  cell{21}{1} = {c=2}{},
  cell{21}{5} = {c=6}{},
  cell{22}{1} = {c=2}{},
  cell{22}{3} = {c=2}{},
  cell{22}{5} = {c=6}{},
  vlines,
  hline{1-2,4,6,8-17,19-23} = {-}{},
  hline{3,5,7} = {2-10}{},
  hline{18} = {1-2}{},
}
                               &                  & {1D Sod\\ -Lax}                          & {Isentro-\\ tropic}        & {Forward\\Facing}             & Scramjet & Airfoil & {Sphere-\\Cone} & {Euler-\\ config6}                                                                              & {Double\\ Mach} \\
Train                          & $T_\text{train}$ & 0.15                                     & 0.3                        & 1                             & 1.6      & 1.2     & 0.0003          & 0.16                                                                                            & 0.02            \\
                               & using            & $\rho, u, p$                             & $\gamma = \LRc{1.2, 1.6} $ &                               &          &         &                 &                                                                                                 &                 \\
{Vali-\\ dation}               & $T_\text{train}$ & 0.15                                     & 0.3                        & 1                             & 1.6      & 1.2     & 0.0003          & 0.16                                                                                            & 0.02            \\
                               & using            & $\rho, u, p$                                      & $\gamma = 1.4$             &                               &          &         &                 &                                                                                                 &                 \\
Test                           & $T_\text{test}$  & 0.25                                     & 1                          & 4                             & 6        & 7.5     & 0.0015          & 0.8                                                                                             & 0.25            \\
                               & using            & $\rho, u, p$                                      & $\gamma = 1.4$             &                               &          &         &                 &                                                                                                 &                 \\
$\dt$                          &                  & 0.0001                                   & 0.002                      & 0.001                         & 0.002    & 0.0015  & 5e-7            & 0.0004                                                                                          & 0.0001          \\
Noise level, $\delta$          &                  & 0.5\%                                    & 0.2\%                      & 2\%                           & 4\%      & 4\%     & 11\%            & 2\%                                                                                             & 1\%             \\
Optimizer                      &                  & ADAM (default settings)                  &                            &                               &          &         &                 &                                                                                                 &                 \\
Learning rate                  &                  & $10^-3$                                  &                            &                               &          &         &                 &                                                                                                 &                 \\
Batch size, s                  &                  & 15                                       & 30                         & 20                            & 16       & 16      & 2               & 2                                                                                               & 2               \\
$\Psi_\text{flux}$             &                  & 1 hidden layer of 128 neurons            &                            &                               &          &         &                 &                                                                                                 &                 \\
$\Psi_\text{vol}$              &                  & None                                     & None                       & 1 hidden layer of 128 neurons &          &         &                 &                                                                                                 &                 \\
Initializers                   &                  & $\mc{N}\LRp{0, 0.01}$ with random seed 0 &                            &                               &          &         &                 &                                                                                                 &                 \\
Activation                     &                  & Tanh                                     &                            &                               &          &         &                 &                                                                                                 &                 \\
Train loss ($L^2$)             &                  & $\LRp{\rho, \rho u, E}$                  &                            &                               &          &         &                 & $\LRp{\rho, \rho u, \rho v, E}$ &                 \\
{Validation\\(relative $L^2$)} &                  &                                          &                            &                               &          &         &                 &                                                                                                 &                 \\
$\rho$ constraint              &                  & $\LRs{0.1, 50}$                          &                            &                               &          &         &                 &                                                                                                 &                 \\
Slope limiter                  &                  & Yes                                      & None                       & Yes                           &          &         &                 &                                                                                                 &                 \\
Element order                  &                  & 1,6                                      & 1,2,3,4                    & 1                             &          &         &                 &                                                                                                 &                 \\
Precision                      &                  & Double                                   &                            & Single                        &          &         &                 &                                                                                                 &                 
\end{tblr}
}
\end{table}

\subsubsection{Data generation}

Unless otherwise stated, the data sets are generated by solving the compressible Euler equations with the Lax-Friedrichs flux and over-integration (to avoid aliasing and improve accuracy) using second-order strong stability Runge-Kutta scheme.
% \hai{I added this to state explicitly that we use 2nd-SSP-RK}. 
Both the training and validation datasets consist of snapshots captured within the same time interval $\left[0, T_{\text{train}}\right]$. However, the training and validation data are derived from different initial conditions, gas densities, or viscosities. Meanwhile, the test data is gathered within the time interval $\left[0, T_{\text{test}}\right]$, where $T_{\text{test}} > T_{\text{train}}$. 
To demonstrate the robust generalization capabilities of the trained \oDGNet, we study it not only on test datasets from the same problem  (same initial conditions, initial conditions, geometries, mesh size, etc) but also on test datasets with different geometry, and mesh size, initial conditions, etc. Additional problem-specific information is given at the beginning of the corresponding section.

% For this study, we opt to generate test data under the same initial conditions, gas density, and viscosity as the validation data. This approach is designed to emulate practical scenarios where current data is available and there is a need to predict future outcomes within the same framework. For simplicity, we generate training, validation, and test data sets with the same time step $\Delta t$, this is not an obligatory practice since our \oDGNet approach is time-invariant by design. Furthermore, it is important to note that training and validation data sets are generated using the same mesh discretization, referred to as Model 1. To demonstrate the robust generalization capabilities of the trained neural networks, we not only test on Model 1, but also on test datasets with different geometry and mesh discretization, referred to as Model 2 (and Model 3 for the 2D Euler Benchmarks in \cref{sect:2d_euler_benchmarks}). Additional problem-specific information is given at the beginning of the corresponding section.

\subsubsection{Training settings}
For training \oDGNet\hspace{-1ex}, we utilize the \texttt{ADAM} optimizer \cite{kingma2014adam} in \texttt{JAX} \cite{jax2018github} with default parameters and a learning rate of $10^{-3}$. Due to GPU memory limitations, we input a random window of consecutive $s$ solution snapshots (batch size) from the training data per epoch. Both the flux neural network and the volume integral correction neural network are composed of a one-hidden layer with 128 neurons with Tanh activation function. Note that we verified that networks with 2, 3, or 4 layers with the same neurons and same training process showed inferior performance compared to the single-layer ones. The possible reason is that deeper neural networks are more complex and thus more challenging to train. Consequently, we consider only one-hidden layer networks. The neural network weights and biases are initialized from random normal distribution $\mc{N}\LRp{0,0.01}$. Throughout the training process for all problems, we enforce a constraint on the prime variable $\rho$, ensuring $\rho \in \LRs{0.1,50}$ to maintain physically reasonable quantity in the early training epochs. We employ double precision for training in the 1D Sod and Lax shock tube problems (\cref{sect:1dsodtube}) and 2D Isentropic Vortex (\cref{sect:isentropic_vortex}) while opting for single precision for the other problems for faster computation and lower memory demand.

\subsubsection{Simplified training for model-constrained approach, \mcDGNet\hspace{-1ex}}
% \hai{adding this part to clarify about what we do in code}
Based on empirical findings from our previous research \cite{nguyen2022model}, which demonstrated that high values of the balance parameter $\alpha$ ($\geq 10^5$) were optimal for weighting model-constrained and machine learning loss terms, we conducted additional validation studies on the Forward Facing Step problem. We found that training performance is comparable whether using large $\alpha$ values or ignoring the machine learning loss term. Given these findings, in order to optimize computational efficiency and memory utilization for larger-scale problems, we opt to train \mcDGNet using only the model-constrained loss term across all problems.

\subsubsection{Selection of the ``best" trained \oDGNet\hspace{-1ex}}
We consider the relative $L^2$-error 
\begin{equation}
	\eqnlab{relative_L2_error}
	\sum_{i=1}^{n_t} \frac{\nor{\ut{i} - \ui{i}}_{L^2\LRp{\Omega}}}{\nor{\ui{i}}_{L^2\LRp{\Omega}}},
\end{equation}
where $n_t = \frac{T_\text{train}}{\dt}$ represents the number of snapshots within the training time interval $\LRs{0, T_\text{train}}$, and $\ut{i}$ and $\ui{i}$ denote the \oDGNet and  Discontinuous Galerkin solutions, respectively. To select the ``optimal" neural network, we first calculate the average relative $L^2$-error \cref{eq:relative_L2_error} for all conservative variables  $\ub$ in the validation datasets. 
%within the time interval $\LRs{0, T_\text{train}}$ between the \oDGNet solutions and Discontinuous Galerkin solutions (or the analytical solutions in the isentropic vortex problem). 
For 2D Forward Facing Step (\cref{sect:forward_facing_conner}), 2D Scramjet problem (\cref{sect:sramjet}), 2D airfoil problem (\cref{sect:airfoil}), and 2D Hypersonic flow through sphere-cone (\cref{sect:Hypersonic_Flow}), we do not compute the error for the $\rho v$ component since the magnitude of $v$ is typically small during initial time steps. Then, the neural network that is selected is the one that yields the lowest average relative $L^2$-error on the validation data set.

\subsubsection{Noise level and slope limiters}
The noise level (standard deviation) $\delta$ in \cref{eq:add_noise} is not the same for all problems and it is chosen heuristically. For each problem, we train with several noise levels, starting from $1\%$ and increasing in $1\%$  until the validation loss degrades. Random noise realizations are then independently drawn from the standard normal distribution $\boldsymbol{\eta} = \mathcal{N}(0, \delta^2 \mathbf{I})$ and are scaled 
% \tanbui{we do not discuss in \cref{eq:add_noise}? How do you actually compute $\delta$ in \cref{eq:add_noise}?} \hai{No, we do not discuss. the noise realization $\epsilon$ is computed as $\epsilon = \mu \circ \ub$, scaling is necessary to make noise equivalently meaningful at all nodal values. In all papers (inverse/forward), we used this way.} \tanbui{My point was that we need to be consistent, we cannot present something in \cref{eq:add_noise} and then talk about something completely different in our numerical result. Pick one, and stick with it throughout.} \hai{updated in theory part for consistency} 
according to the magnitude of the corresponding component of solutions in a nodal-wise manner, i.e., $\boldsymbol{\epsilon} = \boldsymbol{\eta} \odot \ub$. In other words, the noise realization is sampled as $\epsb \sim \LRp{0, \delta^2 \Sigma_{\ub}}$  where $\Sigma_{\ub} = \diag \LRp{\ub^2}$. During training, for each epoch, a new randomized noise realization is generated and added to the noise-free solutions. 
%This minipatch approach is  memory-efficient compared to extracting random minipatch from the set of pre-randomized training samples.

The slope limiter in  \cite{tu2005slope} is employed to stabilize computations for problem with shocks such as the 1D Sod and Lax shock tube problems (\cref{sect:1dsodtube}), 2D Forward Facing Step (\cref{sect:forward_facing_conner}), 2D Scramjet problem (\cref{sect:sramjet}), 2D airfoil problem (\cref{sect:airfoil}), 2D Euler Benchmarks (\cref{sect:2d_euler_benchmarks}), 2D Double Mach Reflection (\cref{sect:2d_euler_double_mach}), and 2D Hypersonic flow through sphere-cone (\cref{sect:Hypersonic_Flow}). Since the slope limiter operator limits the convergence rate of numerical DG simulation, we use solution order $N = 1$ with different mesh discretization for all these problems. An exception is made for the 1D Sod and Lax shock tube problems \cref{sect:1dsodtube}, where higher-order elements are still used. However, predictions from these higher-order elements are not superior to those from lower-order elements. In 2D isentropic vortex \cref{sect:isentropic_vortex} problem, the slope limiter is not applied since the problem does not contain shock waves.

\subsubsection{Implicit Backward Euler scheme}
%\tanbui{We need to provide a reason why we need implicit solver here.} \hai{I added the following sentence}As stated in \cref{sect:trainingDGNetBlock}, we choose to train \oDGNet using explicit time integration to be cost effective. Once trained, \oDGNet can be used with any explicit or implicit time integrator. In particular,
In \cref{sect:1dsodtube},\cref{sect:isentropic_vortex}, and \cref{sect:forward_facing_conner}, we verify that the ee \oDGNet trained with 2nd-SSP-RK scheme can be used together with the implicit Backward Euler scheme for solving Euler equations. The Newton-Raphson method is applied to address the nonlinear system of equations described by
\begin {equation}
    \eqnlab{newton_raphson_implicit_time_integration}
    \ui{i,n+1} = \ui{i,n} - \LRp{\eval{\nabla_{\ub} \mb{F}}_{\ui{i,n}}}^{-1} \eval{\mb{F}}_{\ui{i,n}},
\end{equation}
where the function $\mb{F}$ is defined as
\begin{equation*}
    \mb{F}\LRp{\ub} = \ub - \LRp{\ui{i,0} + \dt \DGGNN\LRp{\ub}} \quad \text{or} \quad \mb{F}\LRp{\ub} = \ub - \LRp{\ui{i,0} + \dt \, \mc{F}\LRp{\ub}},
\end{equation*}
for the \oDGNet and DG methods, respectively. The initial guess $\ui{i,0}$ is assigned to the current converged solution.
%$\mc{F}$ denotes the right-hand side dynamic evolution operator of the traditional DG method. 
It is important to note that computing the Jacobian matrix explicitly in \cref{eq:newton_raphson_implicit_time_integration} is computationally and memory-intensive, and thus impractical. To efficiently handle this, we utilize the capabilities of \texttt{JAX} \cite{jax2018github} to compute the Jacobian-vector product operator. For the solution of the last term on the right-hand side \cref{eq:newton_raphson_implicit_time_integration}, we employ the GMRES algorithm, facilitated by \texttt{JAXopt} \cite{jaxopt_implicit_diff}, to solve large linear system equations.

% \clearpage
\subsection{1D Sod and Lax shock tube problems}
\seclab{1dsodtube}

In this section, we address the 1D Euler equation given as
% \begin{equation} 
%     \begin{aligned}
%         \pp{\rho}{t} + \pp{{\rho u}}{x} &= 0, \\
%         \pp{{\rho u}}{t} + \pp{{\rho u^2 + p}}{x} &= 0, \\
%         \pp{{E}}{t} + \pp{{u\LRp{E + p}}}{x} &= 0,
%     \end{aligned}
% \end{equation}
\begin{equation} 
    \frac{\partial}{\partial t} \underbrace{\begin{pmatrix}
        \rho \\
        \rho u \\
        E
        \end{pmatrix}}_{\ub} + \frac{\partial}{\partial x_1} \underbrace{\begin{pmatrix}
        \rho u \\
        \rho u^2 + p \\
        u(E + p)
        \end{pmatrix}}_{\fb_{1}} = 0,
    \eqnlab{1D_Euler_equation}
\end{equation}
where
\begin{equation*}
	E = \frac{p}{\gamma - 1} + \frac{\rho u^2}{2}, \quad \gamma = 1.4 .
\end{equation*}
The initial condition is presented in the following form:
\begin{equation*}
    \LRp{\rho, u, p} = \begin{cases}
        \LRp{\rho_\text{L}, u_L, p_\text{L}}, & \text{if } 0 \leq x_1 < 0.5 \\
        \LRp{\rho_\text{R}, u_R, p_\text{R}}, & \text{if } 0.5 \leq x_1 \leq 1
    \end{cases}.
    \eqnlab{1D_form}
\end{equation*}

\begin{figure}[htb!]
    \centering
        \begin{tabular*}{\textwidth}{c c}
            \centering
            Case I: Sod shock tube in Model 1 & Case II: Sod shock tube in Model 2
            \\
            \raisebox{-0.5\height}{\includegraphics[width = .48\textwidth]{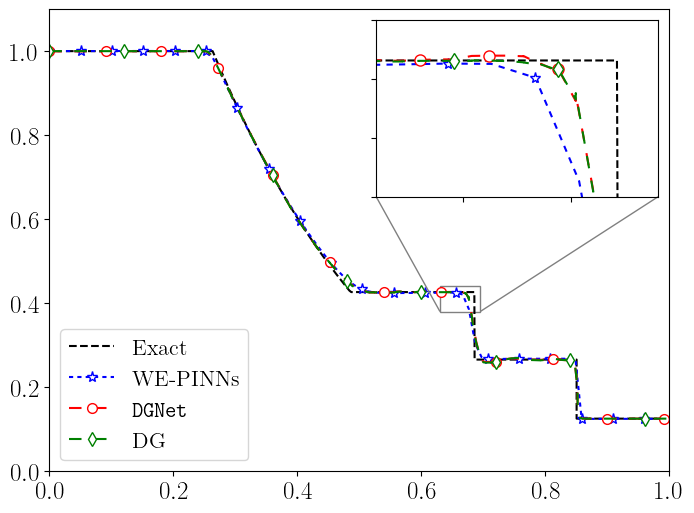}} &
            \raisebox{-0.5\height}{\includegraphics[width = .48\textwidth]{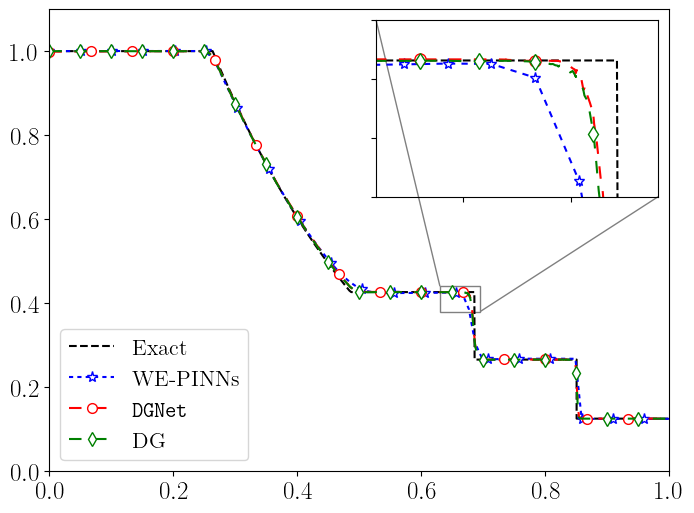}} 
            \vspace{-0.5ex}
            \\ 
            {\small \quad $x_1$} & {\small \quad $x_1$}
            \\~\\
            Case III: Lax shock tube in Model 1 & Case IV: Lax shock tube in Model 2
            \\
            \raisebox{-0.5\height}{\includegraphics[width = .48\textwidth]{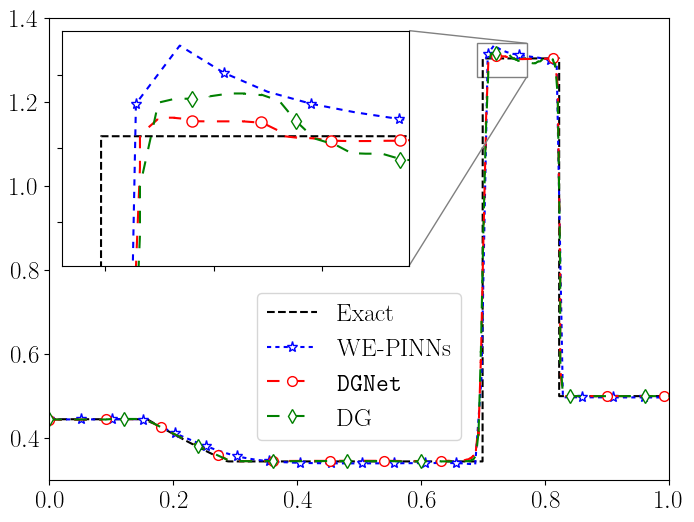}} &
            \raisebox{-0.5\height}{\includegraphics[width = .48\textwidth]{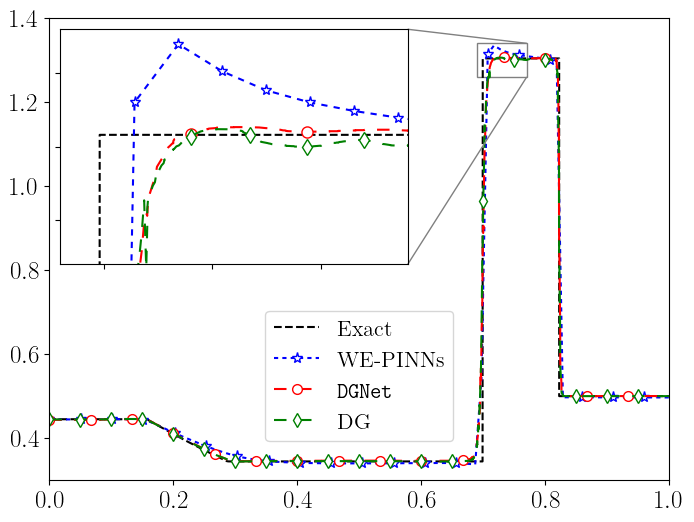}} 
            \vspace{-0.5ex}
            \\ 
            {\small \quad $x_1$} & {\small \quad $x_1$}
        \end{tabular*}
        \caption{
        % \textcolor{red}{Cole: The axes of the plots are not labeled.}
        {\bf 1D Sod and Lax shock tube problems:} predicted density solutions, $\rho$, for four different cases using the \oDGNet network trained with Model 1 Sod shock tube training data. Exact stands for the exact solution, DG for the DG solution, and WE-PINNs for the result from \cite{liu2024discontinuity}.
        {\bf Top Left: } Sod shock tube in  Model 1 at $T = 0.2s$. {\bf Top Right: } Sod shock tube in Model 2 at $T = 0.2s$. {\bf Bottom Left: } Lax shock tube in  Model 1 at $T = 0.13s$. {\bf Bottom Right: } Lax shock tube in Model 2 at $T = 0.13s$.
        }
        \label{fig:1D_sodtube_250_N_1}
\end{figure}

The initial conditions for the Sod shock tube problem \cite{sod1978survey} are defined as $\LRs{\rho_\text{L}, u_L, p_\text{L}} = \LRp{1,0,1}$ for the left side and $\LRs{\rho_\text{L}, u_L, p_\text{L}} = \LRp{0.1,0,0.125}$ for the right side. On the other hand, for the Lax shock tube problem \cite{laney1998computational}, the initial conditions are $\LRs{\rho_\text{L}, u_L, p_\text{L}} = \LRp{0.445,0.698,3.528}$ on the left and $\LRs{\rho_\text{L}, u_L, p_\text{L}} = \LRp{0.5, 0, 0.571}$ on the right. The training data is produced by solving \cref{eq:1D_Euler_equation} using eight different initial conditions, expressed as:
\[\rho_\text{L} \times \rho_\text{R} \times p_\text{L} \times p_\text{R} \times u_L \times u_R = \LRc{0.7, 1.3} \times \LRc{0.05, 0.2} \times  \LRc{0.8, 1.2} \times \LRc{0.1} \times \LRc{0} \times \LRc{0}\]
over the time interval $\LRs{0, 0.15}s$ with $\dt = 10^{-4}s$. Note that Sod and Lax problem settings are not in the training data sets. The computational domain is discretized into two configurations: Model 1 with $K = 250$ elements of order $N=1$,
%\tanbui{we have problem with notation here:  we can not use the same notion of the number of elements and the solution order. Please fix them everywhere!} \hai{I fixed, K should be the number of elements}, 
and Model 2 with $K = 500$ elements of order $N=6$. For validation data, we solve the Sod shock tube problem within the same time interval $\LRs{0, 0.15}s$ with $\dt = 10^{-4}s$ with $K = 250$ elements of order $N=1$. We generate four test datasets by solving \cref{eq:1D_Euler_equation} for four unseen cases, detailed as follows: Case I: Sod shock tube using Model 1 with $T_\text{test} = 0.2s$; Case II: Sod shock tube using Model 2 with $T_\text{test} = 0.2s$; Case III: Lax shock tube using Model 1 with $T_\text{test} = 0.13s$; Case IV: Lax shock tube using Model 2 with $T_\text{test} = 0.13s$.

In this 1D problem, we also emphasize that both the \nDGNet approach and \mcDGNet approach with small noise (less than $0.5\%$) exhibit almost identical performance. The possible reason for this equivalence is that the noise-free normalized training data is sufficiently informative for training, and thus no more extra information is induced by the data randomization. A quantitative analysis on how much enriching training data can be obtained with data randomization %in the normalized data space 
is further elaborated in \cref{sect:P6_2D_Noise_corruption}. Therefore, we present the \mcDGNet approach result as the representative result, denoted as \DGNet. \cref{fig:1D_sodtube_250_N_1} shows the predicted density solutions $\rho$ for four different cases using the trained network from Model 1 Sod shock tube training data. It can be seen that the \oDGNet approach is capable of generalizing for unseen initial conditions and beyond the training time horizon. Indeed, \oDGNet is in excellent agreement with traditional Discontinuous Galerkin (DG) solutions and this is not surprising as we train \oDGNet to learn the underlying DG discretization. Furthermore, the trained network not only successfully solves the same Sod shock tube problem on a finer mesh with higher-order elements in Model 2, but also effectively addresses the Lax shock tube problem in both Model 1 and Model 2 configurations. This out-of-distribution generalizability originates from various strategies that we deployed in the design and training of \oDGNet including randomization and normalization process.
%, which enables the \oDGNet approach to generalize to completely different initial conditions.
We also compare our approach with the WE-PINN approach in \cite{liu2024discontinuity}. It is worth noting that, in the WE-PINN method, solving for each initial condition necessitates launching a separate training process, thus four distinct trainings need to be implemented for four cases. Our \oDGNet approach not only demonstrates superior accuracy, particularly near shocks, but it also offers generalizability by requiring a single trained network to solve all four problems.

\noindent{Implicit \oDGNet\hspace{-1ex}}.

% \tanbui{Did you use an explicit integrator for the above results? if yes, state it.}
% \hai{We used 2nd-SSP-RK scheme. I added information in the data generation general setting section}
Once trained, the \oDGNet neural network (trained with Sod shock tube  Model 1 training data) can be seamlessly integrated within the Backward Euler scheme to solve both Sod and Lax shock tube problems. \cref{fig:1D_sodtube_implicit} illustrates the predicted outcomes from the \oDGNet alongside traditional DG solutions using the Backward Euler scheme on Model 1 at time $T = 0.25s$ for the Sod shock tube and $T = 0.13s$ for the Lax shock tube. We emphasize that the 2nd-SSP-RK is unstable with the corresponding chosen time step $\dt$ in both problems. \cref{fig:1D_sodtube_implicit} shows that the implicit \oDGNet solutions are stable and visibly indistinguishable from the implicit DG solutions. This finding further underscores the robustness and generalization capability of the \oDGNet approach in   effectively handling  diverse initial conditions in the implicit scheme.

\begin{figure}[htb!]
    \centering
        \begin{tabular*}{\textwidth}{c c}
            \centering
            Sod shock tube & Lax shock tube
            \\
            \raisebox{-0.5\height}{\includegraphics[width = .48\textwidth]{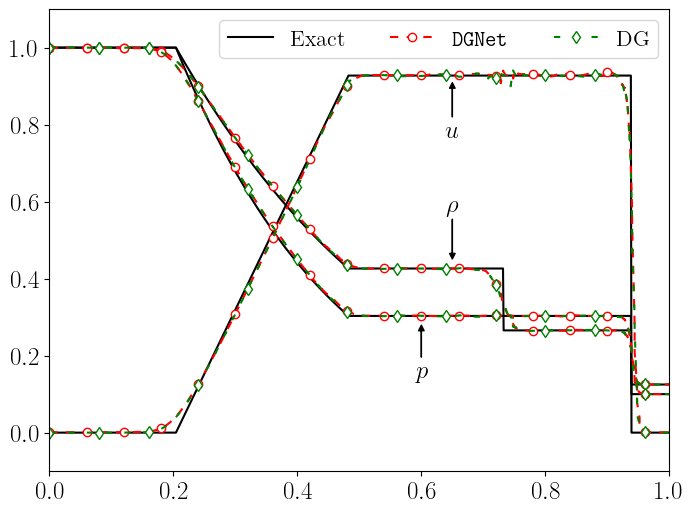}} &
            \raisebox{-0.5\height}{\includegraphics[width = .48\textwidth]{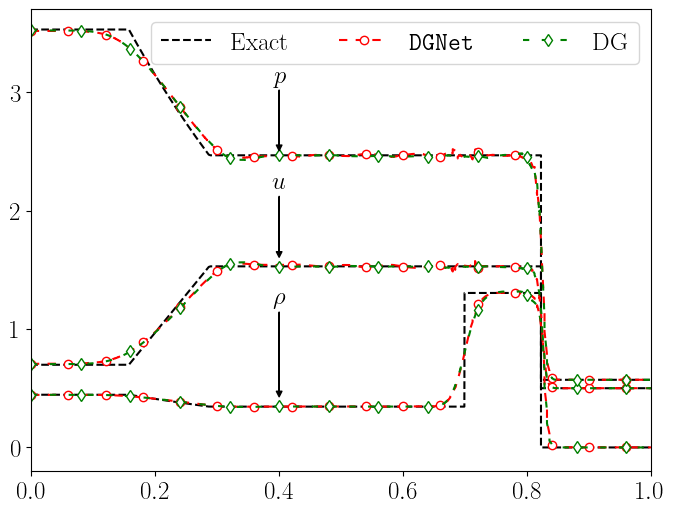}} 
            \vspace{-0.5ex}
            \\
            {\small \quad $x_1$} & {\small \quad $x_1$}
        \end{tabular*}
        \caption{{\bf 1D Sod and Lax shock tube problems:} comparison of the exact solution, implicit DG and implicit \oDGNet solution on a discretization with $K = 250$ and solution order $ N = 1$ (Model 1). {\bf Left:} Sod shock tube predictions at $T = 0.25s, \dt = 0.002s$. {\bf Right:} Lax shock tube predictions at $T = 0.13s, \dt = 0.001s$. Note that, the 2nd-SSP-RK4 scheme is unstable with either of the time stepsizes. As can be seen, DG and \oDGNet yield essentially identical results.}
        % {\bf: confirmed the implicit DG with small $\dt$ performs is identical as explicit DG}
        \figlab{1D_sodtube_implicit}
\end{figure}

% \clearpage

\subsection{Isentropic vortex problem}
\seclab{isentropic_vortex}

\begin{figure}[htb!]
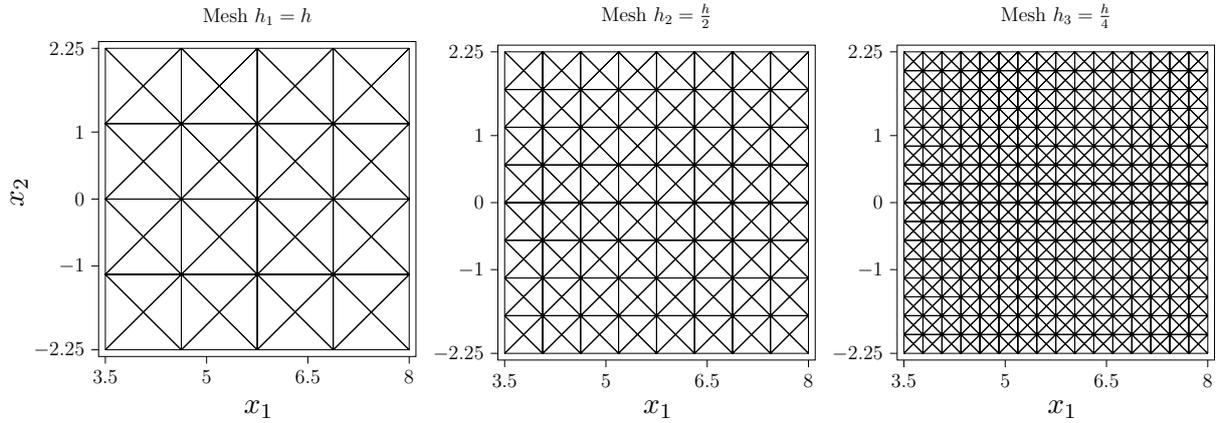

    \centering
        \begin{tabular*}{\textwidth}{c@{\hskip -0.01cm} c@{\hskip -0.01cm} c@{\hskip -0.01cm} c@{\hskip -0.01cm}}
            \centering
            \rotatebox[origin=c]{90}{\small $x_2$} &
            \raisebox{-0.5\height}{\resizebox{0.32\textwidth}{!}{% This file was created with tikzplotlib v0.10.1.
\begin{tikzpicture}

\definecolor{darkgray176}{RGB}{176,176,176}

\begin{axis}[
title={Mesh $h_1 = h$},
tick align=outside,
tick pos=left,
unbounded coords=jump,
x grid style={darkgray176},
xmin=3.4, xmax=8.1,
xtick style={color=black},
xtick = {3.5,5.,6.5,8},
y grid style={darkgray176},
ymin=-2.35, ymax=2.35,
ytick style={color=black},
ytick = {-2.25,-1.,0,1.,2.25},
height = .5\textwidth,
width = .5\textwidth,
]
\addplot [line width=0.08pt, black]
table {%
4.625 -1.125
5.75 -1.125
nan nan
5.1875 -1.6875
5.75 -1.125
nan nan
5.1875 -1.6875
4.625 -1.125
nan nan
6.875 1.125
5.75 1.125
nan nan
6.3125 1.6875
5.75 1.125
nan nan
6.3125 1.6875
6.875 1.125
nan nan
4.625 1.125
4.625 0
nan nan
4.0625 0.5625
4.625 0
nan nan
4.0625 0.5625
4.625 1.125
nan nan
6.875 -1.125
6.875 0
nan nan
7.4375 -0.5625
6.875 0
nan nan
7.4375 -0.5625
6.875 -1.125
nan nan
6.875 -1.125
6.875 -2.25
nan nan
6.3125 -1.6875
6.875 -2.25
nan nan
6.3125 -1.6875
6.875 -1.125
nan nan
4.625 1.125
4.625 2.25
nan nan
5.1875 1.6875
4.625 2.25
nan nan
5.1875 1.6875
4.625 1.125
nan nan
4.625 -1.125
3.5 -1.125
nan nan
4.0625 -0.5625
3.5 -1.125
nan nan
4.0625 -0.5625
4.625 -1.125
nan nan
6.875 1.125
8 1.125
nan nan
7.4375 0.5625
8 1.125
nan nan
7.4375 0.5625
6.875 1.125
nan nan
4.625 -1.125
4.625 -2.25
nan nan
4.0625 -1.6875
4.625 -2.25
nan nan
4.0625 -1.6875
4.625 -1.125
nan nan
6.875 1.125
6.875 2.25
nan nan
7.4375 1.6875
6.875 2.25
nan nan
7.4375 1.6875
6.875 1.125
nan nan
4.625 1.125
3.5 1.125
nan nan
4.0625 1.6875
3.5 1.125
nan nan
4.0625 1.6875
4.625 1.125
nan nan
6.875 -1.125
8 -1.125
nan nan
7.4375 -1.6875
8 -1.125
nan nan
7.4375 -1.6875
6.875 -1.125
nan nan
6.875 -1.125
5.75 -1.125
nan nan
6.3125 -0.5625
5.75 -1.125
nan nan
6.3125 -0.5625
6.875 -1.125
nan nan
4.625 1.125
5.75 1.125
nan nan
5.1875 0.5625
5.75 1.125
nan nan
5.1875 0.5625
4.625 1.125
nan nan
4.625 -1.125
4.625 0
nan nan
5.1875 -0.5625
4.625 0
nan nan
5.1875 -0.5625
4.625 -1.125
nan nan
6.875 1.125
6.875 0
nan nan
6.3125 0.5625
6.875 0
nan nan
6.3125 0.5625
6.875 1.125
nan nan
5.75 6.24500451351651e-17
5.75 -1.125
nan nan
5.1875 -0.5625
5.75 -1.125
nan nan
5.1875 -0.5625
5.75 6.24500451351651e-17
nan nan
5.75 -6.24500451351651e-17
5.75 1.125
nan nan
6.3125 0.5625
5.75 1.125
nan nan
6.3125 0.5625
5.75 -6.24500451351651e-17
nan nan
5.75 0
4.625 0
nan nan
5.1875 0.5625
4.625 0
nan nan
5.1875 0.5625
5.75 0
nan nan
5.75 0
6.875 0
nan nan
6.3125 -0.5625
6.875 0
nan nan
6.3125 -0.5625
5.75 0
nan nan
8 -2.25
6.875 -2.25
nan nan
7.4375 -1.6875
6.875 -2.25
nan nan
7.4375 -1.6875
8 -2.25
nan nan
3.5 2.25
4.625 2.25
nan nan
4.0625 1.6875
4.625 2.25
nan nan
4.0625 1.6875
3.5 2.25
nan nan
3.5 -2.25
3.5 -1.125
nan nan
4.0625 -1.6875
3.5 -1.125
nan nan
4.0625 -1.6875
3.5 -2.25
nan nan
8 2.25
8 1.125
nan nan
7.4375 1.6875
8 1.125
nan nan
7.4375 1.6875
8 2.25
nan nan
5.75 -2.25
4.625 -2.25
nan nan
5.1875 -1.6875
4.625 -2.25
nan nan
5.1875 -1.6875
5.75 -2.25
nan nan
5.75 2.25
6.875 2.25
nan nan
6.3125 1.6875
6.875 2.25
nan nan
6.3125 1.6875
5.75 2.25
nan nan
3.5 -6.24500451351651e-17
3.5 1.125
nan nan
4.0625 0.5625
3.5 1.125
nan nan
4.0625 0.5625
3.5 -6.24500451351651e-17
nan nan
8 6.24500451351651e-17
8 -1.125
nan nan
7.4375 -0.5625
8 -1.125
nan nan
7.4375 -0.5625
8 6.24500451351651e-17
nan nan
5.75 -2.25
5.75 -1.125
nan nan
6.3125 -1.6875
5.75 -1.125
nan nan
6.3125 -1.6875
5.75 -2.25
nan nan
5.75 2.25
5.75 1.125
nan nan
5.1875 1.6875
5.75 1.125
nan nan
5.1875 1.6875
5.75 2.25
nan nan
3.5 0
4.625 0
nan nan
4.0625 -0.5625
4.625 0
nan nan
4.0625 -0.5625
3.5 0
nan nan
8 0
6.875 0
nan nan
7.4375 0.5625
6.875 0
nan nan
7.4375 0.5625
8 0
nan nan
5.75 -1.125
5.75 -2.25
nan nan
5.1875 -1.6875
5.75 -2.25
nan nan
5.1875 -1.6875
5.75 -1.125
nan nan
5.75 1.125
5.75 2.25
nan nan
6.3125 1.6875
5.75 2.25
nan nan
6.3125 1.6875
5.75 1.125
nan nan
4.625 0
3.5 0
nan nan
4.0625 0.5625
3.5 0
nan nan
4.0625 0.5625
4.625 0
nan nan
6.875 0
8 0
nan nan
7.4375 -0.5625
8 0
nan nan
7.4375 -0.5625
6.875 0
nan nan
6.875 -2.25
5.75 -2.25
nan nan
6.3125 -1.6875
5.75 -2.25
nan nan
6.3125 -1.6875
6.875 -2.25
nan nan
4.625 2.25
5.75 2.25
nan nan
5.1875 1.6875
5.75 2.25
nan nan
5.1875 1.6875
4.625 2.25
nan nan
3.5 -1.125
3.5 0
nan nan
4.0625 -0.5625
3.5 0
nan nan
4.0625 -0.5625
3.5 -1.125
nan nan
8 1.125
8 0
nan nan
7.4375 0.5625
8 0
nan nan
7.4375 0.5625
8 1.125
nan nan
4.625 -2.25
3.5 -2.25
nan nan
4.0625 -1.6875
3.5 -2.25
nan nan
4.0625 -1.6875
4.625 -2.25
nan nan
6.875 2.25
8 2.25
nan nan
7.4375 1.6875
8 2.25
nan nan
7.4375 1.6875
6.875 2.25
nan nan
3.5 1.125
3.5 2.25
nan nan
4.0625 1.6875
3.5 2.25
nan nan
4.0625 1.6875
3.5 1.125
nan nan
8 -1.125
8 -2.25
nan nan
7.4375 -1.6875
8 -2.25
nan nan
7.4375 -1.6875
8 -1.125
nan nan
5.75 -1.125
5.75 0
nan nan
6.3125 -0.5625
5.75 0
nan nan
6.3125 -0.5625
5.75 -1.125
nan nan
5.75 1.125
5.75 0
nan nan
5.1875 0.5625
5.75 0
nan nan
5.1875 0.5625
5.75 1.125
nan nan
4.625 0
5.75 0
nan nan
5.1875 -0.5625
5.75 0
nan nan
5.1875 -0.5625
4.625 0
nan nan
6.875 0
5.75 0
nan nan
6.3125 0.5625
5.75 0
nan nan
6.3125 0.5625
6.875 0
nan nan
5.75 -1.125
4.625 -1.125
nan nan
5.1875 -0.5625
4.625 -1.125
nan nan
5.1875 -0.5625
5.75 -1.125
nan nan
5.75 1.125
6.875 1.125
nan nan
6.3125 0.5625
6.875 1.125
nan nan
6.3125 0.5625
5.75 1.125
nan nan
4.625 -6.24500451351651e-17
4.625 1.125
nan nan
5.1875 0.5625
4.625 1.125
nan nan
5.1875 0.5625
4.625 -6.24500451351651e-17
nan nan
6.875 6.24500451351651e-17
6.875 -1.125
nan nan
6.3125 -0.5625
6.875 -1.125
nan nan
6.3125 -0.5625
6.875 6.24500451351651e-17
nan nan
6.875 -2.25
6.875 -1.125
nan nan
7.4375 -1.6875
6.875 -1.125
nan nan
7.4375 -1.6875
6.875 -2.25
nan nan
4.625 2.25
4.625 1.125
nan nan
4.0625 1.6875
4.625 1.125
nan nan
4.0625 1.6875
4.625 2.25
nan nan
3.5 -1.125
4.625 -1.125
nan nan
4.0625 -1.6875
4.625 -1.125
nan nan
4.0625 -1.6875
3.5 -1.125
nan nan
8 1.125
6.875 1.125
nan nan
7.4375 1.6875
6.875 1.125
nan nan
7.4375 1.6875
8 1.125
nan nan
4.625 -2.25
4.625 -1.125
nan nan
5.1875 -1.6875
4.625 -1.125
nan nan
5.1875 -1.6875
4.625 -2.25
nan nan
6.875 2.25
6.875 1.125
nan nan
6.3125 1.6875
6.875 1.125
nan nan
6.3125 1.6875
6.875 2.25
nan nan
3.5 1.125
4.625 1.125
nan nan
4.0625 0.5625
4.625 1.125
nan nan
4.0625 0.5625
3.5 1.125
nan nan
8 -1.125
6.875 -1.125
nan nan
7.4375 -0.5625
6.875 -1.125
nan nan
7.4375 -0.5625
8 -1.125
nan nan
5.75 -1.125
6.875 -1.125
nan nan
6.3125 -1.6875
6.875 -1.125
nan nan
6.3125 -1.6875
5.75 -1.125
nan nan
5.75 1.125
4.625 1.125
nan nan
5.1875 1.6875
4.625 1.125
nan nan
5.1875 1.6875
5.75 1.125
nan nan
4.625 6.24500451351651e-17
4.625 -1.125
nan nan
4.0625 -0.5625
4.625 -1.125
nan nan
4.0625 -0.5625
4.625 6.24500451351651e-17
nan nan
6.875 -6.24500451351651e-17
6.875 1.125
nan nan
7.4375 0.5625
6.875 1.125
nan nan
7.4375 0.5625
6.875 -6.24500451351651e-17
nan nan
};
\end{axis}

\end{tikzpicture}}} &
            \raisebox{-0.5\height}{\resizebox{0.32\textwidth}{!}{\input{Figs/Isentropic_vortex/2D_forth_coarse_no_flux_mesh_1_modified.tex}}} &
            \raisebox{-0.5\height}{\resizebox{0.32\textwidth}{!}{\input{Figs/Isentropic_vortex/2D_forth_coarse_no_flux_mesh_2_modified.tex}}}
            \vspace{-0.5ex}
            \\
            &  \quad \quad {\small $x_1$} &  \quad \quad {\small $x_1$} &  \quad \quad {\small $x_1$} 
        \end{tabular*}
        \caption{{\bf 2D Isentropic vortex:} nested sequence of mesh
        grid $\LRc{h_1, h_2, h_3}$ for convergence rate analysis, the reference length $h = \frac{4.5 \sqrt{2}}{8}$.}
        \figlab{Isentropic_vortex_meshgrid_set}
\end{figure}

\begin{table}[htb!]
    \centering
    \caption{{\bf 2D Isentropic vortex:} $L^2-$errors and convergence rate for density $\rho$ and $x$-momentum $\rho u$ of the proposed \oDGNet approach and  the Discontinuous Galerkin approach with $T = 0.1s, \dt = 0.002s$.}
    \tablab{Isentropic_vortex_convergence_rate_table}
    \begin{tabular}{l|c c c | c | c c c | c|}
    \toprule
    \multicolumn{1}{l|}{} & \multicolumn{4}{c|}{$\rho$}  & \multicolumn{4}{c|}{$\rho u$}  \\
    \midrule
    {} &         h &       h/2 &       h/4 &      Rate &         h &       h/2 &       h/4 &      Rate \\
    \midrule
    \midrule
    \multicolumn{1}{l}{} & \multicolumn{8}{c}{DG}  \\
    \midrule
    N = 1 &  6.32e-02 &  2.51e-02 &  7.51e-03 &  1.55 &  1.09e-01 &  4.62e-02 &  1.11e-02 &  1.66 \\
    N = 2 &  1.95e-02 &  4.43e-03 &  8.58e-04 &  2.25 &  4.30e-02 &  8.55e-03 &  1.48e-03 &  2.43 \\
    N = 3 &  6.71e-03 &  7.58e-04 &  8.20e-05 &  3.18 &  1.39e-02 &  1.45e-03 &  1.12e-04 &  3.48 \\
    N = 4 &  2.58e-03 &  1.46e-04 &  7.65e-06 &  4.20 &  4.93e-03 &  2.06e-04 &  9.61e-06 &  4.50 \\
    \midrule
    \midrule
    \multicolumn{1}{l}{} & \multicolumn{8}{c}{\oDGNet trained with equivalent data $\LRp{N_i, h_j}$}  \\
    \midrule
    N = 1 &  6.32e-02 &  2.49e-02 &  7.22e-03 &  1.57 &  1.11e-01 &  4.72e-02 &  1.12e-02 &  1.66 \\
    N = 2 &  1.96e-02 &  4.35e-03 &  9.29e-04 &  2.20 &  4.39e-02 &  8.33e-03 &  1.48e-03 &  2.45 \\
    N = 3 &  6.77e-03 &  7.53e-04 &  8.20e-05 &  3.18 &  1.35e-02 &  1.45e-03 &  1.12e-04 &  3.46 \\
    N = 4 &  2.77e-03 &  1.46e-04 &  8.89e-06 &  4.15 &  4.96e-03 &  2.07e-04 &  1.06e-05 &  4.44 \\
    \midrule
    \midrule
    \multicolumn{1}{l}{} & \multicolumn{8}{c}{\oDGNet trained with data $\LRp{N=3, \frac{h}{4}}$}  \\
    \midrule
    N = 1 &  6.30e-02 &  2.47e-02 &  7.12e-03 &  1.58 &  1.10e-01 &  4.71e-02 &  1.10e-02 &  1.67 \\
    N = 2 &  1.97e-02 &  4.43e-03 &  8.58e-04 &  2.26 &  4.38e-02 &  8.51e-03 &  1.48e-03 &  2.45 \\
    N = 3 &  6.70e-03 &  7.55e-04 &  8.20e-05 &  3.18 &  1.38e-02 &  1.44e-03 &  1.12e-04 &  3.47 \\
    N = 4 &  2.58e-03 &  1.46e-04 &  1.10e-05 &  3.94 &  4.90e-03 &  2.06e-04 &  1.24e-05 &  4.31 \\
    \midrule
    \midrule
    \multicolumn{1}{l}{} & \multicolumn{8}{c}{\oDGNet trained with data $\LRp{N=3, \frac{h}{2}}$}  \\
    \midrule
    N = 1 &  6.23e-02 &  2.43e-02 &  7.03e-03 &  1.58 &  1.09e-01 &  4.63e-02 &  1.10e-02 &  1.67 \\
    N = 2 &  1.95e-02 &  4.43e-03 &  8.59e-04 &  2.25 &  4.34e-02 &  8.52e-03 &  1.48e-03 &  2.44 \\
    N = 3 &  6.69e-03 &  7.53e-04 &  9.20e-05 &  3.09 &  1.38e-02 &  1.45e-03 &  1.19e-04 &  3.43 \\
    N = 4 &  2.57e-03 &  1.50e-04 &  5.15e-05 &  2.82 &  4.90e-03 &  2.07e-04 &  4.97e-05 &  3.31 \\
    \midrule
    \midrule
    \multicolumn{1}{l}{} & \multicolumn{8}{c}{\oDGNet trained with data $\LRp{N=3, h}$}  \\
    \midrule
    N = 1 &  6.74e-02 &  2.70e-02 &  7.74e-03 &   1.57 &  1.17e-01 &  5.04e-02 &  1.17e-02 &  1.67 \\
    N = 2 &  2.07e-02 &  4.73e-03 &  2.87e-03 &   1.42 &  4.54e-02 &  8.65e-03 &  3.25e-03 &  1.90 \\
    N = 3 &  6.77e-03 &  2.13e-03 &  3.28e-03 &   0.52 &  1.35e-02 &  2.53e-03 &  3.45e-03 &  0.98 \\
    N = 4 &  3.09e-03 &  2.36e-03 &  3.76e-03 &   -0.14 &  5.11e-03 &  2.37e-03 &  3.81e-03 &  0.21 \\
    \bottomrule
\end{tabular}

\end{table}

\begin{figure}[htb!]
    \centering
        \begin{tabular*}{\textwidth}{c c}
            \centering
            $\quad \quad N = 1$ & $\quad \quad N = 2$ 
            \\
            \raisebox{-0.5\height}{\includegraphics[width = .48\textwidth]{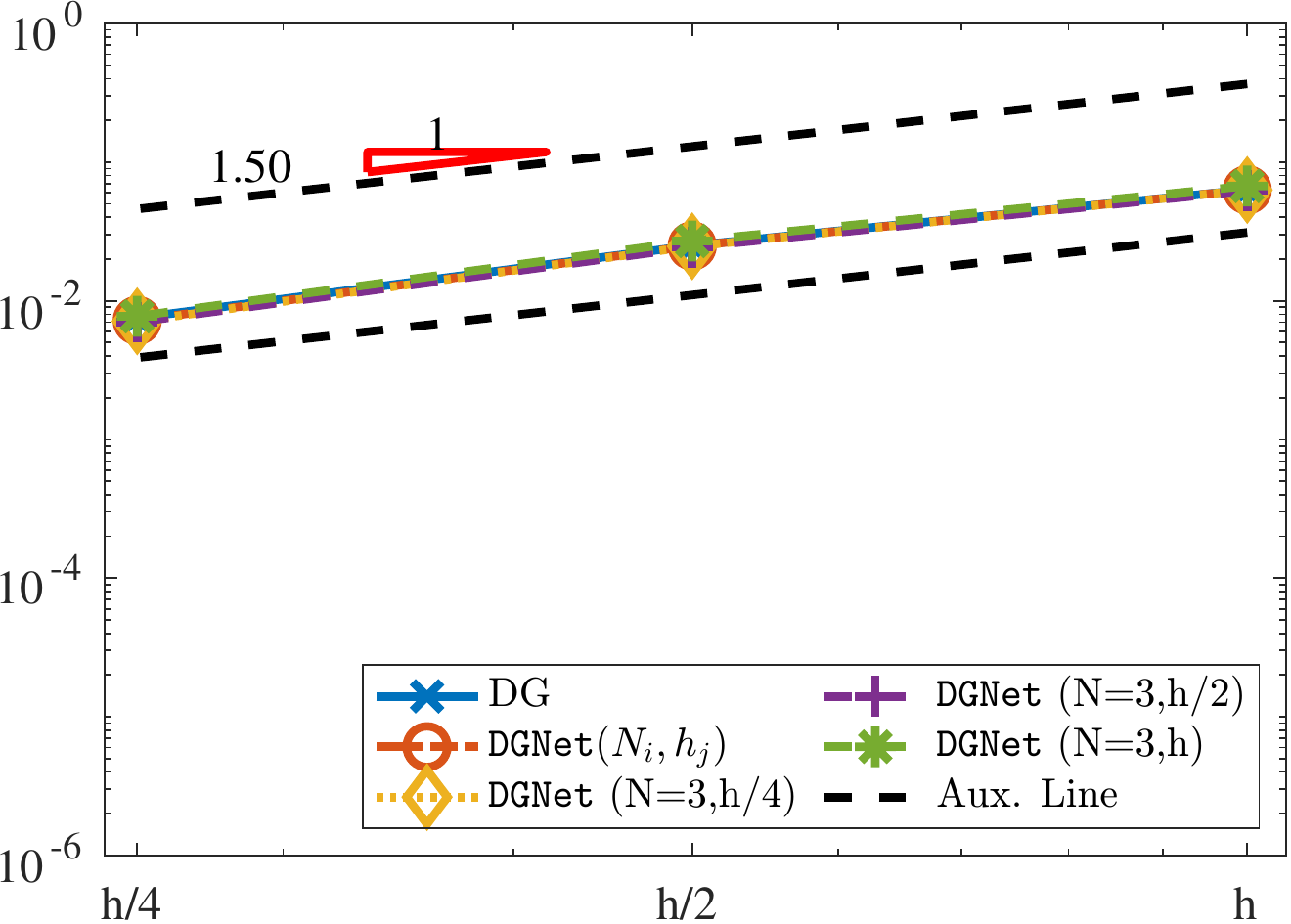}} &
            \raisebox{-0.5\height}{\includegraphics[width = .48\textwidth]{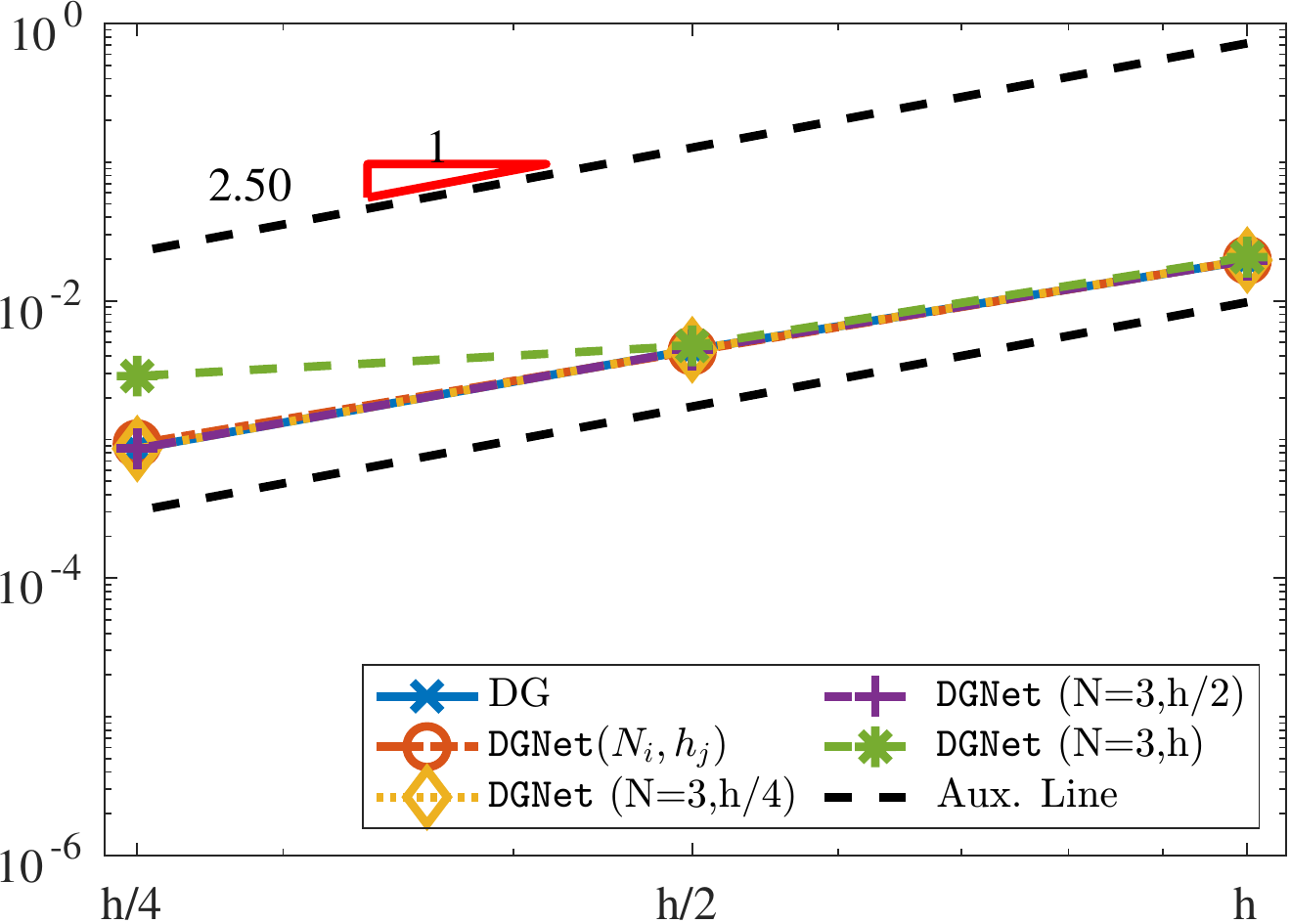}} 
            \\ ~ \\
            $\quad \quad N = 3$ & $\quad \quad N = 4$
            \\
            \raisebox{-0.5\height}{\includegraphics[width = .48\textwidth]{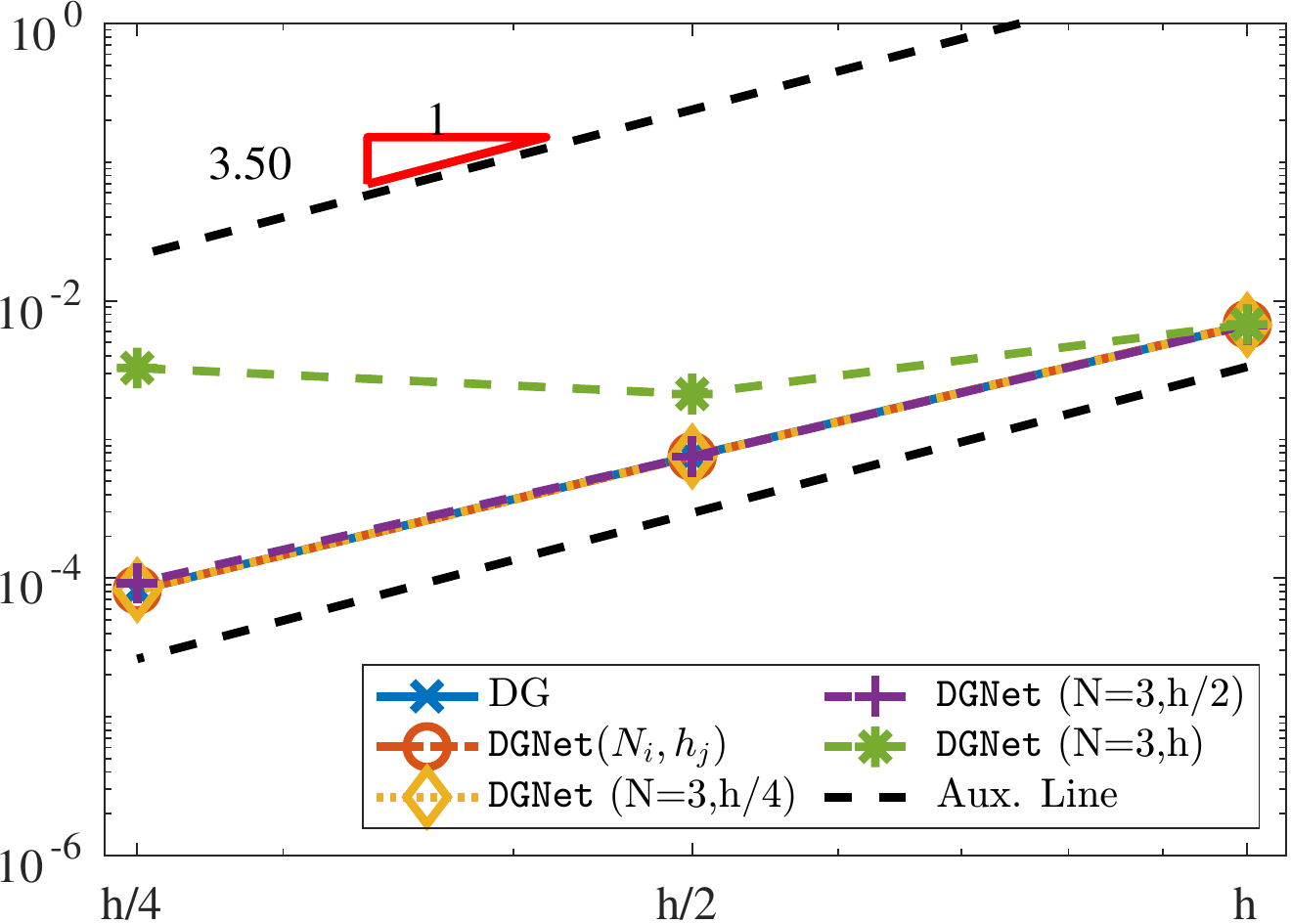}} &
            \raisebox{-0.5\height}{\includegraphics[width = .48\textwidth]{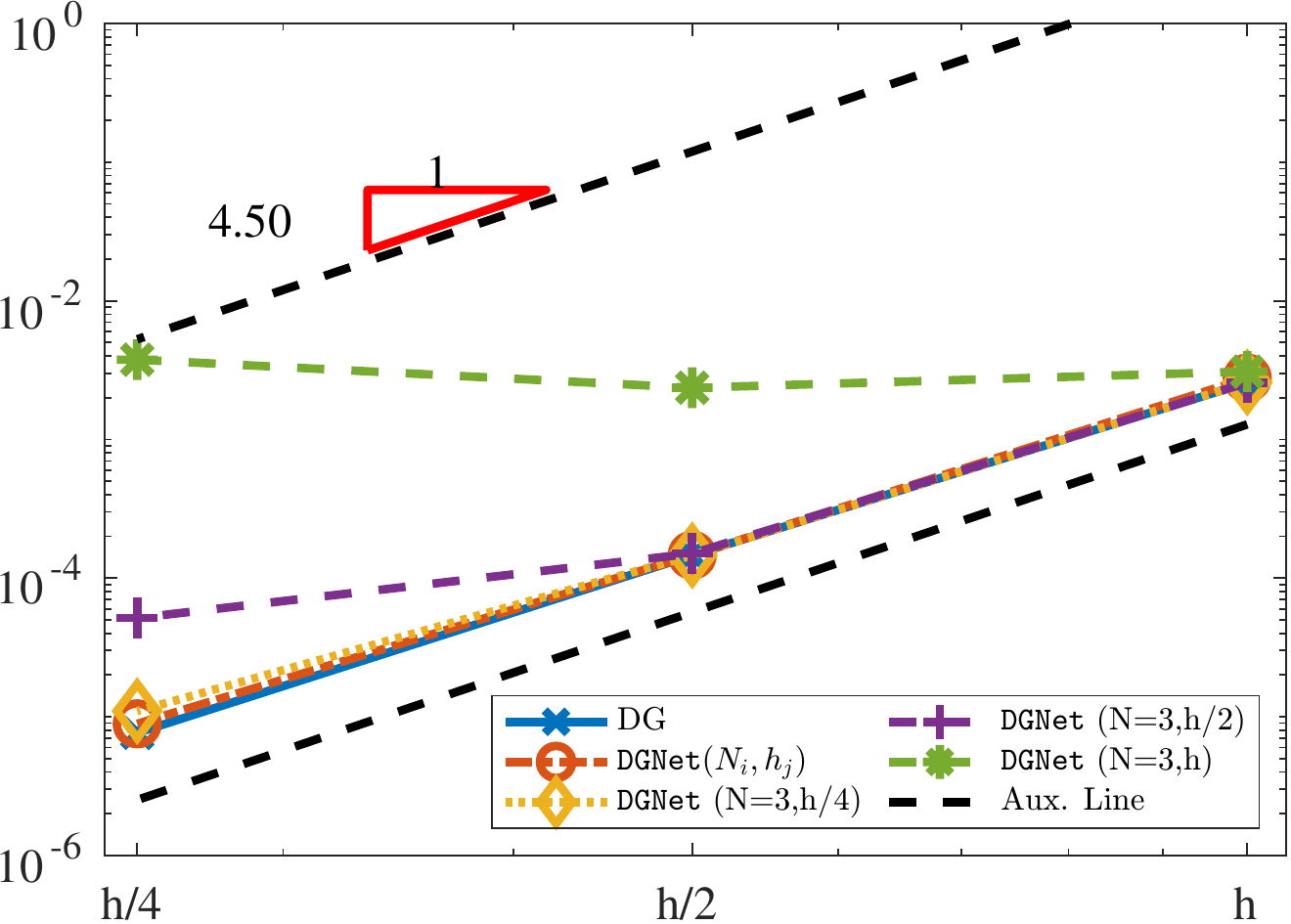}}
        \end{tabular*}
        \caption{{\bf 2D Isentropic vortex:} A comparison for the convergence rate of density $\rho$ from DG and different trained \oDGNet neural networks. {\bf Top left:} $\LRp{N=1, h_j}$ discretization. {\bf Top right:} $\LRp{N=2, h_j}$  discretization. {\bf Bottom left:} $\LRp{N=3, h_j}$ discretization. {\bf Bottom right:} $\LRp{N=4, h_j}$  discretization.}
        \figlab{Isentropic_vortex_convergen_rate}
\end{figure}

In this section, we focus on studying the convergence rate of the proposed \oDGNet approach for the 2D isentropic vortex problem \cite{yee1999low} for which the closed-form solution is available. The 2D compressible Euler equations are given by
\begin{equation*}
    \begin{aligned}
        \frac{\partial}{\partial t} 
            \underbrace{\begin{pmatrix}
            \rho \\
            \rho u \\
            \rho v \\
            E
            \end{pmatrix}}_{\ub} 
        + \frac{\partial}{\partial x_1} 
            \underbrace{\begin{pmatrix}
            \rho u \\
            \rho u^2 + p \\
            \rho u v \\
            u(E + p)
            \end{pmatrix}}_{\fb_1} 
        + \frac{\partial}{\partial x_2} 
            \underbrace{\begin{pmatrix}
            \rho v \\
            \rho u v \\
            \rho v^2 + p \\
            v(E + p)
            \end{pmatrix}}_{\fb_2} = 0,
    \end{aligned}
    \eqnlab{2D_Euler_equations}
\end{equation*}
where
\begin{equation*}
	E = \frac{p}{\gamma - 1} + \frac{\rho}{2} \LRp{u^2 + v^2}.
\end{equation*}
An exact solution for this problem is given by
\begin{equation*}
    \eqnlab{isentropic_vortex_exact_solution}
    \begin{aligned}
        u &=1-\beta e^{(1-r^{2})}\frac{{x_2}}{2\pi}, \\
        v &=\beta e^{(1-r^{2})}\frac{x_1-5}{2\pi},\\
        \rho &=\left(1-\left(\frac{\gamma-1}{16\gamma\pi^{2}}\right)\beta^{2}e^{2\left(1-r^{2}\right)}\right)^{\frac{1}{\gamma-1}}, \\
        p &= \rho^\gamma,
    \end{aligned}
\end{equation*}
where $r = \sqrt{\LRp{x_1-t-5}^2 + {{x_2}}^2},\beta = 5 $ and $\gamma$ is the gas constant (whose value varies for different data sets). The initial and boundary conditions are determined by evaluating the exact solution at the appropriate time and spatial coordinate points. Convergence analysis is conducted using a set of nested mesh grids, denoted as $\LRc{h_1, h_2, h_3}$, as shown in \cref{fig:Isentropic_vortex_meshgrid_set}. The coarsest mesh $h_1$ features the longest reference edge length $h = \frac{4.5 \sqrt{2}}{8}$, while the finer meshes are $h_2 = \frac{h}{2}$ and $h_3 = \frac{h}{4}$. The solution order is set to $N \in \LRc{1, 2, 3, 4}$. Therefore, we have $12$ different discretization settings with different element orders and mesh sizes: $\LRp{N_i, h_j}, i = 1,\hdots, 4, j = 1,2,3$. For data generation, we solve 2D Euler for each combination $\LRp{N_i, h_j}$ with $\gamma = \LRp{1.2, 1.6}$ for training data and with $\gamma = 1.4$ for validation data within the time interval $\LRs{0,0.3}s$. The test data set is obtained directly from the analytical solutions.

In this problem, no slope limiter is applied as no shock waves are present. For simplicity, a uniform time step $\dt = 0.002s$ is used 
for all 12 different discretization settings.
%across all mesh grids, adhering to the CFL condition. 
For this problem, both the \nDGNet approach and \mcDGNet approach with a small noise level (less than $0.2\%$) show the same performance. Therefore, we opt to present the \mcDGNet approach results as the representative and denoted as \DGNet. All four components $\rho$, $\rho u$, $\rho v$, and $E$ show a similar convergence rate, thus we only present the results for $\rho$ and $\rho u$ for brevity. The convergence rate computed from the $L^2-$errors for both the proposed \oDGNet approach and the Discontinuous Galerkin (DG) approach at $T = 0.1s$ is presented in \cref{tab:Isentropic_vortex_convergence_rate_table}, and is further illustrated in \cref{fig:Isentropic_vortex_convergen_rate}. As can be seen, the convergence rates obtained from \oDGNet and DG trained with data from the same pairs $\LRp{N_i, h_j}$ are comparable. This is in agreement with the expected theoretical convergence rate in \cref{eq:errorDGNet}. \cref{fig:Isentropic_vortex_convergen_rate} reveals an interesting feature of \oDGNet\hspace{-1ex}: the numerical results suggest that \oDGNet be discretization-invariant \cite{kovachki2023neural, li2020fourier,ong2022integral}, that is, \oDGNet trained on one discretization does not produce higher error for other discretization (finer or coarser). In fact, numerically \oDGNet is more than discretization-invariant as when it is trained on higher solution order (see the purple lines of the first row and the bottom left subfigures in \cref{fig:Isentropic_vortex_convergen_rate}) it can maintain the convergence rate for lower-order solutions even with finer mesh sizes. 
%As presented in \cref{sect:Error_estimation}, the total error is bounded by the sum of the error between machine learning predictions versus numerical test data $\LRp{N_i,h_j}$ and numerical simulation with $\LRp{N_i,h_j}$ versus analytical solutions. Here, it can be interpreted that the former error is sufficiently small to preserve the convergence rate very well, but not completely. Intuitively, in this problem, the prime fields are simply translated to the right with a slight change in profile, thus normalized training, validation, and test datasets are almost similar. 
%However, using the neural network trained with $\LRp{N = 3, \frac{h}{4}}$ for solving other $\LRp{N_i, h_j}$ combinations maintains the convergence only for the same or lower order elements and the same or coarser mesh discretization, but not for the higher order element $N = 4$ with the same mesh discretization $\frac{h}{4}$. 
This is expected since our \oDGNet approach accuracy is aligned with the information of training data and the well-known fact that higher solution order typically yields an exponential convergence rate for smooth solutions versus a polynomial convergence rate with mesh size.
%In this case, the error between machine learning prediction and numerical test data is larger due to more discrepancy between training data $\LRp{N = 3, \frac{h}{4}}$ and numerical test data $\LRp{N = 4, \frac{h}{4}}$. For the same reason, similar patterns can be observed for the case $\LRp{N =3, \frac{h}{2}}$ and $\LRp{N = 3, h}$ trained \oDGNet neural networks. Interestingly, the neural network trained with $\LRp{N = 3, h}$ can preserve the convergence rate for $\LRp{N=1, h_j}, j = 1,2,3$. It is likely because the discretization $\LRp{N = 3, h}$ is as fine as $\LRp{N = 1, h_j}$ discretization, hence the training data information $\LRp{N = 3, h}$ is sufficiently close to numerical test data $\LRp{N = 1, h_j}$.

%\tanbui{starting from here}

For testing long-term prediction capacity, we employ different \oDGNet neural networks trained with $\LRp{N = 3, \frac{h}{4}}$, $\LRp{N = 3, \frac{h}{2}}$, $\LRp{N = 3, h}$  to solve 2D Euler problems on the $\LRp{N = 3, \frac{h}{4}}$ mesh discretization up to $T_\text{test} = 2$s. The left subfigure of \cref{fig:Isentropic_vortex_relative_error_comparison} shows the relative $L^2-$error of \oDGNet and  DG solutions. As can be seen, \oDGNet trained with $\LRp{N = 3, \frac{h}{4}}$ can provide as accurate predictions as the DG approach. We see that \oDGNet trained with $\LRp{N = 3, \frac{h}{2}}$ yields slightly less accurate prediction, and \oDGNet trained with $\LRp{N = 3, h}$ results in significantly more erroneous predictions. This outcome is expected since the training data from $\LRp{N = 3, \frac{h}{4}}$ is sufficiently informative for training \oDGNet neural networks to predict long-term solutions on the same discretization $\LRp{N = 3, \frac{h}{4}}$. The other networks, trained on $\LRp{N = 3, \frac{h}{2}}$ and $\LRp{N = 3, h}$, have more discrepancy between training data and numerical test data. 

\noindent{\bf Implicit \oDGNet\hspace{-1ex}.} We also implement the trained \oDGNet in conjunction with implicit time integration to solve the 2D Euler problem on the $\LRp{N = 3, \frac{h}{4}}$ mesh discretization up to $T_\text{test} = 2s$ with time step size of $\dt = 0.02s$. The right subfigure of \cref{fig:Isentropic_vortex_relative_error_comparison} presents the relative $L^2-$error of \oDGNet solutions and  DG solutions, both using implicit time integration. As shown, the implicit \oDGNet solutions are in agreement with the implicit DG method. The results highlight that the \oDGNet approach can be combined with implicit time integration to provide long-term predictions for the 2D Euler problem as good as the implicit DG method.

% \begin{figure}[htb!]
%     \centering
%     \includegraphics[width = .8\textwidth]{Figs/Isentropic_vortex/2D_DG_relative_error_comparison.png}
%     \caption{{\bf 2D Isentropic vortex:} Comparison of $L^2-$relative error test predictions upto $T=2s$ between different approaches.} 
%     \figlab{Isentropic_vortex_relative_error_comparison}
% \end{figure}

% \begin{figure}[htb!]
%     \centering
%     \resizebox{.9\textwidth}{!}{\input{Figs/Isentropic_vortex/2D_DG_relative_error_comparison_modified.tex}}
%     % \resizebox{.9\textwidth}{!}{\input{Figs/Isentropic_vortex/2D_DG_relative_error_comparison.tex}}
%     \caption{{\bf 2D Isentropic vortex:} comparison of relative $L^2-$error of density predictions up to $T=2s$ between \oDGNet neural networks and traditional DG method integration with both 2nd-SSP-RK and implicit Backward Euler scheme on the $\LRp{N = 3, \frac{h}{4}}$ mesh discretization.} 
%     \figlab{Isentropic_vortex_relative_error_comparison}
% \end{figure}

\begin{figure}[htb!]
    \centering
    \begin{tabular*}{\textwidth}{c@{\hskip -0.01cm} c@{\hskip -0.01cm}}
        \centering
        \quad \quad  Explicit 2nd-SSP-RK scheme & \quad \quad Implicit Backward Euler scheme \\
        \raisebox{-0.5\height}{\resizebox{0.5\textwidth}{!}{\input{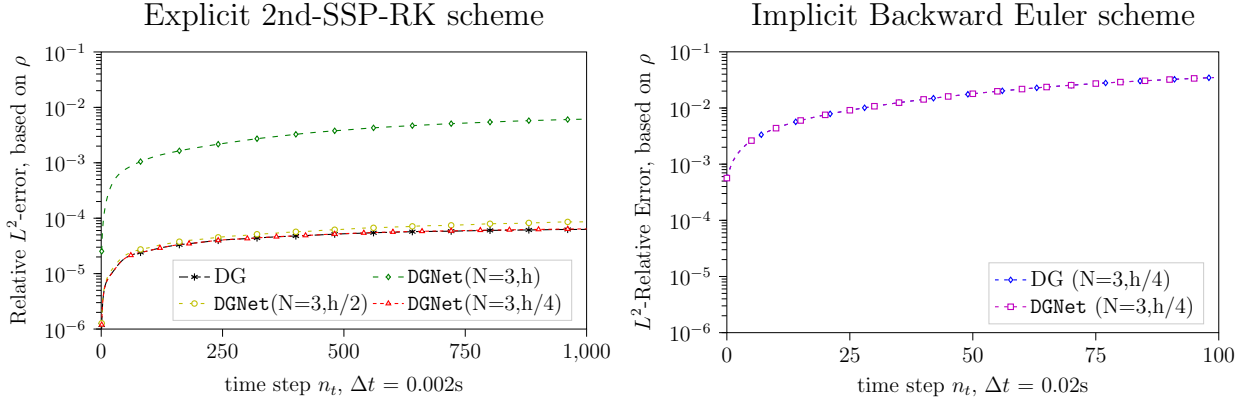}}} &
        \raisebox{-0.5\height}{\resizebox{0.5\textwidth}{!}{% This file was created with tikzplotlib v0.10.1.
\begin{tikzpicture}

\definecolor{darkgray176}{RGB}{176,176,176}
\definecolor{darkviolet1910191}{RGB}{191,0,191}
\definecolor{lightgray204}{RGB}{204,204,204}

\begin{axis}[
legend cell align={left},
legend columns=1,
legend style={
  fill opacity=0.8,
  draw opacity=1,
  text opacity=1,
  at={(0.97,0.03)},
  anchor=south east,
  draw=lightgray204
},
log basis y={10},
tick align=outside,
tick pos=left,
x grid style={darkgray176},
xlabel={time step \(\displaystyle n_t\), \(\displaystyle \Delta t\) = 0.02s},
xmin=0, xmax=100,
xtick={0,25, 50, 75, 100},
xtick style={color=black},
y grid style={darkgray176},
ylabel={\(\displaystyle L^2\)-Relative Error, based on \(\displaystyle \rho\)},
ymin=1e-06, ymax=0.1,
ymode=log,
ytick style={color=black},
width = 11cm,
height = 7.cm,
]
\addplot [semithick, blue, dash pattern=on 2pt off 3pt, mark=diamond*, mark size=1.5, mark repeat=7, mark options={solid,fill=white}]
table {%
-1 0
0 0.000565165700309101
1 0.00104356614641167
2 0.00147046653204191
3 0.00187013987816672
4 0.00225354383799601
5 0.00262573823133804
6 0.00298866667660278
7 0.00334412819080574
8 0.0036929273891378
9 0.00403564557958158
10 0.00437286897493124
11 0.00470492094410329
12 0.00503229990951652
13 0.0053554951562465
14 0.00567543143239884
15 0.00599231589316697
16 0.0063067552344469
17 0.00661944180657065
18 0.00693081609451739
19 0.00724149915966873
20 0.00755213420997541
21 0.00786310388785777
22 0.0081748880114372
23 0.00848798771516378
24 0.0088027919696828
25 0.00911958892077993
26 0.00943873357082405
27 0.00976052967249374
28 0.0100850929616461
29 0.0104125388295014
30 0.0107430422986948
31 0.0110767604915213
32 0.0114136162648438
33 0.0117537116512006
34 0.0120969457942934
35 0.0124431082713382
36 0.0127924216999967
37 0.0131445400644791
38 0.0134993565307209
39 0.0138568573831324
40 0.0142168837254297
41 0.0145793649980596
42 0.0149439440663703
43 0.0153106888655793
44 0.0156792825597179
45 0.0160497078195693
46 0.0164217928148474
47 0.0167954091980248
48 0.0171703337891938
49 0.0175465362813655
50 0.0179237849828934
51 0.0183018589087202
52 0.0186808793117632
53 0.0190606027116924
54 0.0194407743624594
55 0.0198213964565454
56 0.0202021005616274
57 0.0205828134063415
58 0.0209633862307016
59 0.0213436243689247
60 0.0217232988283649
61 0.0221021983758259
62 0.0224802658236408
63 0.0228572014138662
64 0.0232327887348779
65 0.0236065711618912
66 0.0239787724816166
67 0.0243489427376332
68 0.0247170061975583
69 0.0250828804988803
70 0.0254462772227264
71 0.0258069739918446
72 0.0261649290218034
73 0.026519982278287
74 0.0268721005722507
75 0.027221188938024
76 0.0275670694838965
77 0.0279099638961698
78 0.0282497324935141
79 0.0285862215661984
80 0.028919871320021
81 0.0292504513050025
82 0.0295781396273561
83 0.029902958903139
84 0.0302252069653903
85 0.0305450143782007
86 0.030862313079254
87 0.0311772287478294
88 0.0314903174285365
89 0.0318013763099413
90 0.0321108014545578
91 0.0324186397691658
92 0.0327251303113126
93 0.0330304175337018
94 0.0333345817829418
95 0.0336376400667613
96 0.0339397930621112
97 0.0342410704113602
98 0.0345414762532172
99 0.0348410656453561
};
\addlegendentry{DG (N=3,h/4)}
\addplot [semithick, darkviolet1910191, dash pattern=on 2pt off 3pt, mark=square*, mark size=1.5, mark repeat=5, mark options={solid,fill=white}]
table {%
-1 0
0 0.000565165700309101
1 0.00104356614641167
2 0.00147046653204191
3 0.00187013987816672
4 0.00225354383799601
5 0.00262573823133804
6 0.00298866667660278
7 0.00334412819080574
8 0.0036929273891378
9 0.00403564557958158
10 0.00437286897493124
11 0.00470492094410329
12 0.00503229990951652
13 0.0053554951562465
14 0.00567543143239884
15 0.00599231589316697
16 0.0063067552344469
17 0.00661944180657065
18 0.00693081609451739
19 0.00724149915966873
20 0.00755213420997541
21 0.00786310388785777
22 0.0081748880114372
23 0.00848798771516378
24 0.0088027919696828
25 0.00911958892077993
26 0.00943873357082405
27 0.00976052967249374
28 0.0100850929616461
29 0.0104125388295014
30 0.0107430422986948
31 0.0110767604915213
32 0.0114136162648438
33 0.0117537116512006
34 0.0120969457942934
35 0.0124431082713382
36 0.0127924216999967
37 0.0131445400644791
38 0.0134993565307209
39 0.0138568573831324
40 0.0142168837254297
41 0.0145793649980596
42 0.0149439440663703
43 0.0153106888655793
44 0.0156792825597179
45 0.0160497078195693
46 0.0164217928148474
47 0.0167954091980248
48 0.0171703337891938
49 0.0175465362813655
50 0.0179237849828934
51 0.0183018589087202
52 0.0186808793117632
53 0.0190606027116924
54 0.0194407743624594
55 0.0198213964565454
56 0.0202021005616274
57 0.0205828134063415
58 0.0209633862307016
59 0.0213436243689247
60 0.0217232988283649
61 0.0221021983758259
62 0.0224802658236408
63 0.0228572014138662
64 0.0232327887348779
65 0.0236065711618912
66 0.0239787724816166
67 0.0243489427376332
68 0.0247170061975583
69 0.0250828804988803
70 0.0254462772227264
71 0.0258069739918446
72 0.0261649290218034
73 0.026519982278287
74 0.0268721005722507
75 0.027221188938024
76 0.0275670694838965
77 0.0279099638961698
78 0.0282497324935141
79 0.0285862215661984
80 0.028919871320021
81 0.0292504513050025
82 0.0295781396273561
83 0.029902958903139
84 0.0302252069653903
85 0.0305450143782007
86 0.030862313079254
87 0.0311772287478294
88 0.0314903174285365
89 0.0318013763099413
90 0.0321108014545578
91 0.0324186397691658
92 0.0327251303113126
93 0.0330304175337018
94 0.0333345817829418
95 0.0336376400667613
96 0.0339397930621112
97 0.0342410704113602
98 0.0345414762532172
99 0.0348410656453561
};
\addlegendentry{\oDGNet (N=3,h/4)}
\end{axis}

\end{tikzpicture}}} 
        \vspace{-0.5ex}
    \end{tabular*}
    \caption{{\bf 2D Isentropic vortex:} comparison of relative $L^2-$error of density predictions up to $T=2s$ between \oDGNet neural networks and traditional DG method integration with explicit 2nd-SSP-RK scheme with $\dt = 0.002s$ ({\bf Left})and implicit Backward Euler scheme with $\dt = 0.02s$ ({\bf Right}) on the $\LRp{N = 3, \frac{h}{4}}$ mesh discretization.} 
    \figlab{Isentropic_vortex_relative_error_comparison}
\end{figure}

% \clearpage

\subsection{Forward facing step problem}
\seclab{forward_facing_conner}

\begin{figure}[htb!]
    \centering
        \begin{tabular*}{\textwidth}{c c}
            \centering
            Model 1 & Model 2
            \\
            \raisebox{-0.5\height}{\includegraphics[width = .48\textwidth]{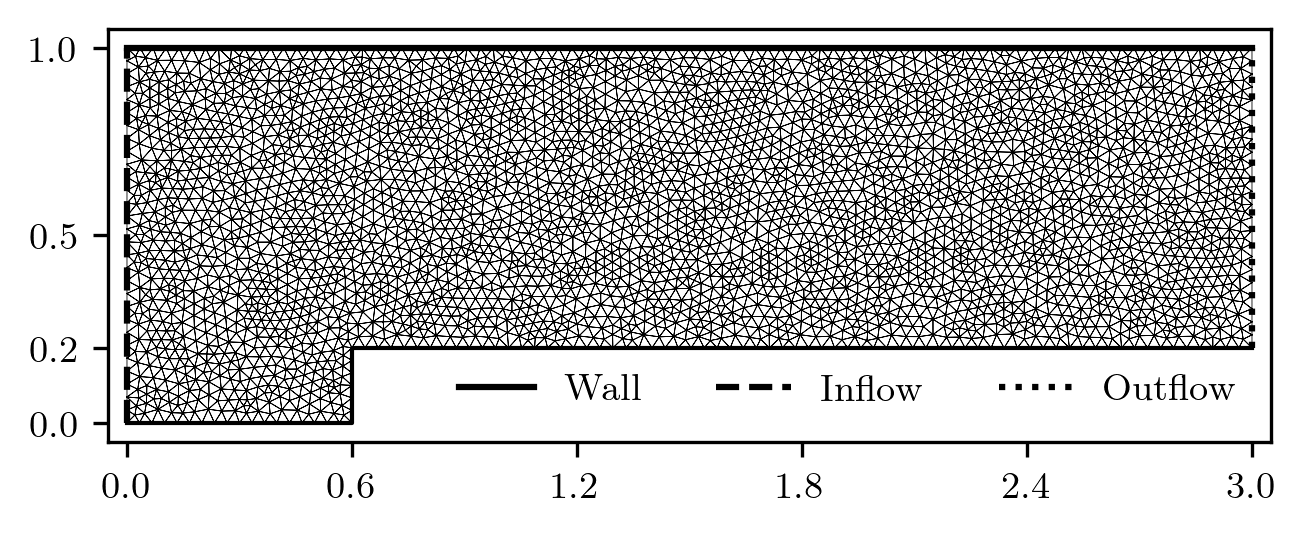}} &
            \raisebox{-0.5\height}{\includegraphics[width = .48\textwidth]{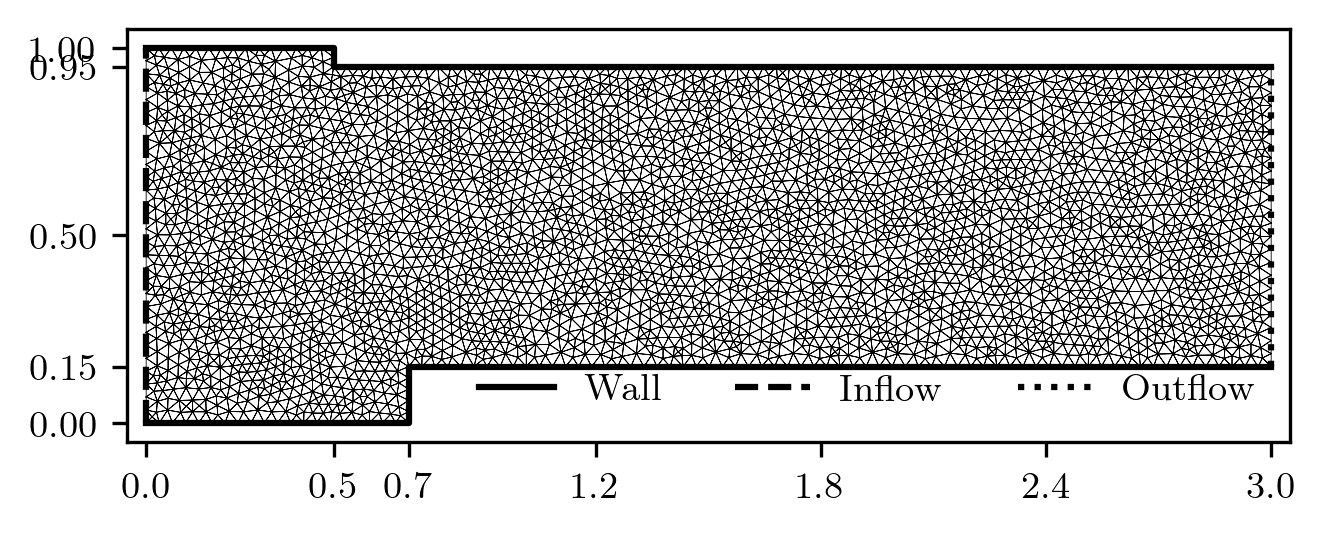}} 
        \end{tabular*}
        \caption{{\bf 2D forward facing step:} domain discretization for Model 1 (K = 7039 elements) and Model 2 (K = 7008 elements) mesh girds.}
        \figlab{forward_facing_mesh_grid_set}
\end{figure}

\begin{figure}[htb!]
    \centering
        \raisebox{-0.5\height}{\resizebox{0.8\textwidth}{!}{% This file was created with tikzplotlib v0.10.1 and corrected manually
\begin{tikzpicture}

    \definecolor{darkgray176}{RGB}{176,176,176}
    \definecolor{lightgray204}{RGB}{204,204,204}
    
    \begin{axis}[
    width=14cm,
    height=5cm,
    legend cell align={left},
    legend style={
        fill opacity=0.8, 
        draw opacity=1, 
        text opacity=1, 
        draw=lightgray204,
        at={(0.98,0.98)},
        anchor=north east,
        font=\footnotesize  % Reduced font size for legend
    },
    tick align=outside,
    tick pos=left,
    x grid style={darkgray176},
    xlabel={Noise level $\delta$},
    xlabel style={font=\small},  % Slightly reduced font size for x-label
    xmin=-0.008, xmax=0.168,
    xtick style={color=black, font=\footnotesize},  % Reduced font size for x-ticks
    xtick={0., 0.04, 0.08, 0.12, 0.16},
    xticklabel={\pgfmathprintnumber[fixed, fixed zerofill, precision=2]{\tick}},  % Format x-ticks as decimals
    y grid style={darkgray176},
    ylabel={Relative $L^2$-error},
    ylabel style={font=\small},  % Slightly reduced font size for y-label
    ymin=0, ymax=0.025,
    ytick style={color=black, font=\footnotesize},  % Reduced font size for y-ticks
    ytick={0,0.005,0.01,0.015,0.02,0.025},
    ]
    \addplot [semithick, black, mark=asterisk, mark size=3, mark options={solid}]
    table {%
    0 0.0212913937866688
    0.01 0.00868068542331457
    0.02 0.00594778079539537
    0.03 0.00606754561886191
    0.04 0.00605273945257068
    0.05 0.00626298086717725
    0.06 0.0062317824922502
    0.07 0.00648258021101356
    0.08 0.00724800489842892
    0.09 0.00728429853916168
    0.1 0.00804220233112574
    0.11 0.00865828990936279
    0.12 0.00953815970569849
    0.13 0.0126465847715735
    0.14 0.0139803998172283
    0.15 0.0158460736274719
    0.16 0.0190907958894968
    };
    \addlegendentry{Average relative error for validation data}
    \end{axis}
    
\end{tikzpicture}}}
        \caption{{\bf 2D forward facing step:} a survey of different noise levels for data randomization versus validation data relative $L^2$-error average over three conservative components $\LRp{\rho, \rho u, E}$ obtained by \mcDGNet approach.}
        \figlab{Relative_error_validation_data_model1_versus_noise}
\end{figure}
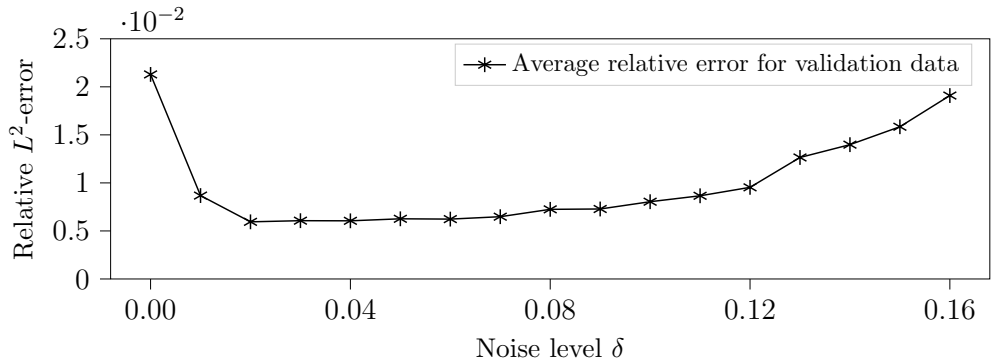

\begin{figure}[htb!]
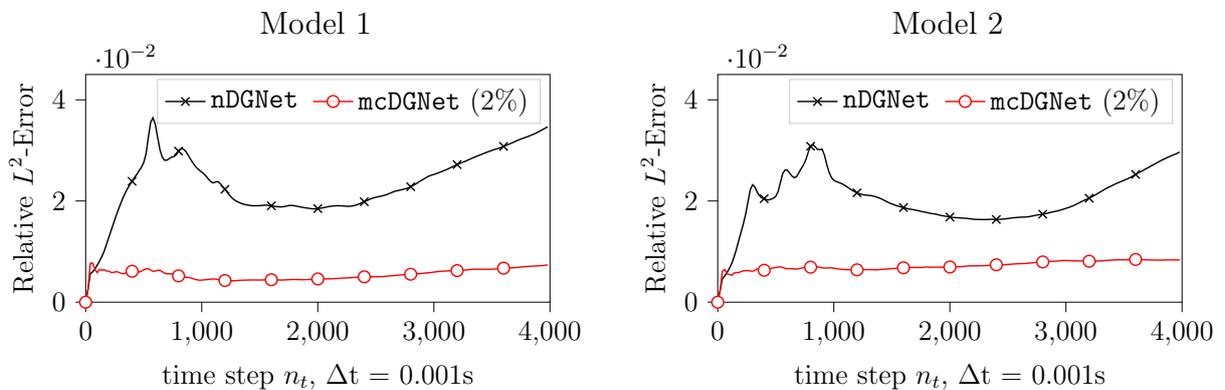

    \centering
        \begin{tabular*}{\textwidth}{c c}
            \centering
            \quad Model 1 & \quad  Model 2
            \\
            \raisebox{-0.5\height}{\resizebox{0.48\textwidth}{!}{\input{Figs/2D_Euler_Forward_facing/Relative_error_test_data_model1_mesh_modified.tex}}} &
            \raisebox{-0.5\height}{\resizebox{0.48\textwidth}{!}{\input{Figs/2D_Euler_Forward_facing/Relative_error_test_data_model2_mesh_modified.tex}}} 
        \end{tabular*}
        \caption{{\bf 2D forward facing step:}  relative average $L^2$-error of \nDGNet and \mcDGNet approaches on test data for three conservative variables $\LRp{\rho, \rho u, E}$   at different time steps for Model 1 ({\bf Left}) and Model 2 ({\bf Right}: this is complete out-of-distribution case).}
        \figlab{Relative_error_test_data_model1_model2_mesh}
\end{figure}

\begin{figure}[htb!]
    \centering
        \begin{tabular*}{\textwidth}{c c@{\hskip -0.0001cm} c@{\hskip -0.002cm} c@{\hskip -0.002cm} c@{\hskip -0.002cm}}
            \centering
            & & DG & \nDGNet  & \mcDGNet  ($2\%$)
            \\
            \multirow{2}{*}{\rotatebox[origin=l]{90}{Model 1 \quad  }} &
            \rotatebox[origin=c]{90}{Pred} &
            \raisebox{-0.5\height}{\includegraphics[width = .31\textwidth]{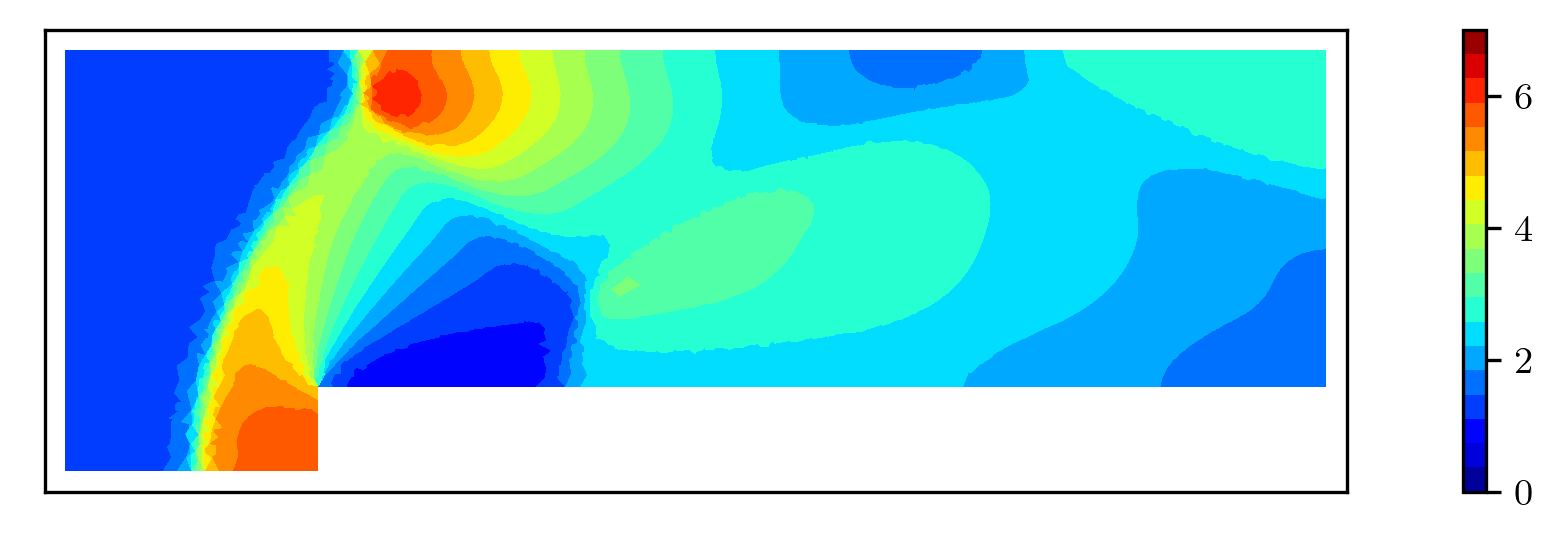}} &
            \raisebox{-0.5\height}{\includegraphics[width = .31\textwidth]{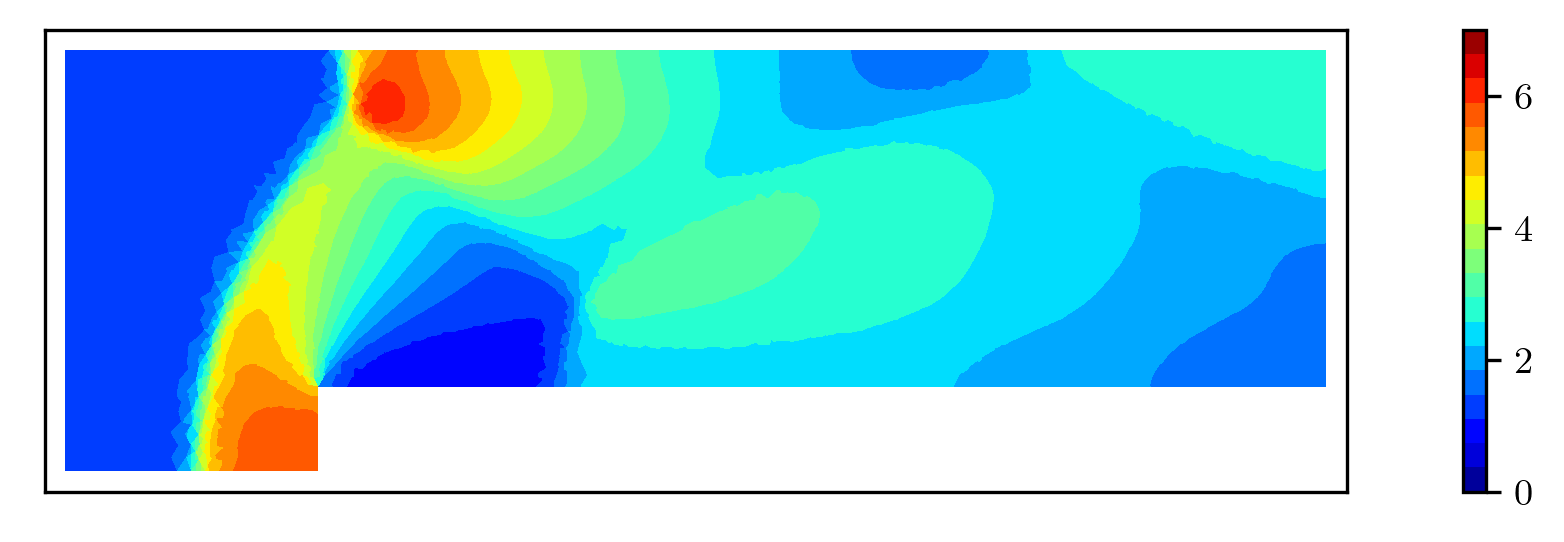}} &
            \raisebox{-0.5\height}{\includegraphics[width = .31\textwidth]{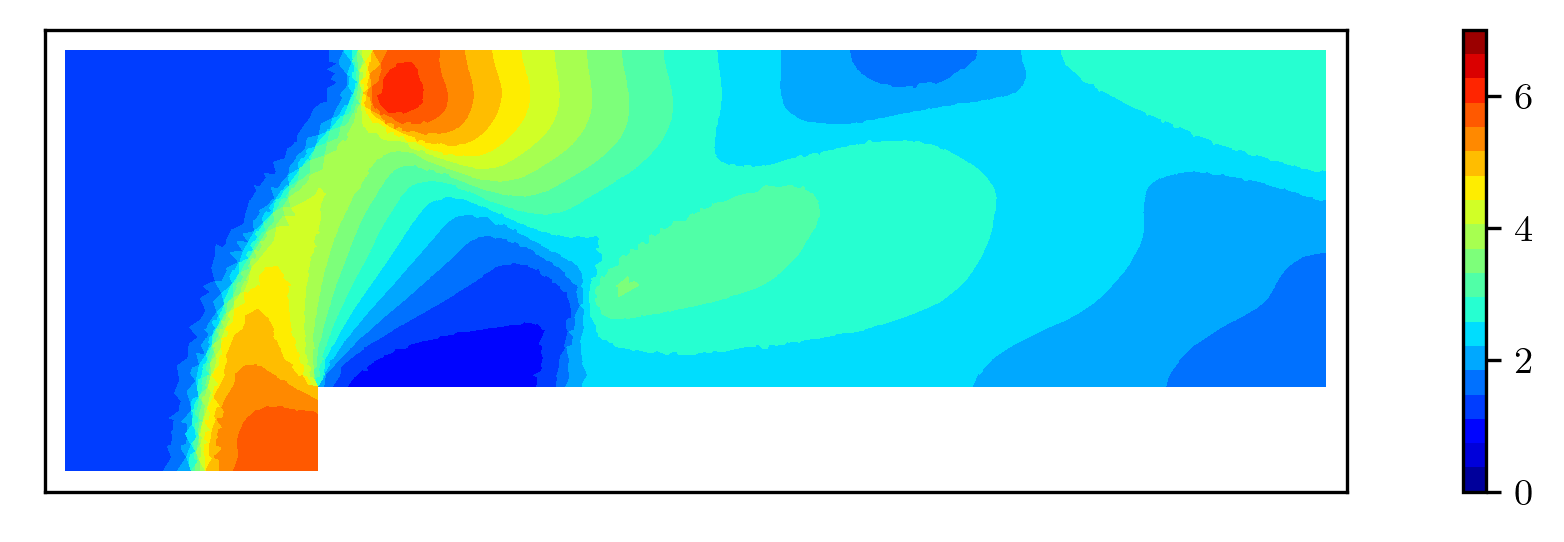}}
            \\
            &
            \rotatebox[origin=c]{90}{Error} &
            \raisebox{-0.5\height}{} &
            \raisebox{-0.5\height}{\includegraphics[width = .31\textwidth]{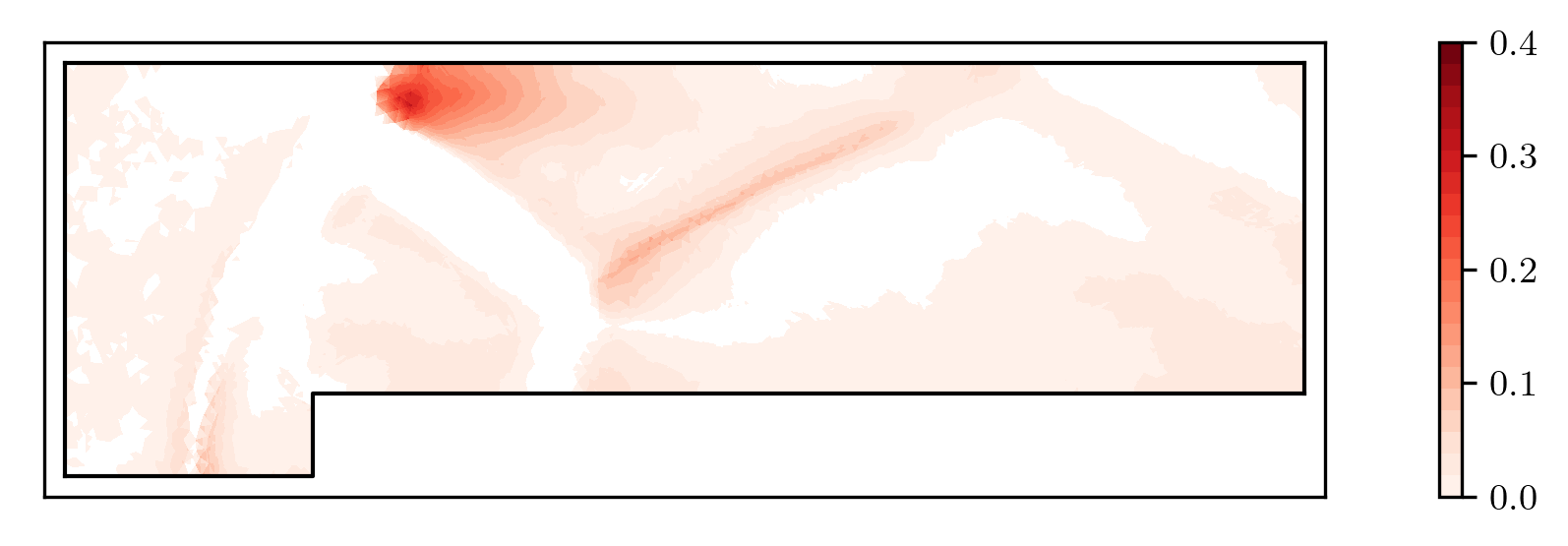}} &
            \raisebox{-0.5\height}{\includegraphics[width = .31\textwidth]{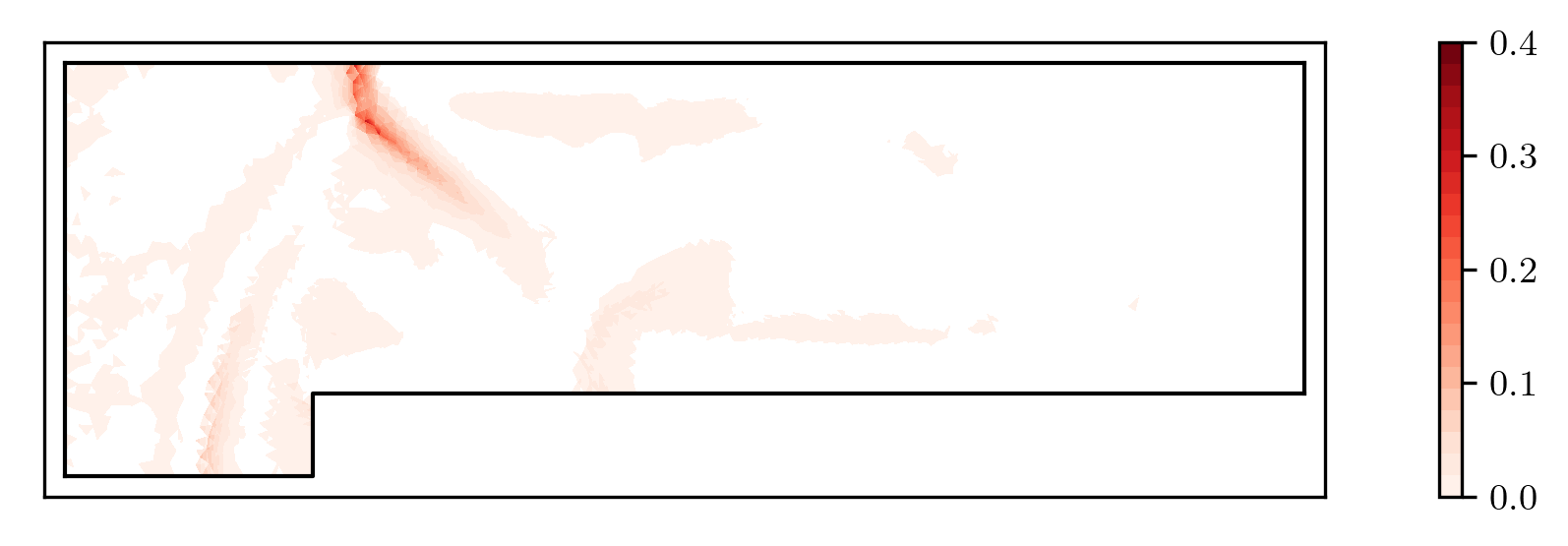}}
            \\
            \multirow{2}{*}{\rotatebox[origin=l]{90}{Model 2 \quad  }} &
            \rotatebox[origin=c]{90}{Pred} &
            \raisebox{-0.5\height}{\includegraphics[width = .31\textwidth]{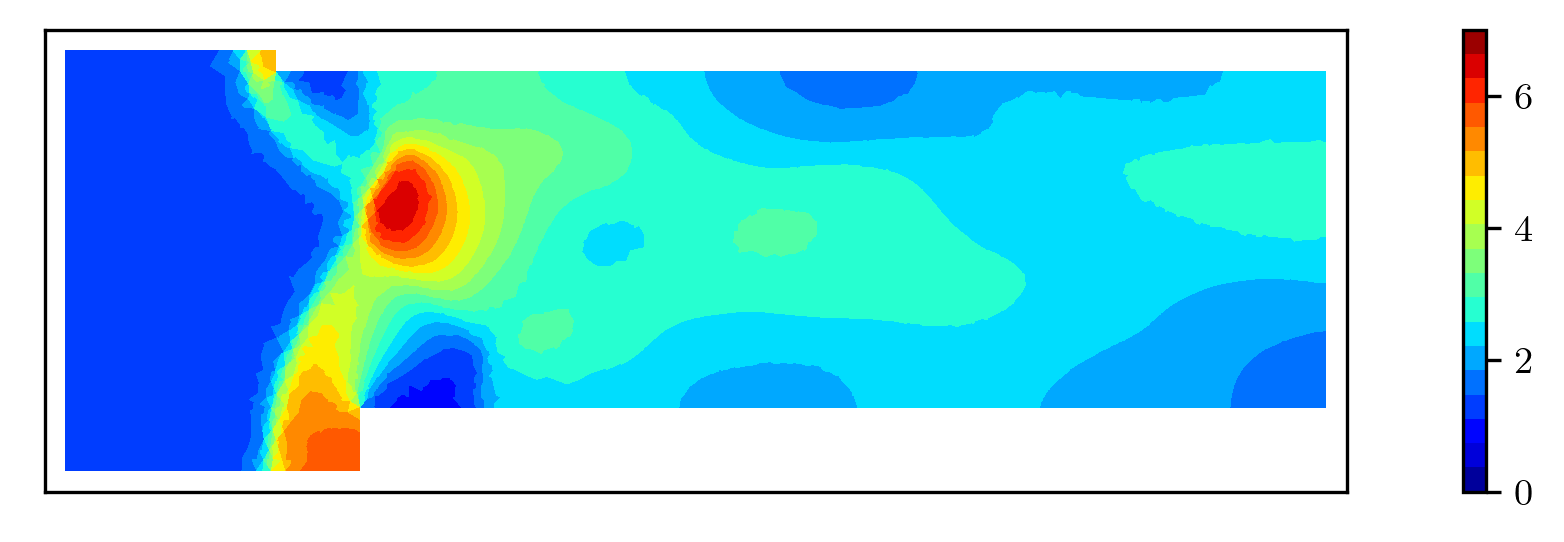}} &
            \raisebox{-0.5\height}{\includegraphics[width = .31\textwidth]{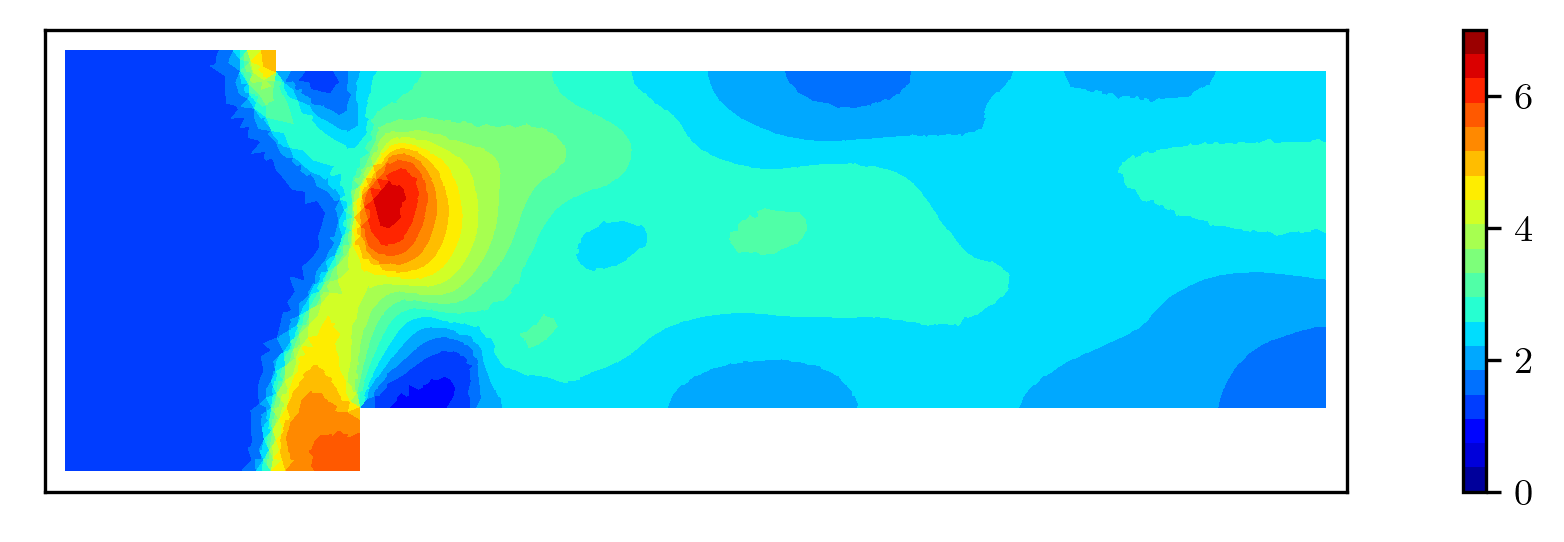}} &
            \raisebox{-0.5\height}{\includegraphics[width = .31\textwidth]{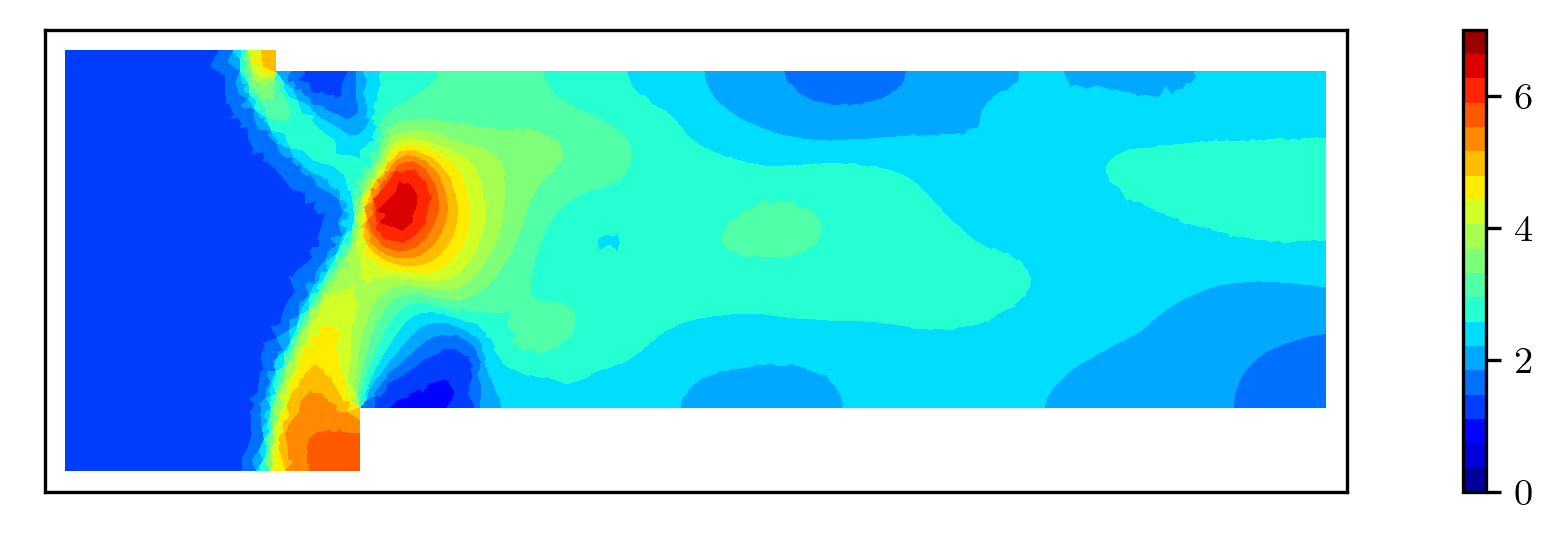}}
            \\
            &
            \rotatebox[origin=c]{90}{Error} &
            \raisebox{-0.5\height}{} &
            \raisebox{-0.5\height}{\includegraphics[width = .31\textwidth]{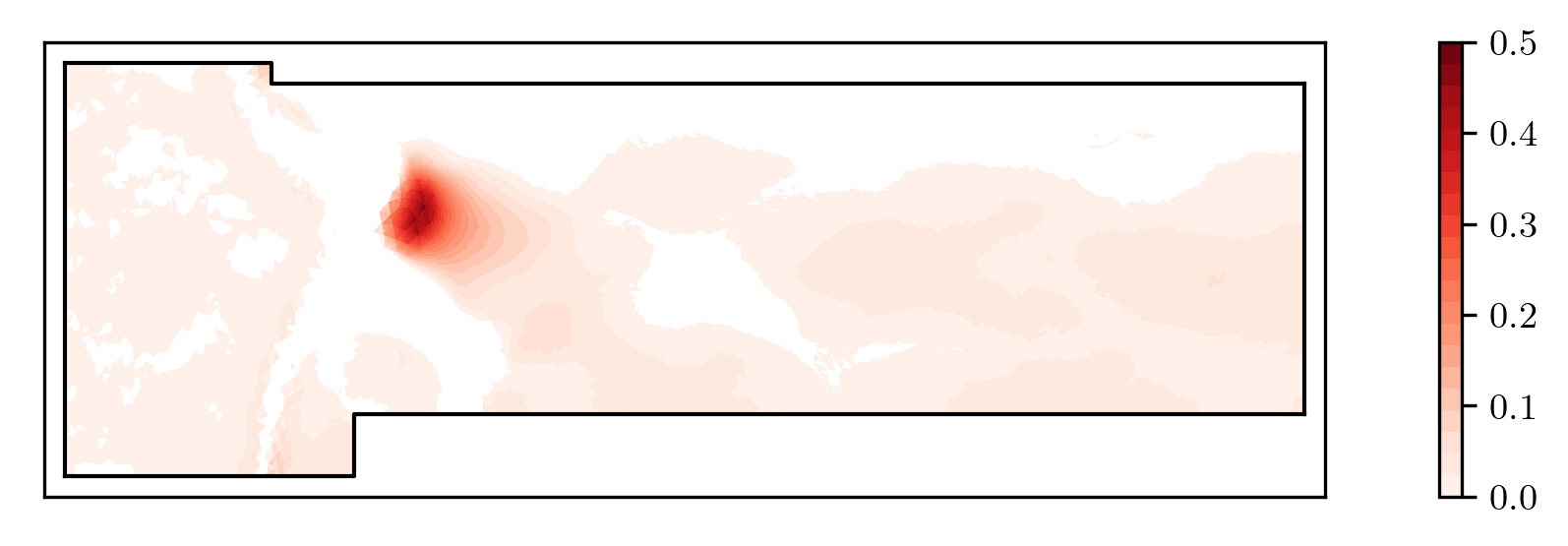}} &
            \raisebox{-0.5\height}{\includegraphics[width = .31\textwidth]{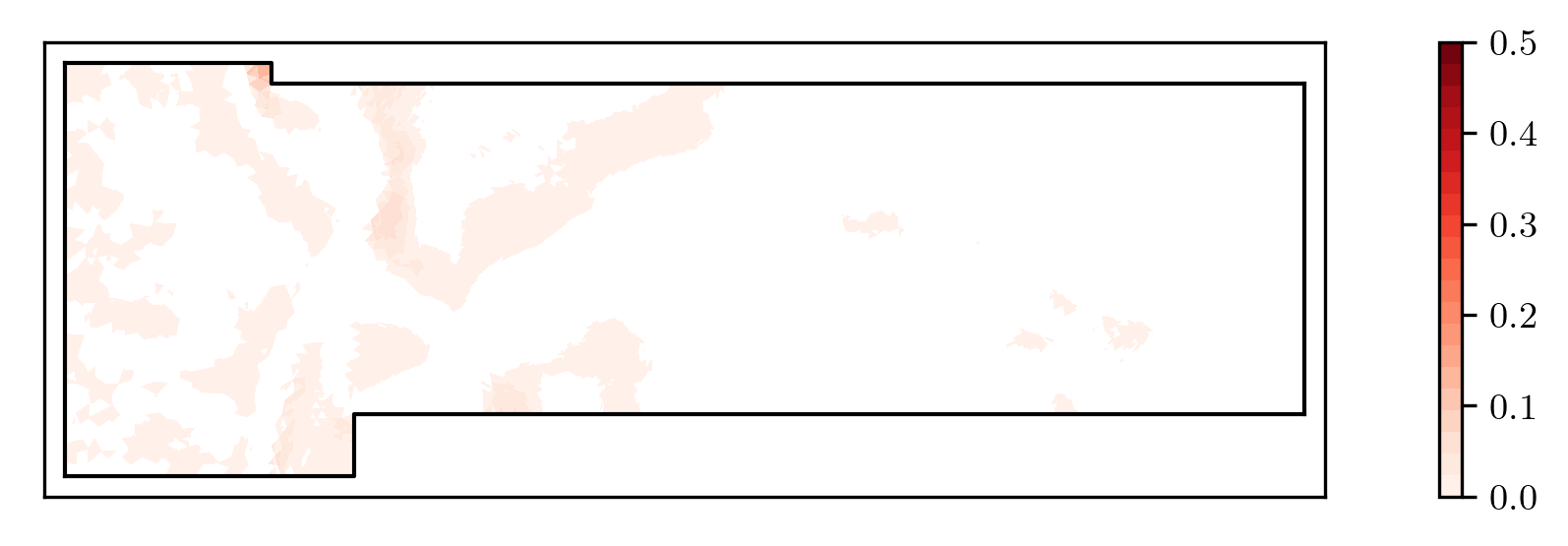}}
        \end{tabular*}
        \caption{{\bf 2D forward facing step:} predicted density field obtained by \nDGNet and \mcDGNet  and corresponding  pointwise error $\rho_{\text{DG}} - \rho_{\text{pred}}$  at large testing time $T_\text{test} = 4s$ and different gas constant of $\gamma = 1.4$.} 
        \figlab{prediction_solution_model1_mesh}
\end{figure}

In this problem, we consider the 2D Euler equation where a supersonic uniform flow with Mach number $M = 3$ approaches a forward-facing step \cite{hesthaven2007nodal}. The inflow boundary condition is the uniform flow, the outflow boundary condition is free, and the walls are modeled using the reflective boundary condition. The initial condition is defined as $\rho_0 = \gamma$, $\rho_0 u_0 = M \gamma$, $\rho_0 v_0 = 0$, $p_0 = 1$, and $E_0 = \frac{p_0}{\gamma - 1} + \frac{\rho_0}{2}\LRp{u_0^2 + v_0^2}$. Two different 
settings
%meshes \tanbui{Do you mean one mesh for one geometry? In other words, the right figure has less number of elements?} \hai{I meant one model = one mesh = one geometry. I did not create different meshes for the same geometry.} 
are presented in \cref{fig:forward_facing_mesh_grid_set}: Model 1 with $K = 7039$ elements and Model 2 with $K = 7008$ elements. For data generation, the training data is generated with the Model 1 by solving the 2D Euler equation with $\gamma = \LRc{1.2, 1.6}$ within the time interval $\LRs{0, 1}s$. The validation data is produced in the same settings except with a different gas constant of  $\gamma = 1.4$. The test data is generated for both Model 1 and Model 2 using $\gamma = 1.4$ with the larger time horizon $\LRs{0, 4}s$. For simplicity, we use a fixed time step size of $\dt = 0.001s$ for all data sets.

\nDGNet network is trained with clean data, while \mcDGNet is trained with training data corrupted by different noise levels in $\LRc{1,2,\hdots,16}\%$.
%to $16\%$ with a gap of $1\%$. 
As shown in \cref{fig:Relative_error_validation_data_model1_versus_noise}, \mcDGNet trained with $2\%$ noise level provides the lowest average relative $L^2$-error in the validation time interval $\LRs{0,1}s$. This noise level is considered as the ``optimal" value for this problem, and the robustness of the resulting \oDGNet will be studied against  Roe flux data in \cref{sect:forward_facing_conner_Roe_flux}, various random initialization for weights/biases in \cref{sect:Random_initializers_and_noise_seeds}, and data randomization in \cref{sect:P6_2D_Noise_corruption}.

The average relative $L^2$-errors of three conservative variables $\LRp{\rho, \rho u, E}$ between predicted \oDGNet solutions and DG solution on the test data with large time interval $\LRs{0,4}s$ are presented in \cref{fig:Relative_error_test_data_model1_model2_mesh}. 
\mcDGNet approach is superior over \nDGNet in both Model 1 and Model 2 throughout all time steps. Not only
is \mcDGNet error lower, but it is also almost constant over the whole testing time horizon. Again, the only difference between \mcDGNet and \nDGNet is the data randomization for the former.
The results in \cref{fig:Relative_error_test_data_model1_model2_mesh} verify the implicit regularization effect induced by randomization that is proved in \cref{theo:data_rand} (and consequently the error control in \cref{thm:mainTheo_computable}).
% It is thanks to the \mcDGNet approach regularizes the learning better than the \nDGNet approach via implicit regularization terms. 
From a data enrichment point of view, as discussed in \cref{sect:P6_2D_Noise_corruption}, \mcDGNet covers bigger data space via randomization. 
Since, compared to Model 1 that is used for training \nDGNet and \mcDGNet\hspace{-1ex}, Model 2 is a completely different geometry and mesh, the results in the right subfigure of  \cref{fig:Relative_error_test_data_model1_model2_mesh} also reveal that both \nDGNet and \mcDGNet is able to generalize well for this {\bf out-of-distribution} case. 

\cref{fig:prediction_solution_model1_mesh} shows that the \nDGNet struggles to capture shocks in the density field for both Model 1 and Model 2 (higher pointwise error $\rho_{\text{DG}} - \rho_{\text{pred}}$ relative to the DG method). In contrast, \mcDGNet provides a superior shock-capturing ability (smaller error in the shock locations relative to the DG method).
%, thus solutions are more accurate, especially at shock locations. 
We also note that both \nDGNet and \mcDGNet performance is comparable to the DG method approach in  regions with smooth solutions.

\noindent{\bf Implicit \mcDGNet\hspace{-1ex}.} We implement the Backward Euler scheme (see \cref{sect:1dsodtube}) for the DG method and the trained \mcDGNet neural network. 
%The Backward Euler algorithm is presented in \cref{sect:1dsodtube}. 
We take  $\dt = 0.0005s$ for the time step size, for which the 2nd-RK is unstable. The density solutions from  DG and \mcDGNet for Model 1 at $T_\text{test} = 4s$ are shown in \cref{fig:prediction_solution_model1_implicitt}. As can be seen, \mcDGNet with implicit method gives predictions as good as those obtained from DG counterpart. 

\begin{figure}[htb!]
    \centering
        \begin{tabular*}{\textwidth}{c c}
            \centering
            DG & \mcDGNet  ($2\%$)
            \\
            \raisebox{-0.5\height}{\includegraphics[width = .48\textwidth]{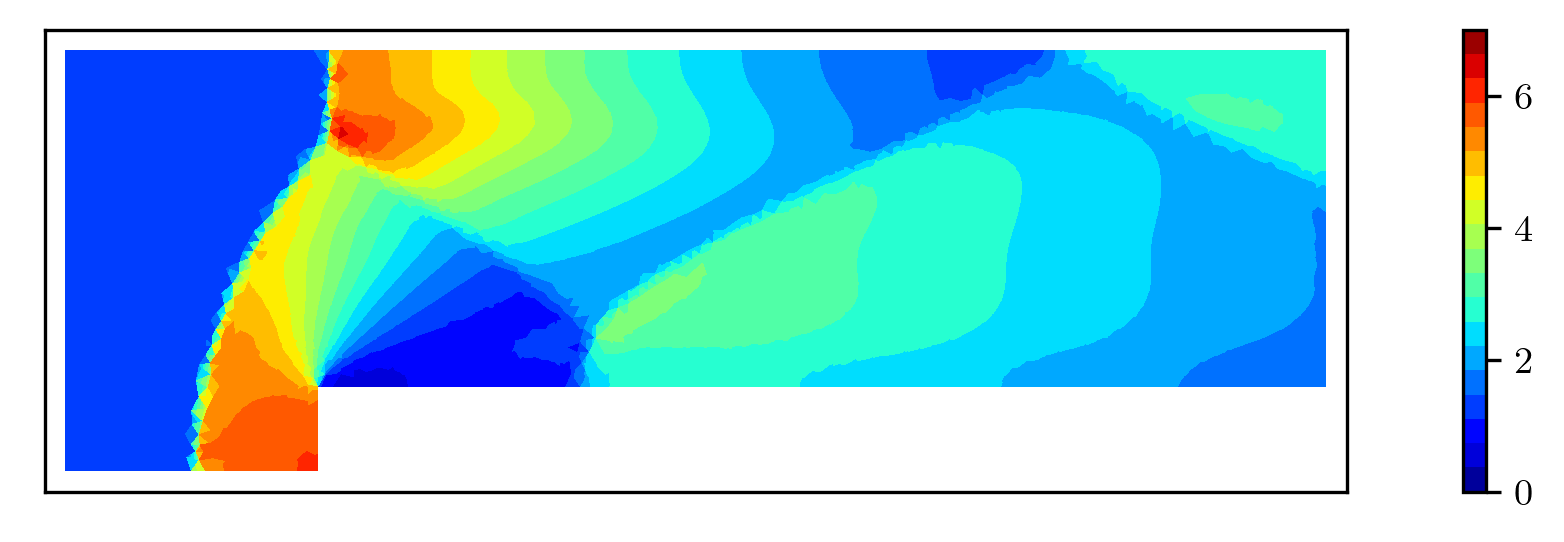}} &
            \raisebox{-0.5\height}{\includegraphics[width = .48\textwidth]{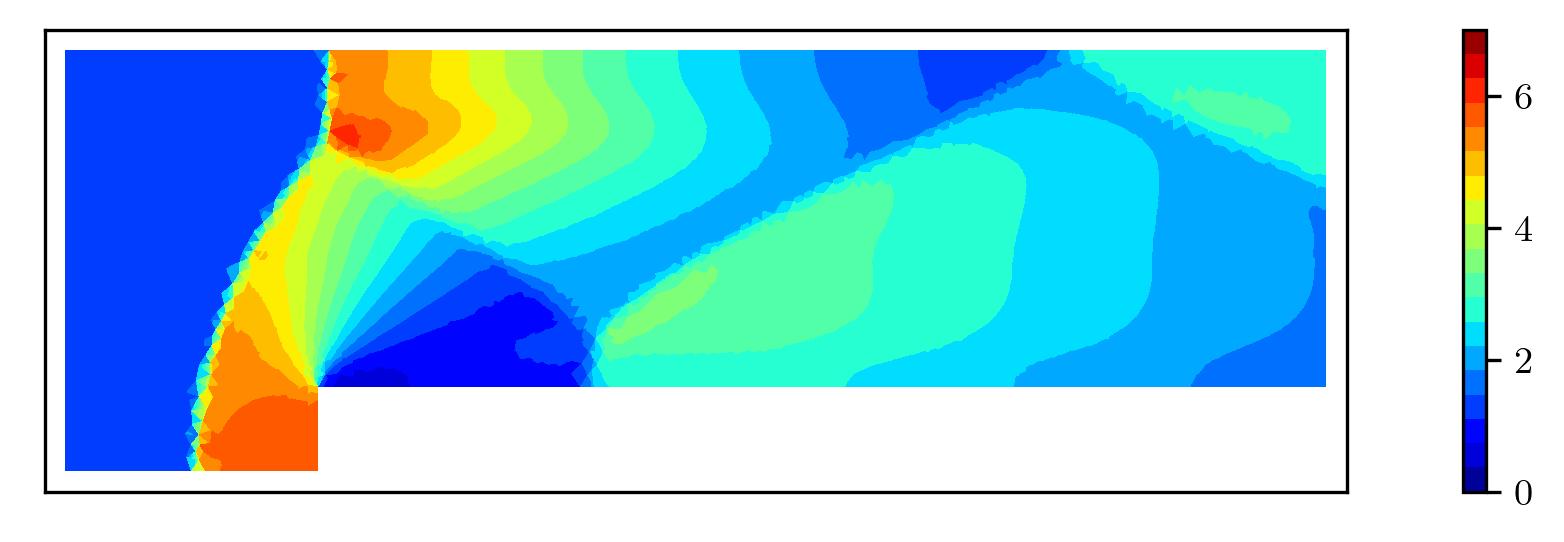}}
        \end{tabular*}
        \caption{{\bf 2D forward facing step:} Implicit solutions by DG approach and \mcDGNet  at time step $T_\text{test} = 4s$ with $\dt = 0.0005s$ in Model 1. The results are visibly indistinguishable.}
        \figlab{prediction_solution_model1_implicitt}
\end{figure}

%\clearpage

\subsection{Scramjet problem}
\seclab{sramjet}

\begin{figure}[htb!]
    \centering
        \begin{tabular*}{\textwidth}{c c}
            \centering
            Model 1 & Model 2
            \\
            \raisebox{-0.5\height}{\includegraphics[width = .48\textwidth]{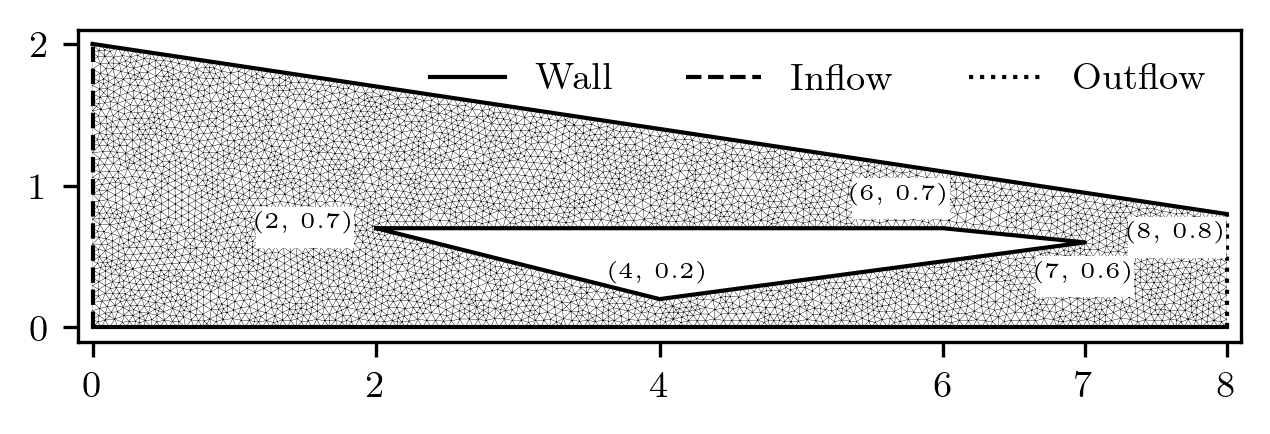}} &
            \raisebox{-0.5\height}{\includegraphics[width = .48\textwidth]{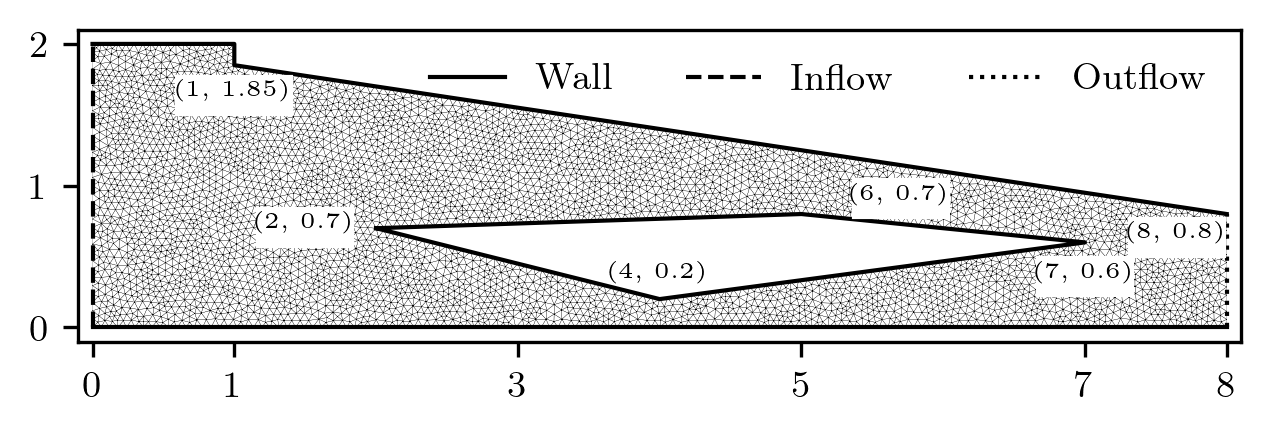}} 
        \end{tabular*}
        \caption{{\bf 2D Scramjet:} domain discretization for Model 1 (K = 9038 elements) and Model 2 (K = 8635 elements) mesh grids.}
        \figlab{scramjet_grid_set}
\end{figure}

\begin{figure}[htb!]
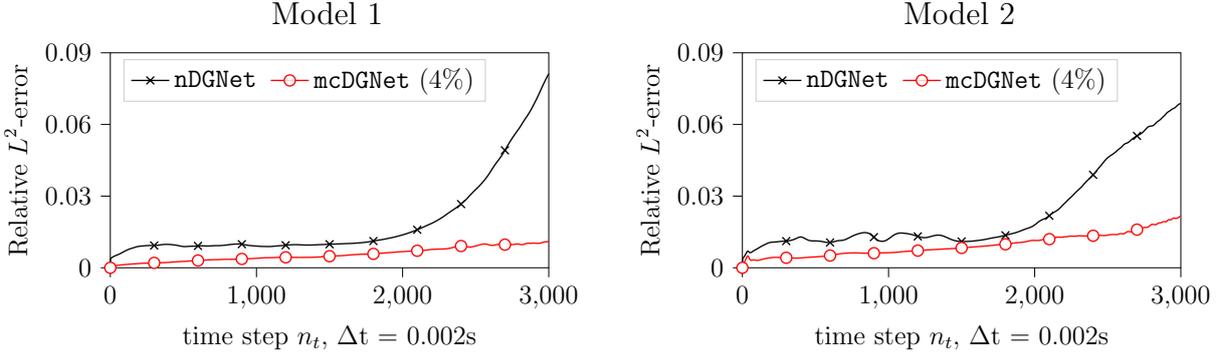

    \centering
        \begin{tabular*}{\textwidth}{c c}
            \centering
            \quad \quad Model 1 & \quad \quad Model 2
            \\
            \raisebox{-0.5\height}{\resizebox{0.48\textwidth}{!}{\input{Figs/2D_Scramjet/Relative_error_test_data_model1_mesh_modified.tex}}} &
            \raisebox{-0.5\height}{\resizebox{0.48\textwidth}{!}{\input{Figs/2D_Scramjet/Relative_error_test_data_model2_mesh_modified.tex}}} 
        \end{tabular*}
        \caption{{\bf 2D Scramjet:} Average relative $L^2$-error average over three conservative components $\LRp{\rho, \rho u, E}$ predictions obtained by \nDGNet and \mcDGNet approaches at different time steps for Model 1 ({\bf Left}) and Model 2 ({\bf Right}) mesh grids.}
        \figlab{scramjet_Relative_error_test_data_model1_model2_mesh}
\end{figure}

\begin{figure}[htb!]
    \centering
        \begin{tabular*}{\textwidth}{c c@{\hskip -0.0001cm} c@{\hskip -0.002cm} c@{\hskip -0.002cm} c@{\hskip -0.002cm}}
            \centering
            & & DG & \nDGNet & \mcDGNet  ($4\%$)
            \\
            \multirow{2}{*}{\rotatebox[origin=l]{90}{Model 1 \quad  }} &
            \rotatebox[origin=c]{90}{Pred} &
            \raisebox{-0.5\height}{\includegraphics[width = .31\textwidth]{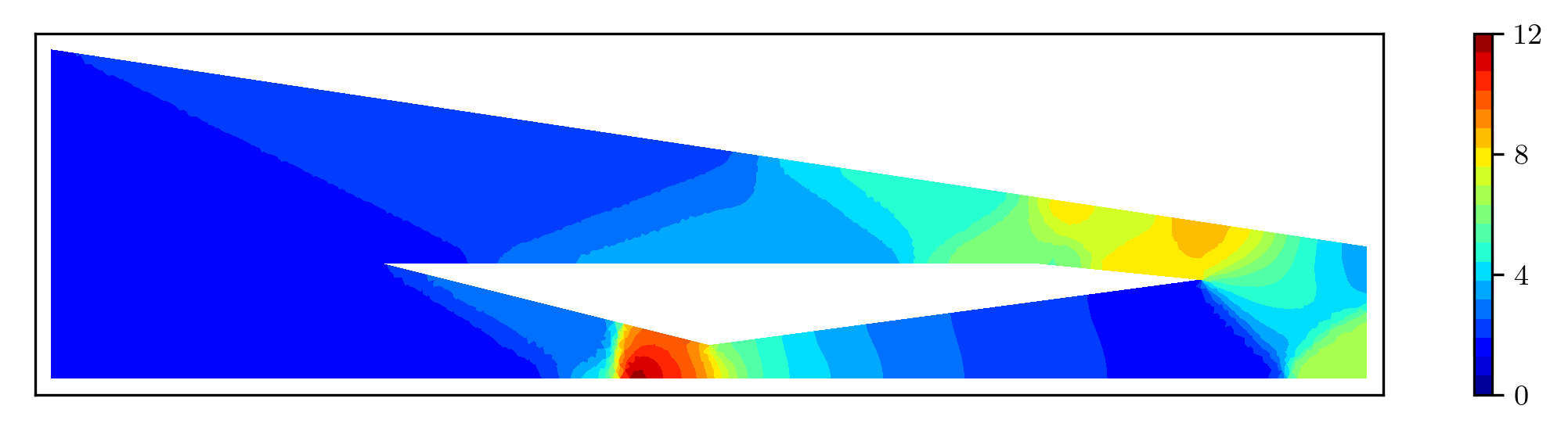}} &
            \raisebox{-0.5\height}{\includegraphics[width = .31\textwidth]{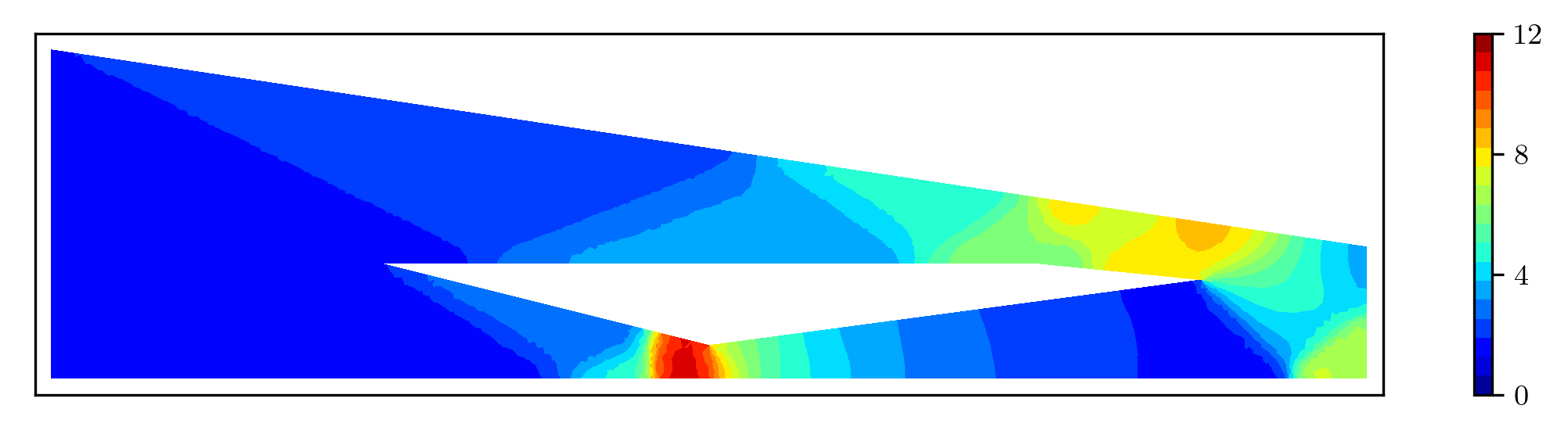}} &
            \raisebox{-0.5\height}{\includegraphics[width = .31\textwidth]{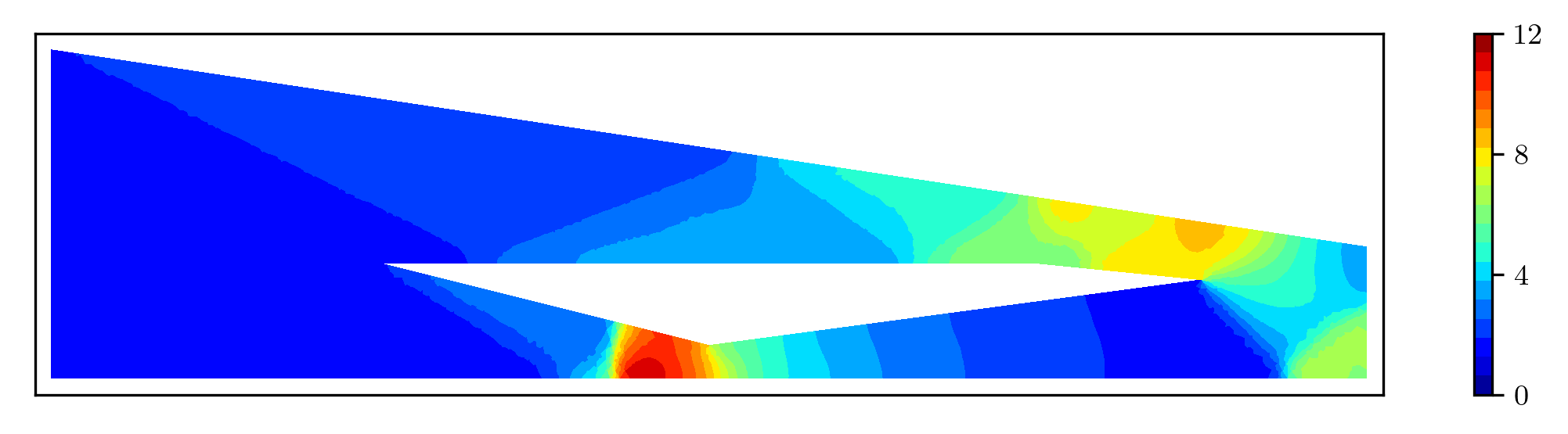}}
            \\
            &
            \rotatebox[origin=c]{90}{Error} &
            \raisebox{-0.5\height}{} &
            \raisebox{-0.5\height}{\includegraphics[width = .31\textwidth]{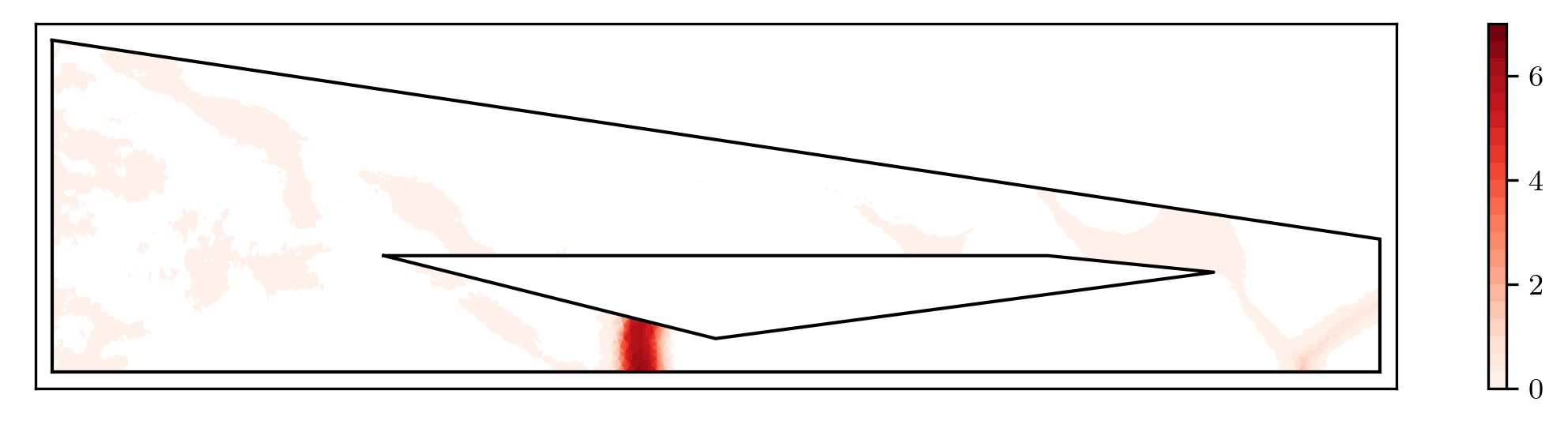}} &
            \raisebox{-0.5\height}{\includegraphics[width = .31\textwidth]{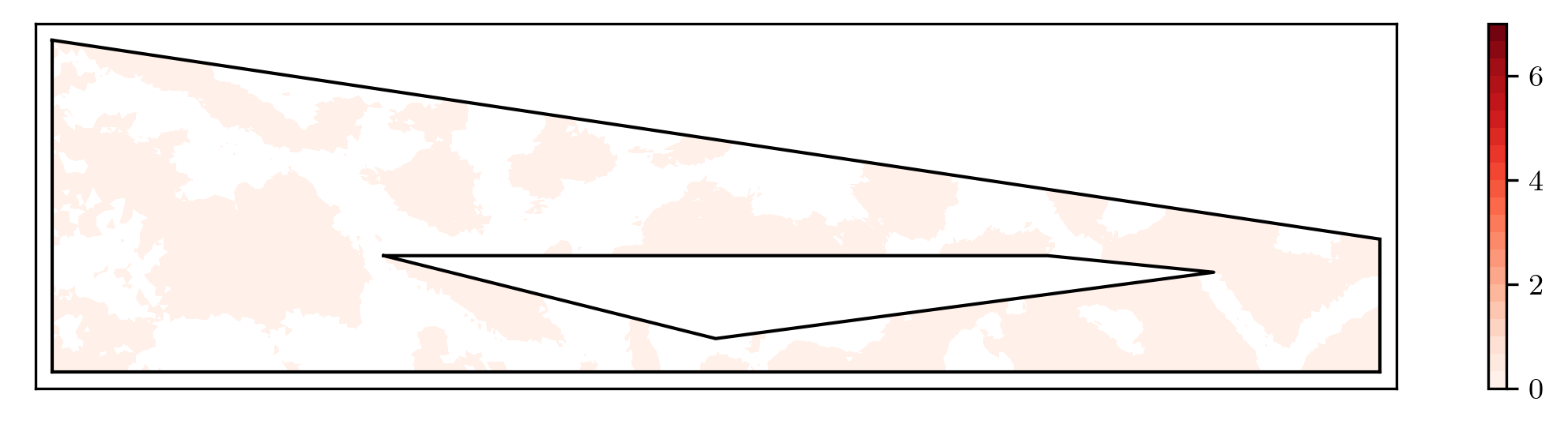}}
            \\
            \multirow{2}{*}{\rotatebox[origin=l]{90}{Model 2 \quad  }} &
            \rotatebox[origin=c]{90}{Pred} &
            \raisebox{-0.5\height}{\includegraphics[width = .31\textwidth]{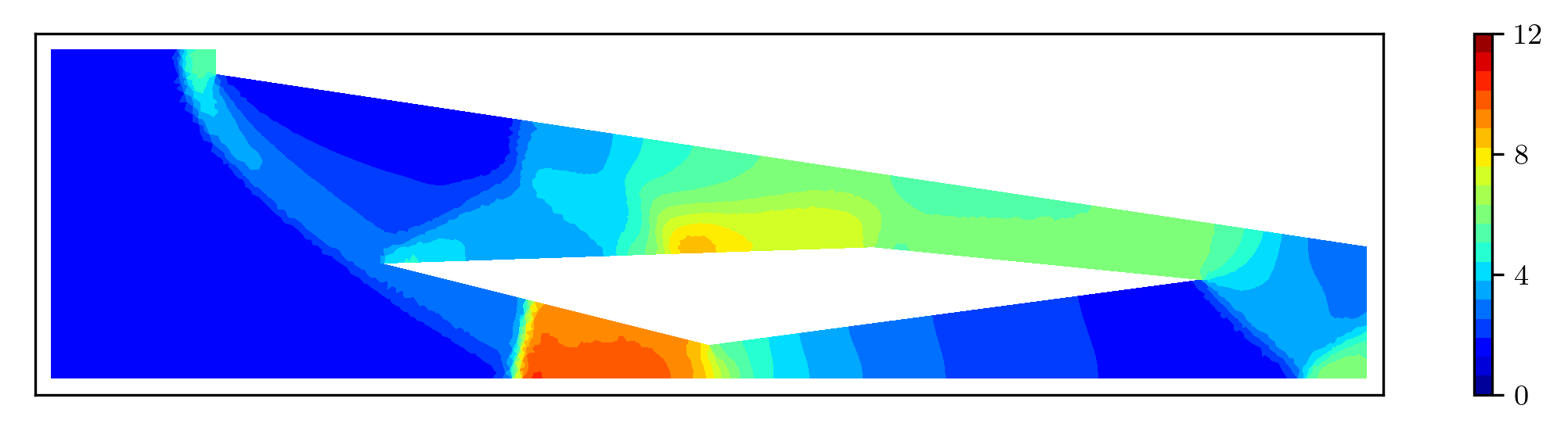}} &
            \raisebox{-0.5\height}{\includegraphics[width = .31\textwidth]{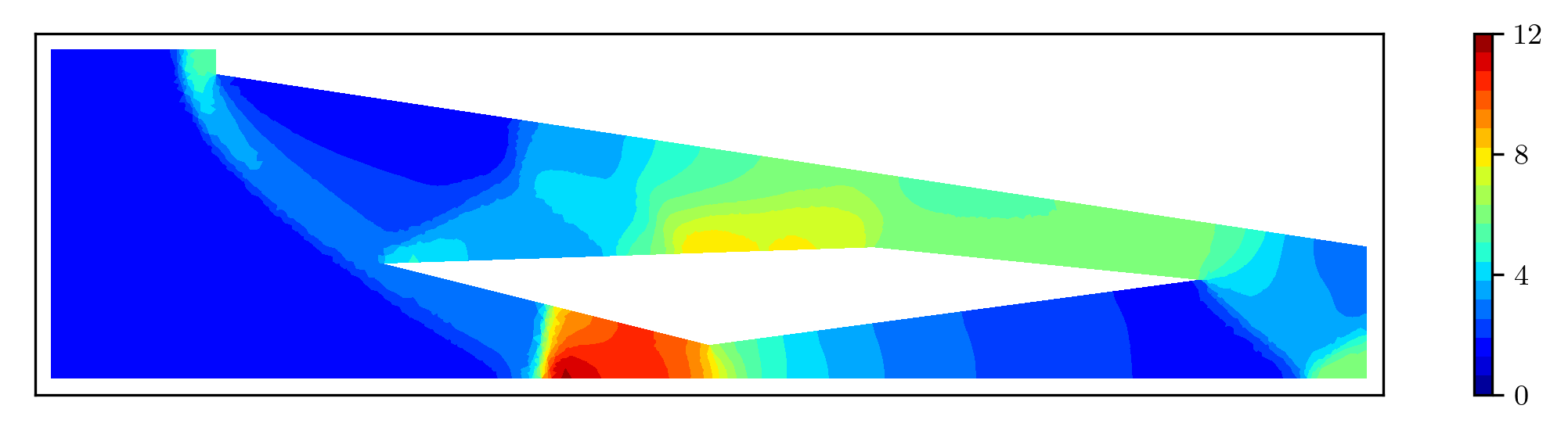}} &
            \raisebox{-0.5\height}{\includegraphics[width = .31\textwidth]{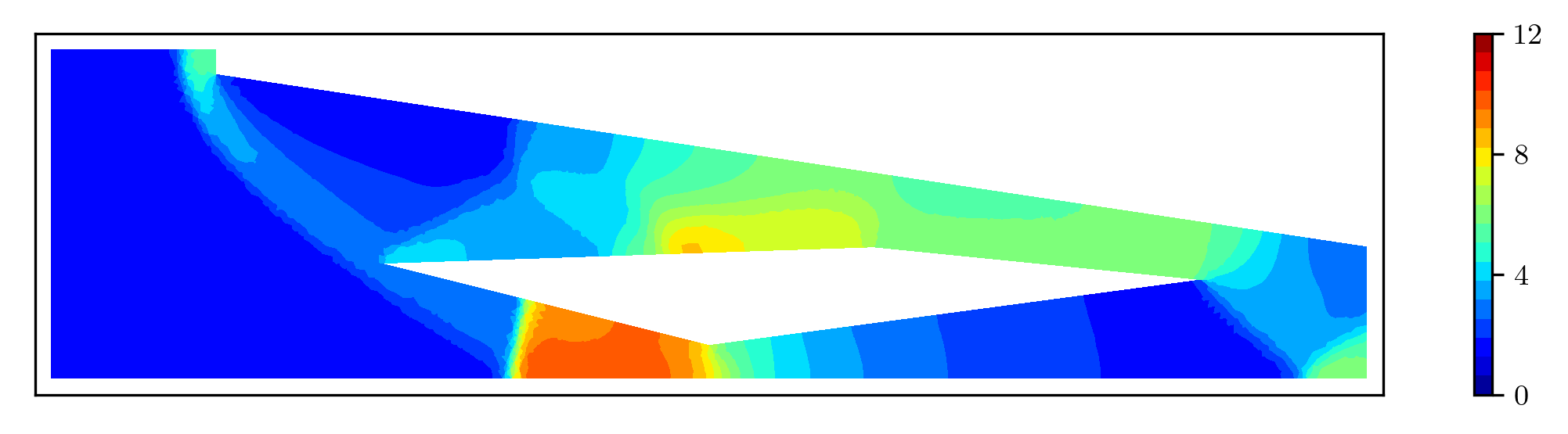}}
            \\
            &
            \rotatebox[origin=c]{90}{Error} &
            \raisebox{-0.5\height}{} &
            \raisebox{-0.5\height}{\includegraphics[width = .31\textwidth]{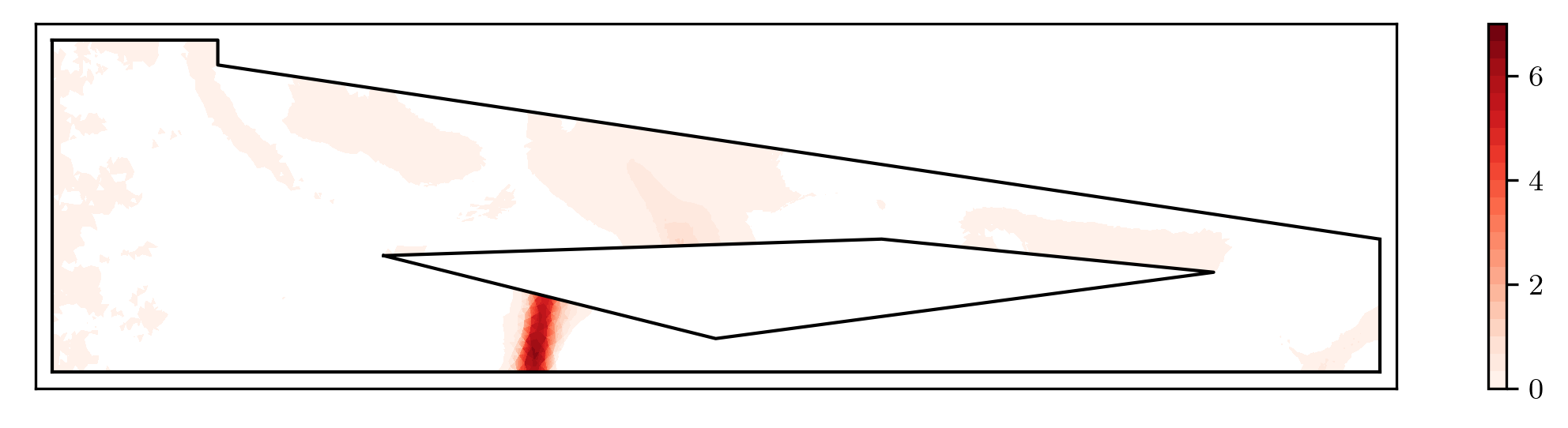}} &
            \raisebox{-0.5\height}{\includegraphics[width = .31\textwidth]{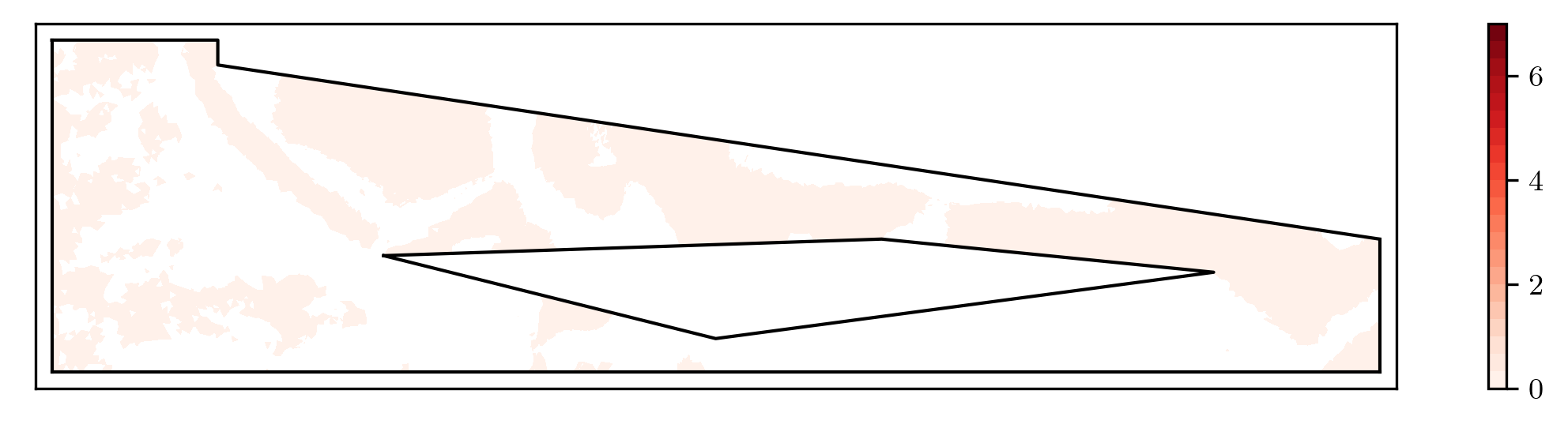}}
        \end{tabular*}
        \caption{{\bf 2D Scramjet:} predicted density field obtained by \nDGNet and \mcDGNet approaches and corresponding prediction pointwise error $\rho_{\text{DG}} - \rho_{\text{pred}}$ at final time step $T_\text{test} = 6s$ for both Model 1 and Model 2. As can be seen,  \mcDGNet is comparable to DG, while \nDGNet has large errors in the shock region.} 
        \figlab{scramjet_prediction_solution_model1_mesh}
\end{figure}

We consider the standard scramjet engine example with a superonic upstream flow with Mach number $M = 3$ \cite{hesthaven2007nodal}. The boundary conditions are as follows: uniform flow for the inlet,  free for the outlet, and the reflective boundary condition for the wall. The initial condition is defined as $\rho_0 = \gamma$, $\rho_0 u_0 = M \gamma$, $\rho_0 v_0 = 0$, $p_0 = 1$, and $E_0 = \frac{p_0}{\gamma - 1} + \frac{\rho_0}{2}\LRp{u_0^2 + v_0^2}$. We investigate the accuracy and the generalization of the \oDGNet approach on two different settings in \cref{fig:scramjet_grid_set}: Model 1 with $K = 9038$ elements and Model 2 with $K = 8635$ elements. For data generation, we solve the 2D Euler equation with $\gamma \in \LRc{1.2,1.6}$ for training data and with $\gamma = 1.4$ for validation data over the time interval $\LRs{0, 1.6}$ using Model 1. The test data is produced with $\gamma = 1.4$ and larger time interval of $\LRs{0, 6}s$ for both Model 1 and Model 2. Note that Model 2 setting is a complete out-of-distribution. A uniform time stepsize of $\Delta t = 0.002s$ is used  for generating all data sets.

The \nDGNet approach is trained with clean training data, while the \mcDGNet approach is trained with noisy training data with $4\%$ noise. \cref{fig:scramjet_Relative_error_test_data_model1_model2_mesh} presents the  average relative $L^2$-error (against the corresponding DG solutions) on the test data for three conservative variables $\LRp{\rho, \rho u, E}$ from \nDGNet\hspace{-1ex} and \mcDGNet\hspace{-1ex} approaches over the test time interval $\LRs{0,6}s$. As can be seen, \mcDGNet is much more accurate than \nDGNet{\hspace{-1ex}}, especially close to the end of the time interval where the shock profile becomes stiffer. This observation reveals the benefits of implicit regularization terms induced in \mcDGNet framework and also the data enrichment from data randomization technique as discussed in \cref{sect:P6_2D_Noise_corruption}. For the generalization capability, it can be seen that both \nDGNet and \mcDGNet approaches are capable of generalizing to solve the problem in Model 2 (with completely different geometry and mesh), despite both networks being trained using data generated from Model 1. However, the \nDGNet approach exhibits larger error in the snapshot density field at the final time step $T_\text{test} = 6$s than the \mcDGNet{\hspace{-1ex}}, as presented in \cref{fig:scramjet_prediction_solution_model1_mesh}. To be more specific, \nDGNet solutions present higher prediction pointwise error $\rho_{\text{DG}} - \rho_{\text{pred}}$ than those obtained by the \mcDGNet approach at shock locations for both Model 1 and Model 22.

% \clearpage

\subsection{Airfoil problem}
\seclab{airfoil}

\begin{figure}[htb!]
    \centering
    \includegraphics[width = .5\textwidth]{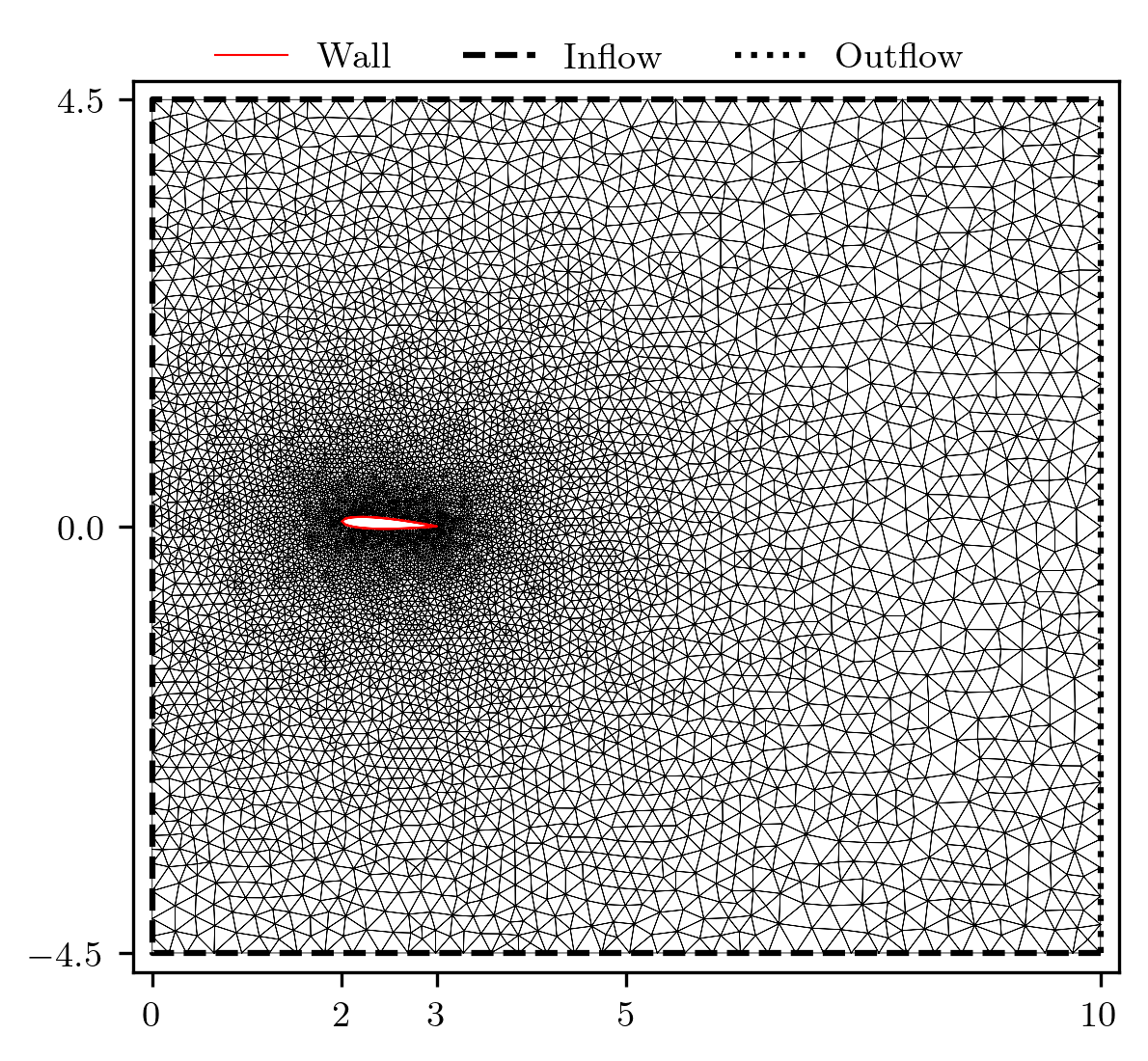}
    \caption{{\bf 2D Airfoil:} Domain, mesh, and boundary conditions for airfoil problems for Model 1: AOA = $3^o$, Mach = 0.8, K = 13400 elements.} 
    \figlab{airfoil_configuration}
\end{figure}

\begin{figure}[htb!]
    \centering
        \begin{tabular*}{\textwidth}{c@{\hskip -0.0cm} c@{\hskip -0.0cm}}
            \centering
            \quad \quad Model 1 & \quad \quad Model 2
            \\
            \raisebox{-0.5\height}{\resizebox{0.48\textwidth}{!}{\input{Figs/2D_Airfoil/Relative_error_test_data_M08_AOA3_modified.tex}}} &
            \raisebox{-0.5\height}{\resizebox{0.48\textwidth}{!}{\input{Figs/2D_Airfoil/Relative_error_test_data_M1p2_AOA5_modified.tex}}} 
        \end{tabular*}
        \caption{{\bf 2D Airfoil:} Average eelative $L^2$-error over three conservative components $\LRp{\rho, \rho u, E}$ for test data obtained by \nDGNet and \mcDGNet approaches at different time steps for Model 1 ({\bf Left}) and Model 2 ({\bf Right})..}
        \figlab{airfoil_Relative_error_test_data_model1_model2_mesh}
\end{figure}

\begin{figure}[htb!]
    \centering
        \begin{tabular*}{\textwidth}{c@{\hskip 0.01cm} c@{\hskip -0.1cm} c@{\hskip -0.1cm} c@{\hskip -0.1cm} c@{\hskip -0.1cm} c@{\hskip -0.1cm}}
            \centering
            & \multirow{2}{*}{DG \quad \quad}  &  \multicolumn{2}{c}{\mcDGNet ($4\%$)}  & \multicolumn{2}{c}{\nDGNet}
            \\
            &  & Pred \quad \quad  & Error \quad \quad & Pred \quad \quad & Error \quad \quad
            \\
            \rotatebox[origin=c]{90}{$T = 1.2$} &
            \raisebox{-0.5\height}{\includegraphics[width = .20\textwidth]{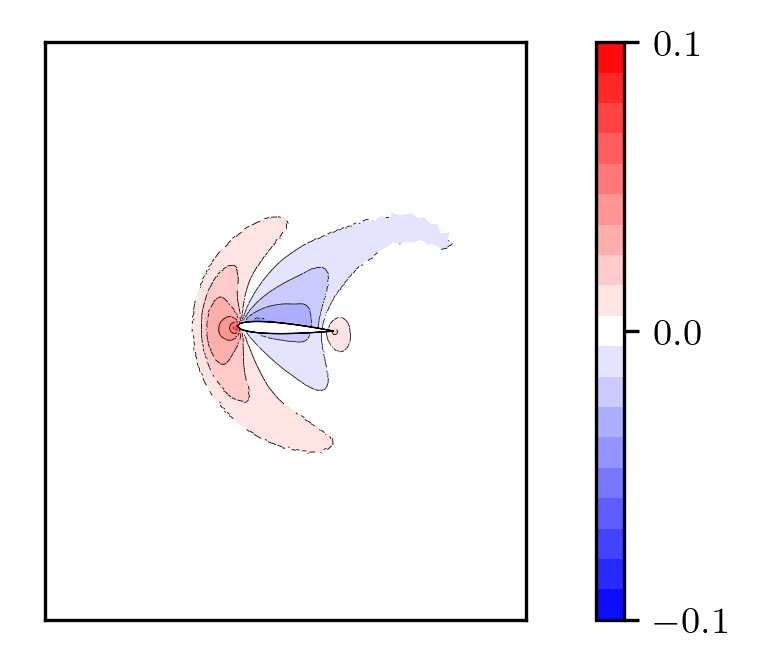}} &
            \raisebox{-0.5\height}{\includegraphics[width = .20\textwidth]{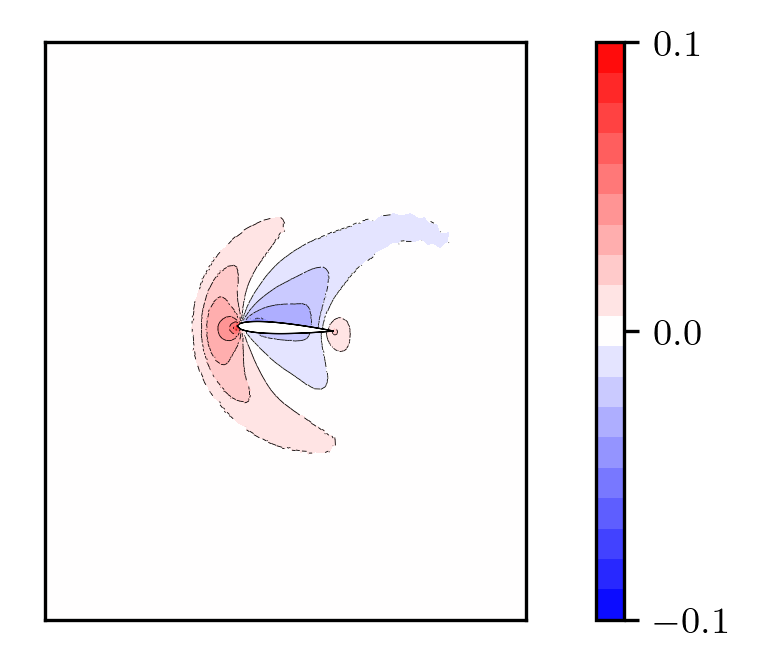}} &
            \raisebox{-0.5\height}{\includegraphics[width = .20\textwidth]{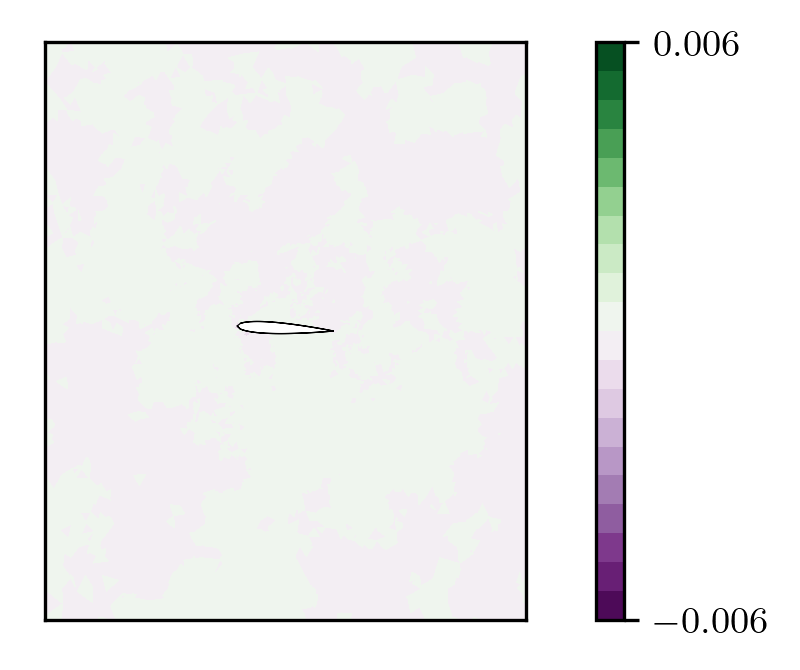}} &
            \raisebox{-0.5\height}{\includegraphics[width = .20\textwidth]{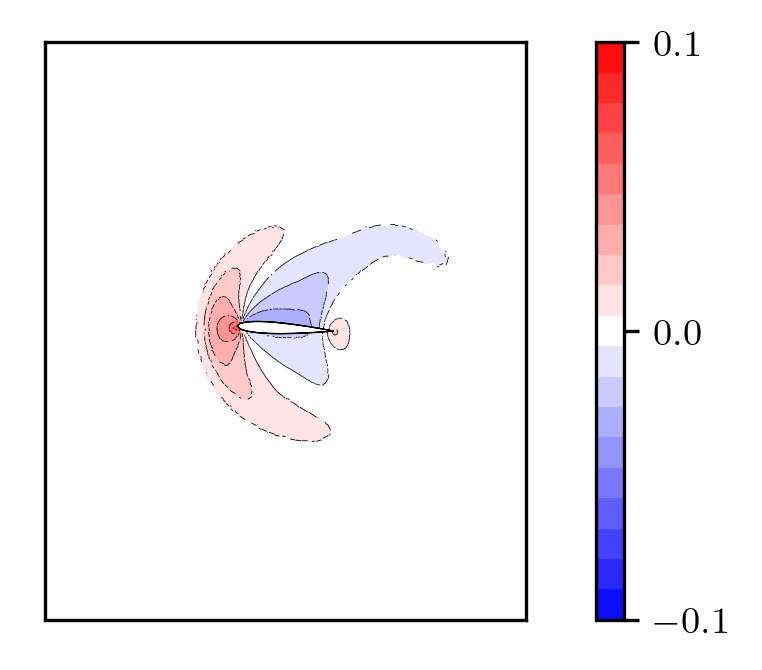}} &
            \raisebox{-0.5\height}{\includegraphics[width = .20\textwidth]{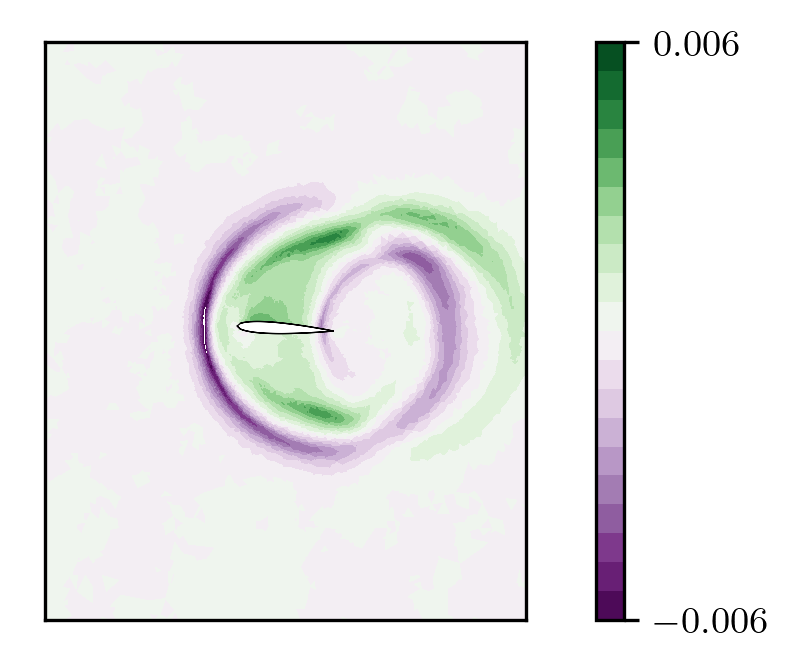}} 
            \\
            \rotatebox[origin=c]{90}{$T = 3$} &
            \raisebox{-0.5\height}{\includegraphics[width = .20\textwidth]{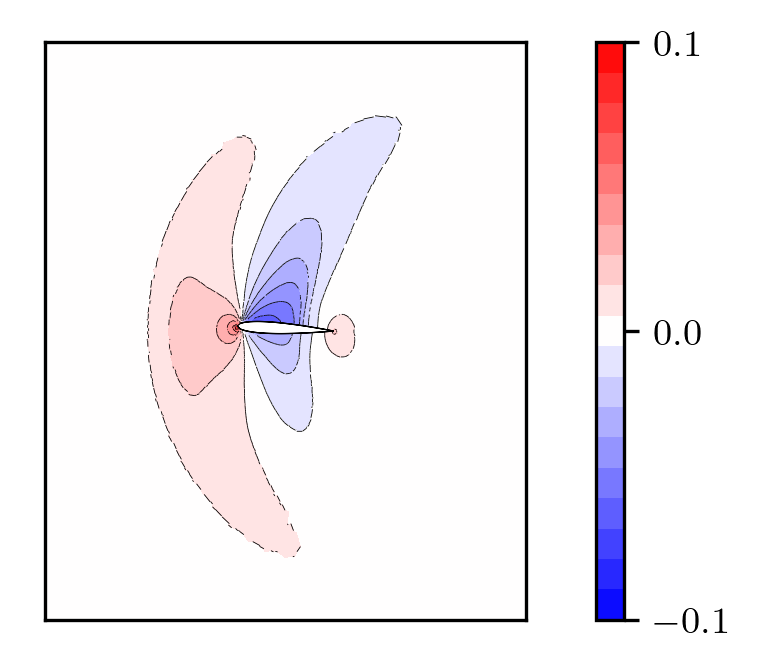}} &
            \raisebox{-0.5\height}{\includegraphics[width = .20\textwidth]{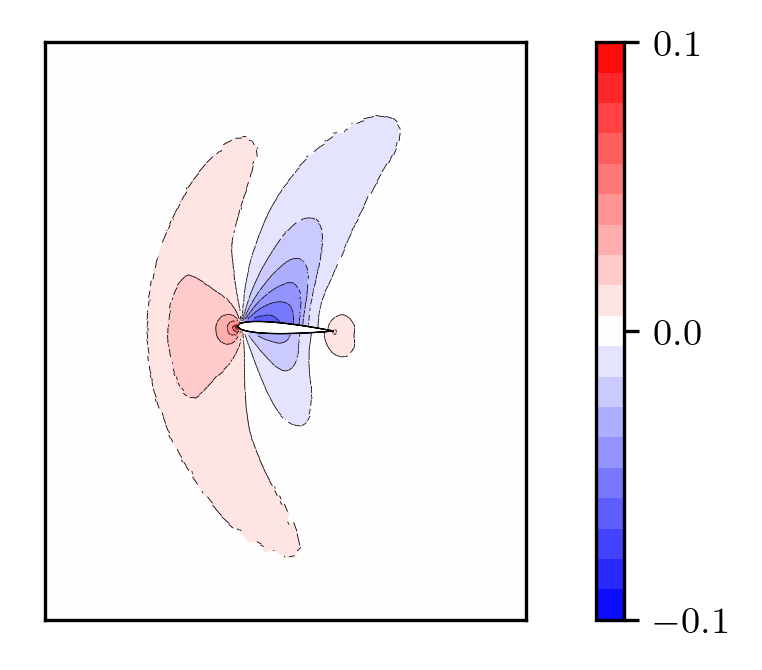}} &
            \raisebox{-0.5\height}{\includegraphics[width = .20\textwidth]{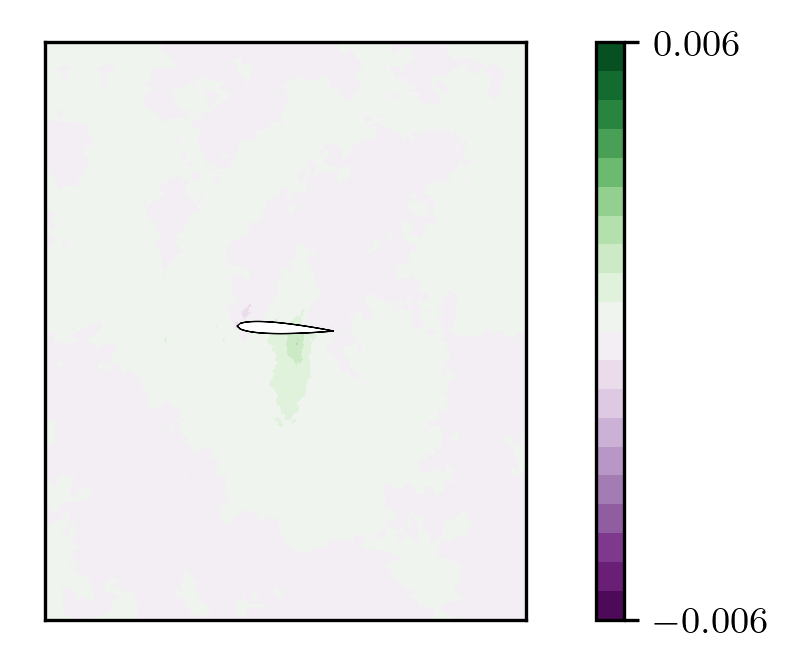}} &
            \raisebox{-0.5\height}{\includegraphics[width = .20\textwidth]{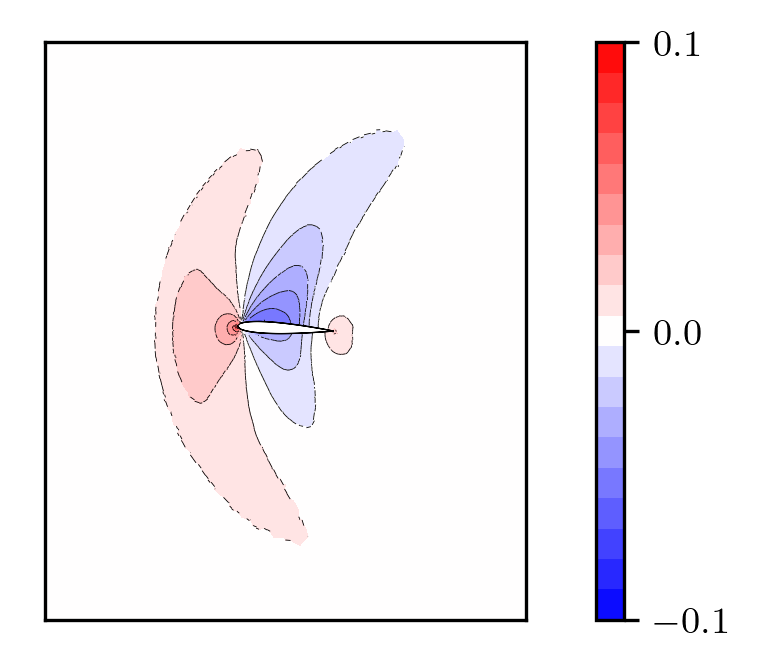}} &
            \raisebox{-0.5\height}{\includegraphics[width = .20\textwidth]{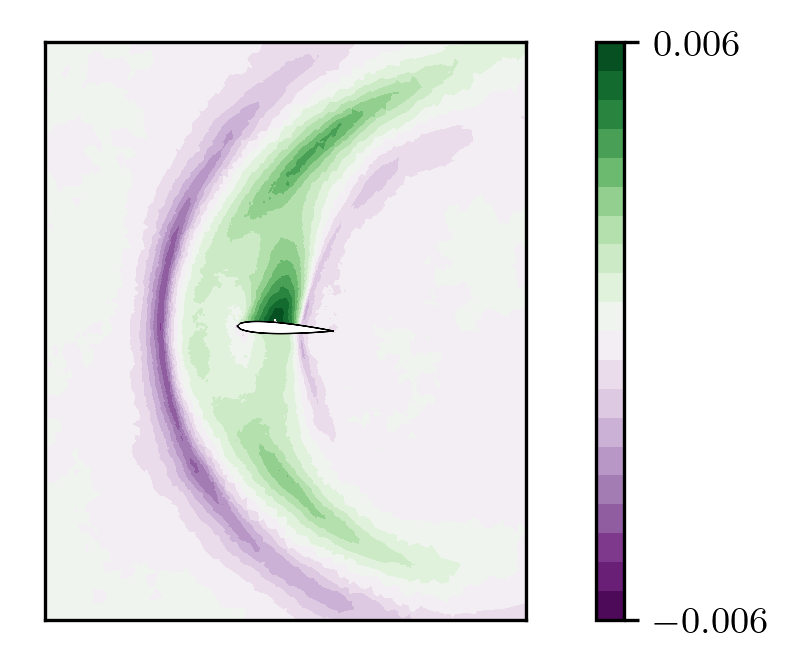}}
            \\
            \rotatebox[origin=c]{90}{$T = 5.25$} &
            \raisebox{-0.5\height}{\includegraphics[width = .20\textwidth]{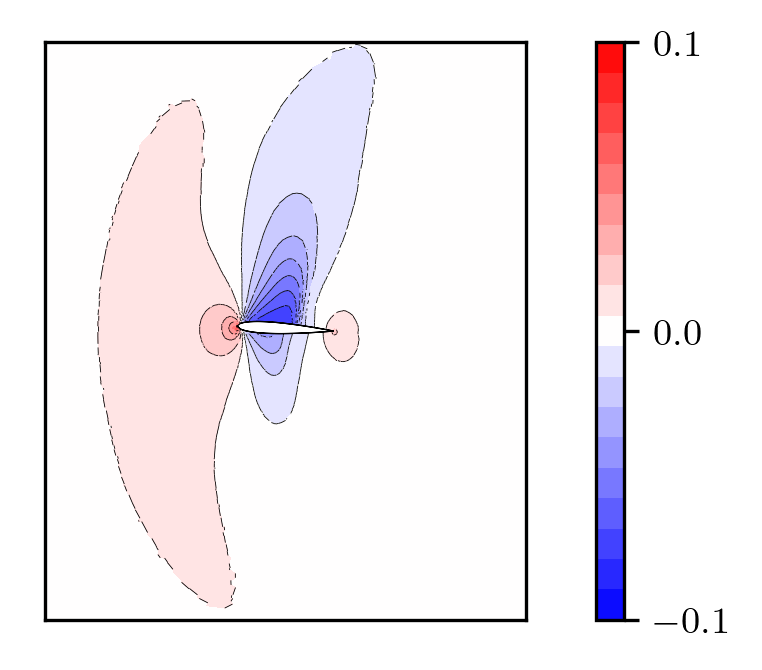}} &
            \raisebox{-0.5\height}{\includegraphics[width = .20\textwidth]{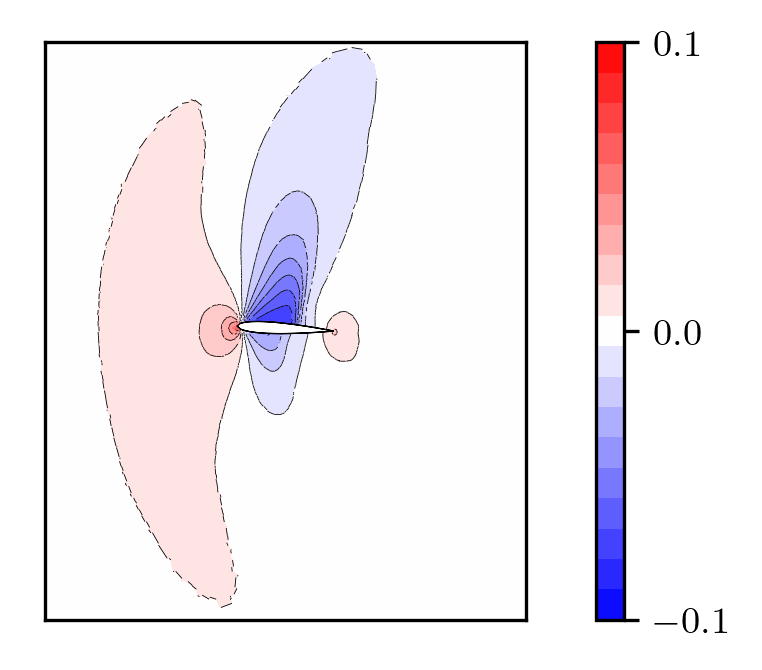}} &
            \raisebox{-0.5\height}{\includegraphics[width = .20\textwidth]{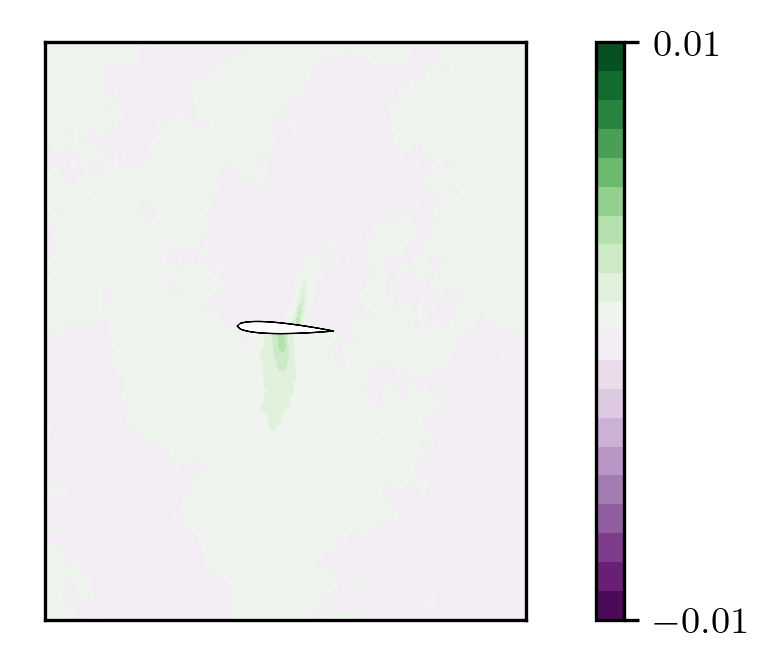}} &
            \raisebox{-0.5\height}{\includegraphics[width = .20\textwidth]{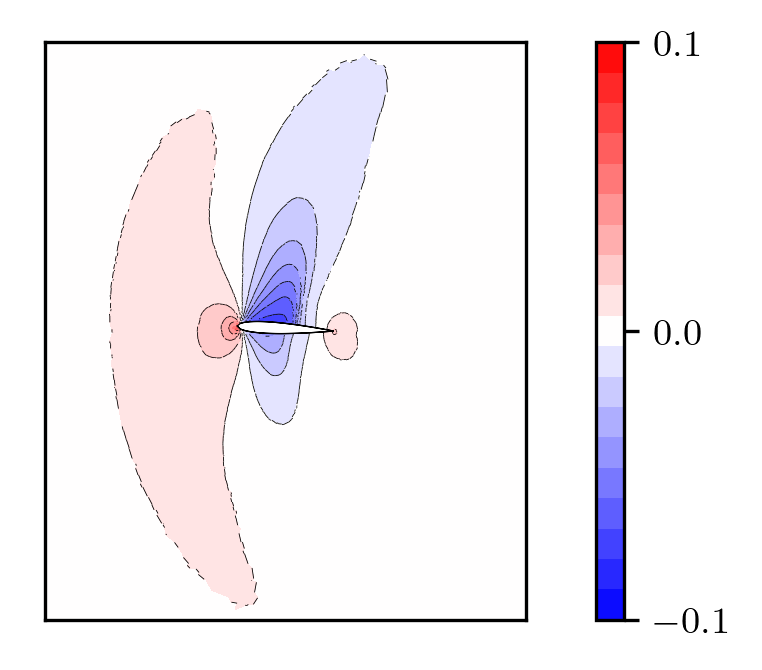}} &
            \raisebox{-0.5\height}{\includegraphics[width = .20\textwidth]{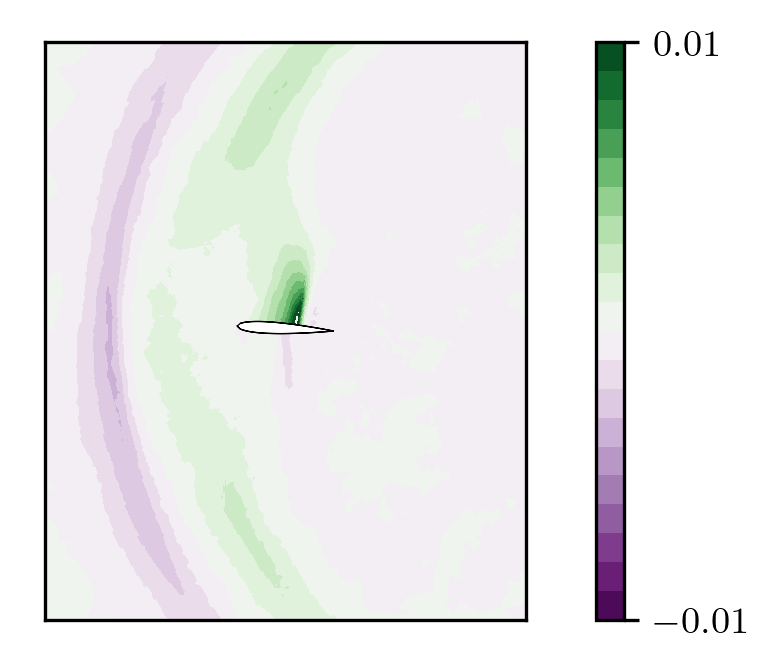}}
            \\
            \rotatebox[origin=c]{90}{$T = 7.5$} &
            \raisebox{-0.5\height}{\includegraphics[width = .20\textwidth]{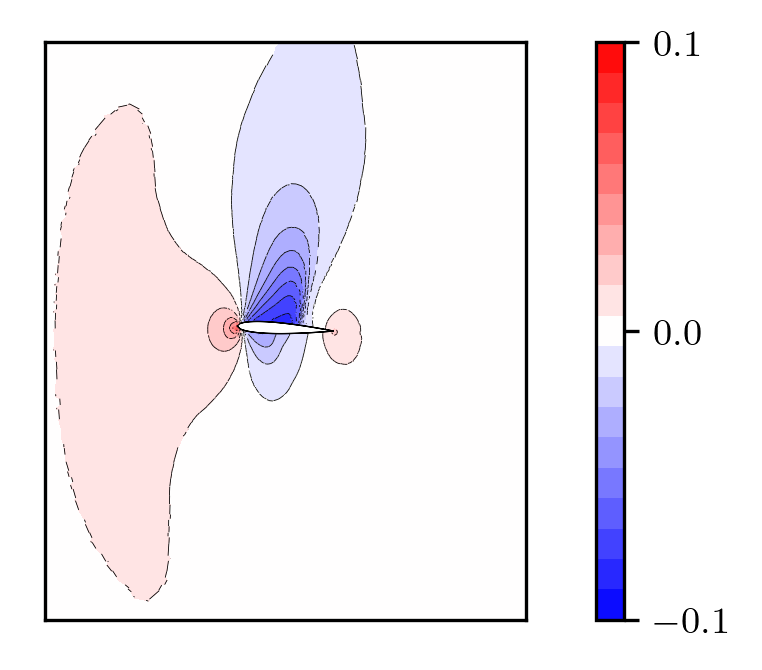}} & 
            \raisebox{-0.5\height}{\includegraphics[width = .20\textwidth]{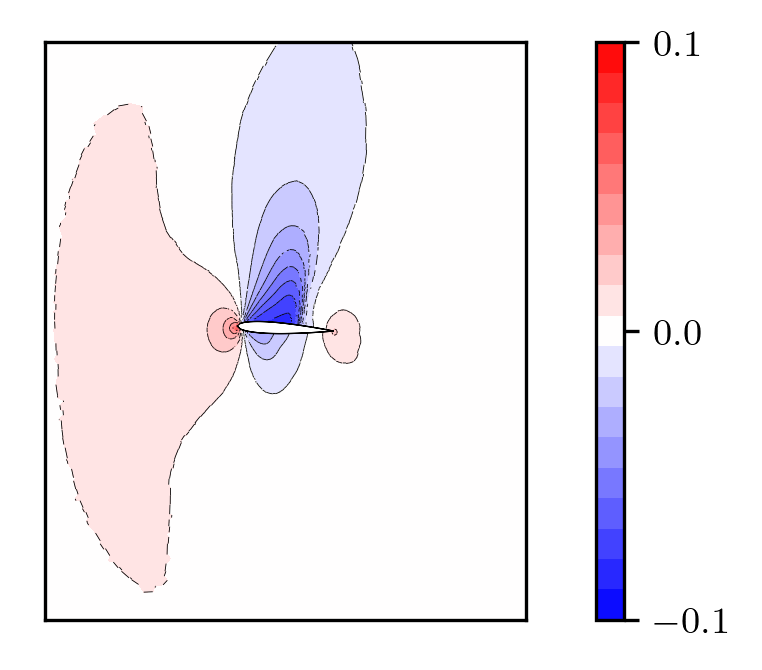}} & 
            \raisebox{-0.5\height}{\includegraphics[width = .20\textwidth]{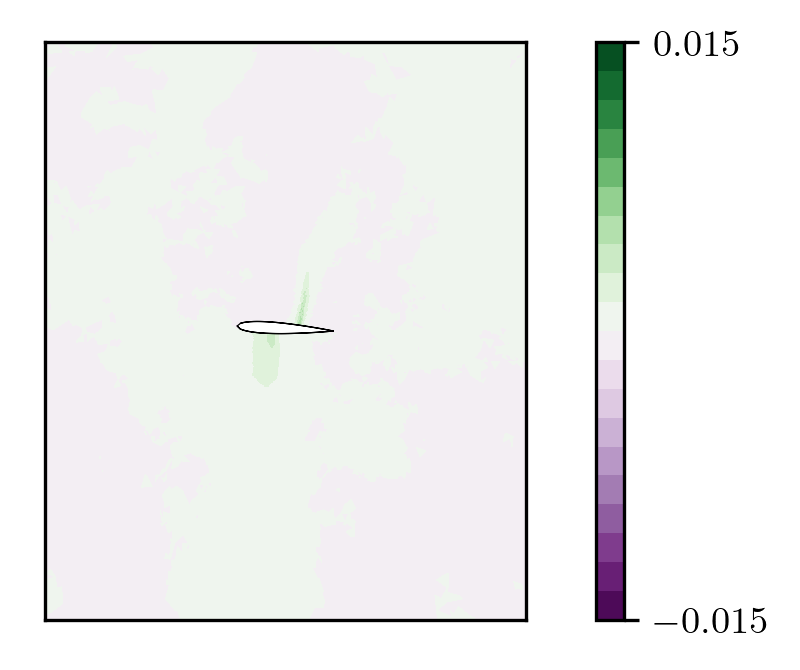}} & 
            \raisebox{-0.5\height}{\includegraphics[width = .20\textwidth]{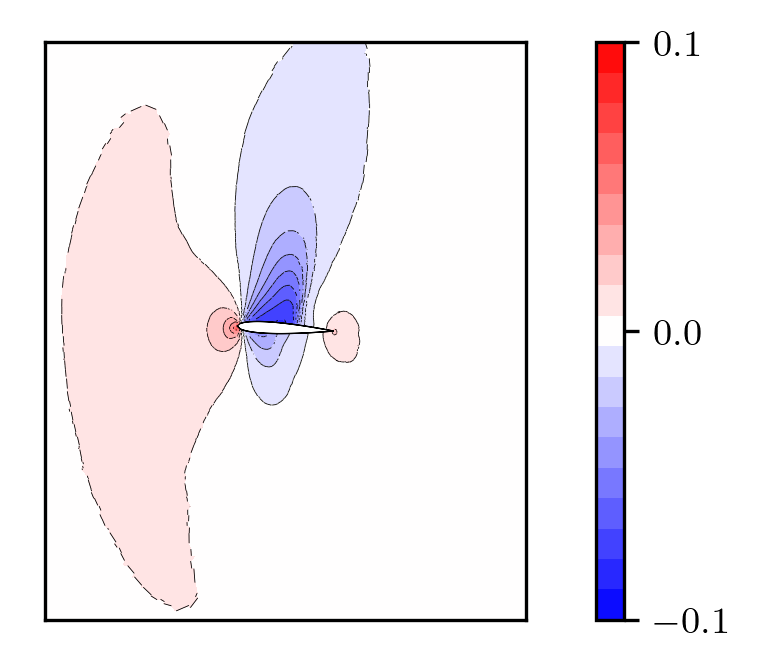}} &
            \raisebox{-0.5\height}{\includegraphics[width = .20\textwidth]{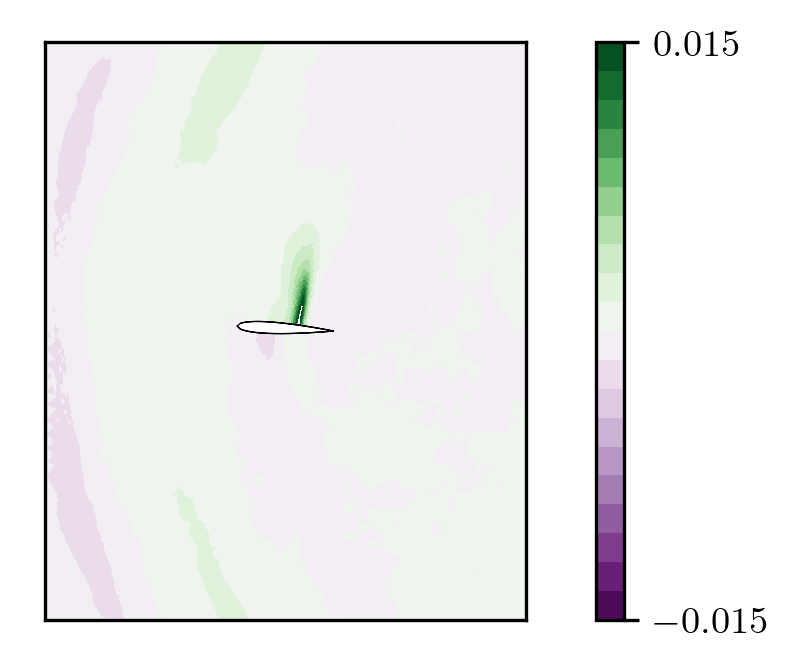}}
        \end{tabular*}
        \caption{{\bf 2D Airfoil:} predicted pressure coefficient field obtained by \nDGNet and \mcDGNet approaches and corresponding prediction pointwise error $C_{p,\text{DG}} - C_{p,\text{pred}}$ at time $T \in \LRc{1.2, 3, 5.25, 7.5}s$ for the case of Mach $M = 0.8$ and AOA = $3^o$ in Model 1. Plots are cropped over the domain $\LRs{-3, 3} \times \LRs{0,5}$ for zoomed-in views.} 
        \figlab{airfoil_prediction_solution_model1_mesh}
\end{figure}

\begin{figure}[htb!]
    \centering
        \begin{tabular*}{\textwidth}{c@{\hskip 0.01cm} c@{\hskip -0.1cm} c@{\hskip -0.1cm} c@{\hskip -0.1cm} c@{\hskip -0.1cm} c@{\hskip -0.1cm}}
            \centering
            & \multirow{2}{*}{DG \quad \quad} & \multicolumn{2}{c}{\mcDGNet ($4\%$)} &  \multicolumn{2}{c}{\nDGNet} 
            \\
            &  & Pred \quad \quad  & Error \quad \quad & Pred \quad \quad & Error \quad \quad
            \\
            \rotatebox[origin=c]{90}{$T = 1.2$} &
            \raisebox{-0.5\height}{\includegraphics[width = .20\textwidth]{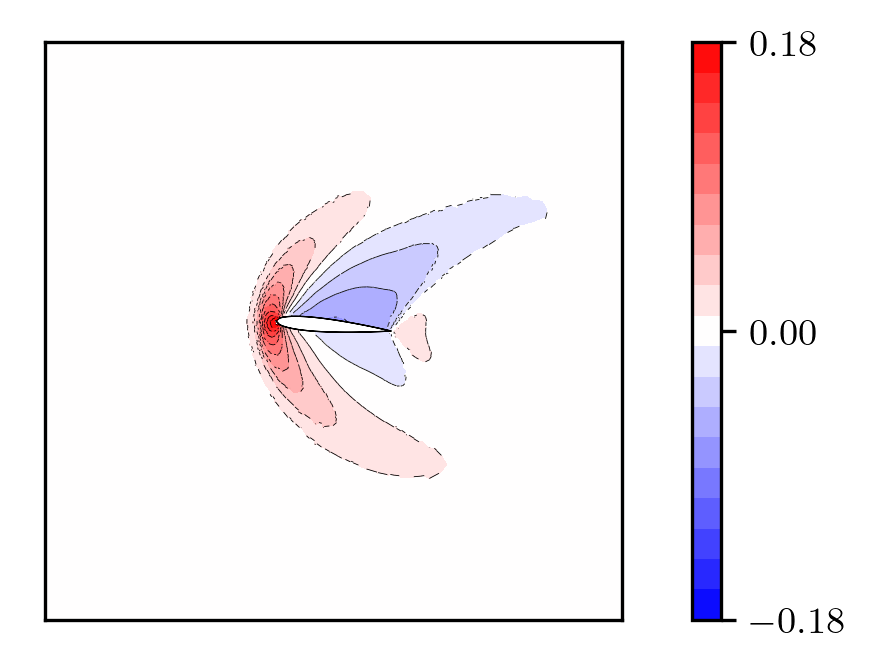}} &
            \raisebox{-0.5\height}{\includegraphics[width = .20\textwidth]{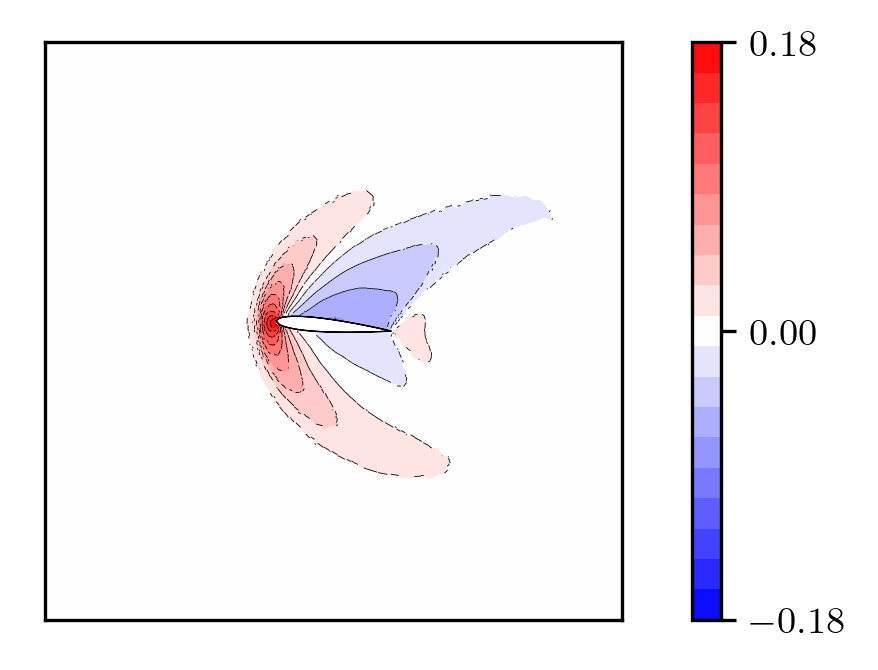}} &
            \raisebox{-0.5\height}{\includegraphics[width = .20\textwidth]{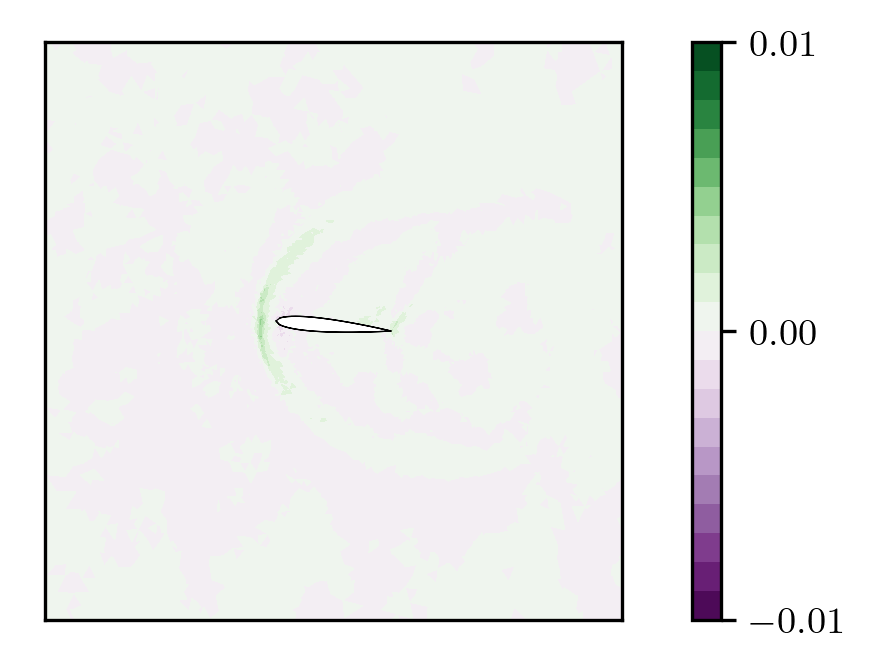}} &
            \raisebox{-0.5\height}{\includegraphics[width = .20\textwidth]{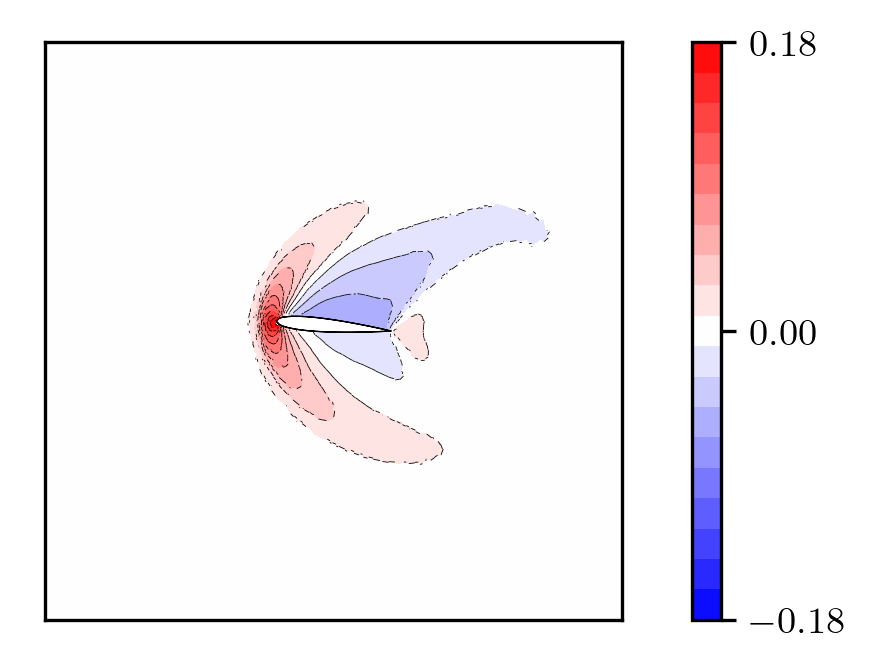}} &
            \raisebox{-0.5\height}{\includegraphics[width = .20\textwidth]{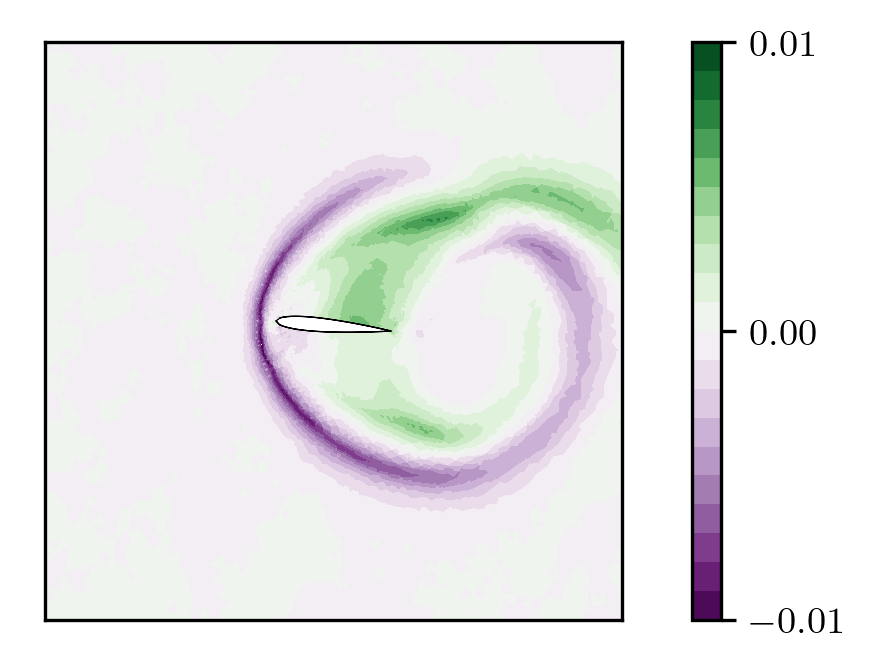}} 
            \\
            \rotatebox[origin=c]{90}{$T = 3$} &
            \raisebox{-0.5\height}{\includegraphics[width = .20\textwidth]{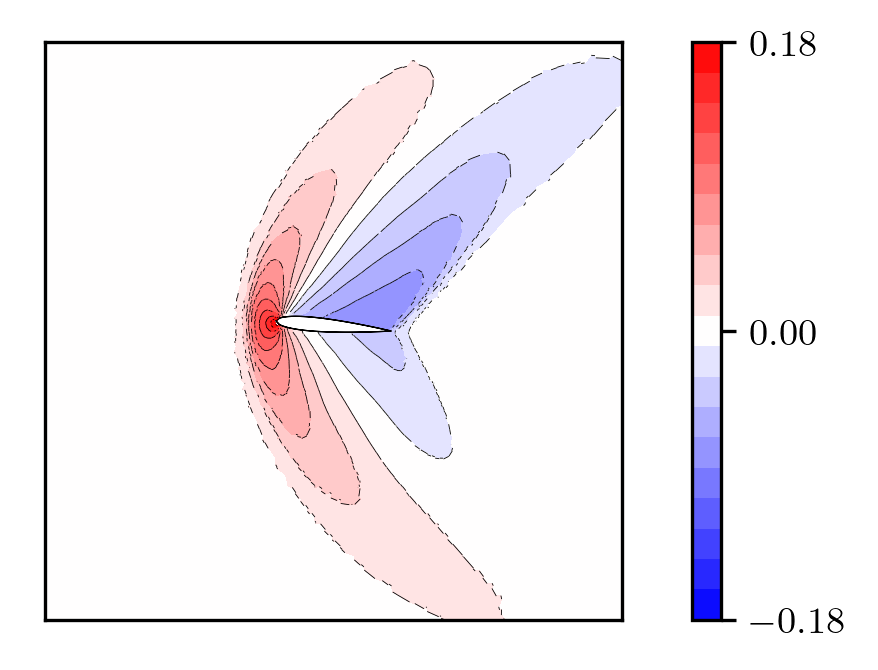}} &
            \raisebox{-0.5\height}{\includegraphics[width = .20\textwidth]{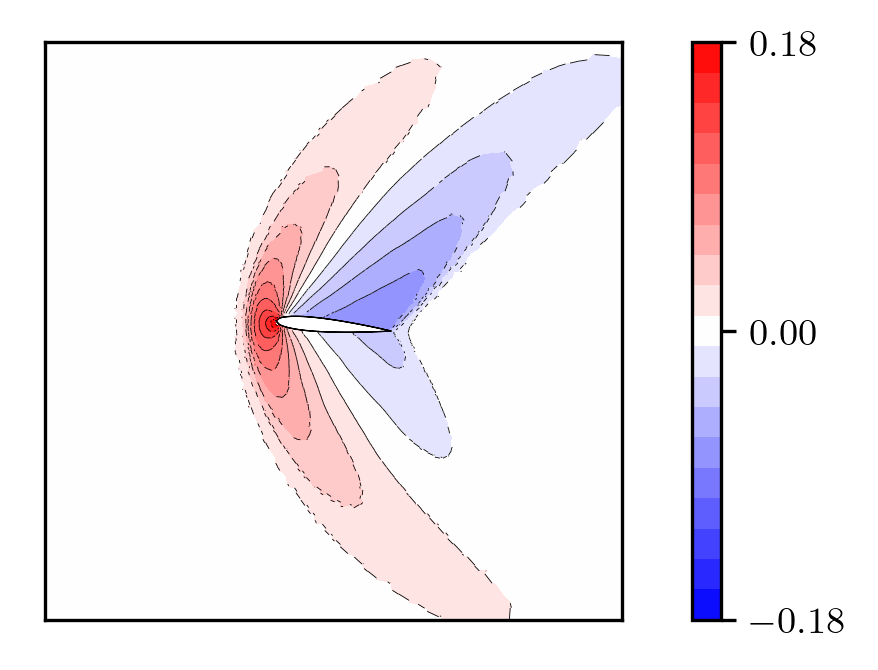}} &
            \raisebox{-0.5\height}{\includegraphics[width = .20\textwidth]{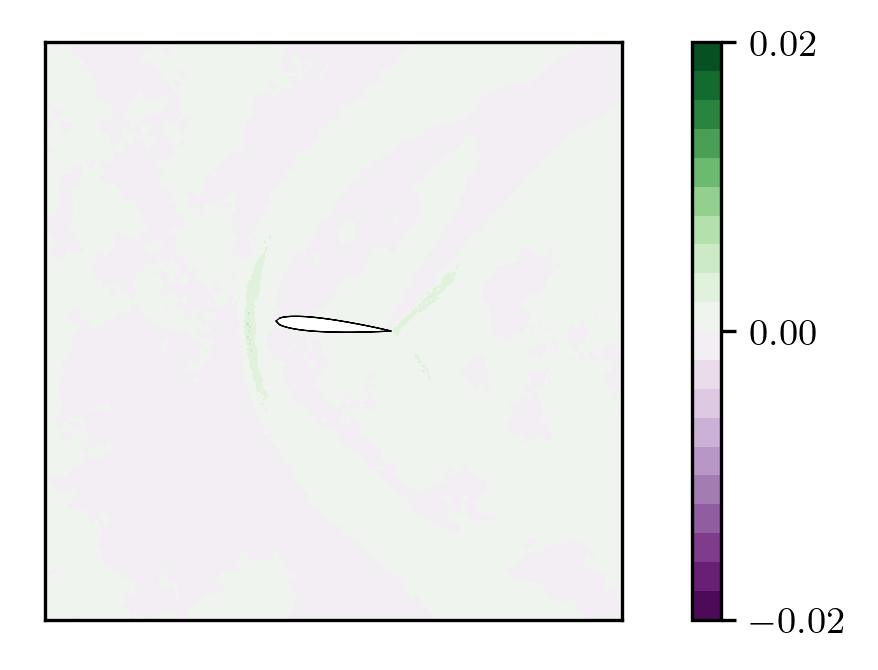}} &
            \raisebox{-0.5\height}{\includegraphics[width = .20\textwidth]{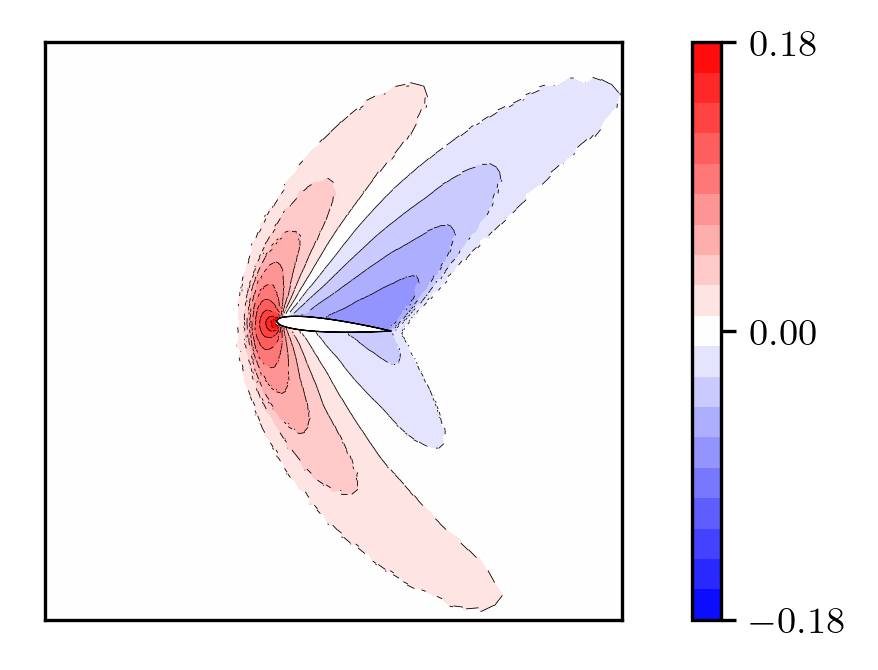}} &
            \raisebox{-0.5\height}{\includegraphics[width = .20\textwidth]{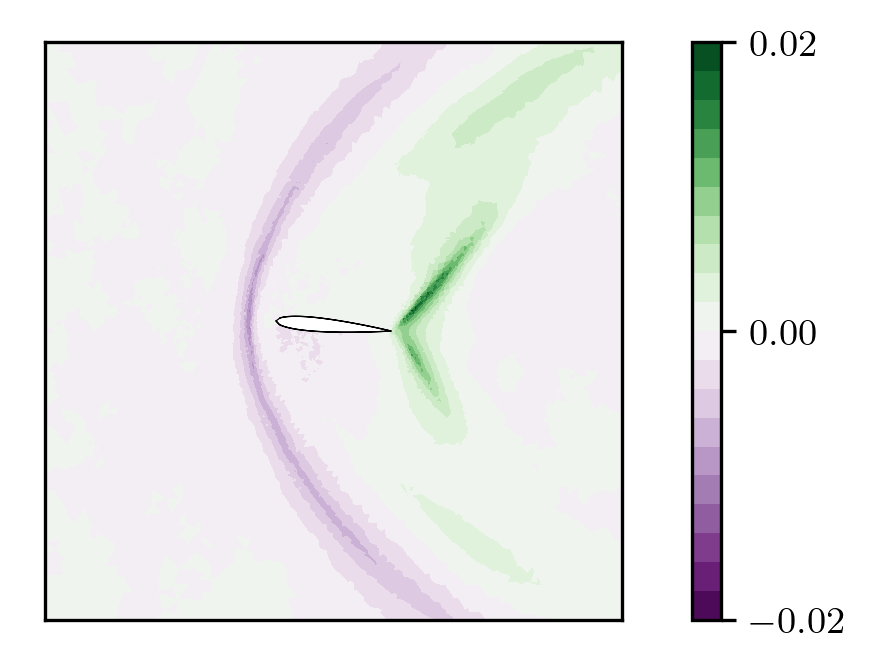}}
            \\
            \rotatebox[origin=c]{90}{$T = 5.25$} &
            \raisebox{-0.5\height}{\includegraphics[width = .20\textwidth]{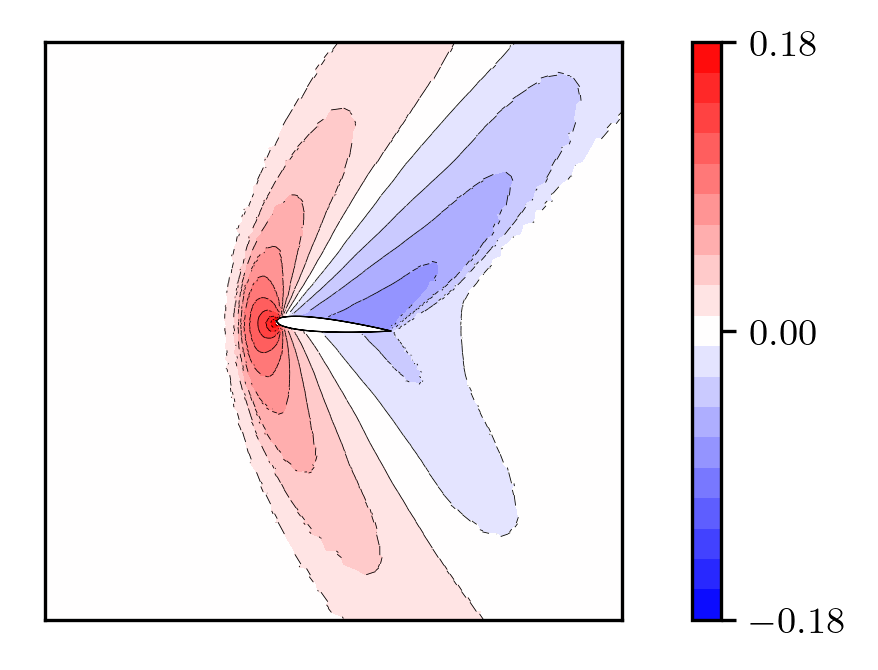}} &
            \raisebox{-0.5\height}{\includegraphics[width = .20\textwidth]{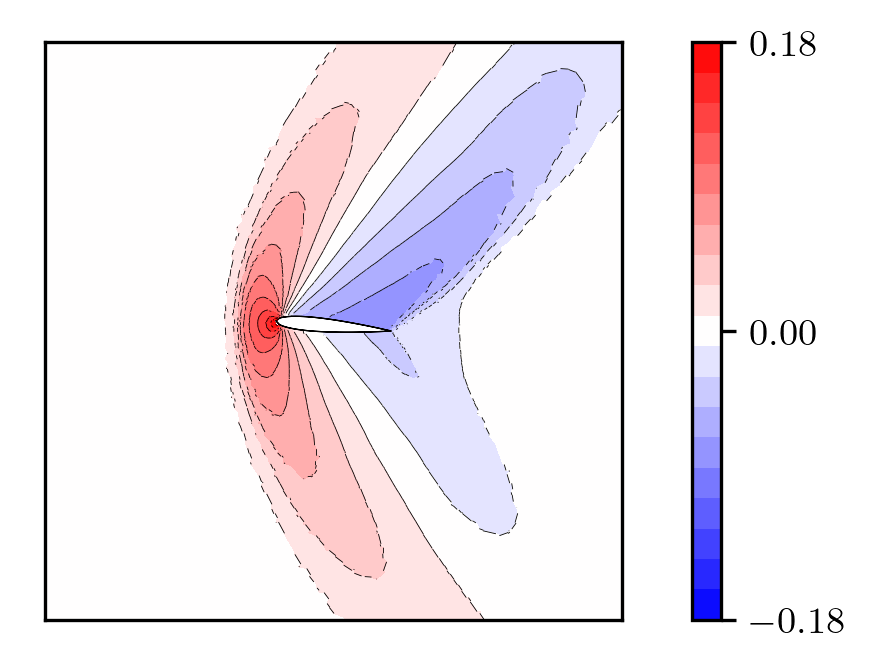}} &
            \raisebox{-0.5\height}{\includegraphics[width = .20\textwidth]{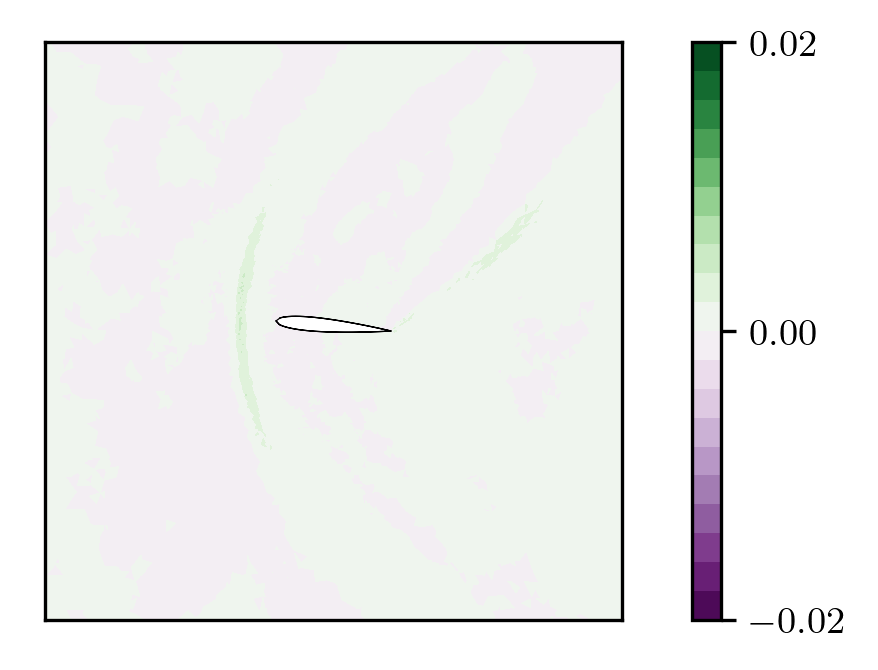}} &
            \raisebox{-0.5\height}{\includegraphics[width = .20\textwidth]{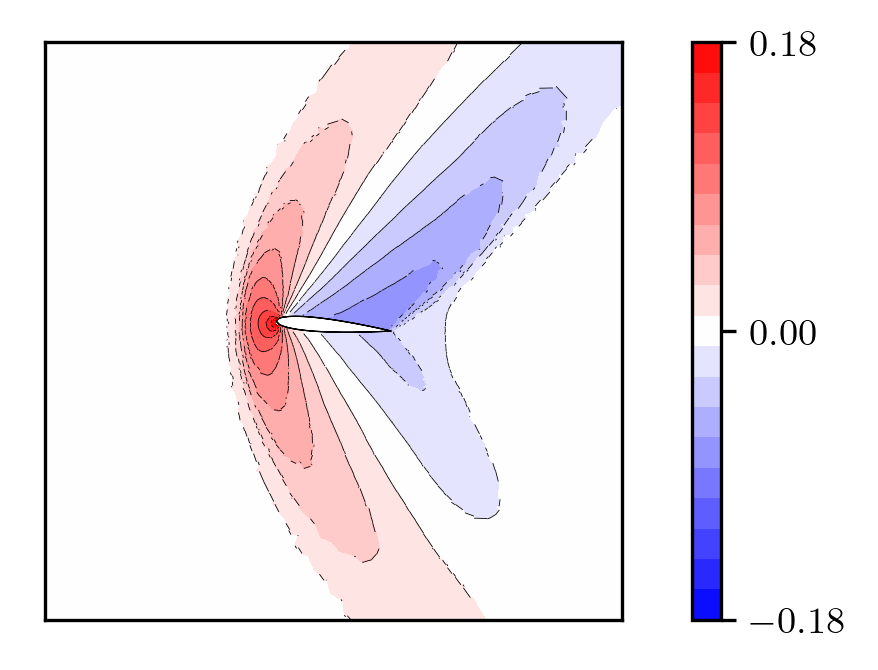}} &
            \raisebox{-0.5\height}{\includegraphics[width = .20\textwidth]{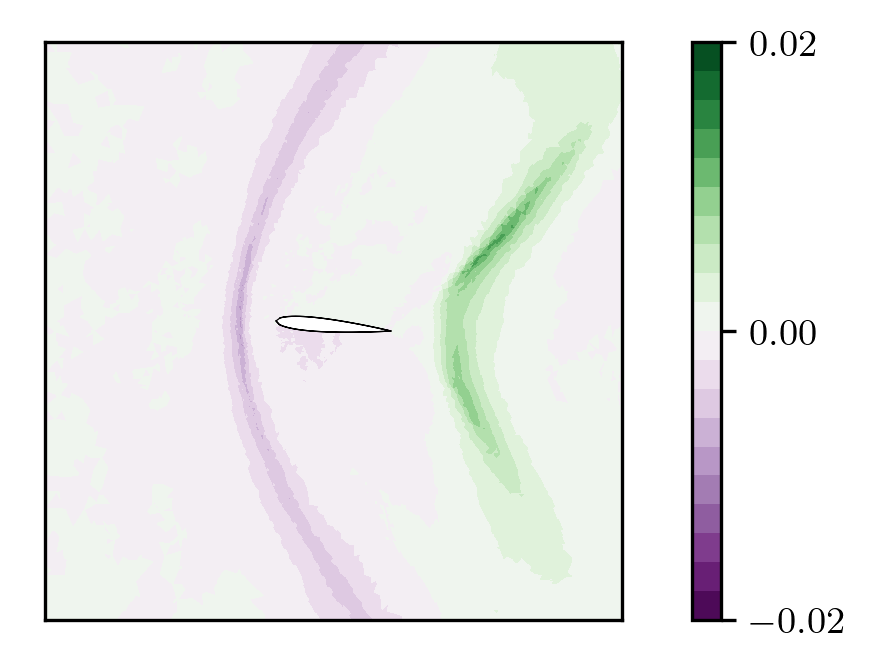}}
            \\
            \rotatebox[origin=c]{90}{$T = 7.5$} &
            \raisebox{-0.5\height}{\includegraphics[width = .20\textwidth]{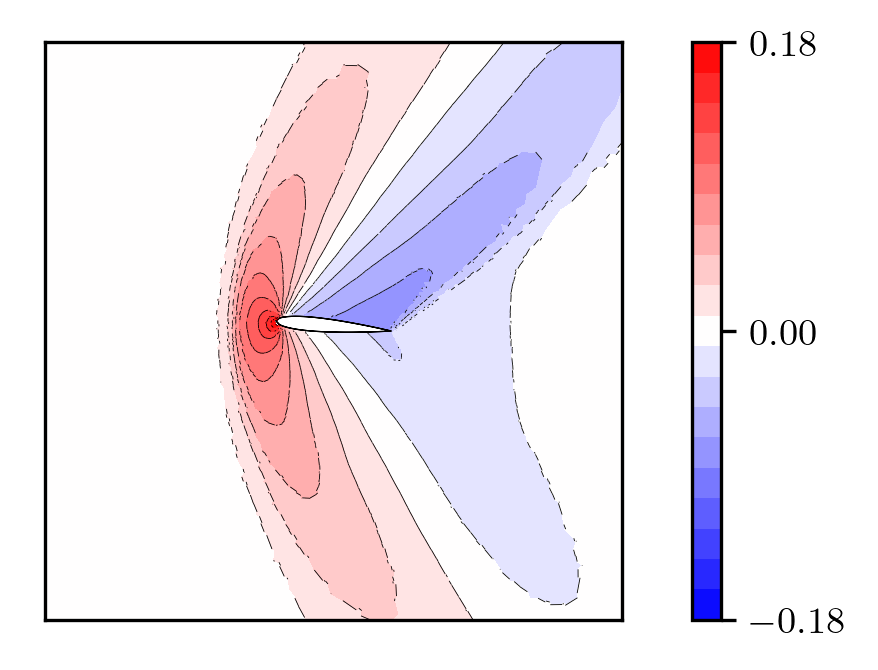}} & 
            \raisebox{-0.5\height}{\includegraphics[width = .20\textwidth]{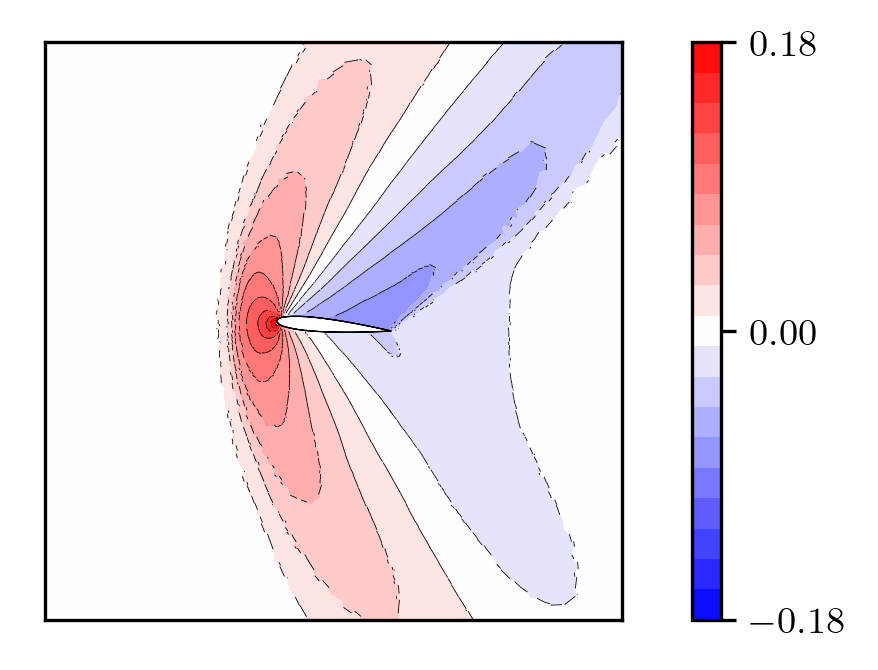}} & 
            \raisebox{-0.5\height}{\includegraphics[width = .20\textwidth]{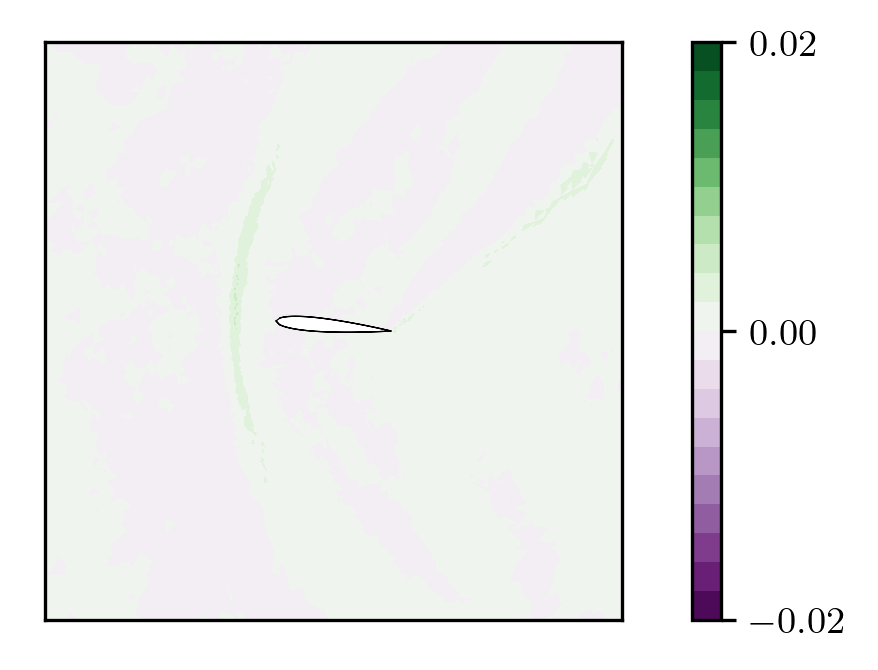}} & 
            \raisebox{-0.5\height}{\includegraphics[width = .20\textwidth]{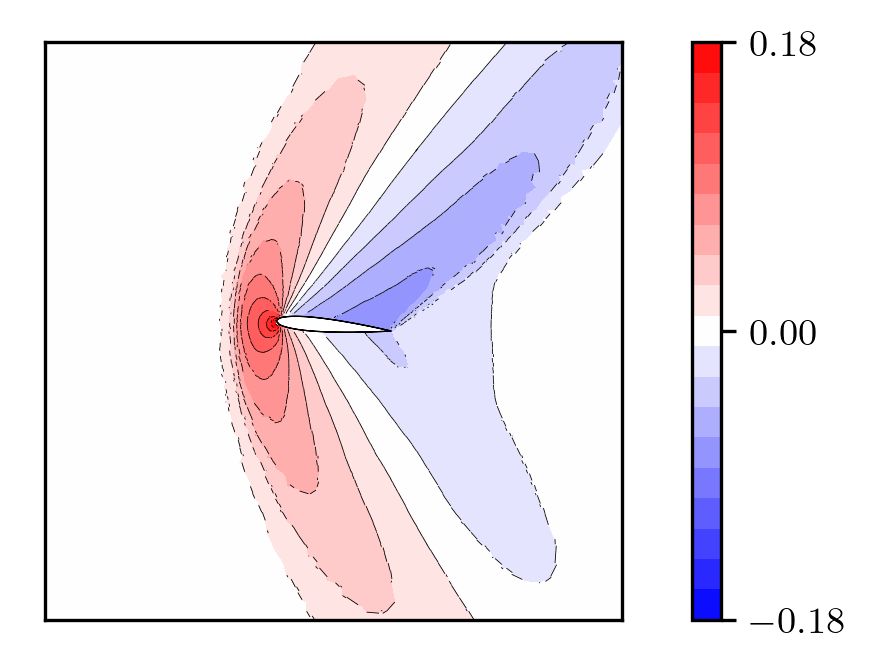}} &
            \raisebox{-0.5\height}{\includegraphics[width = .20\textwidth]{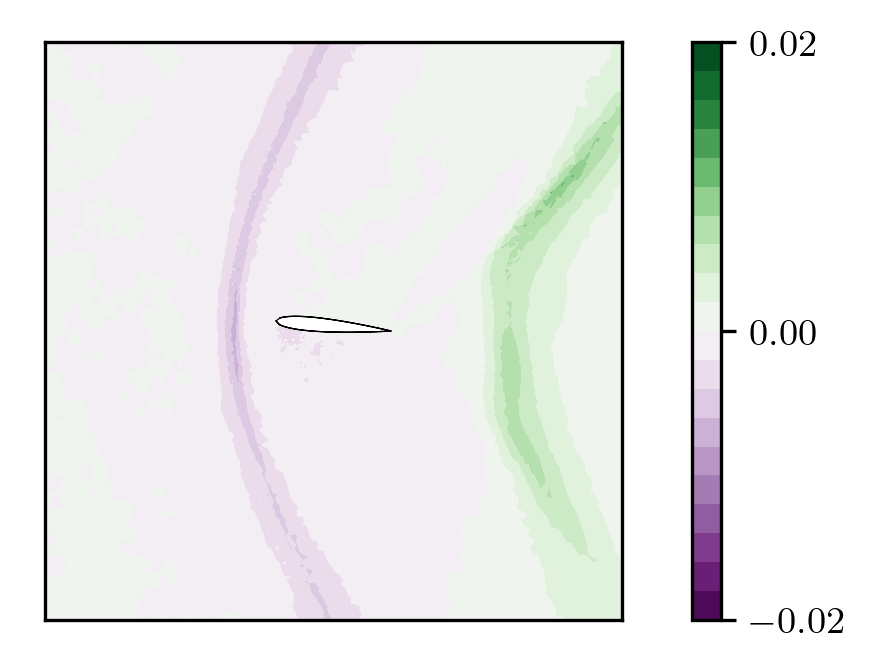}}
        \end{tabular*}
        \caption{{\bf 2D Airfoil:} predicted pressure coefficient field obtained by \nDGNet and \mcDGNet approaches and corresponding prediction pointwise error $C_{p,\text{DG}} - C_{p,\text{pred}}$ at time steps $T = \LRc{1.2, 3, 5.25, 7.5}s$ for the case of Mach $M = 1.2$ and AOA = $5^o$ in Model 2 mesh grid. Plots are cropped over the domain $\LRs{-2.5, 2.5} \times \LRs{0,5}$.} 
        \figlab{airfoil_prediction_solution_model3_mesh}
\end{figure}

\begin{figure}[htb!]
    \centering
        \begin{tabular*}{\textwidth}{c@{\hskip -0.0cm} c@{\hskip -0.0cm}}
            \centering
            \quad \quad Model 1 & \quad \quad Model 2
            \\
            \raisebox{-0.5\height}{\includegraphics[width = .48\textwidth]{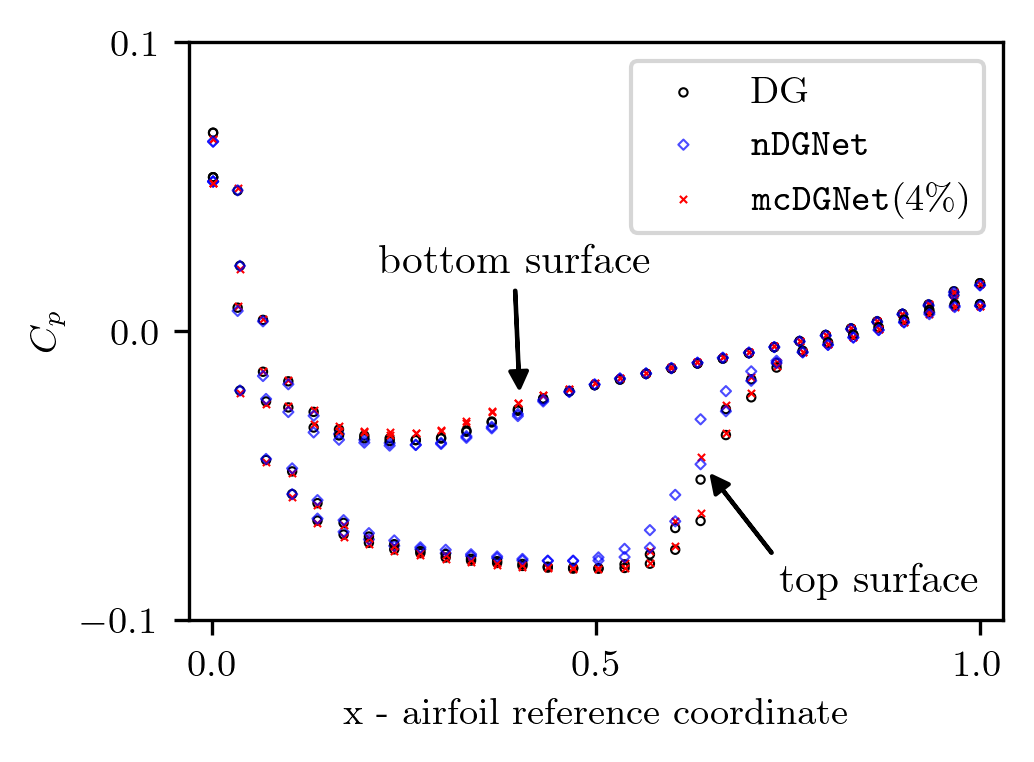}} &
            \raisebox{-0.5\height}{\includegraphics[width = .48\textwidth]{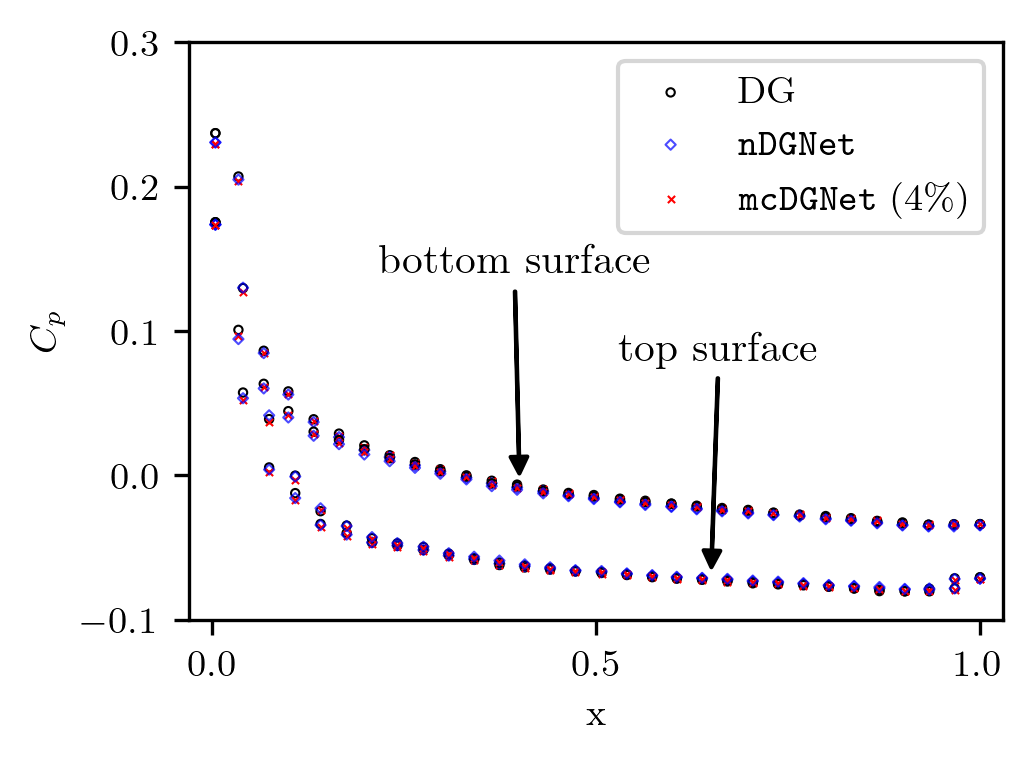}} 
        \end{tabular*}
        \caption{{\bf 2D Airfoil:} predicted surface pressure coefficients at $T_\text{test} = 7.5s$ obtained by \nDGNet and \mcDGNet approaches for Model 1 ({\bf Left}), Model 2 ({{\bf Right}}) configurations.}
        \figlab{airfoil_pressure_on_surfaces_model1_model2_mesh}
\end{figure}

In this problem, we consider the uniform upstream flow passing the NACA0012 airfoil using 2D Euler equations. The problem is modeled over a large domain, $\LRs{-4.5,4.5} \times \LRs{0, 10}$, in order to sufficiently capture the effects away from the airfoil surface. On the boundary of the domain, the outflow boundary condition is imposed on the right edge and the inflow boundary condition is assigned to the other edges. The airfoil surface is modeled as a wall boundary condition, as shown in \cref{fig:airfoil_configuration}. The initial condition is defined as $\rho_0 = \gamma$, $\rho_0 u_0 = M \gamma$, $\rho_0 v_0 = 0$, $p_0 = 1$, and $E_0 = \frac{p_0}{\gamma - 1} + \frac{\rho_0}{2}\LRp{u_0^2 + v_0^2}$. Two models are considered: Model 1 with a Mach number of 0.8, AOA = $3^o$, K = 13400 elements, and Model 2 with a Mach number of 1.2, AOA = $5^o$, K = 13441 elements. For data generalization, we generate the training data by solving the 2D Euler equation with $\gamma \in \LRc{1.2, 1.6}$ and validation data with $\gamma = 1.4$ over the time interval $\LRs{0,1.2}s$ in Model 1. The test data sets for both Model 1 and Model 2 are produced over the test period of $\LRs{0,7.5}s$. The uniform time step size of $\dt = 0.0015s$ is adopted to generate all data sets.

The \nDGNet approach is trained with noise-free training data, while the \mcDGNet approach is trained with corrupted training data with $4\%$ noise. The average relative $L^2$-error  over conservative variables $\LRp{\rho, \rho u, E}$ between \oDGNet predictions and traditional DG method is presented in \cref{fig:airfoil_Relative_error_test_data_model1_model2_mesh} for the unseen test data.
For both models, the \nDGNet predictions are less accurate than the ones obtained from \mcDGNet approach. The reason is that the \mcDGNet approach offers implicit regularization on tangent slope surrogate models (see \cref{sect:Data_rand}). Additionally, the improvement in \mcDGNet stems from the fact that data randomization technique enriches training data significantly, as shown in \cref{sect:P6_2D_Noise_corruption}. On the other hand, we can see the generalization capability of \oDGNet approaches as being trained from Model 1 settings and tested for out-of-distribution test cases with different Mach numbers and AOAs in Model 2. 

The pressure coefficient field is a dimensionless number that describes how the pressure deviates from the freestream condition, normalized by the freestream dynamic pressure. It is computed as
\begin{equation*}
    C_p = \frac{p-p_0}{\frac{1}{2}\rho_0 \LRp{u_0^2+v_0^2}}.
\end{equation*}
\cref{fig:airfoil_prediction_solution_model1_mesh} and \cref{fig:airfoil_prediction_solution_model3_mesh} present the pressure coefficient field and corresponding pointwise error $C_{p,\text{DG}} - C_{p,\text{pred}}$ at time steps $T \in \LRc{1.2, 3, 5.25, 7.5}s$ on Model 1 and Model 2, respectively, as predicted by the \nDGNet and \mcDGNet approaches. The \mcDGNet approach is consistently superior to the \nDGNet approach in forecasting solutions for the reasons discussed above. The airfoil surface pressure coefficient distribution profiles at $T_\text{test} = 7.5s$ for \oDGNet and DG approaches are presented in \cref{fig:airfoil_pressure_on_surfaces_model1_model2_mesh}. It is not surprising that \mcDGNet predictions have a better agreement with DG solutions compared to \nDGNet \hspace{-1ex}, especially at locations with sharp changes.

% \clearpage

\subsection{2D Euler Benchmarks}
\seclab{2d_euler_benchmarks}

In this section, we consider the 2D Euler equations with benchmark configuration 6 and configuration 12 in \cite{kurganov2002solution}, over the domain $\Omega = [0,1]^2$. The initial conditions for two different configurations on four quadrants are given in \cref{tab:2D_Euler_Benchmarks_Initial_Conditions}.

\begin{table}[htb!]
    \centering
    \caption{{\bf 2D Euler Benchmarks:} Initial conditions for Euler Benchmark configurations 6 and 12 \cite{kurganov2002solution}.}
    \tablab{2D_Euler_Benchmarks_Initial_Conditions}
    \begin{tblr}{
  cells = {c},
  cell{1}{1} = {r=2}{},
  cell{1}{2} = {c=4}{},
  cell{1}{6} = {c=4}{},
  vlines,
  hline{1,4-7} = {-}{},
  hline{2-3} = {2-9}{},
}
Quadrant                      & Configuration 6  &                 &                 &                    & Configuration 12 &                 &                 &                    \\
                              & $\quad u \quad $ & $\quad v \quad$ & $\quad p \quad$ & $\quad \rho \quad$ & $ u $            & $\quad v \quad$ & $\quad p \quad$ & $\quad \rho \quad$ \\
Q1 $[0, 0.5] \times [0, 0.5]$ & -0.75            & 0.5             & 1               & 1                  & 0                & 0               & 0.4             & 0.5313             \\
Q2 $[0, 0.5] \times [0.5, 1]$ & 0.75             & 0.5             & 1               & 2                  & 0.7276           & 0               & 1               & 1                  \\
Q3 $[0.5, 1] \times [0, 0.5]$ & -0.75            & -0.5            & 1               & 3                  & 0                & 0               & 1               & 0.8                \\
Q4 $[0.5, 1] \times [0.5, 1]$ & 0.75             & -0.5            & 1               & 1                  & 0                & 0.7276          & 1               & 1                  
\end{tblr}
\end{table}

\begin{figure}[htb!]
    \centering
        \begin{tabular*}{\textwidth}{c@{\hskip -0.0cm} c@{\hskip -0.0cm}  c@{\hskip -0.0cm}}
            \centering
            Configuration 6 - Model 1  & Configuration 6 - Model 2  & Configuration 12 - Model 3  
            \\
            \raisebox{-0.5\height}{\resizebox{0.33\textwidth}{!}{% This file was created with tikzplotlib v0.10.1.
\begin{tikzpicture}

\definecolor{darkgray176}{RGB}{176,176,176}
\definecolor{green01270}{RGB}{0,127,0}
\definecolor{lightgray204}{RGB}{204,204,204}

\begin{axis}[
legend cell align={left},
legend style={
  fill opacity=0.8,
  draw opacity=1,
  text opacity=1,
  at={(0.03,0.97)},
  anchor=north west,
  draw=lightgray204
},
tick align=outside,
tick pos=left,
x grid style={darkgray176},
xlabel={time step \(\displaystyle n_t\), \(\displaystyle \Delta\)t = 0.0004s},
xmin=0, xmax=2000,
xtick style={color=black},
y grid style={darkgray176},
ylabel={Relative \(\displaystyle L^2\)-error},
ymin=0, ymax=0.2,
ytick={0, 0.1, 0.2},
ytick style={color=black}
]
\addplot [semithick, black, mark=x, mark size=2, mark repeat=8, mark options={solid}]
table {%
0 0
20 0.00461156526580453
40 0.00555365718901157
60 0.00663204491138458
80 0.00771695794537663
100 0.00876978971064091
120 0.00982067082077265
140 0.0108308903872967
160 0.0118138287216425
180 0.0127962185069919
200 0.0137701602652669
220 0.0147348726168275
240 0.0156954675912857
260 0.0166504755616188
280 0.0176095217466354
300 0.0185761544853449
320 0.0195406749844551
340 0.0205127783119678
360 0.0214794166386127
380 0.0224385280162096
400 0.0233816374093294
420 0.0243043880909681
440 0.0252265390008688
460 0.026149483397603
480 0.0270884931087494
500 0.0280517674982548
520 0.0290515199303627
540 0.0300970803946257
560 0.0311709176748991
580 0.0322857424616814
600 0.0334341675043106
620 0.0346169546246529
640 0.0358384624123573
660 0.0371110625565052
680 0.0384169220924377
700 0.0397532284259796
720 0.041115939617157
740 0.0425184294581413
760 0.043965719640255
780 0.0454530082643032
800 0.0469809137284756
820 0.0485423542559147
840 0.050135251134634
860 0.051746342331171
880 0.0533867701888084
900 0.0550518333911896
920 0.0567606016993523
940 0.0585374943912029
960 0.060385774821043
980 0.0623116455972195
1000 0.0643238499760628
1020 0.0664018839597702
1040 0.0685433968901634
1060 0.0707496255636215
1080 0.0730143338441849
1100 0.075331375002861
1120 0.0776869431138039
1140 0.0800641030073166
1160 0.0824404805898666
1180 0.0848008692264557
1200 0.0871508568525314
1220 0.0895078182220459
1240 0.0918925702571869
1260 0.094312384724617
1280 0.0967602133750916
1300 0.0992255061864853
1320 0.101696573197842
1340 0.104173049330711
1360 0.106668181717396
1380 0.109174758195877
1400 0.111676529049873
1420 0.11417131125927
1440 0.116660252213478
1460 0.119118541479111
1480 0.121539890766144
1500 0.123916432261467
1520 0.126248955726624
1540 0.128551602363586
1560 0.130849212408066
1580 0.133157923817635
1600 0.135463580489159
1620 0.137738108634949
1640 0.139950603246689
1660 0.142058983445168
1680 0.144012242555618
1700 0.145742252469063
1720 0.147211015224457
1740 0.148421749472618
1760 0.149412006139755
1780 0.15023572742939
1800 0.150955140590668
1820 0.151637390255928
1840 0.152356132864952
1860 0.153184115886688
1880 0.154177725315094
1900 0.155372262001038
1920 0.156785607337952
1940 0.15842005610466
1960 0.160267248749733
1980 0.162309139966965
};
\addlegendentry{\nDGNet}
\addplot [semithick, red, mark=*, mark size=2, mark repeat=6, mark options={solid,fill=white}]
table {%
0 0
20 0.00291397259570658
40 0.00266743125393987
60 0.00259728590026498
80 0.00255837989971042
100 0.00254868250340223
120 0.00257153762504458
140 0.0025829472579062
160 0.00258729537017643
180 0.00259909359738231
200 0.00263607874512672
220 0.00270621827803552
240 0.00278772483579814
260 0.00288799148984253
280 0.00298750796355307
300 0.0030756953638047
320 0.00316925905644894
340 0.00326431263238192
360 0.00336425774730742
380 0.00347861647605896
400 0.00359684973955154
420 0.00371151464059949
440 0.00382503191940486
460 0.00393726769834757
480 0.00405053235590458
500 0.00416765734553337
520 0.00429506320506334
540 0.00443261349573731
560 0.00458146631717682
580 0.00473516900092363
600 0.00488459970802069
620 0.0050301244482398
640 0.00519980816170573
660 0.00537753105163574
680 0.00555406417697668
700 0.00573523808270693
720 0.00593724707141519
740 0.00616618478670716
760 0.00642146496102214
780 0.0066796881146729
800 0.00693130027502775
820 0.00717768725007772
840 0.00741355028003454
860 0.00764051545411348
880 0.00786795187741518
900 0.00810554437339306
920 0.00835383683443069
940 0.00860721431672573
960 0.00885044597089291
980 0.00909008271992207
1000 0.00933202262967825
1020 0.00958029273897409
1040 0.0098326625302434
1060 0.010085760615766
1080 0.0103397993370891
1100 0.0105963014066219
1120 0.0108579266816378
1140 0.0111205149441957
1160 0.0113784875720739
1180 0.011640815064311
1200 0.0119258239865303
1220 0.0122345238924026
1240 0.0125551735982299
1260 0.0128737781196833
1280 0.0131861632689834
1300 0.0134883020073175
1320 0.0137772839516401
1340 0.0140508664771914
1360 0.0143044022843242
1380 0.0145423915237188
1400 0.0147783719003201
1420 0.0150198694318533
1440 0.0152645735070109
1460 0.0155070209875703
1480 0.0157384891062975
1500 0.0159517172724009
1520 0.0161472633481026
1540 0.0163303688168526
1560 0.0165052711963654
1580 0.0166755504906178
1600 0.0168390274047852
1620 0.0169972218573093
1640 0.0171607490628958
1660 0.0173486098647118
1680 0.0175759233534336
1700 0.0178444590419531
1720 0.0181408114731312
1740 0.0184565149247646
1760 0.0187845192849636
1780 0.0191196128726006
1800 0.0194586794823408
1820 0.0198045894503593
1840 0.0201624128967524
1860 0.0205347910523415
1880 0.0209217332303524
1900 0.0213179979473352
1920 0.0217156186699867
1940 0.0221075229346752
1960 0.0224875882267952
1980 0.0228529721498489
};
\addlegendentry{\mcDGNet ($2\%$)}
\end{axis}

\end{tikzpicture}}} &
            \raisebox{-0.5\height}{\resizebox{0.33\textwidth}{!}{%This file was created with tikzplotlib v0.10.1.
\begin{tikzpicture}
\definecolor{darkgray176}{RGB}{176,176,176}
\definecolor{green01270}{RGB}{0,127,0}
\definecolor{lightgray204}{RGB}{204,204,204}

\begin{axis}[
legend cell align={left},
legend style={
    fill opacity=0.8,
    draw opacity=1,
    text opacity=1,
    at={(0.03,0.97)},
    anchor=north west,
    draw=lightgray204
},
tick align=outside,
tick pos=left,
x grid style={darkgray176},
xlabel={time step \(\displaystyle n_t\), \(\displaystyle \Delta\)t = 0.0002s},
xmin=0, xmax=4000,
xtick style={color=black},
y grid style={darkgray176},
ylabel={Relative \(\displaystyle L^2\)-error},
ymin=0, ymax=0.6,
ytick={0, 0.2, 0.4, 0.6},
ytick style={color=black}
]
\addplot [semithick, black, mark=x, mark size=2.5, mark repeat=8, mark options={solid}]
table {%
0 0
40 0.00257597211748362
80 0.00454433728009462
120 0.00642677303403616
160 0.00826675072312355
200 0.0100768879055977
240 0.0118719972670078
280 0.0136663457378745
320 0.0154623026028275
360 0.0172718446701765
400 0.0191014111042023
440 0.0209565386176109
480 0.0228452086448669
520 0.0247693248093128
560 0.0267378948628902
600 0.0287551153451204
640 0.030827147886157
680 0.0329585894942284
720 0.0351532623171806
760 0.0374174453318119
800 0.0397538878023624
840 0.0421696566045284
880 0.044669583439827
920 0.0472621284425259
960 0.0499518513679504
1000 0.0527475886046886
1040 0.0556581169366837
1080 0.0586905106902122
1120 0.0618459433317184
1160 0.065128281712532
1200 0.0685450732707977
1240 0.0721041783690453
1280 0.0758095532655716
1320 0.0796516984701157
1360 0.0835855901241302
1400 0.0875847637653351
1440 0.0916617959737778
1480 0.0958183109760284
1520 0.100050434470177
1560 0.104357950389385
1600 0.108744367957115
1640 0.113217577338219
1680 0.117790587246418
1720 0.122477620840073
1760 0.12729249894619
1800 0.13224820792675
1840 0.137359917163849
1880 0.14263179898262
1920 0.148062154650688
1960 0.153648257255554
2000 0.159376084804535
2040 0.165281817317009
2080 0.171440273523331
2120 0.177860662341118
2160 0.184467956423759
2200 0.191176623106003
2240 0.197958886623383
2280 0.20483136177063
2320 0.211830615997314
2360 0.218986034393311
2400 0.226281404495239
2440 0.233683198690414
2480 0.241167351603508
2520 0.248713120818138
2560 0.256292372941971
2600 0.263897001743317
2640 0.271553933620453
2680 0.2793088555336
2720 0.287182867527008
2760 0.295206844806671
2800 0.303393125534058
2840 0.311700850725174
2880 0.320096731185913
2920 0.328603684902191
2960 0.337228357791901
3000 0.345922291278839
3040 0.354579389095306
3080 0.36311462521553
3120 0.371585011482239
3160 0.380057066679001
3200 0.388559877872467
3240 0.397139966487885
3280 0.405873447656631
3320 0.414796769618988
3360 0.423881381750107
3400 0.433052867650986
3440 0.442211896181107
3480 0.451246589422226
3520 0.460049390792847
3560 0.468602895736694
3600 0.477014720439911
3640 0.485463291406631
3680 0.494046837091446
3720 0.502731084823608
3760 0.511446297168732
3800 0.520119607448578
3840 0.528670430183411
3880 0.53704160451889
3920 0.545213282108307
3960 0.553190112113953
};
\addlegendentry{\nDGNet}
\addplot [semithick, red, mark=*, mark size=2.5, mark repeat=8, mark options={solid,fill=white}]
table {%
0 0
40 0.00101258815266192
80 0.00109702907502651
120 0.00136240874417126
160 0.00171993533149362
200 0.0021446559112519
240 0.00260282540693879
280 0.00307934638112783
320 0.00356900924816728
360 0.00406829360872507
400 0.00457746814936399
440 0.00509512005373836
480 0.00562193663790822
520 0.00615742895752192
560 0.00670153088867664
600 0.00725603988394141
640 0.00782107934355736
680 0.00839737430214882
720 0.00898589193820953
760 0.00958966556936502
800 0.0102072823792696
840 0.0108391921967268
880 0.0114862211048603
920 0.0121490824967623
960 0.0128285400569439
1000 0.0135253379121423
1040 0.0142403254285455
1080 0.0149733852595091
1120 0.0157224200665951
1160 0.0164858810603619
1200 0.0172624289989471
1240 0.0180512182414532
1280 0.0188503693789244
1320 0.0196465272456408
1360 0.0204321872442961
1400 0.0212094932794571
1440 0.02197540178895
1480 0.0227284505963326
1520 0.0234698448330164
1560 0.0242040120065212
1600 0.0249365456402302
1640 0.0256729610264301
1680 0.026419572532177
1720 0.0271792896091938
1760 0.0279529448598623
1800 0.0287372618913651
1840 0.0295082870870829
1880 0.0301997046917677
1920 0.0308994762599468
1960 0.0317639298737049
2000 0.032719723880291
2040 0.0337447635829449
2080 0.0348372608423233
2120 0.035960428416729
2160 0.0370941199362278
2200 0.038251094520092
2240 0.0394183546304703
2280 0.0405874252319336
2320 0.0418326258659363
2360 0.0431790947914124
2400 0.0445890836417675
2440 0.0460418164730072
2480 0.0475320890545845
2520 0.0490546971559525
2560 0.0506074279546738
2600 0.0521950796246529
2640 0.0538167841732502
2680 0.055473268032074
2720 0.0571738891303539
2760 0.058923427015543
2800 0.060724139213562
2840 0.0625765770673752
2880 0.0644484311342239
2920 0.0662780404090881
2960 0.0680395811796188
3000 0.0697294324636459
3040 0.0713505744934082
3080 0.0728792101144791
3120 0.0742484778165817
3160 0.0754837170243263
3200 0.0767004787921906
3240 0.0780241563916206
3280 0.0795343965291977
3320 0.0812422335147858
3360 0.0830717086791992
3400 0.0849274769425392
3440 0.0867478623986244
3480 0.0885265320539474
3520 0.090317115187645
3560 0.0921604335308075
3600 0.0940652117133141
3640 0.0960226729512215
3680 0.0980325415730476
3720 0.100101493299007
3760 0.102226890623569
3800 0.104398906230927
3840 0.106608614325523
3880 0.108839340507984
3920 0.111068591475487
3960 0.113296493887901
};
\addlegendentry{\mcDGNet ($2\%$)}
\end{axis}

\end{tikzpicture}}} &
            \raisebox{-0.5\height}{\resizebox{0.33\textwidth}{!}{% This file was created with tikzplotlib v0.10.1.
\begin{tikzpicture}

\definecolor{darkgray176}{RGB}{176,176,176}
\definecolor{green01270}{RGB}{0,127,0}
\definecolor{lightgray204}{RGB}{204,204,204}

\begin{axis}[
legend cell align={left},
legend style={
  fill opacity=0.8,
  draw opacity=1,
  text opacity=1,
  at={(0.03,0.97)},
  anchor=north west,
  draw=lightgray204
},
tick align=outside,
tick pos=left,
x grid style={darkgray176},
xlabel={time step \(\displaystyle n_t\), \(\displaystyle \Delta\)t = 0.00025s},
xmin=0, xmax=1000,
xtick style={color=black},
y grid style={darkgray176},
ylabel={Relative \(\displaystyle L^2\)-error},
ymin=0, ymax=0.2,
ytick={0, 0.1, 0.2},
ytick style={color=black}
]
\addplot [semithick, black, mark=x, mark size=2.5, mark repeat=8, mark options={solid}]
table {%
0 0
10 0.00241171009838581
20 0.00410336116328835
30 0.00562425190582871
40 0.00709810201078653
50 0.00854750350117683
60 0.00998231209814548
70 0.0114034041762352
80 0.0128145283088088
90 0.0142162628471851
100 0.0156107135117054
110 0.0170002039521933
120 0.0183870531618595
130 0.0197750981897116
140 0.02116609364748
150 0.0225602798163891
160 0.0239542406052351
170 0.0253448467701674
180 0.0267289709299803
190 0.0281050819903612
200 0.0294728875160217
210 0.0308319590985775
220 0.0321823358535767
230 0.0335220694541931
240 0.0348510071635246
250 0.0361692830920219
260 0.0374782830476761
270 0.0387784615159035
280 0.040069542825222
290 0.041351281106472
300 0.0426225587725639
310 0.0438822954893112
320 0.0451300330460072
330 0.0463659539818764
340 0.0475909560918808
350 0.0488060042262077
360 0.0500130355358124
370 0.051213163882494
380 0.0524066761136055
390 0.0535915791988373
400 0.0547668933868408
410 0.0559319257736206
420 0.057086706161499
430 0.0582305081188679
440 0.0593633837997913
450 0.0604850351810455
460 0.0615950264036655
470 0.0626938492059708
480 0.0637819916009903
490 0.0648599416017532
500 0.065927118062973
510 0.0669834613800049
520 0.0680287331342697
530 0.0690632313489914
540 0.0700874105095863
550 0.0711020976305008
560 0.0721082389354706
570 0.0731070935726166
580 0.074099637567997
590 0.0750857219099998
600 0.076064795255661
610 0.0770360678434372
620 0.077999010682106
630 0.07895328104496
640 0.0798986107110977
650 0.0808354020118713
660 0.0817635804414749
670 0.0826837718486786
680 0.0835960358381271
690 0.0845013484358788
700 0.08539979159832
710 0.0862918347120285
720 0.0871774256229401
730 0.088057205080986
740 0.0889309346675873
750 0.0897987186908722
760 0.0906611382961273
770 0.0915191024541855
780 0.092372789978981
790 0.0932218357920647
800 0.0940662547945976
810 0.0949062258005142
820 0.0957415848970413
830 0.0965721383690834
840 0.0973982363939285
850 0.0982203632593155
860 0.0990386679768562
870 0.0998533964157104
880 0.100664481520653
890 0.101471930742264
900 0.102275408804417
910 0.103074826300144
920 0.103870190680027
930 0.104661643505096
940 0.105449318885803
950 0.106233336031437
960 0.107014328241348
970 0.107792586088181
980 0.108568988740444
990 0.109343774616718
};
\addlegendentry{\nDGNet}
\addplot [semithick, red, mark=*, mark size=2.5, mark repeat=8, mark options={solid,fill=white}]
table {%
0 0
10 0.00104747992008924
20 0.00142091023735702
30 0.00162214296869934
40 0.0017609631177038
50 0.00188017007894814
60 0.00198982213623822
70 0.00209068367257714
80 0.00218658428639174
90 0.00228007300756872
100 0.00237275799736381
110 0.00246594124473631
120 0.00255977222695947
130 0.00265490869060159
140 0.0027514360845089
150 0.00284913368523121
160 0.0029479197692126
170 0.00304750585928559
180 0.00314729055389762
190 0.00324723380617797
200 0.00334720034152269
210 0.00344721018336713
220 0.00354728102684021
230 0.00364782428368926
240 0.00374843180179596
250 0.00384958693757653
260 0.0039509404450655
270 0.00405274610966444
280 0.00415468774735928
290 0.00425712112337351
300 0.0043595521710813
310 0.00446229055523872
320 0.00456483103334904
330 0.00466745300218463
340 0.00477056251838803
350 0.00487422943115234
360 0.00497838482260704
370 0.00508346315473318
380 0.0051889568567276
390 0.00529490411281586
400 0.0054013105109334
410 0.00550815975293517
420 0.00561547512188554
430 0.00572365988045931
440 0.00583229027688503
450 0.00594067107886076
460 0.0060484642162919
470 0.00615475047379732
480 0.00625966116786003
490 0.00636309152469039
500 0.00646606460213661
510 0.00656842207536101
520 0.00667137280106544
530 0.00677464855834842
540 0.00687905494123697
550 0.00698427483439445
560 0.00709083210676908
570 0.00719799939543009
580 0.00730588659644127
590 0.00741403130814433
600 0.00752199348062277
610 0.00762945786118507
620 0.00773669965565205
630 0.00784374121576548
640 0.00795107055455446
650 0.0080588087439537
660 0.00816719606518745
670 0.00827612914144993
680 0.00838587246835232
690 0.00849597901105881
700 0.00860713236033916
710 0.0087189543992281
720 0.00883152149617672
730 0.00894399359822273
740 0.00905558373779058
750 0.00916586816310883
760 0.00927404128015041
770 0.00938058085739613
780 0.00948499515652657
790 0.00958914123475552
800 0.00969232060015202
810 0.00979665853083134
820 0.00990128796547651
830 0.0100080724805593
840 0.0101155117154121
850 0.0102249626070261
860 0.0103346984833479
870 0.0104453507810831
880 0.0105555038899183
890 0.0106652891263366
900 0.0107745714485645
910 0.0108837801963091
920 0.0109936399385333
930 0.011104129254818
940 0.0112162251025438
950 0.0113295475021005
960 0.0114447716623545
970 0.011561538092792
980 0.0116805471479893
990 0.0118011459708214
};
\addlegendentry{\mcDGNet ($2\%$)}
\end{axis}

\end{tikzpicture}}}
        \end{tabular*}
    \caption{{\bf 2D Euler Benchmarks:} Average relative $L^2$-error over four conservative components predictions for test data obtained by \nDGNet and \mcDGNet approaches at different time steps for configuration 6 with Model 1 ({\bf Left}), configuration 6 with Model 2 ({\bf Middle}) and configuration 12 with Model 3 ({\bf Right}).} 
    \figlab{2D_euler_configuration_6_relative_error}
\end{figure}
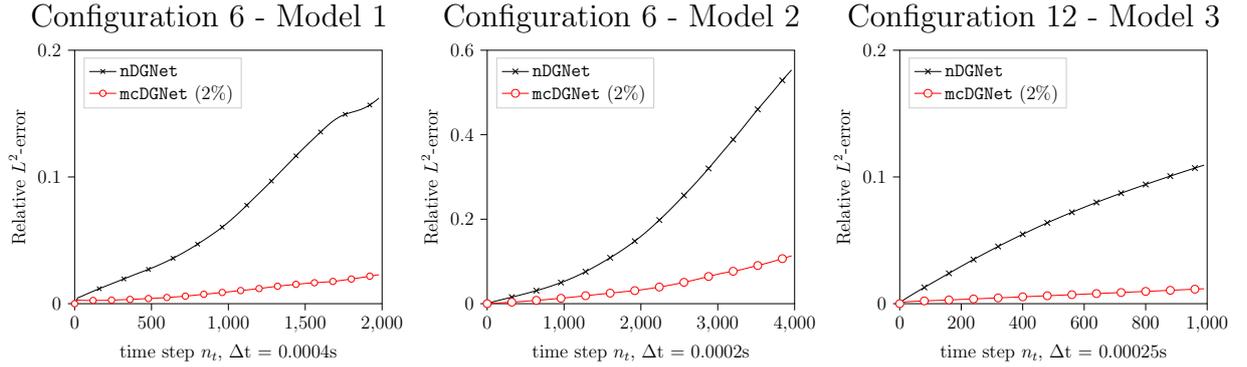

\begin{figure}[htb!]
    \centering
        \begin{tabular*}{\textwidth}{c@{\hskip -0.0cm} c@{\hskip -0.0cm}  c@{\hskip -0.0cm}}
            \centering
            DG $\quad \quad $ & \nDGNet $\quad \quad $ & $\rho_\text{DG} - \rho_\text{\nDGNet}$  $\quad \quad $
            \\
            \raisebox{-0.5\height}{\includegraphics[width = .33\textwidth]{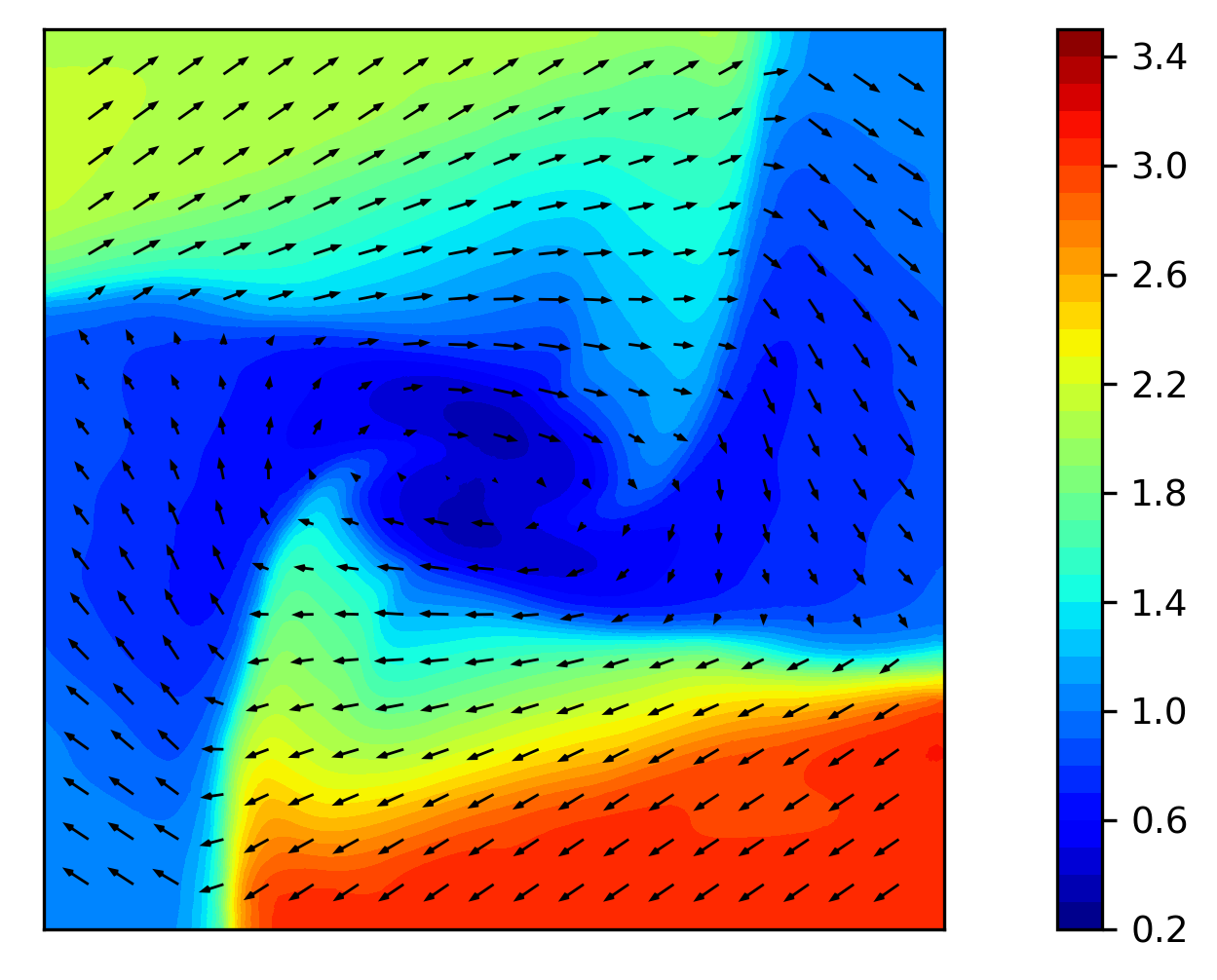}} & 
            \raisebox{-0.5\height}{\includegraphics[width = .33\textwidth]{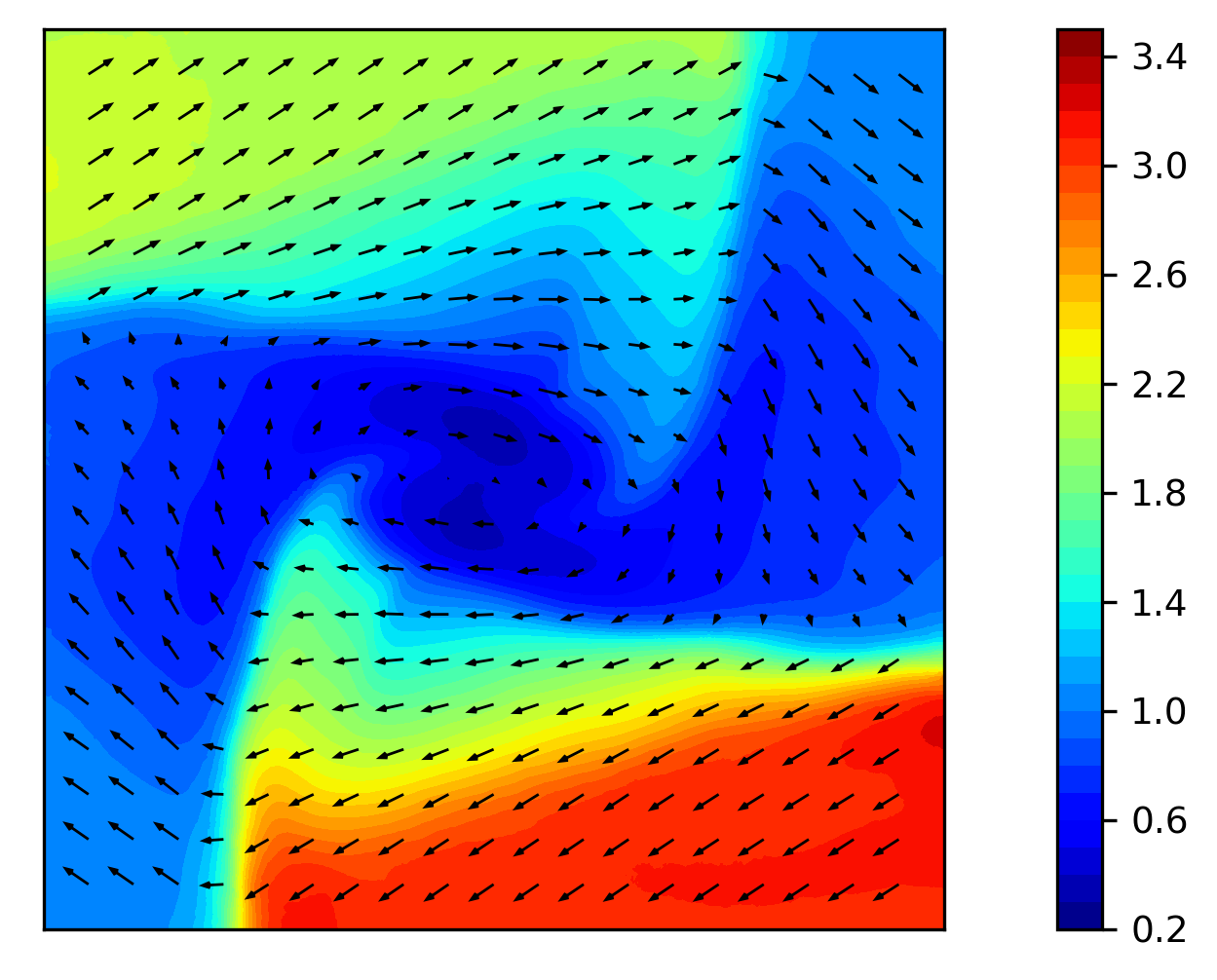}} &
            \raisebox{-0.5\height}{\includegraphics[width = .33\textwidth]{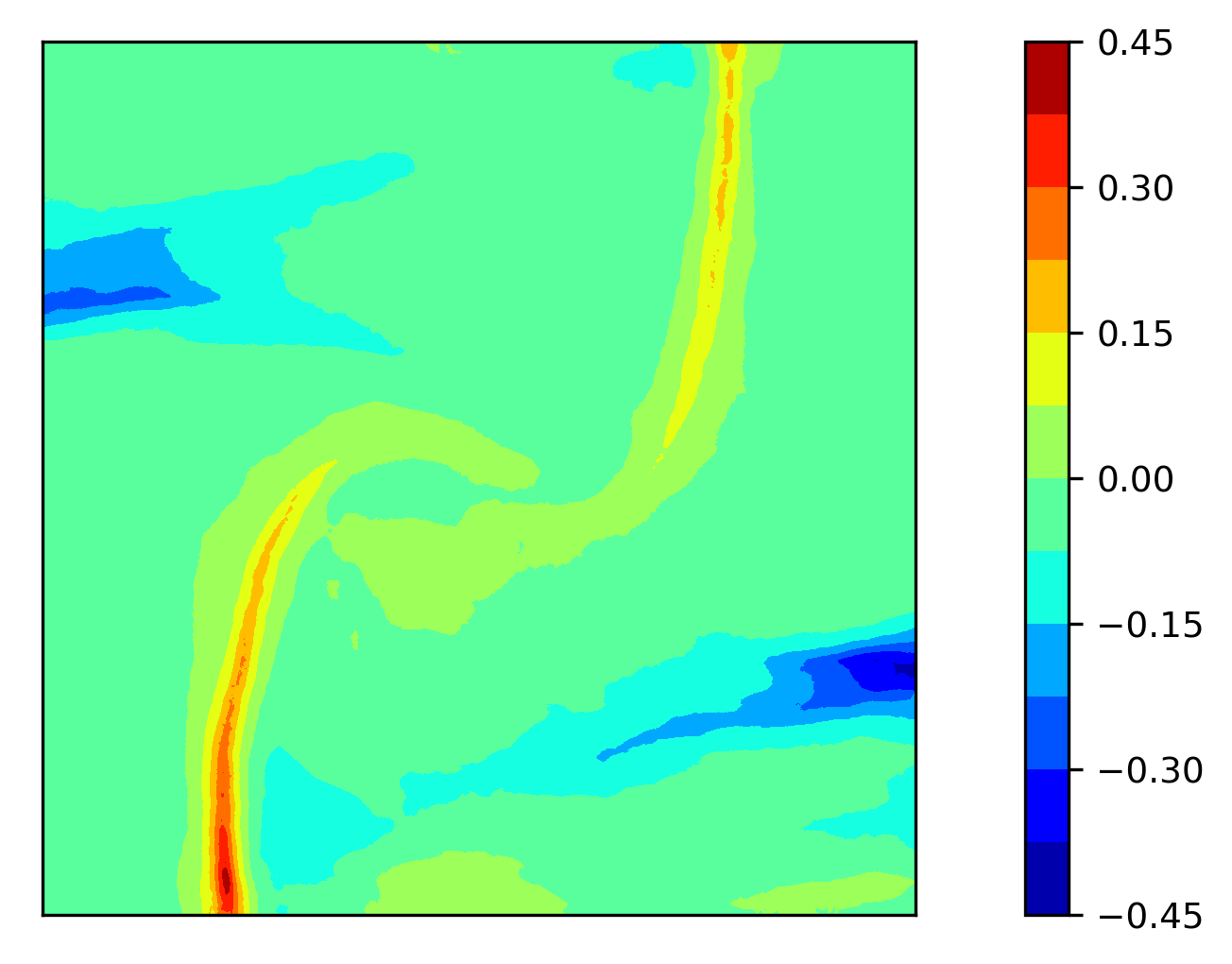}}
            \\ ~ \\
            WE-PINN  $\quad \quad $ & \mcDGNet ($2\%$)  $\quad \quad $ & $\rho_\text{DG} - \rho_\text{\mcDGNet}$  $\quad \quad $
            \\
            \raisebox{-0.5\height}{\includegraphics[width = .33\textwidth]{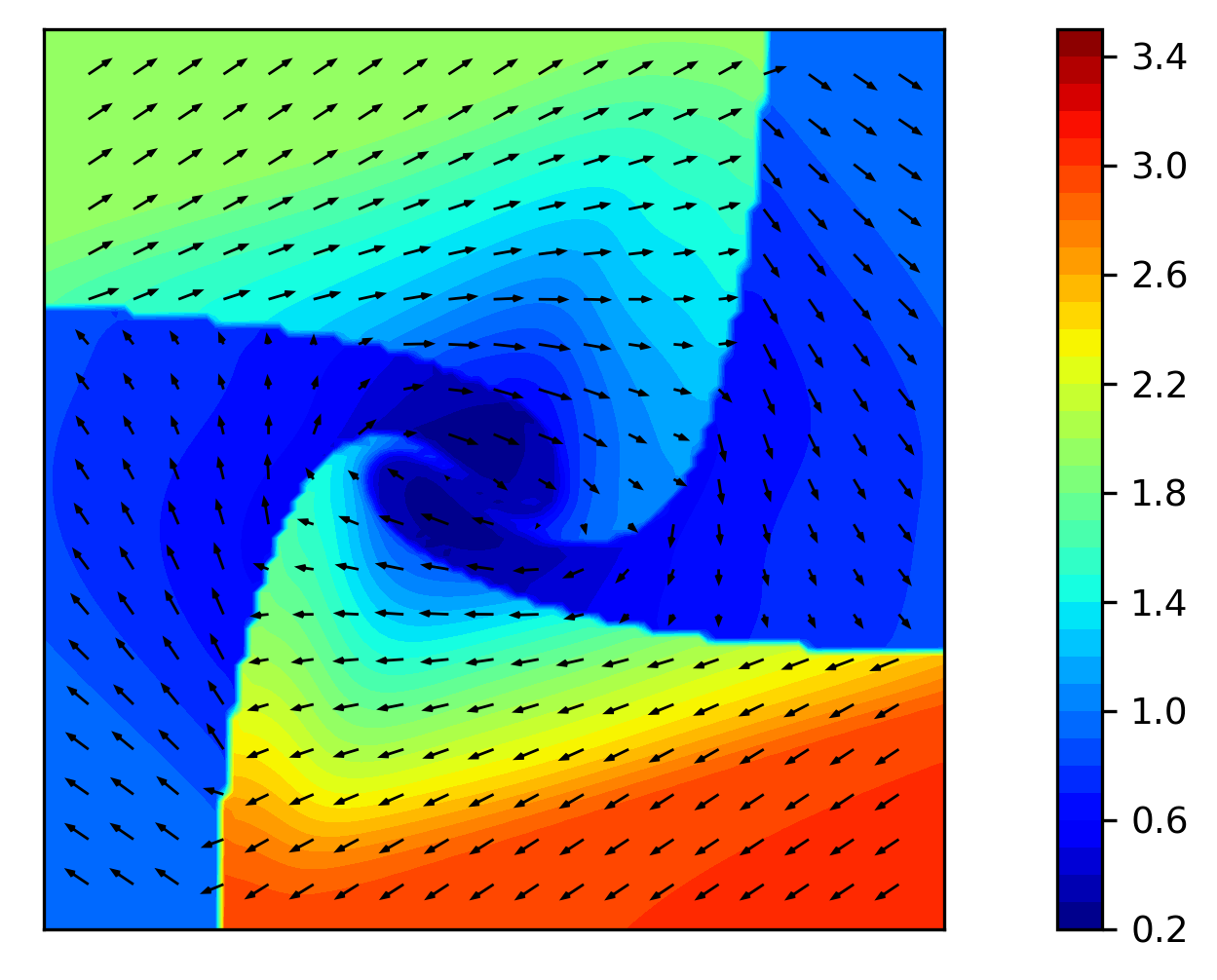}} &
            \raisebox{-0.5\height}{\includegraphics[width = .33\textwidth]{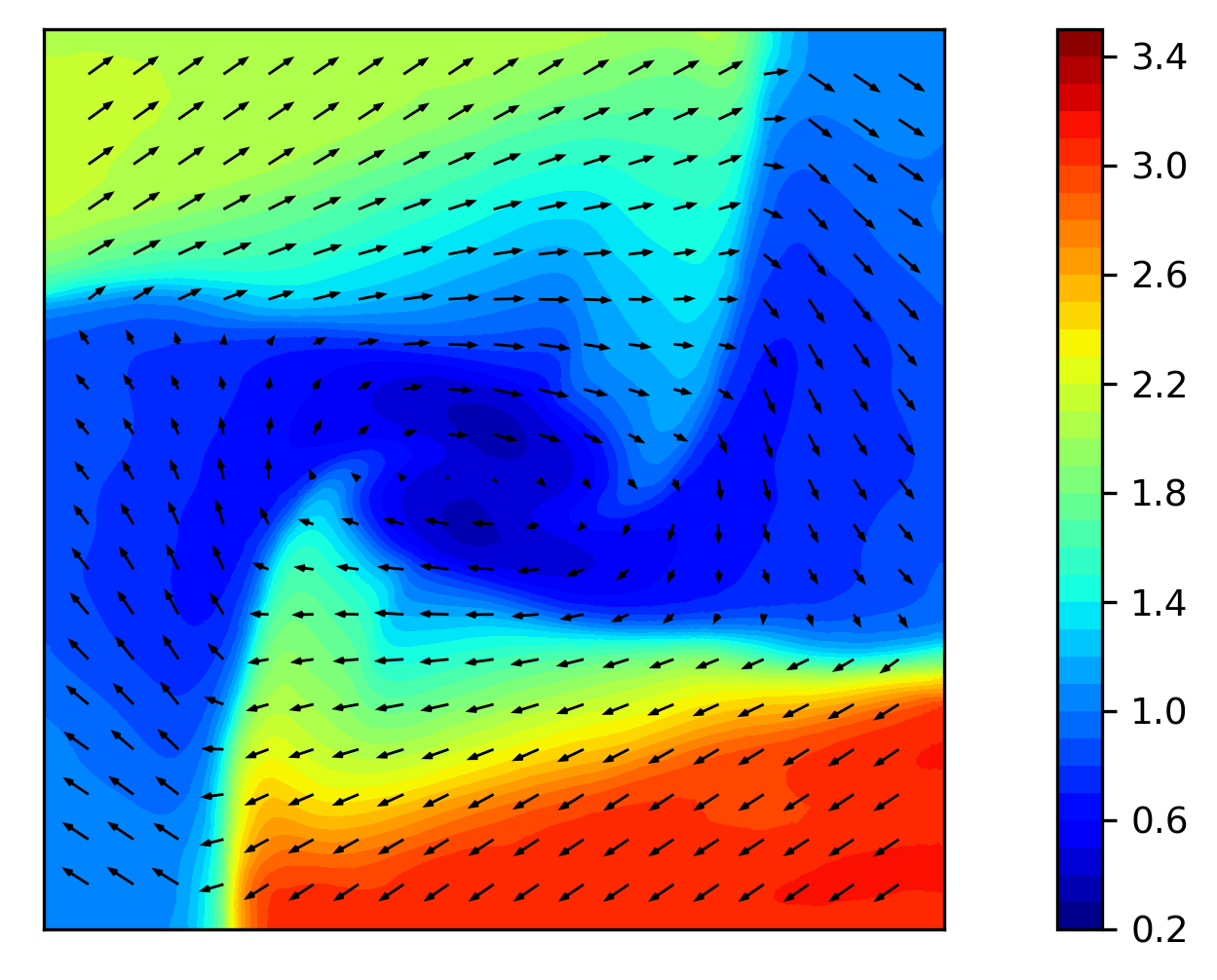}} &
            \raisebox{-0.5\height}{\includegraphics[width = .33\textwidth]{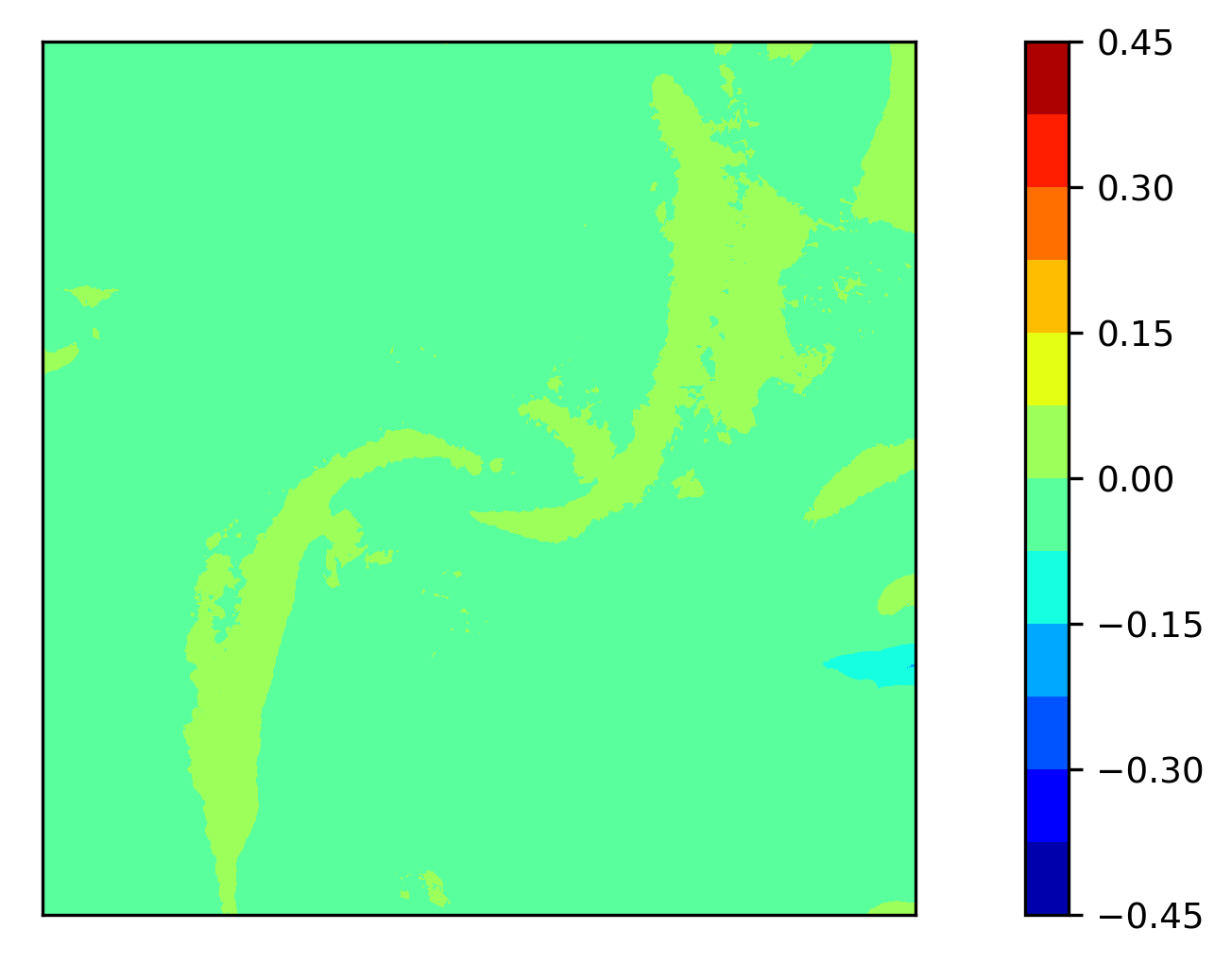}}
        \end{tabular*}
        \caption{{\bf 2D Euler Benchmarks:} predicted density fields obtained by \nDGNet and \mcDGNet approaches and corresponding prediction pointwise error $\rho_{\text{DG}} - \rho_{\text{pred}}$ and DG and WE-PINN solutions at time $T = 0.4s$ for configuration 6 -  Model 1.}
        \figlab{2D_euler_configuration_6_solutions}
\end{figure}

\begin{figure}[htb!]
    \centering
        \begin{tabular*}{\textwidth}{c@{\hskip -0.0cm} c@{\hskip -0.0cm}  c@{\hskip -0.0cm}}
            \centering
            DG $\quad \quad $ & \nDGNet $\quad \quad $ & $\rho_\text{DG} - \rho_\text{\nDGNet}$  $\quad \quad $
            \\
            \raisebox{-0.5\height}{\includegraphics[width = .33\textwidth]{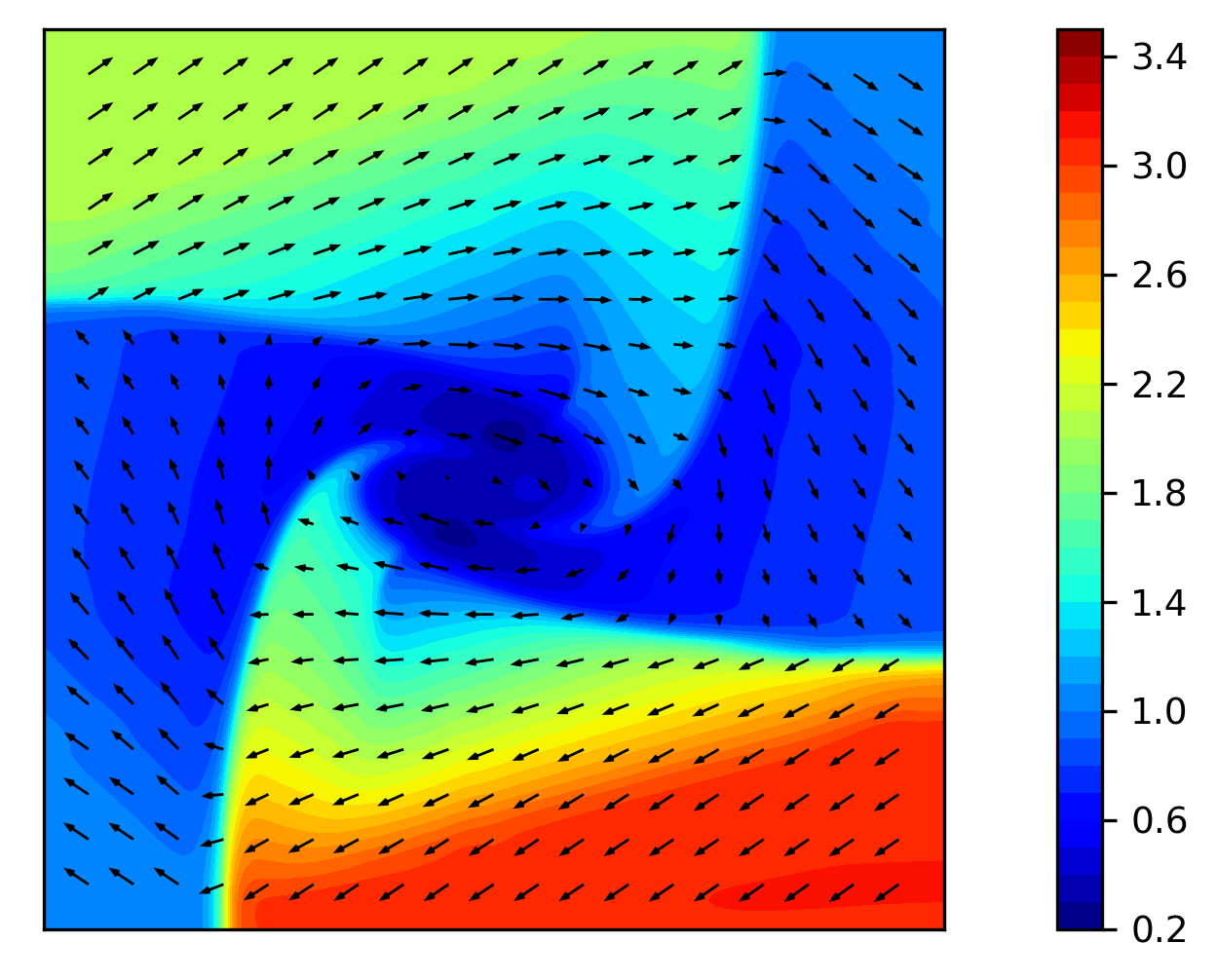}} & 
            \raisebox{-0.5\height}{\includegraphics[width = .33\textwidth]{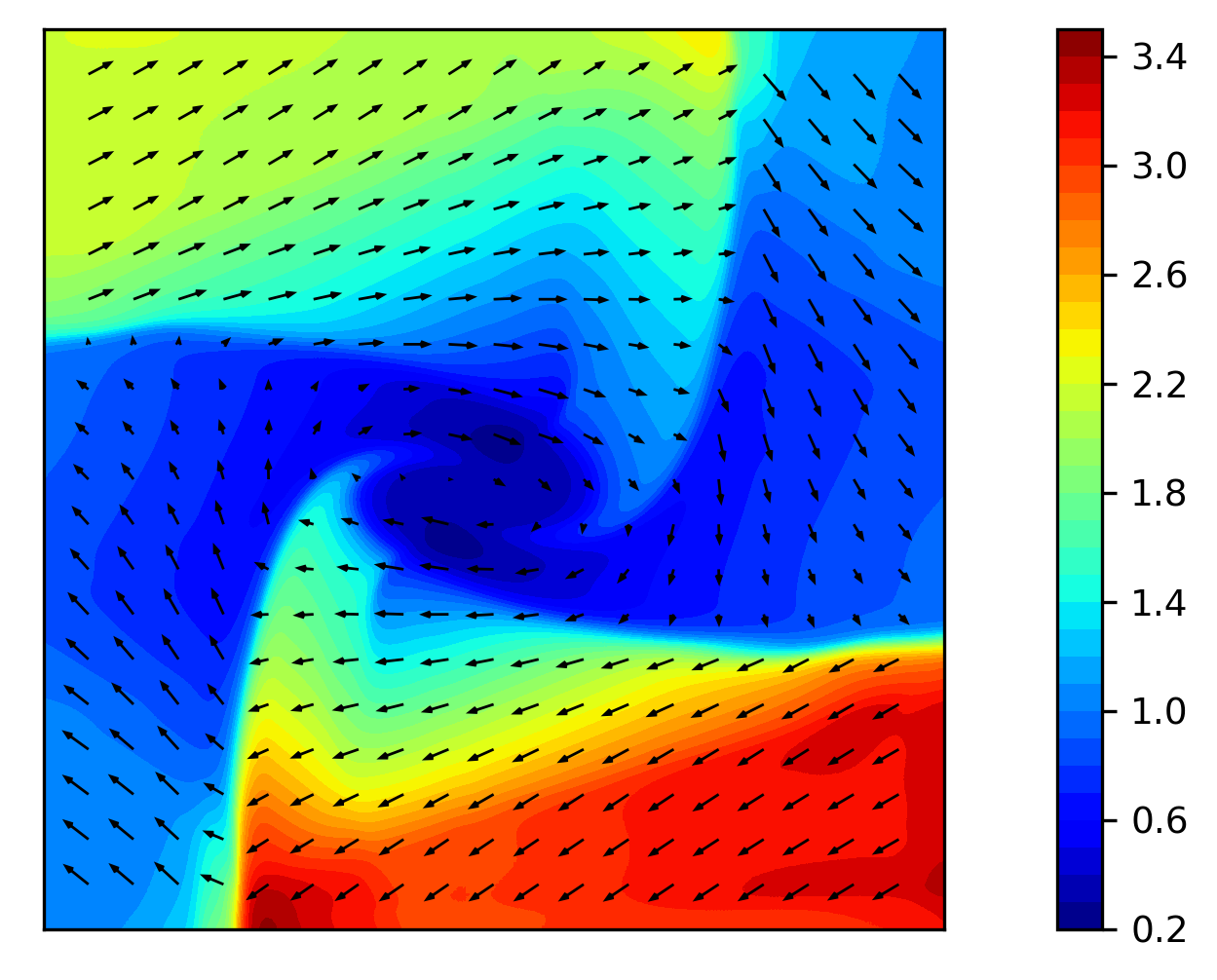}} &
            \raisebox{-0.5\height}{\includegraphics[width = .33\textwidth]{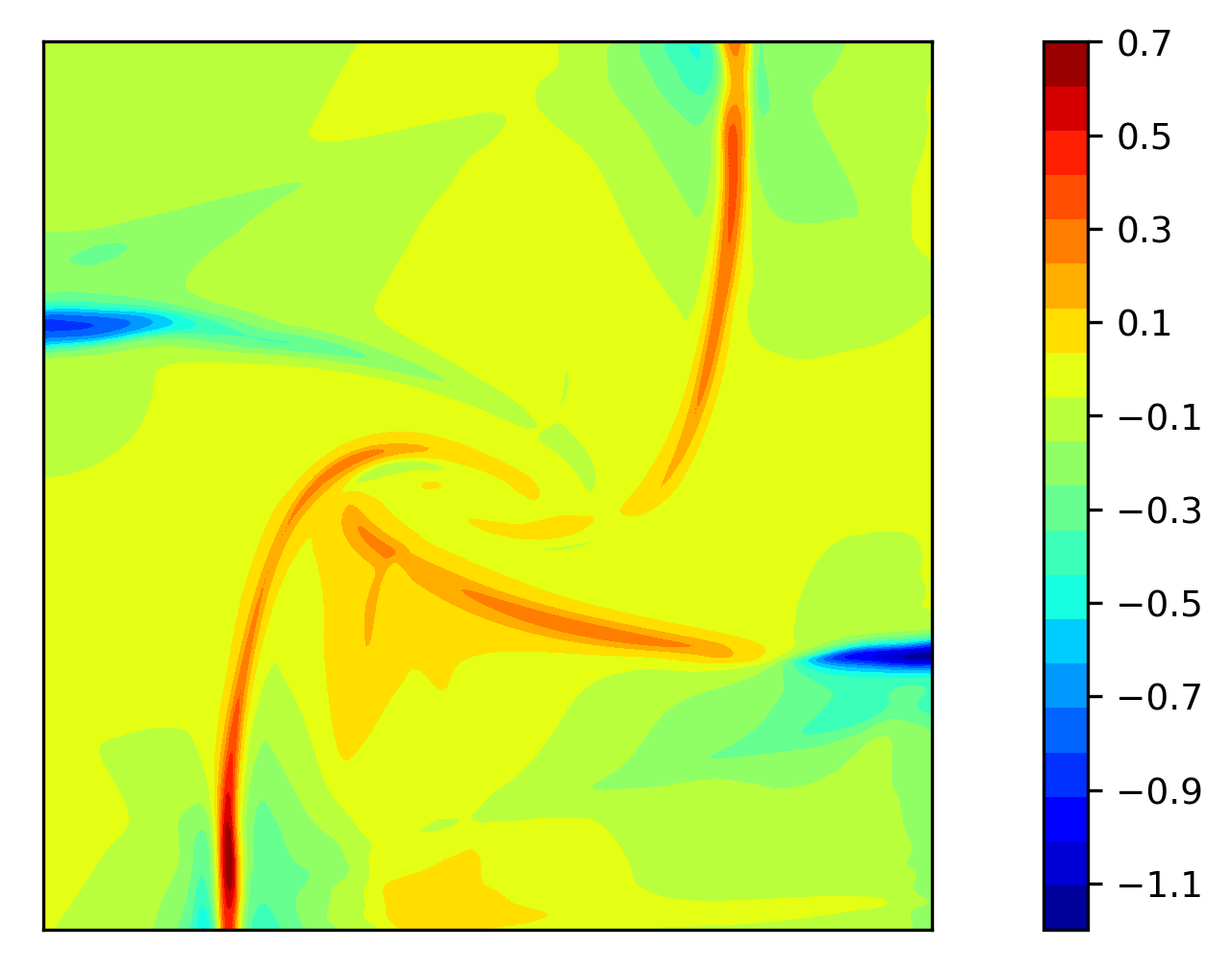}}
            \\ ~ \\
            WE-PINN  $\quad \quad $ & \mcDGNet ($2\%$) $\quad \quad $ & $\rho_\text{DG} - \rho_\text{\mcDGNet}$  $\quad \quad $
            \\
            \raisebox{-0.5\height}{\includegraphics[width = .33\textwidth]{Figs/2D_BenchMarks/configuration6_WE_PINNs.png}} &
            \raisebox{-0.5\height}{\includegraphics[width = .33\textwidth]{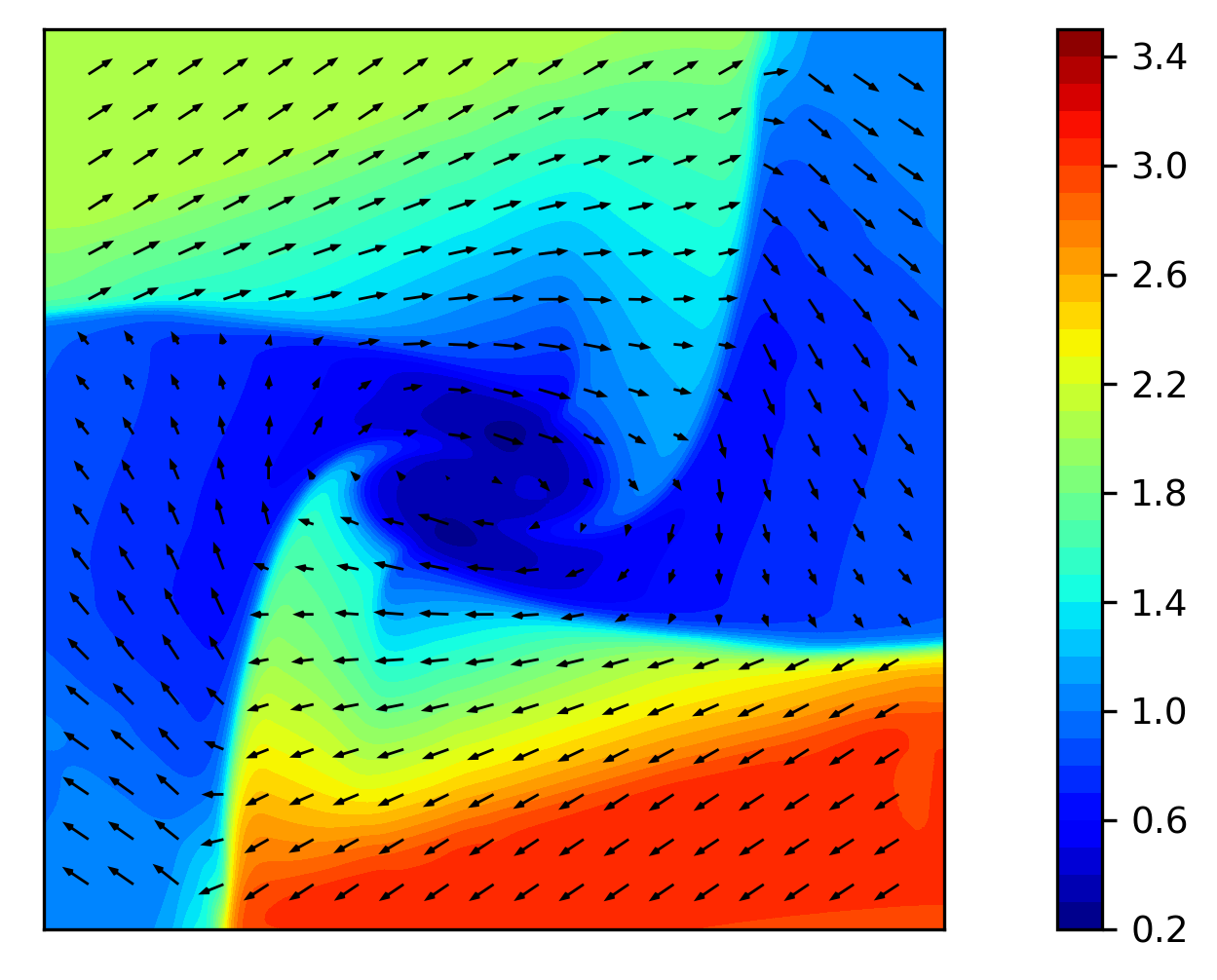}} &
            \raisebox{-0.5\height}{\includegraphics[width = .33\textwidth]{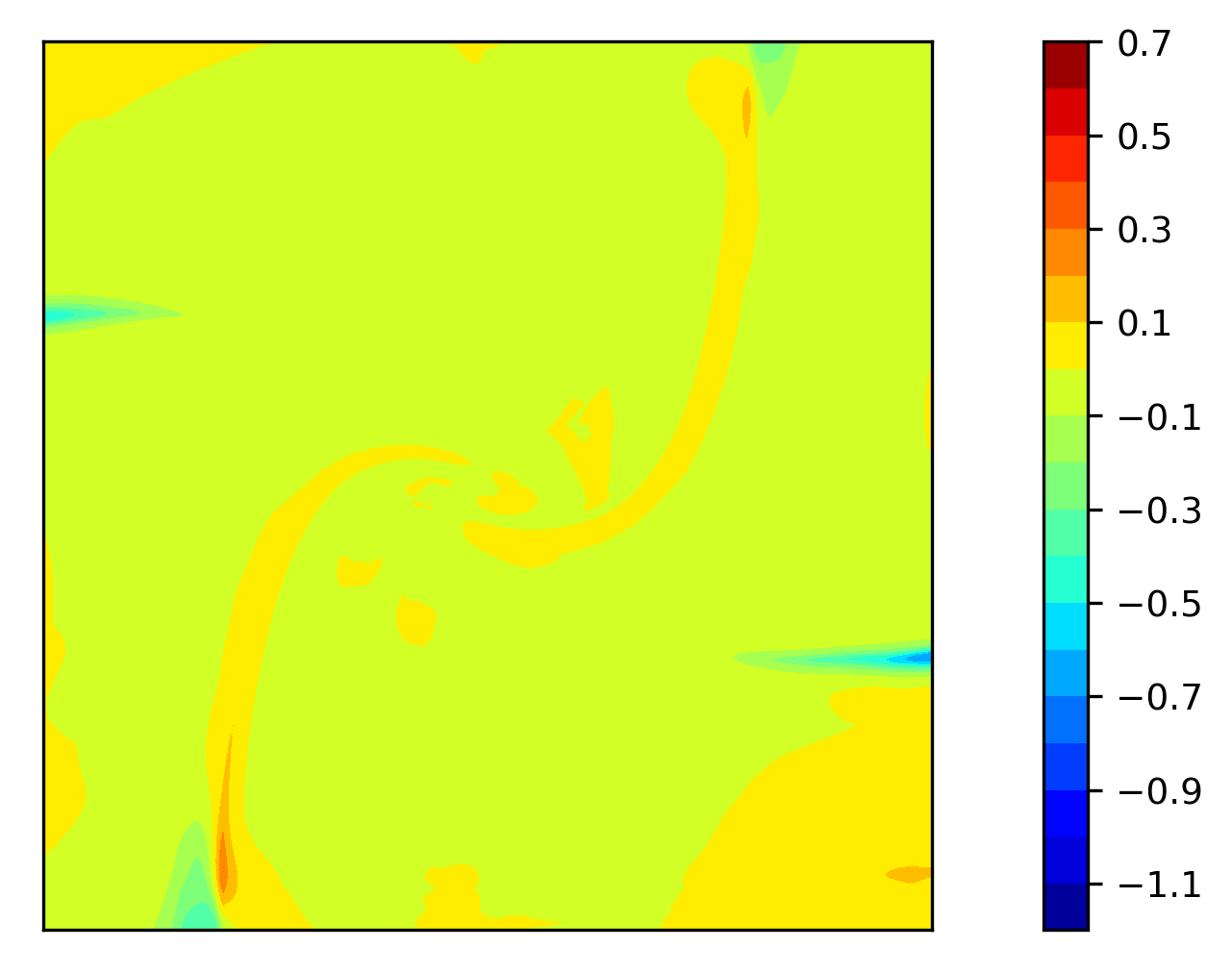}}
        \end{tabular*}
        \caption{{\bf 2D Euler Benchmarks:} predicted density fields obtained by \nDGNet and \mcDGNet approaches and corresponding prediction pointwise error $\rho_{\text{DG}} - \rho_{\text{pred}}$ and DG and WE-PINN solutions at time $T = 0.4s$ for configuration 6 -  Model 2.}
        \figlab{2D_euler_configuration_6_solutions_model2}
\end{figure}

\begin{figure}[htb!]
    \centering
        \begin{tabular*}{\textwidth}{c@{\hskip -0.0cm} c@{\hskip -0.0cm}  c@{\hskip -0.0cm}}
            \centering

            DG $\quad \quad $ & \nDGNet $\quad \quad $ & $\rho_\text{DG} - \rho_\text{\nDGNet}$  $\quad \quad $
            \\
            \raisebox{-0.5\height}{\includegraphics[width = .33\textwidth]{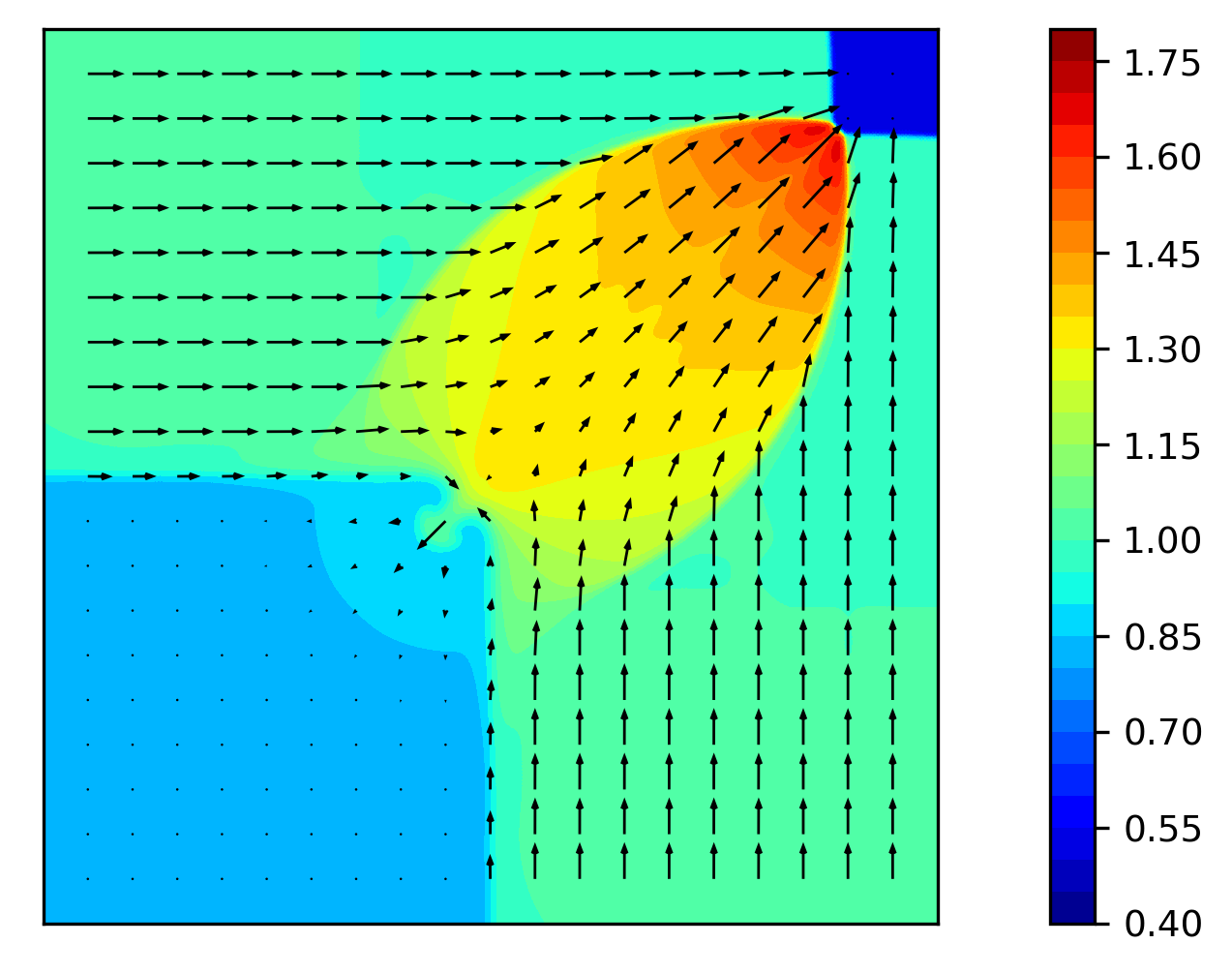}} & 
            \raisebox{-0.5\height}{\includegraphics[width = .33\textwidth]{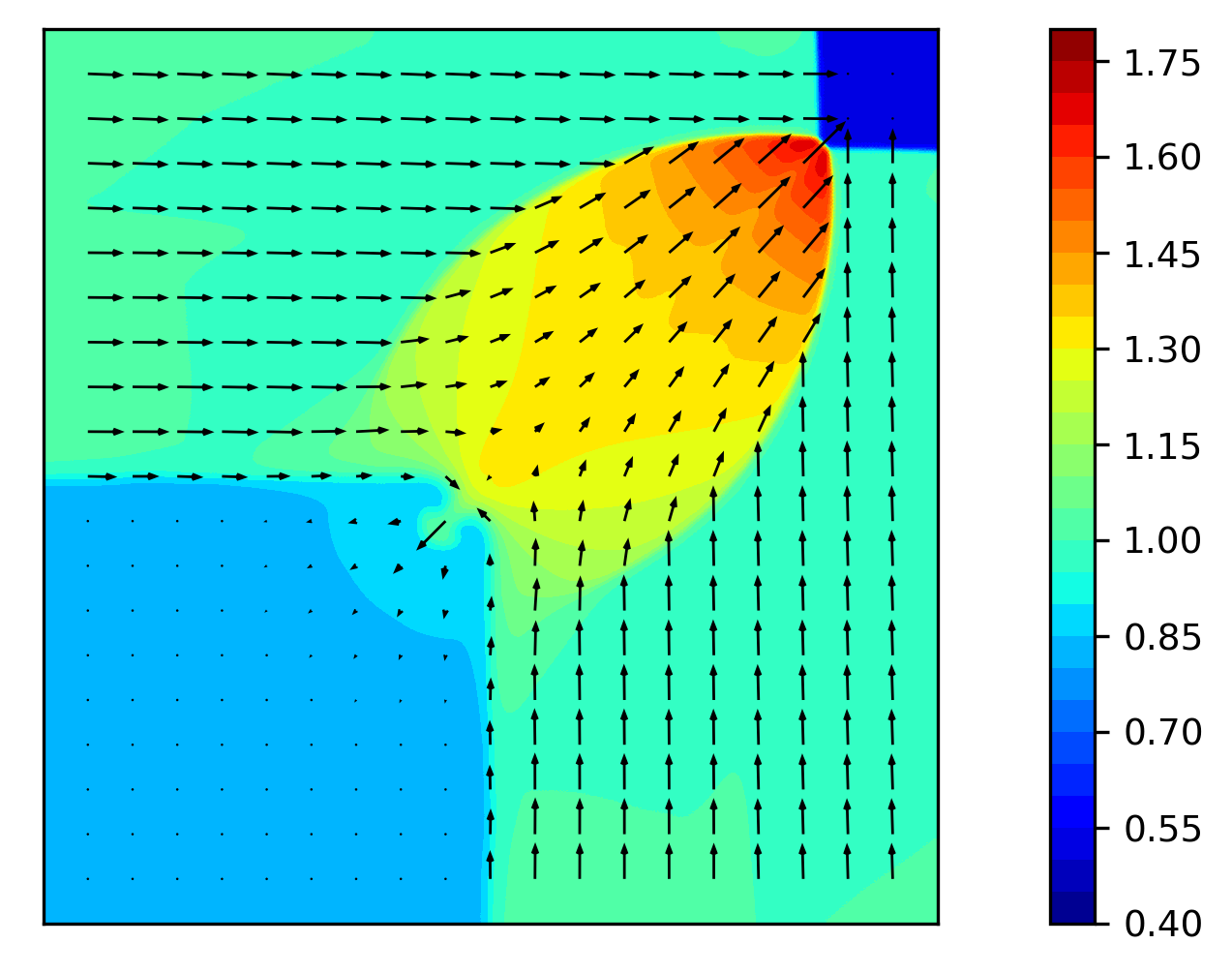}} &
            \raisebox{-0.5\height}{\includegraphics[width = .33\textwidth]{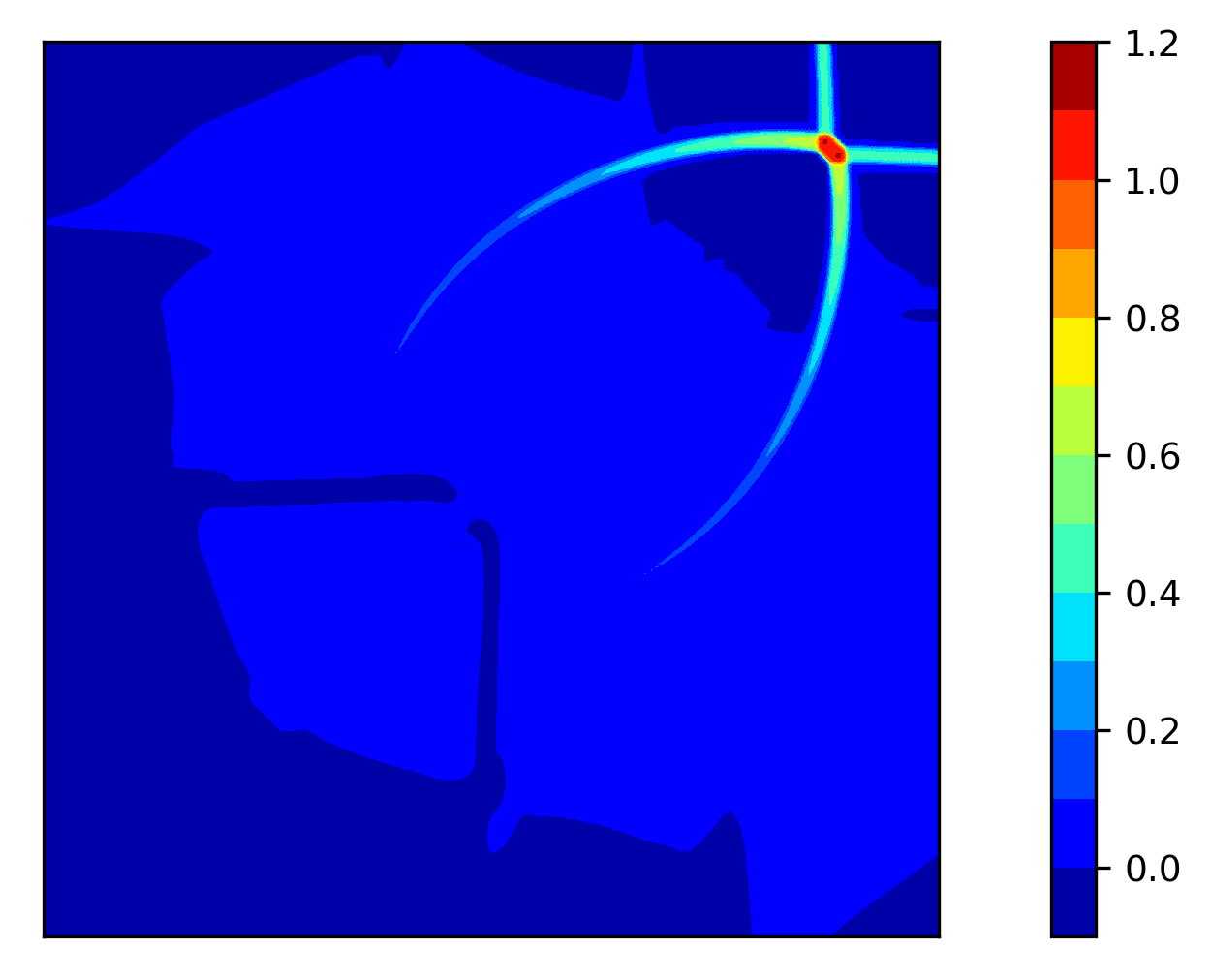}}
            \\ ~ \\
            WE-PINN  $\quad \quad $ & \mcDGNet ($2\%$) $\quad \quad $ & $\rho_\text{DG} - \rho_\text{\mcDGNet}$  $\quad \quad $
            \\
            \raisebox{-0.5\height}{{\bf unobtainable}  $\quad \quad $} &
            \raisebox{-0.5\height}{\includegraphics[width = .33\textwidth]{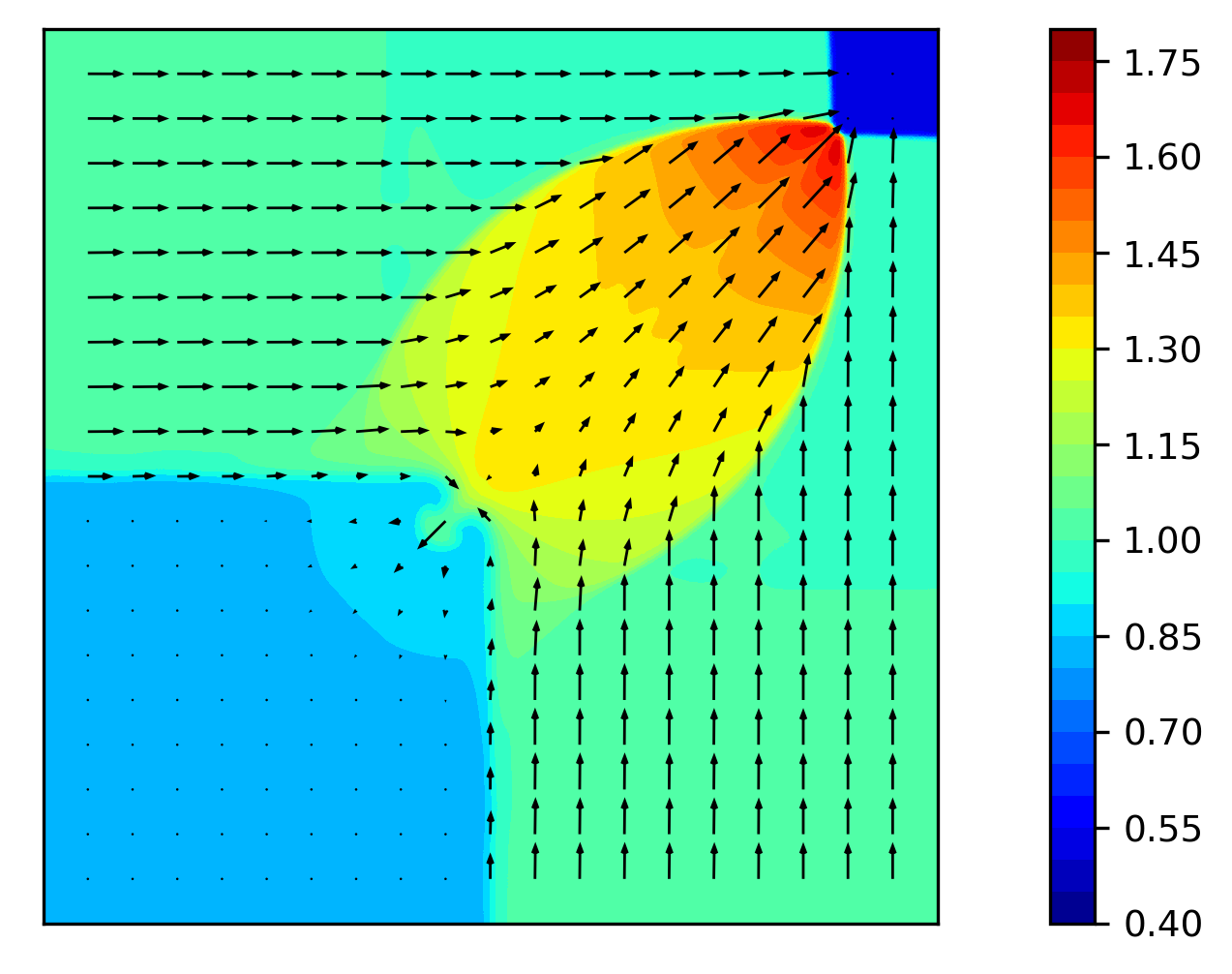}} &
            \raisebox{-0.5\height}{\includegraphics[width = .33\textwidth]{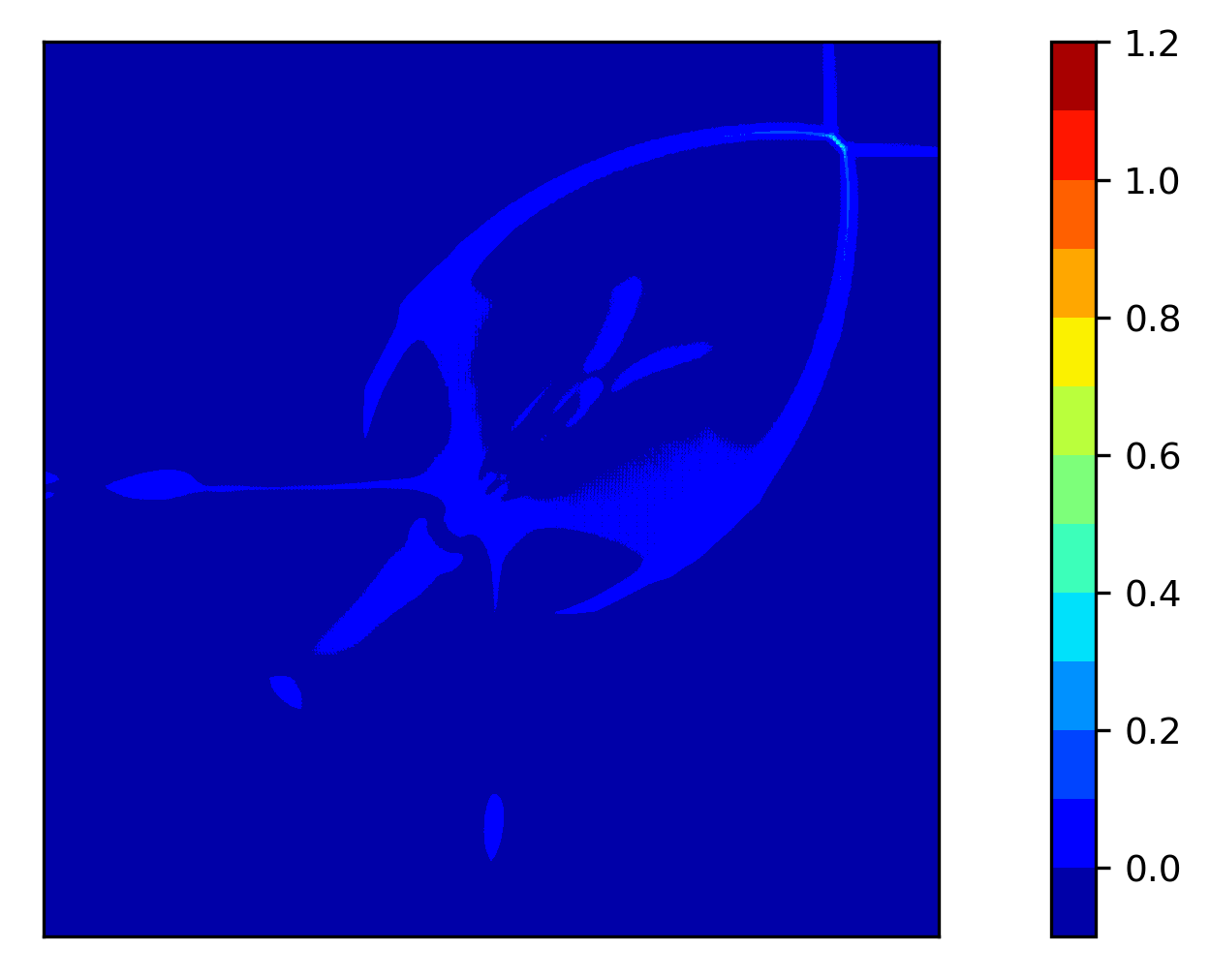}}
        \end{tabular*}
        \caption{{\bf 2D Euler Benchmarks} predicted density fields obtained by \nDGNet and \mcDGNet approaches and corresponding prediction pointwise error $\rho_{\text{DG}} - \rho_{\text{pred}}$ and DG and WE-PINN solutions at time $T = 0.25s$ for configuration 12 -  Model 3.}
        \figlab{2D_euler_configuration_12_solutions_model2}
\end{figure}

For configuration 6  we deploy unstructured triangular meshes with $K = 60108$ elements and $K = 262144$ elements, referred to as Model 1 and Model 2, respectively. Similarly, the domain for configuration 12  is decomposed with $K = 262144$ elements, and named as Model 3. For data generation, we generate training data and validation data by solving the 2D Euler equations with $\gamma \in \LRc{1.2,1.6}$ and with $\gamma = 1.4$ for Model 1  within the training time interval $\LRs{0,0.16}s$. Three test data sets are produced for Model 1 and Model 2 with configuration 6 for a test period of $\LRc{0, 0.8}s$ and for Model 3 with configuration 12 for the test time interval $\LRc{0,0.25}s$. Fixed time step sizes of $\dt = 0.0004s$, $\dt = 0.0002s$, and $\dt = 0.00025s$ are employed for solving 2D Euler equations for data generation in Model 1, Model 2, and Model 3, respectively.

\cref{fig:2D_euler_configuration_6_relative_error} shows the average relative $L^2-$error of conservative components $\LRp{\rho, \rho u, \rho v, E}$ obtained by \nDGNet and \mcDGNet (with $2\%$ noise) approaches at different time steps for Model 1, Model 2 and Model 3. It can be seen that \mcDGNet consistently provides better solutions than \nDGNet at all time steps for all three models. Again, this is because \mcDGNet approach implicitly regularizes the matching between tangent slope surrogate models and ground truth up to second-order derivative during training, thus the tangent slope surrogate models are more stable and generalizable for long-term predictions as discussed in \cref{sect:Data_rand}. For generalization, both \nDGNet and \mcDGNet approaches are capable of generalizing well for different configurations and meshes. This feature originates from the normalization step in the \oDGNet framework, where the neural network receives the normalized data, rather than the physical data. Moreover, in terms of shock-capturing capability, \mcDGNet method is again superior to \nDGNet for all scenarios, as presented in \cref{fig:2D_euler_configuration_6_solutions}, \cref{fig:2D_euler_configuration_6_solutions_model2} and \cref{fig:2D_euler_configuration_12_solutions_model2}. Specifically, the density prediction by the \nDGNet has higher pointwise error  $\rho_{\text{DG}} - \rho_{\text{pred}}$ than that of the \mcDGNet \hspace{-1ex}, especially at the shock locations.

Additionally, we compare the \oDGNet approaches against the WE-PINN approach \cite{liu2024discontinuity}. Although  WE-PINN demonstrates the capability of capturing a sharper shock, it is unable to either predict beyond the training time horizon or generalize to different configurations and meshes. WE-PINN  is implemented for configuration 6 with $400 \times 400$ mesh grids up to $T = 4s$, as shown in \cref{fig:2D_euler_configuration_6_solutions}, \cref{fig:2D_euler_configuration_6_solutions_model2}.  Beyond that time point, no solution is achievable. Additionally, due to the nature of  PINN approaches, only a specific instance of a problem is trained and solved, and thus the solution for configuration 12 is unobtainable unless a separate PINN is trained. Finally, it is worth noting that our DG and \mcDGNet solutions are aligned with the Lax-Friedrichs numerical flux scheme, which is known for introducing extra dissipation in solutions. This is the reason why our \oDGNet solutions at shock locations have less sharp profiles compared to WE-PINN solutions.

% \clearpage

\subsection{2D Double Mach Reflection} 
\seclab{2d_euler_double_mach}

In this section, we consider the  Double Mach Reflection problem as elaborated in \cite{woodward1984numerical}. The problem models a horizontal Mach 10 shocked flow impinging on a ramp, or wedge, at an angle of 60 degrees relative to the horizontal direction. The geometry and boundary conditions are shown in \cref{fig:double_mach_mesh}. The horizontal axis spans 4 units. The lower edge domain is assigned with an inflow for the first $1/6$ unit, while the remainder is designated as a reflective wall. On the upper boundary ($x_2=1$), time-dependent boundary conditions are applied to allow the shock to propagate into the domain as though it extended to infinity. The vertical axis is 1 unit in length. The left boundary maintains post-shock condition values consistently, and the right boundary is modeled as free outflow. The pre-shock conditions are P = 1.0, $\gamma$ = 1.4, $\bar{v} = 0$, and post-shock conditions are P = 116.5, $\gamma$ = 8, $\bar{v} = 8.25$ ($u = \bar{v} \cos(\frac{\pi}{3}), v = \bar{v} \sin(\frac{\pi}{3})$).

\begin{figure}[htb!]
    \centering
    \includegraphics[width = .8\textwidth]{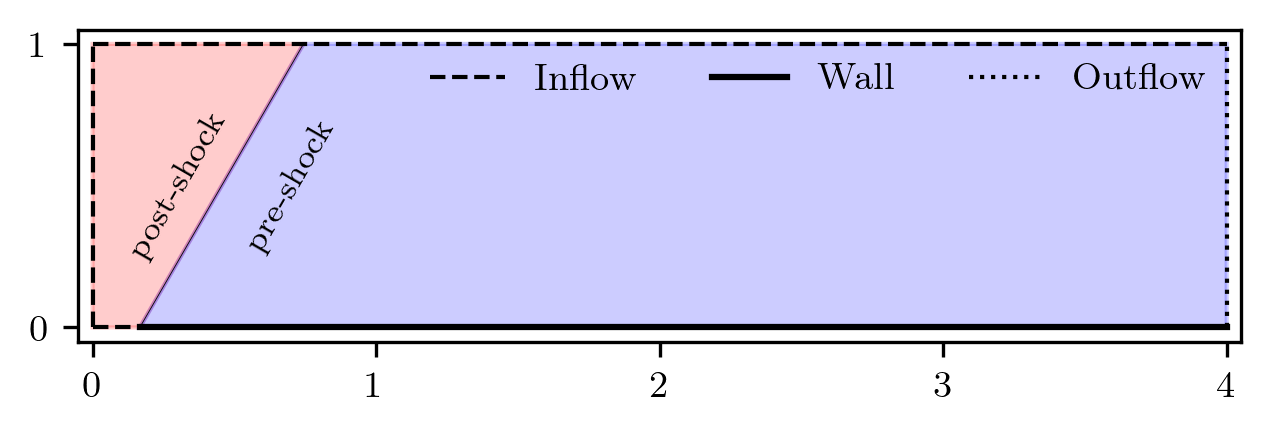}
    \caption{{\bf 2D Double Mach Reflection:} domain discretization and boundary conditions for Double Mach Reflection problem.} 
    \figlab{double_mach_mesh}
\end{figure}

\begin{figure}[htb!]
    \centering
        \begin{tabular*}{\textwidth}{c@{\hskip -0.0cm} c@{\hskip -0.0cm}}
            \centering
            Model 1 & Model 2
            \\
            \raisebox{-0.5\height}{\resizebox{.48\textwidth}{!}{\input{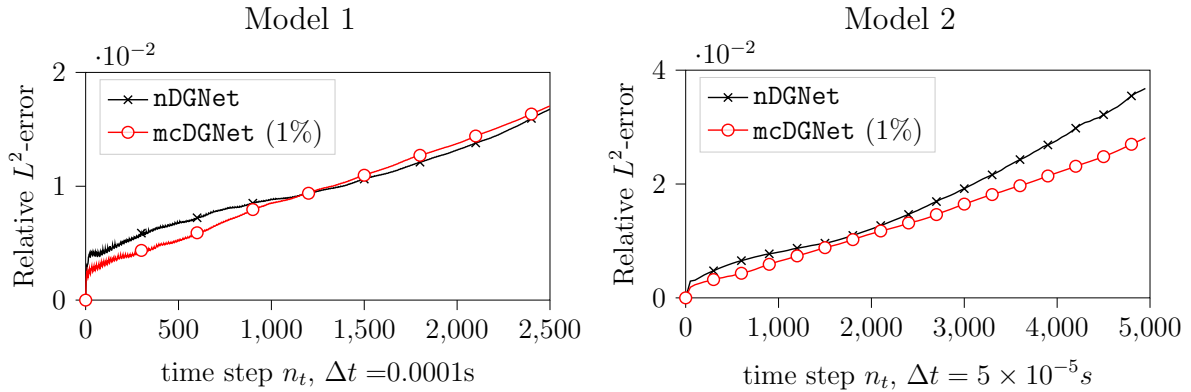}}} &
            \raisebox{-0.5\height}{\resizebox{.48\textwidth}{!}{% This file was created with tikzplotlib v0.10.1.
\begin{tikzpicture}

\definecolor{darkgray176}{RGB}{176,176,176}
\definecolor{green01270}{RGB}{0,127,0}
\definecolor{lightgray204}{RGB}{204,204,204}

\begin{axis}[
width=8.5cm,  % Set the width of the plot
height=5cm,  % Set the height of the plot
legend cell align={left},
legend style={
  fill opacity=0.8,
  draw opacity=1,
  text opacity=1,
  at={(0.03,0.97)},
  anchor=north west,
  draw=lightgray204
},
tick align=outside,
tick pos=left,
x grid style={darkgray176},
xlabel={time step \(\displaystyle n_t\), \(\displaystyle \Delta t = 5 \times  10^{-5}s\)},
xmin=0, xmax=5000,
xtick style={color=black},
y grid style={darkgray176},
ylabel={Relative \(\displaystyle L^2\)-error},
ymin=0, ymax=0.04,
ytick={0., 0.02, 0.04},
ytick style={color=black}
]
\addplot [semithick, black, mark=x, mark size=2.5, mark repeat=6, mark options={solid}]
table {%
0 0
50 0.00294220796786249
100 0.00310109369456768
150 0.00355917681008577
200 0.00403173547238111
250 0.00441270181909204
300 0.00475558172911406
350 0.0051233284175396
400 0.00544784311205149
450 0.00575496070086956
500 0.00604415033012629
550 0.0063104135915637
600 0.00653785048052669
650 0.00675459951162338
700 0.00698118703439832
750 0.00720017589628696
800 0.00738467928022146
850 0.00758580677211285
900 0.00773714855313301
950 0.00787047669291496
1000 0.00802975799888372
1050 0.00817164964973927
1100 0.00834531709551811
1150 0.00856535043567419
1200 0.00871903635561466
1250 0.00886312499642372
1300 0.00900760479271412
1350 0.00918563827872276
1400 0.00931850075721741
1450 0.00946636777371168
1500 0.00960441119968891
1550 0.00976783968508244
1600 0.00998095329850912
1650 0.0102059300988913
1700 0.0104526644572616
1750 0.0107061825692654
1800 0.0109720434993505
1850 0.0112438406795263
1900 0.0115380957722664
1950 0.0118134450167418
2000 0.0120947323739529
2050 0.012392308562994
2100 0.0127017628401518
2150 0.0130107961595058
2200 0.0133494790643454
2250 0.0136803891509771
2300 0.0139968134462833
2350 0.0143418526276946
2400 0.0146536473184824
2450 0.0150393079966307
2500 0.0153717938810587
2550 0.0157747622579336
2600 0.0161400754004717
2650 0.0165644437074661
2700 0.0169205814599991
2750 0.0173531789332628
2800 0.017662838101387
2850 0.017992565408349
2900 0.0183664746582508
2950 0.0188158769160509
3000 0.0191853474825621
3050 0.0196219589561224
3100 0.0200128816068172
3150 0.0204640757292509
3200 0.0208942666649818
3250 0.0212533511221409
3300 0.0216090250760317
3350 0.0220200922340155
3400 0.022441528737545
3450 0.0230131819844246
3500 0.023436551913619
3550 0.0238826647400856
3600 0.0242913626134396
3650 0.0246553681790829
3700 0.0250850170850754
3750 0.0256106201559305
3800 0.0260268151760101
3850 0.0265570487827063
3900 0.0268622655421495
3950 0.0272474996745586
4000 0.0276729576289654
4050 0.0282306578010321
4100 0.0286415759474039
4150 0.0291949398815632
4200 0.0298067331314087
4250 0.0303649567067623
4300 0.0308037921786308
4350 0.0310109481215477
4400 0.0314263626933098
4450 0.0316609777510166
4500 0.0321795120835304
4550 0.032650001347065
4600 0.0331396609544754
4650 0.0335585102438927
4700 0.034173309803009
4750 0.0348190516233444
4800 0.0354771502315998
4850 0.0360066331923008
4900 0.0363688617944717
4950 0.0367858186364174
};
\addlegendentry{\nDGNet }
\addplot [semithick, red, mark=*, mark size=2.5, mark repeat=6, mark options={solid,fill=white}]
table {%
0 0
50 0.00185945606790483
100 0.00229884358122945
150 0.00254673417657614
200 0.00281214038841426
250 0.00299473735503852
300 0.00321188336238265
350 0.00342873530462384
400 0.00355484010651708
450 0.00376454228535295
500 0.00388175901025534
550 0.00411993265151978
600 0.00432093441486359
650 0.00450918078422546
700 0.00473797041922808
750 0.00500950403511524
800 0.0053280838765204
850 0.00561950216069818
900 0.00589195359498262
950 0.00616638921201229
1000 0.00641738018020988
1050 0.00663307681679726
1100 0.00685751205310225
1150 0.00711154285818338
1200 0.0073708794079721
1250 0.00759987253695726
1300 0.00782570149749517
1350 0.00807228125631809
1400 0.00831832736730576
1450 0.00858076941221952
1500 0.00881906878203154
1550 0.00902944058179855
1600 0.00922852661460638
1650 0.00944296177476645
1700 0.00969916768372059
1750 0.00995008088648319
1800 0.0102288443595171
1850 0.0105136577039957
1900 0.0107655646279454
1950 0.0110342456027865
2000 0.0112655553966761
2050 0.0114914821460843
2100 0.0117502501234412
2150 0.0119383623823524
2200 0.0121518122032285
2250 0.0123711880296469
2300 0.0125963995233178
2350 0.0128607451915741
2400 0.0131563367322087
2450 0.0133499316871166
2500 0.0135231548920274
2550 0.0137665895745158
2600 0.014021928422153
2650 0.0143108898773789
2700 0.0146317053586245
2750 0.0149617847055197
2800 0.0152548458427191
2850 0.0155422613024712
2900 0.0158653873950243
2950 0.0161785557866096
3000 0.0164714008569717
3050 0.0167608167976141
3100 0.0170463230460882
3150 0.0173423998057842
3200 0.017627477645874
3250 0.0179015751928091
3300 0.0181468781083822
3350 0.0184178221970797
3400 0.0186411757022142
3450 0.0189240649342537
3500 0.0192035213112831
3550 0.0194631740450859
3600 0.0197016708552837
3650 0.0199628174304962
3700 0.0202321521937847
3750 0.0205118283629417
3800 0.0208009146153927
3850 0.0211157649755478
3900 0.0214102119207382
3950 0.0217018574476242
4000 0.0219809412956238
4050 0.0222612544894218
4100 0.0225387178361416
4150 0.0228168442845345
4200 0.0231170579791069
4250 0.0233991853892803
4300 0.0236762836575508
4350 0.023942107334733
4400 0.0242279022932053
4450 0.0245044250041246
4500 0.0248083546757698
4550 0.0251319222152233
4600 0.0254796165972948
4650 0.0258323084563017
4700 0.0262092649936676
4750 0.0265972074121237
4800 0.0269786976277828
4850 0.0273515842854977
4900 0.027725463733077
4950 0.0281170681118965
};
\addlegendentry{\mcDGNet ($1\%$) }
\end{axis}

\end{tikzpicture}}} 
        \end{tabular*}
        \caption{{\bf 2D Double Mach Reflection:} average relative $L^2$-error over four conservative components $\LRp{\rho, \rho u, \rho v, E}$ for test data obtained by \nDGNet and \mcDGNet approaches at different time steps for Model 1 ({\bf Left}) and Model 2 ({\bf Right}).}
        \figlab{double_mach_error}
\end{figure}

\begin{figure}[htb!]
    \centering
        \begin{tabular*}{\textwidth}{c c@{\hskip -0.0001cm} c@{\hskip -0.002cm} c@{\hskip -0.002cm} c@{\hskip -0.002cm}}
            \centering
            & & DG & \nDGNet  & \mcDGNet  ($1\%$)
            \\
            \multirow{2}{*}{\rotatebox[origin=l]{90}{Model 1 \quad  }} &
            \rotatebox[origin=c]{90}{Pred} &
            \raisebox{-0.5\height}{\includegraphics[width = .31\textwidth]{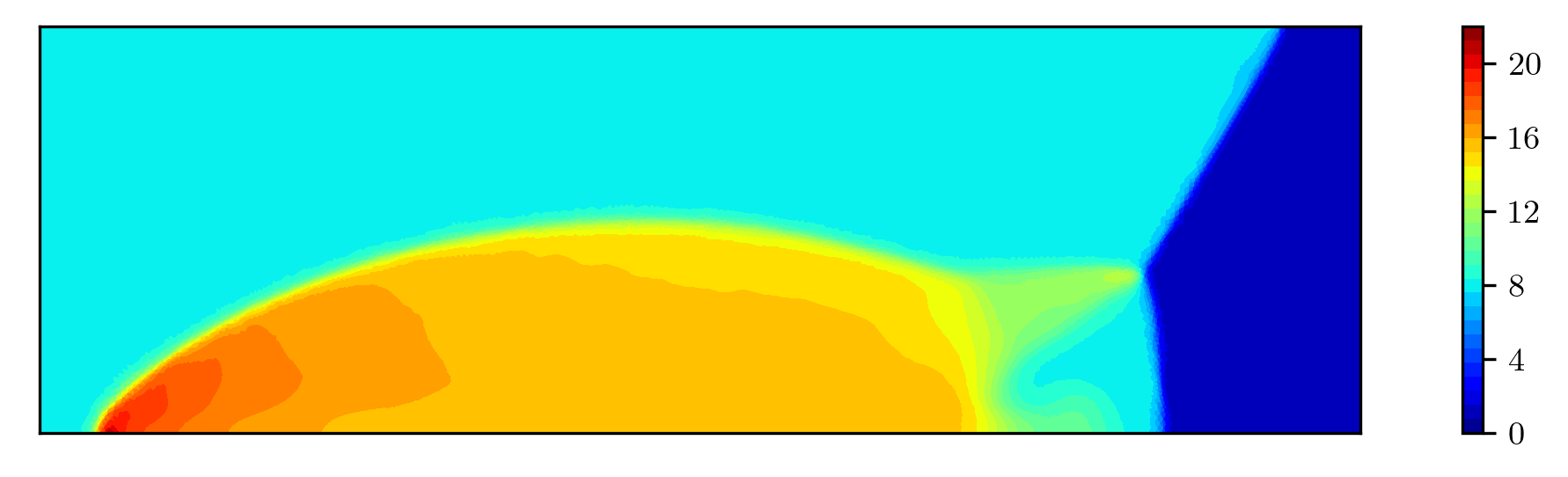}} &
            \raisebox{-0.5\height}{\includegraphics[width = .31\textwidth]{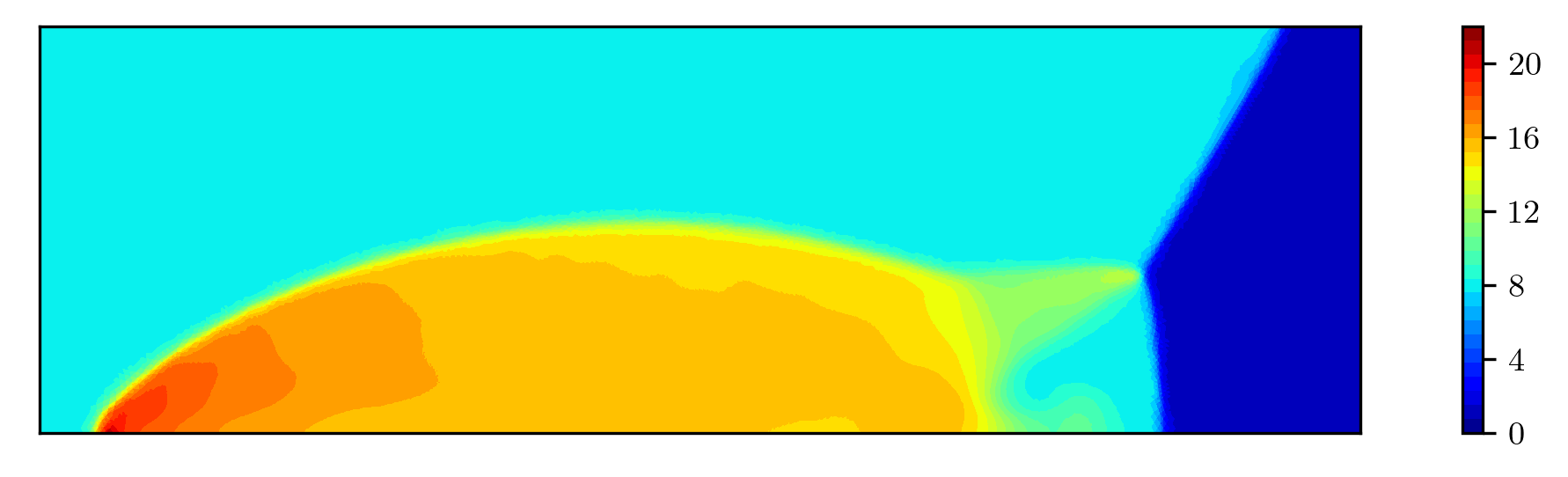}} &
            \raisebox{-0.5\height}{\includegraphics[width = .31\textwidth]{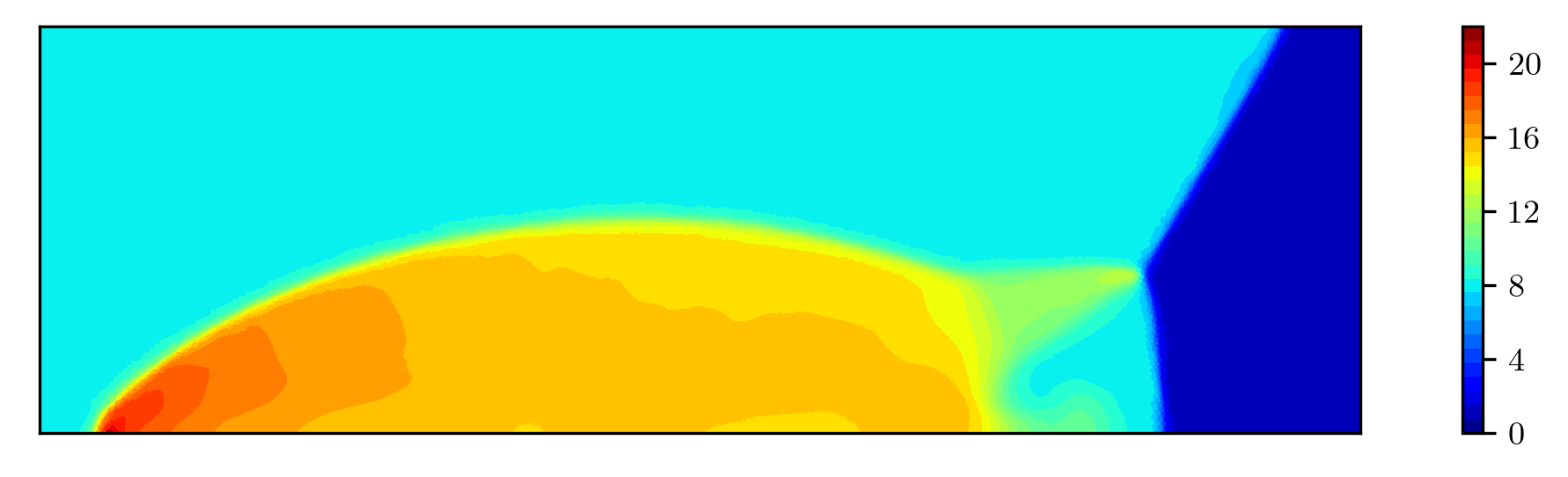}}
            \\
            &
            \rotatebox[origin=c]{90}{Error} &
            \raisebox{-0.5\height}{} &
            \raisebox{-0.5\height}{\includegraphics[width = .31\textwidth]{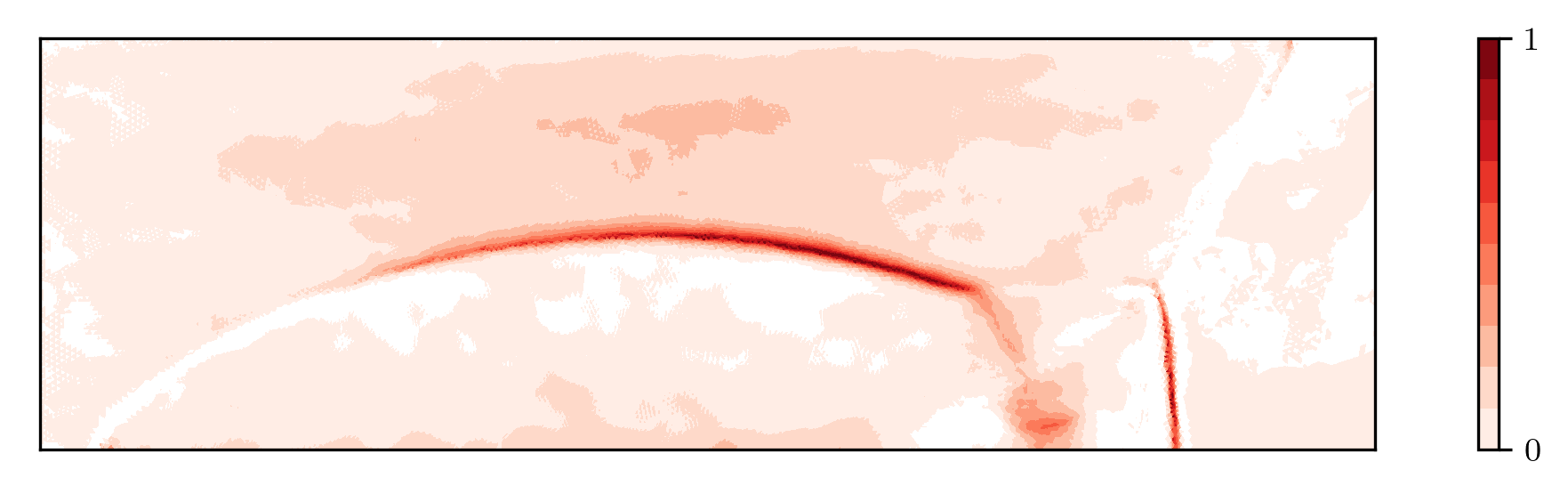}} &
            \raisebox{-0.5\height}{\includegraphics[width = .31\textwidth]{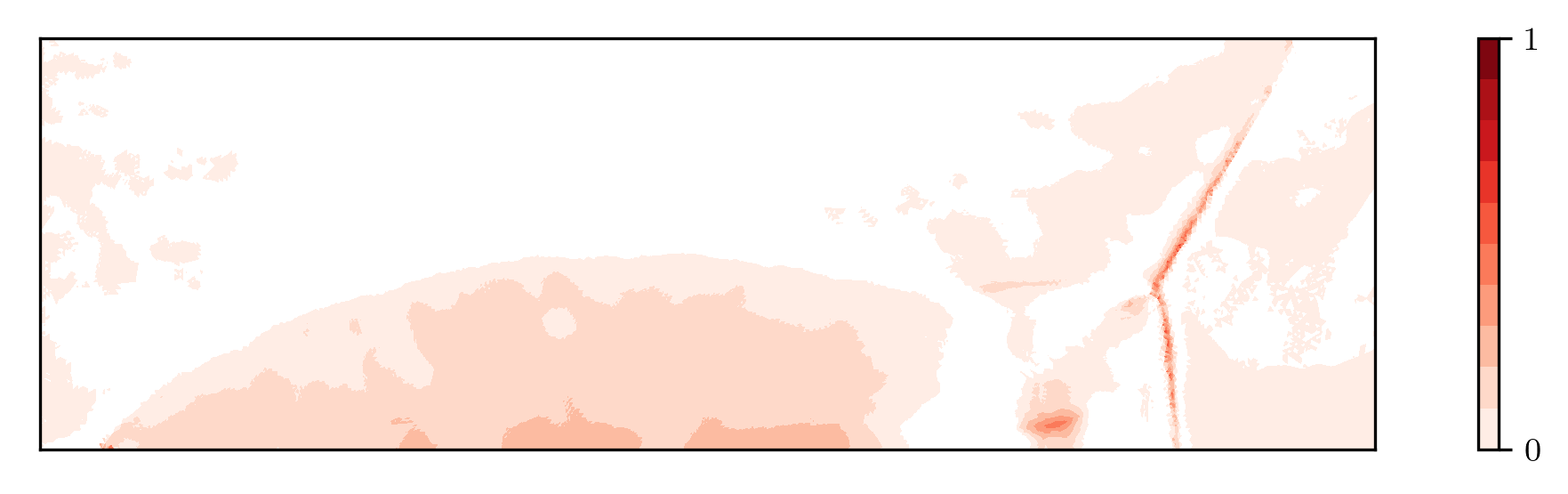}}
            \\
            \multirow{2}{*}{\rotatebox[origin=l]{90}{Model 2 \quad  }} &
            \rotatebox[origin=c]{90}{Pred} &
            \raisebox{-0.5\height}{\includegraphics[width = .31\textwidth]{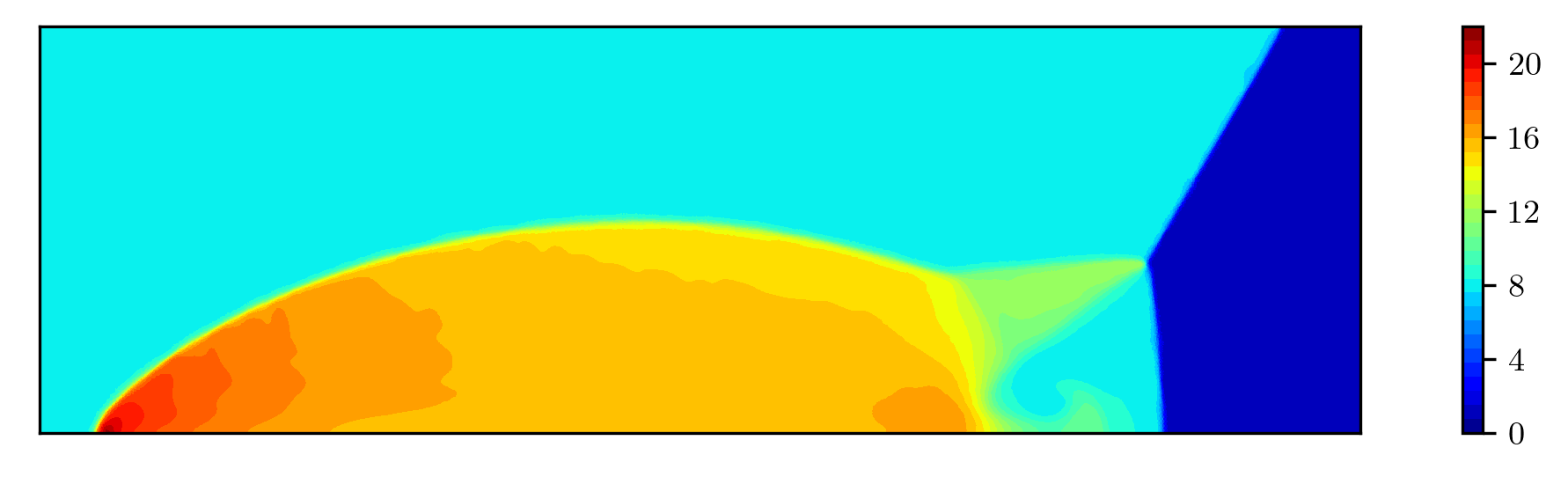}} &
            \raisebox{-0.5\height}{\includegraphics[width = .31\textwidth]{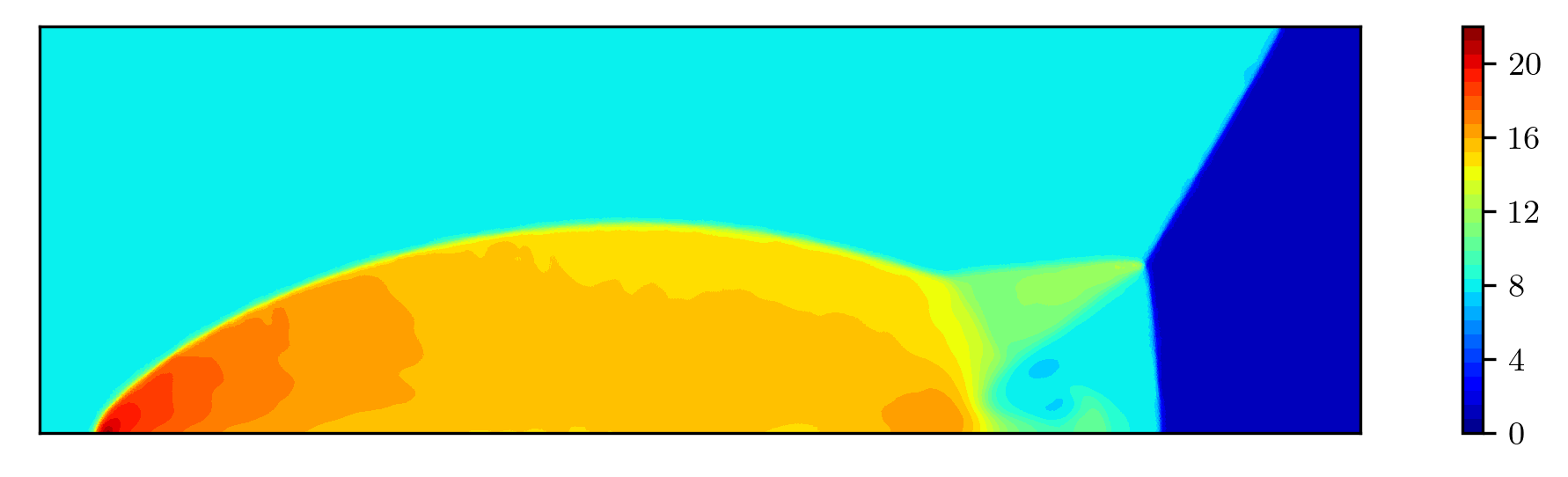}} &
            \raisebox{-0.5\height}{\includegraphics[width = .31\textwidth]{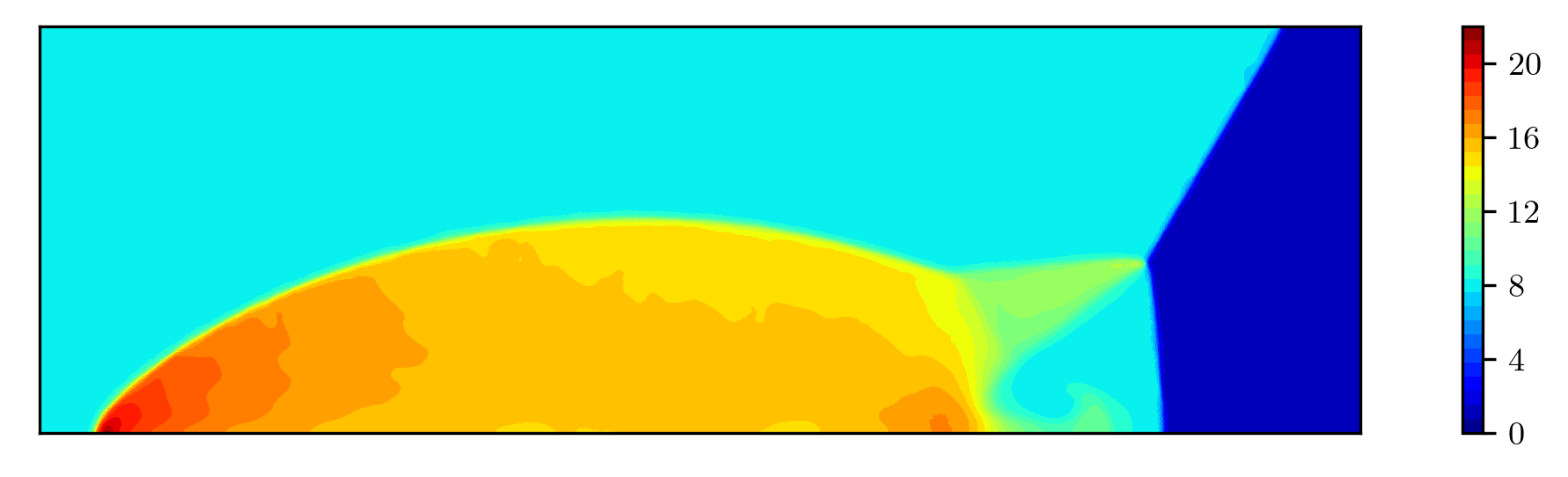}}
            \\
            &
            \rotatebox[origin=c]{90}{Error} &
            \raisebox{-0.5\height}{} &
            \raisebox{-0.5\height}{\includegraphics[width = .31\textwidth]{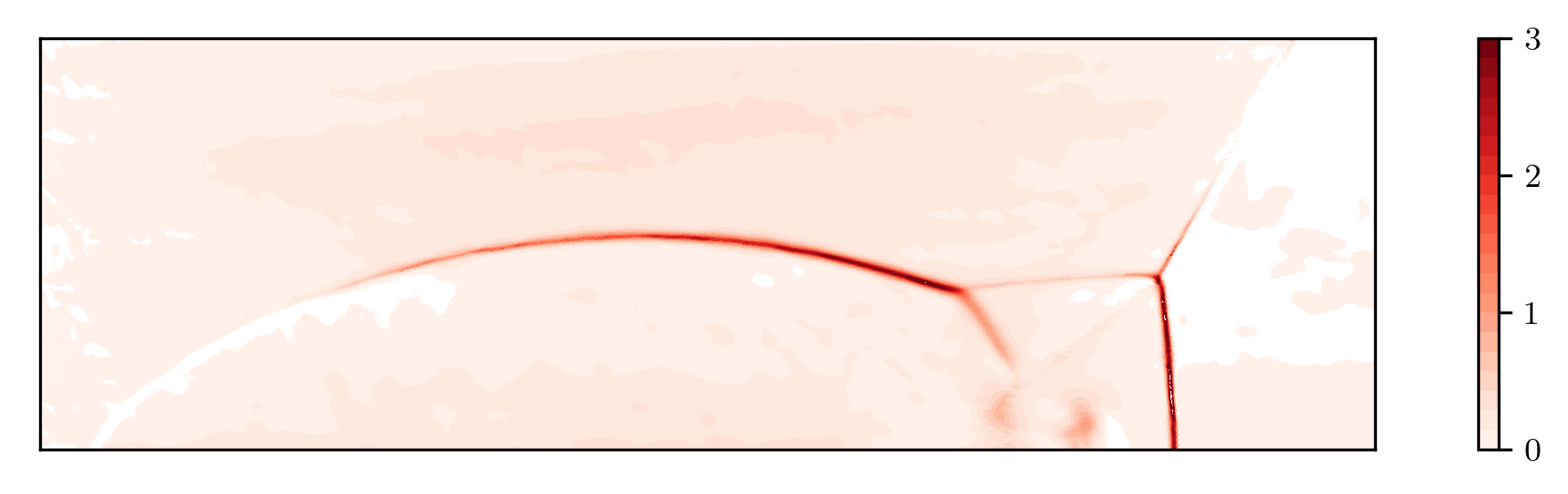}} &
            \raisebox{-0.5\height}{\includegraphics[width = .31\textwidth]{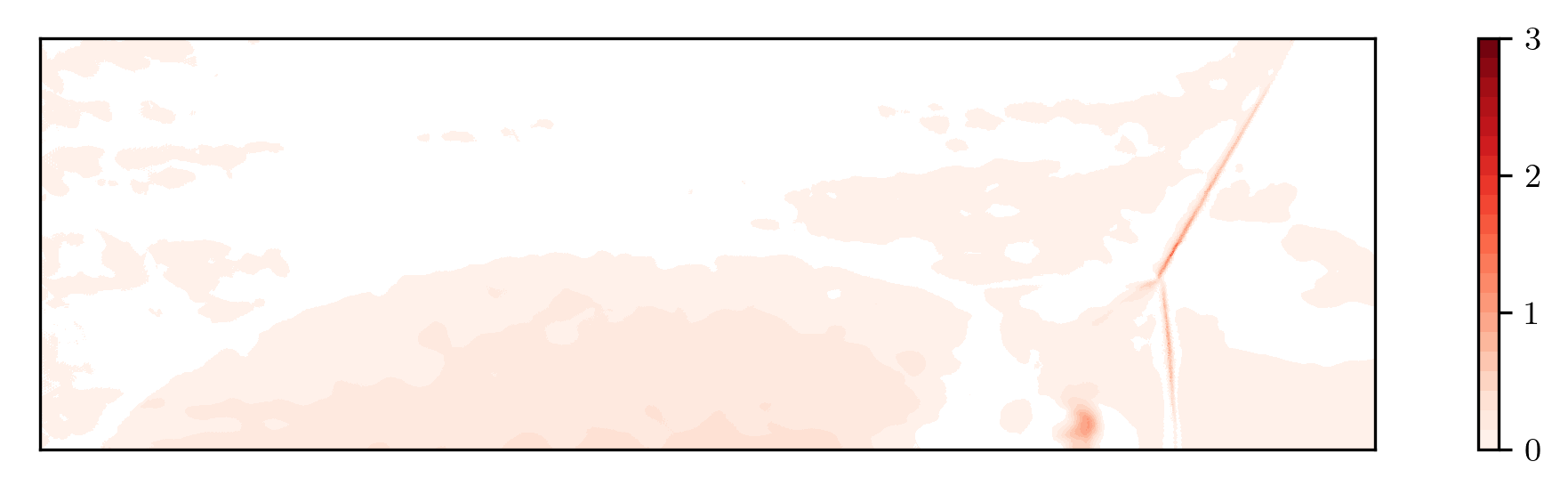}}
        \end{tabular*}
        \caption{{\bf 2D Double Mach Reflection:} predicted density field obtained by \nDGNet and \mcDGNet approaches and corresponding prediction pointwise error $\rho_{\text{DG}} - \rho_{\text{pred}}$  at time step $T_\text{test} = 0.2s$.} 
        \figlab{double_prediction_solution_model1_mesh}
\end{figure}

We discretize the computational domain into $K = 60192$ non-uniform triangular elements, referred to as Model 1, and $K = 240768$ non-uniform triangular elements, denoted as Model 2. For data generation, by solving the 2D Euler equation in Model 1, we generate train data with $\gamma \in \LRc{1.2,1.6}$  and validation data with $\gamma = 1.4$ for the pre-shock condition domain within the training time interval $\LRs{0, 0.02}s$. Two test data sets are produced for both Model 1 and Model 2 over the test period of $\LRs{0, 0.25}s$ with $\gamma = 1.4$. A uniform time step size of $\dt = 0.0001s$ is adopted when solving with  Model 1, and $\dt = 5\times 10^{-5}s$ when solving with Model 2.

The average relative $L^2-$error over conservative components for the test data obtained from \nDGNet and \mcDGNet approaches over the test period is presented in \cref{fig:double_mach_error}. Both two \oDGNet methods exhibit nearly identical performance on Model 1. This can be attributed to, as will be discussed in greater detail in \cref{sect:P6_2D_Noise_corruption}, the training data itself spans the entire potential normalized data space, thus the data randomization only offers regularization effects on training, but not the data enrichment effect. Indeed, \mcDGNet regularization induced by data randomization demonstrates its role in providing superior prediction performance for the \mcDGNet approach as applied to the finer mesh settings - Model 2. This test also reveals the generalizability of \oDGNet approaches to finer discretization mesh configuration. Although \nDGNet and \mcDGNet methods yield comparable relative $L^2$-errors across the entire domain,  \mcDGNet method exhibits a greater ability in capturing shocks compared to \nDGNet \hspace{-1ex}, as illustrated in \cref{fig:double_prediction_solution_model1_mesh}. The predicted density field from \nDGNet method at time $T = 0.2s$ has a larger pointwise error along the sharp shock curve compared to the solution obtained from the \mcDGNet counterpart.

% \clearpage

\begin{figure}[htbp]
    \centering
        \begin{tabular*}{\textwidth}{c c@{\hskip -0.01cm} c@{\hskip -0.1cm} c@{\hskip -0.1cm} c@{\hskip -0.1cm} c@{\hskip -0.1cm} c@{\hskip -0.1cm}}
            \centering
            & & \quad Forward & \quad Scramjet & \quad Airfoil & \quad Euler-config6 & \quad Double Mach
            \\
            \multirow{3}{*}{\rotatebox[origin=l]{90}{noise-free sets used for \nDGNet }} &
            \rotatebox[origin=c]{90}{$\snor{\overline{n_1 \fbxaver}} = 1$} &
            \raisebox{-0.5\height}{\includegraphics[width = .18\textwidth]{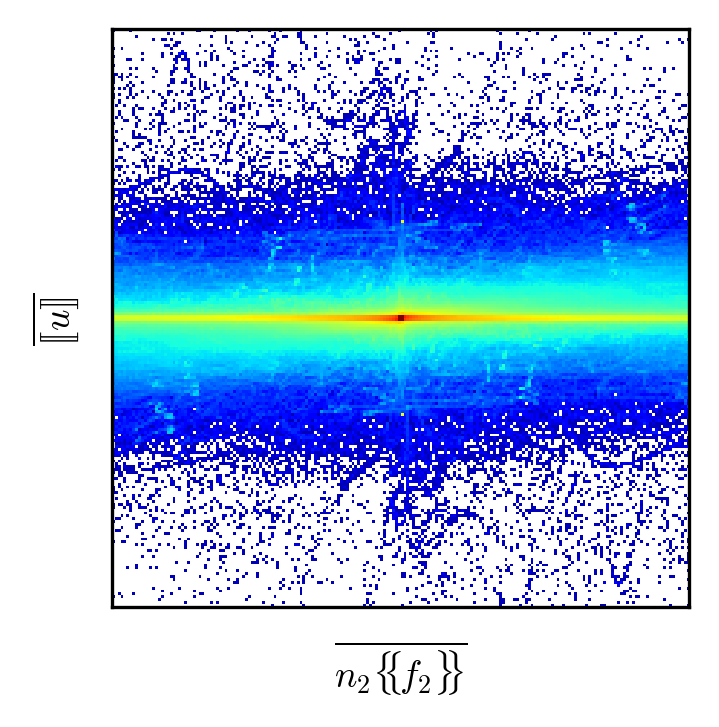}} &
            \raisebox{-0.5\height}{\includegraphics[width = .18\textwidth]{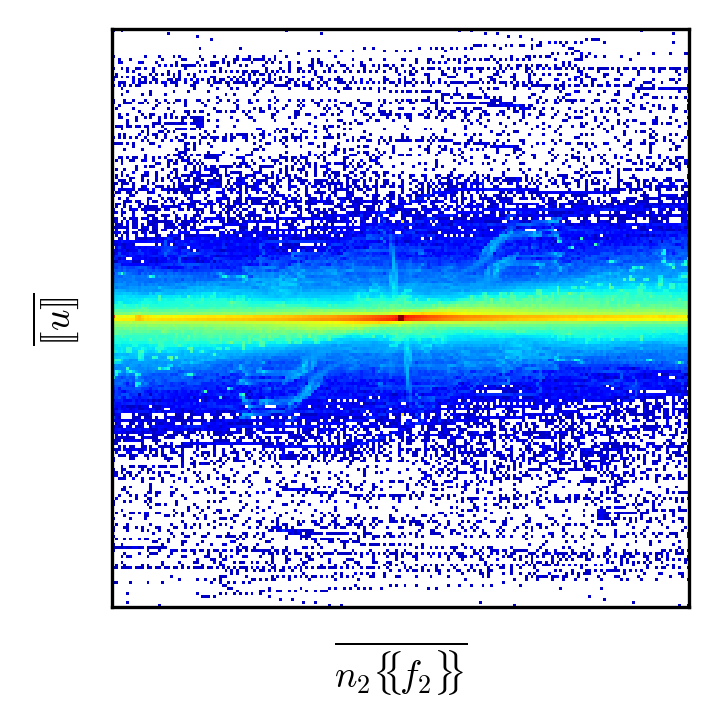}} &
            \raisebox{-0.5\height}{\includegraphics[width = .18\textwidth]{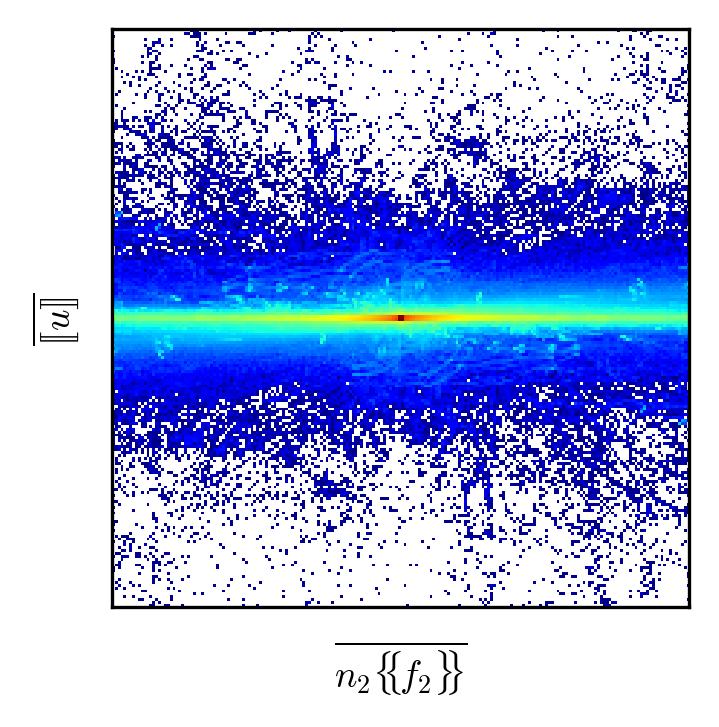}} &
            \raisebox{-0.5\height}{\includegraphics[width = .18\textwidth]{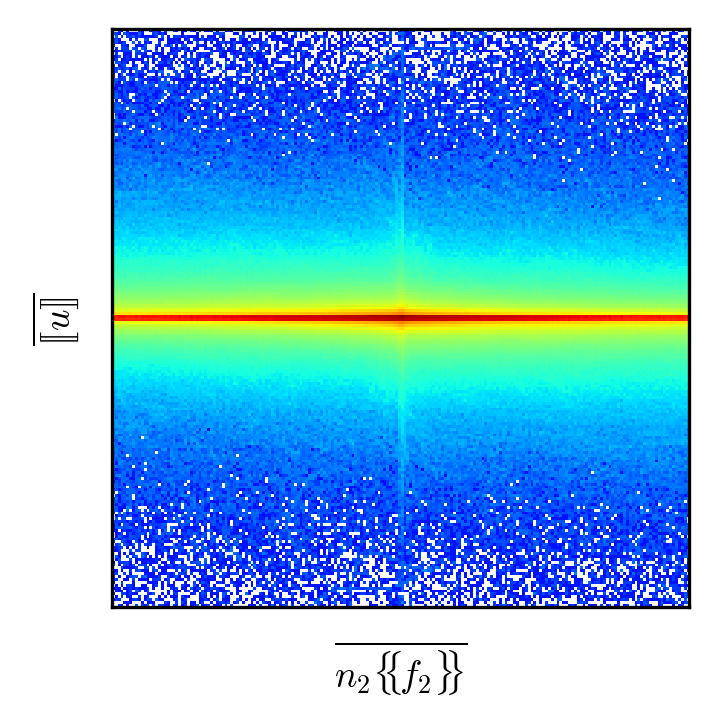}} &
            \raisebox{-0.5\height}{\includegraphics[width = .18\textwidth]{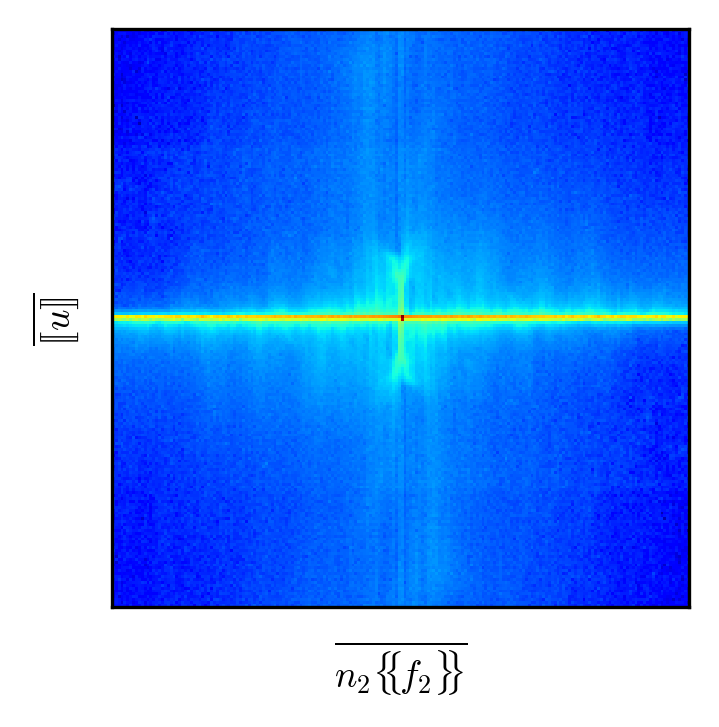}} 
            \\
            &
            \rotatebox[origin=c]{90}{$\snor{\overline{n_2 \fbyaver}} = 1$} &
            \raisebox{-0.5\height}{\includegraphics[width = .18\textwidth]{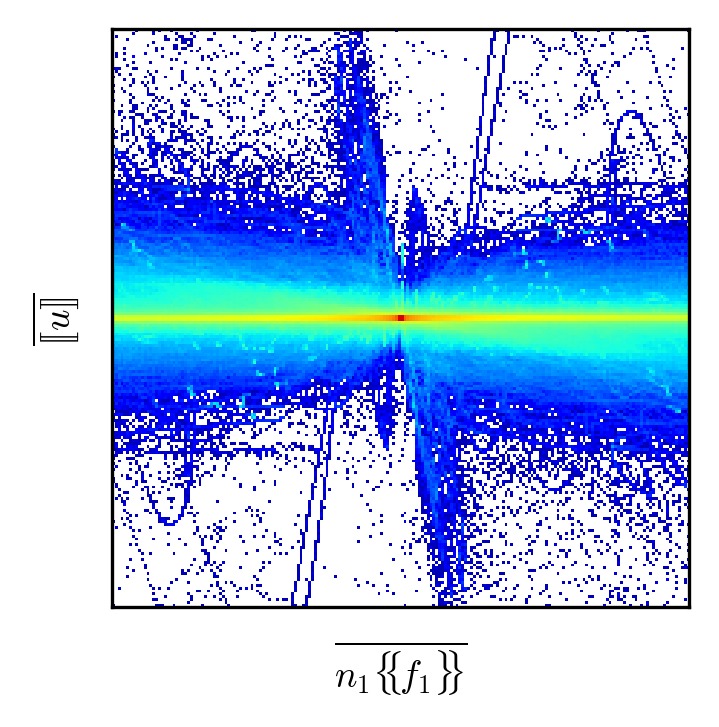}} &
            \raisebox{-0.5\height}{\includegraphics[width = .18\textwidth]{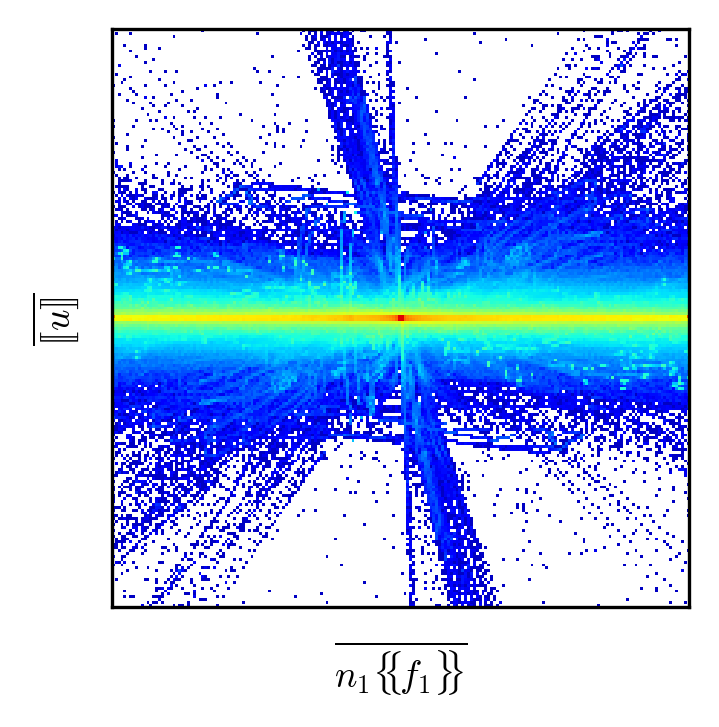}} &
            \raisebox{-0.5\height}{\includegraphics[width = .18\textwidth]{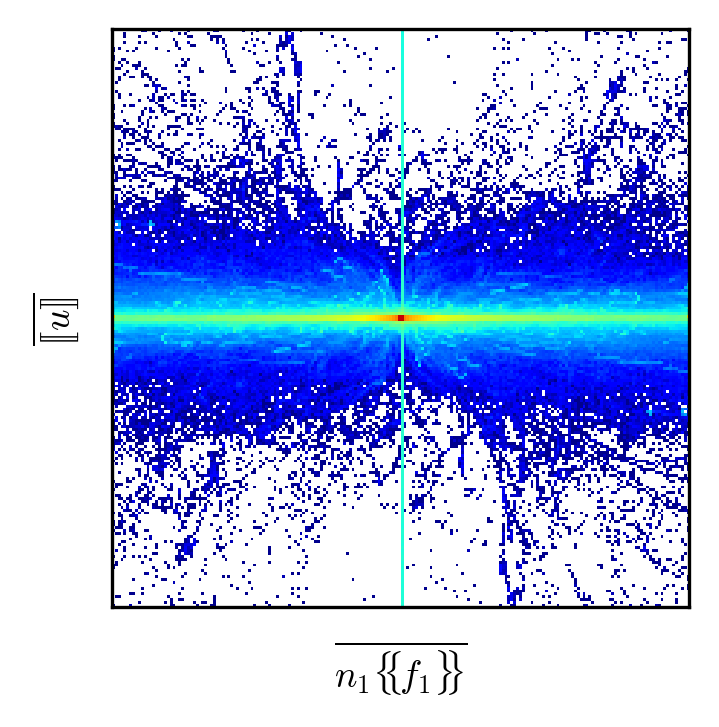}} &
            \raisebox{-0.5\height}{\includegraphics[width = .18\textwidth]{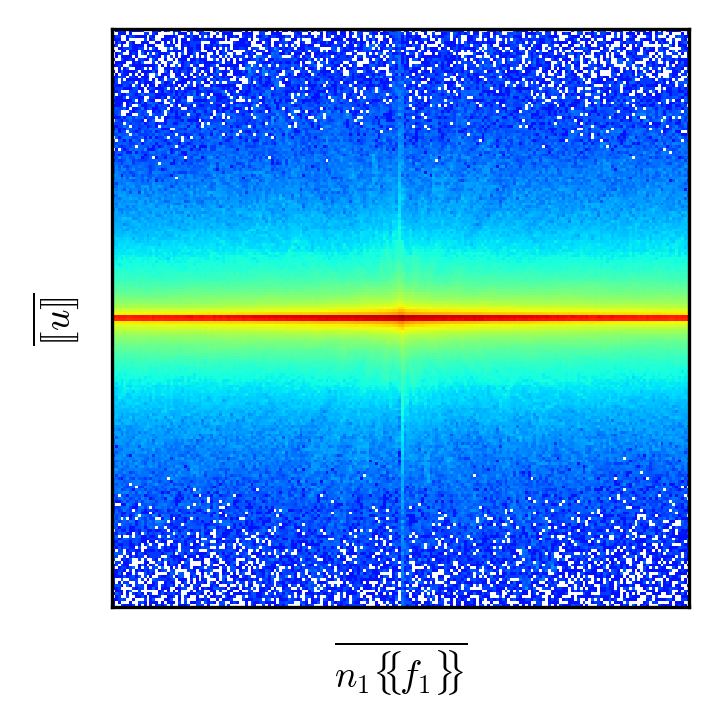}} &
            \raisebox{-0.5\height}{\includegraphics[width = .18\textwidth]{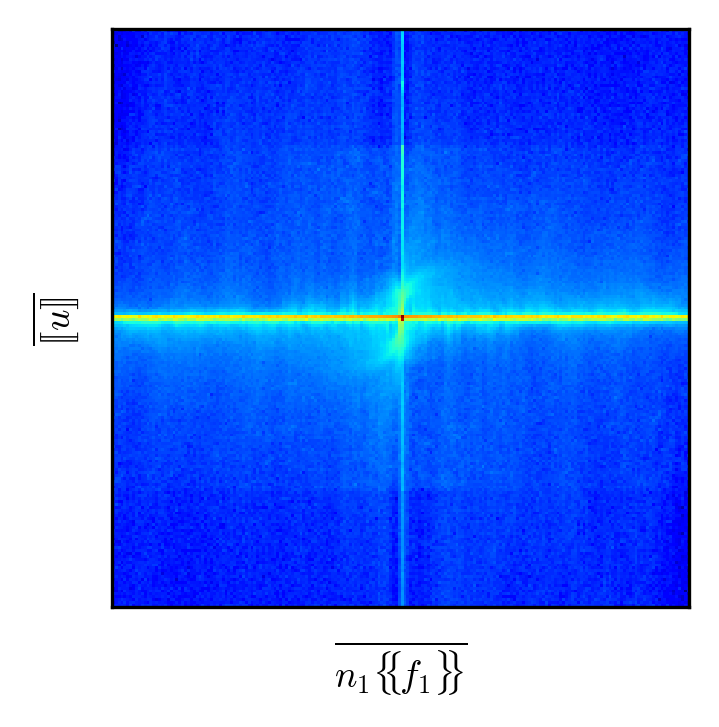}} 
            \\
            &
            \rotatebox[origin=c]{90}{$\snor{\overline{\ubjump}} = 1$} &
            \raisebox{-0.5\height}{\includegraphics[width = .18\textwidth]{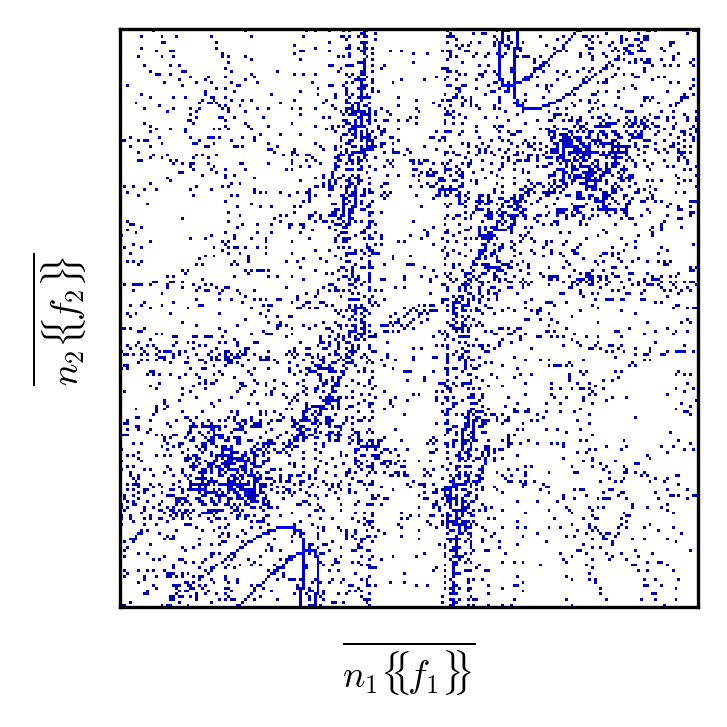}} &
            \raisebox{-0.5\height}{\includegraphics[width = .18\textwidth]{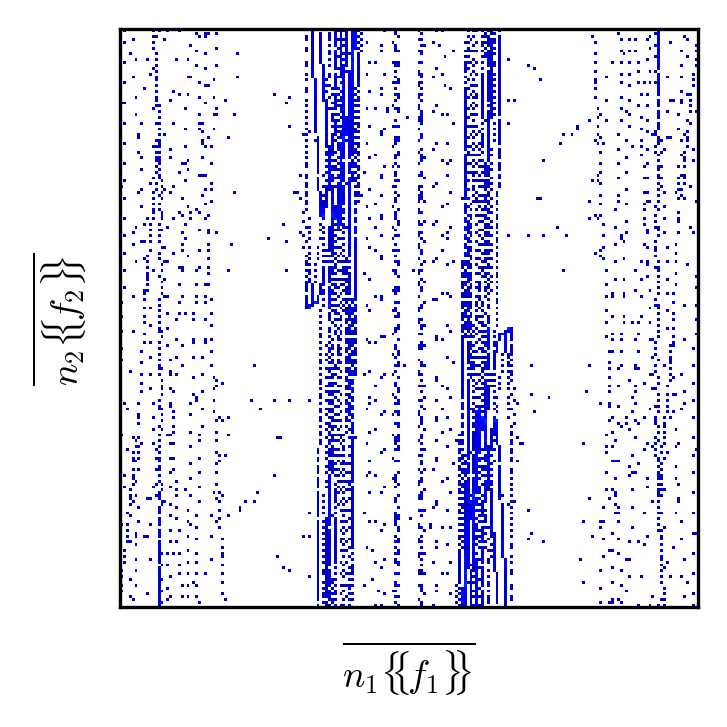}} &
            \raisebox{-0.5\height}{\includegraphics[width = .18\textwidth]{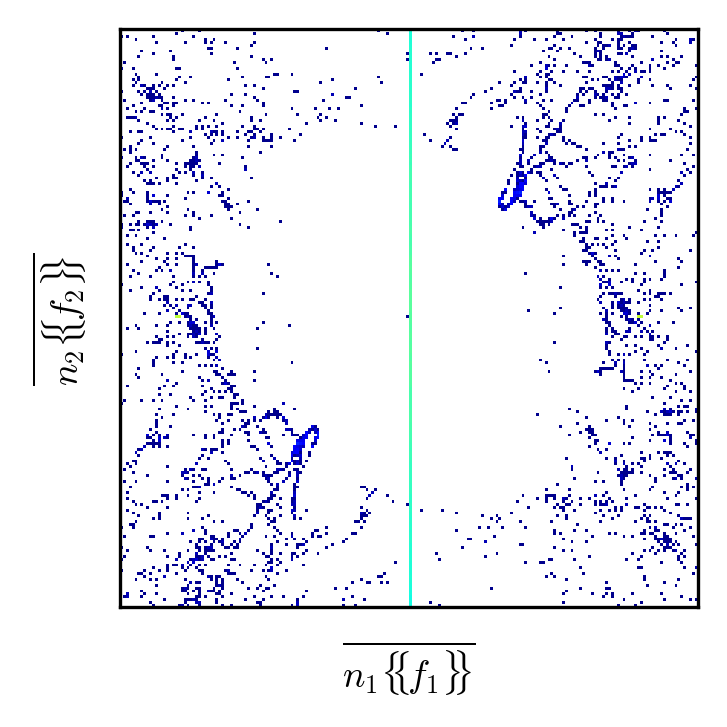}} &
            \raisebox{-0.5\height}{\includegraphics[width = .18\textwidth]{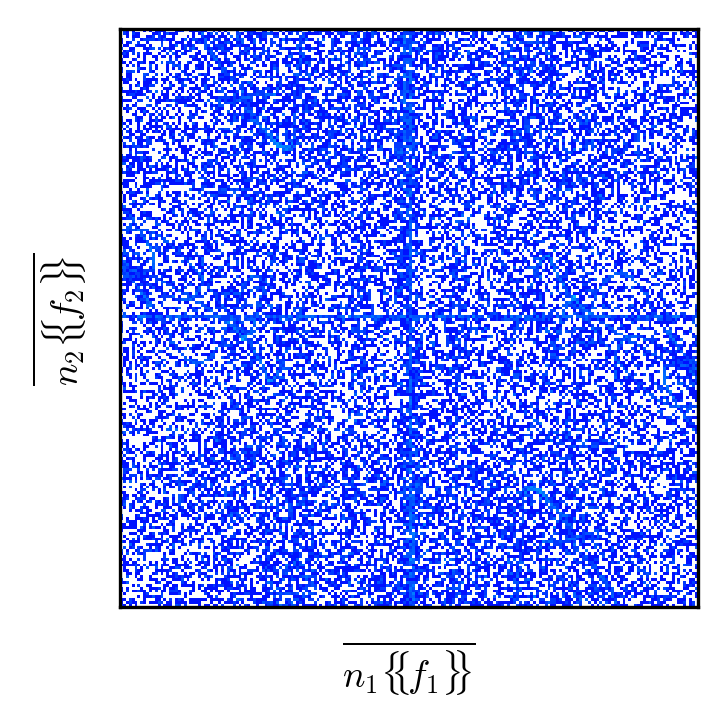}} &
            \raisebox{-0.5\height}{\includegraphics[width = .18\textwidth]{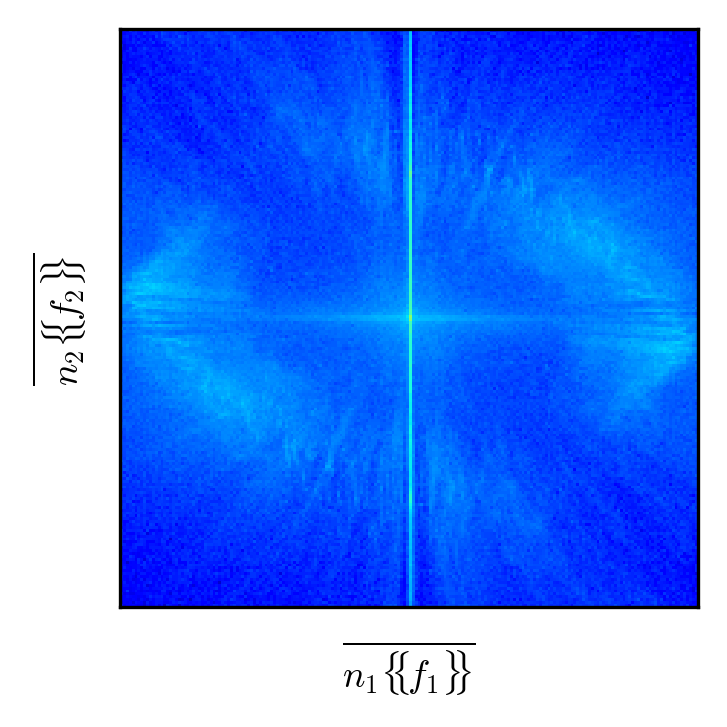}} 
            \\ ~ \\
            \multirow{3}{*}{\rotatebox[origin=l]{90}{randomized sets used for \mcDGNet}} &
            \rotatebox[origin=c]{90}{$\snor{\overline{n_1 \fbxaver}} = 1$} &
            \raisebox{-0.5\height}{\includegraphics[width = .18\textwidth]{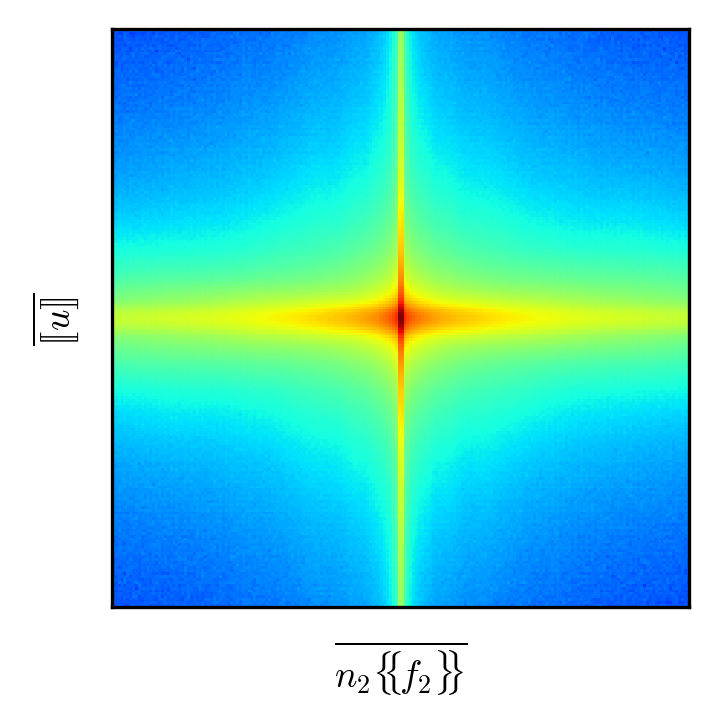}} &
            \raisebox{-0.5\height}{\includegraphics[width = .18\textwidth]{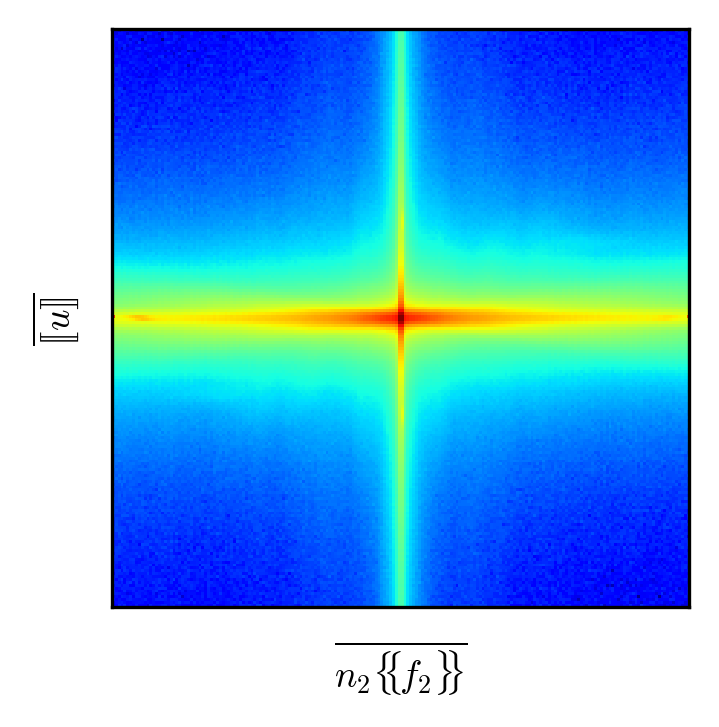}} &
            \raisebox{-0.5\height}{\includegraphics[width = .18\textwidth]{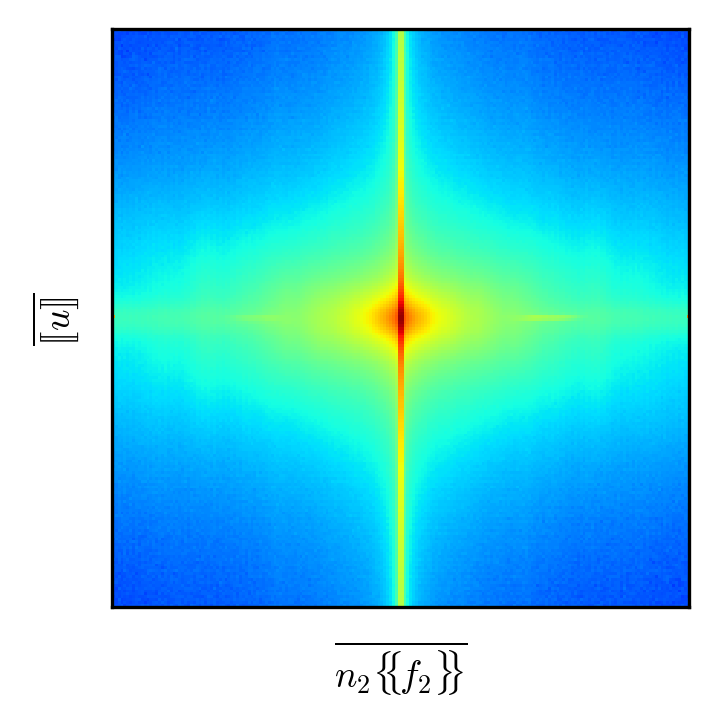}} &
            \raisebox{-0.5\height}{\includegraphics[width = .18\textwidth]{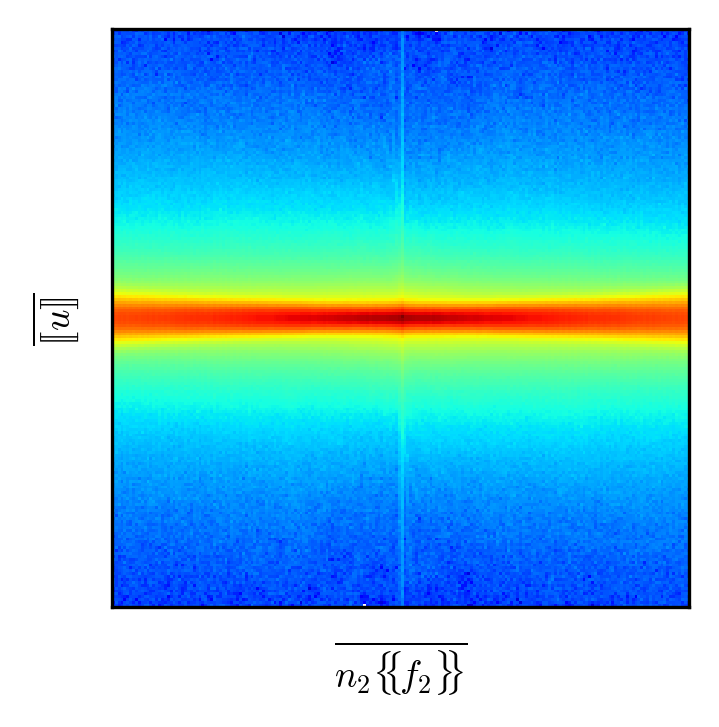}} &
            \raisebox{-0.5\height}{\includegraphics[width = .18\textwidth]{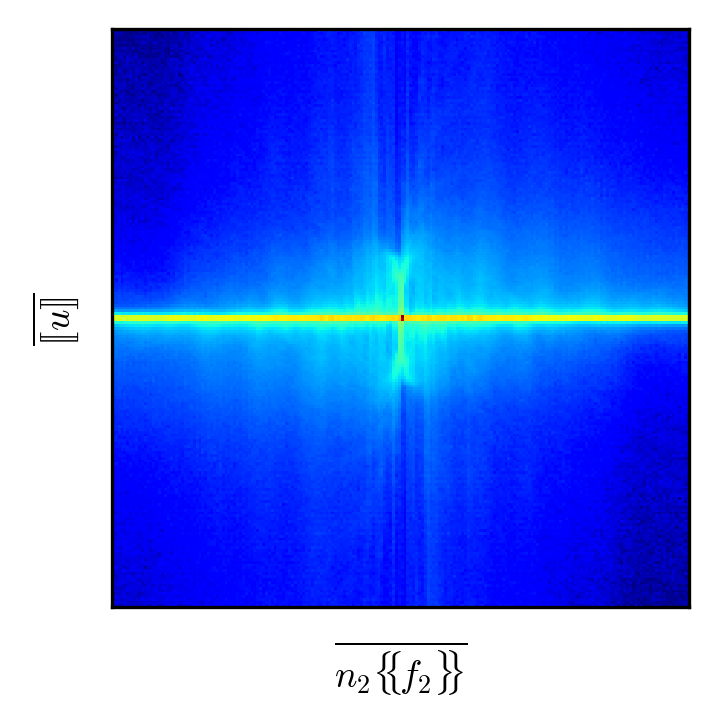}} 
            \\
            &
            \rotatebox[origin=c]{90}{$\snor{\overline{n_2 \fbyaver}} = 1$} &
            \raisebox{-0.5\height}{\includegraphics[width = .18\textwidth]{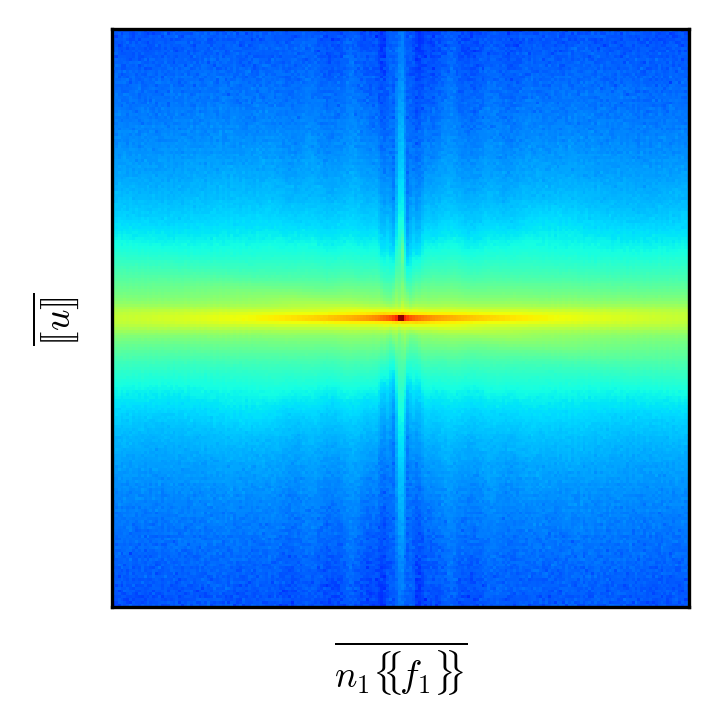}} &
            \raisebox{-0.5\height}{\includegraphics[width = .18\textwidth]{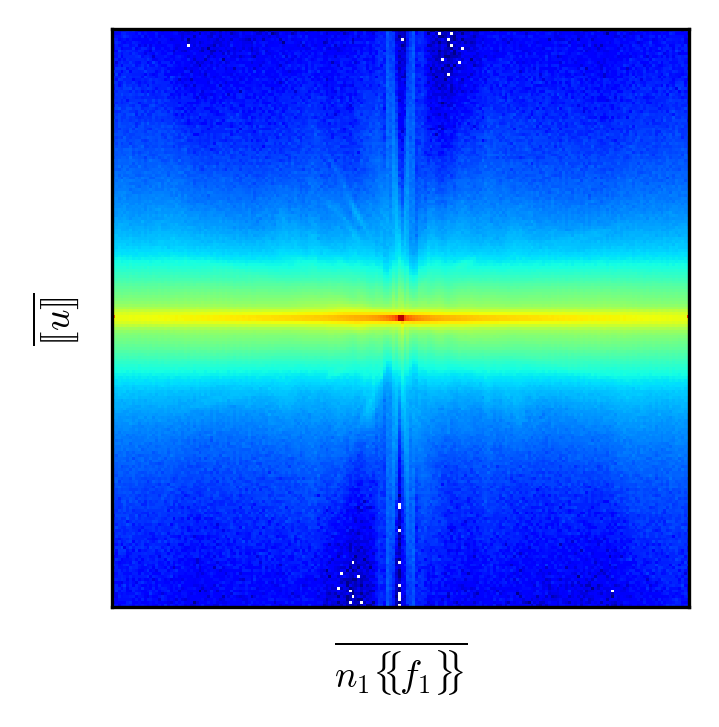}} &
            \raisebox{-0.5\height}{\includegraphics[width = .18\textwidth]{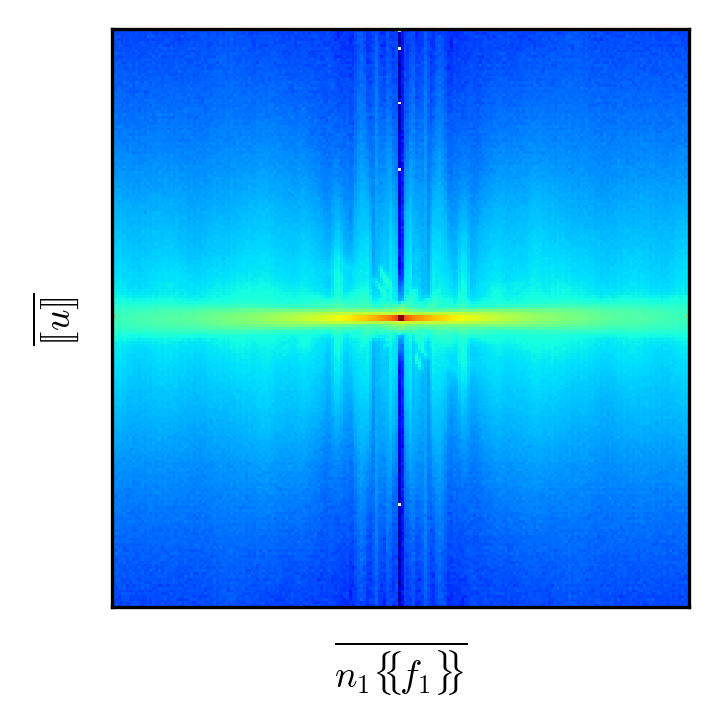}} &
            \raisebox{-0.5\height}{\includegraphics[width = .18\textwidth]{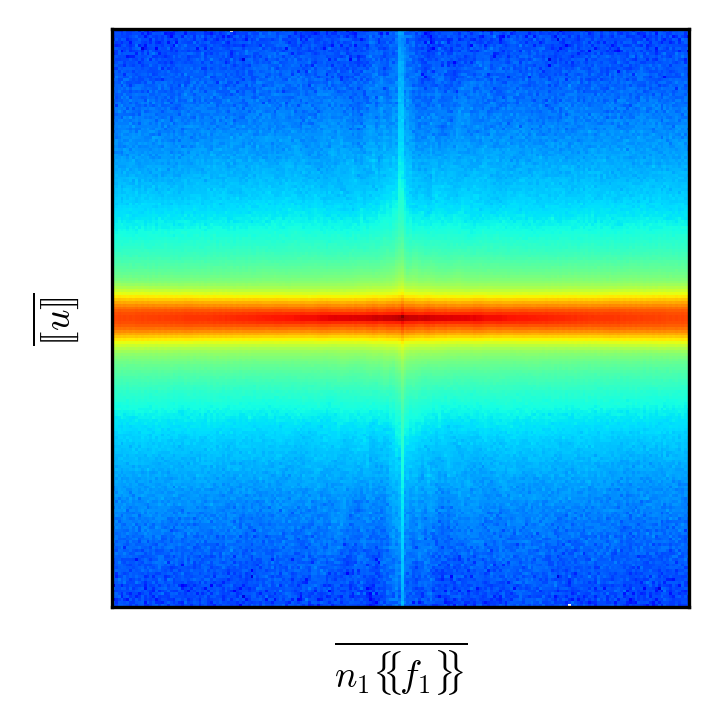}} &
            \raisebox{-0.5\height}{\includegraphics[width = .18\textwidth]{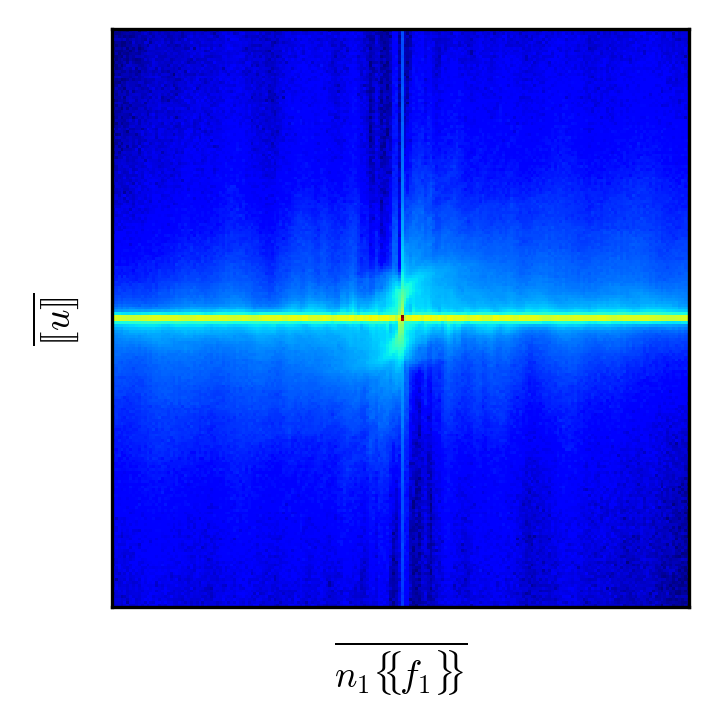}} 
            \\
            &
            \rotatebox[origin=c]{90}{$\snor{\overline{\ubjump}} = 1$} &
            \raisebox{-0.5\height}{\includegraphics[width = .18\textwidth]{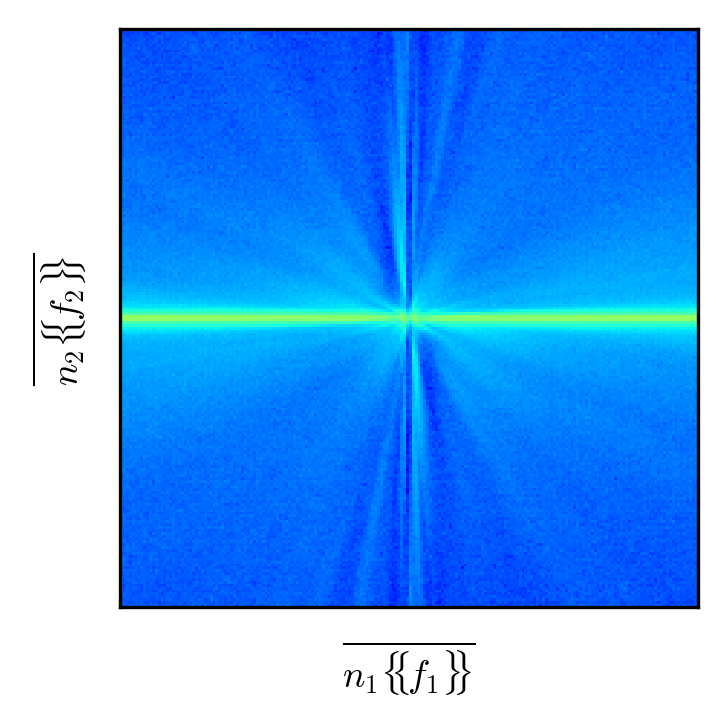}} &
            \raisebox{-0.5\height}{\includegraphics[width = .18\textwidth]{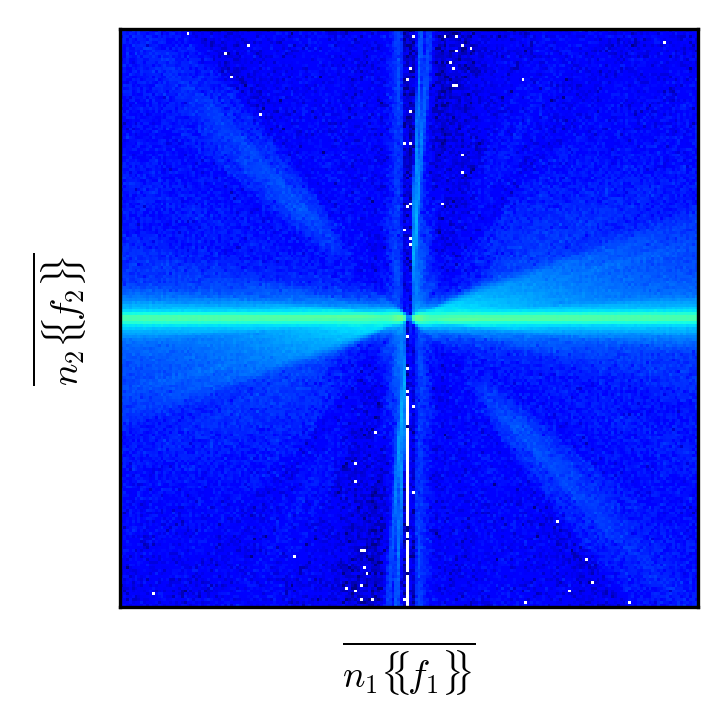}} &
            \raisebox{-0.5\height}{\includegraphics[width = .18\textwidth]{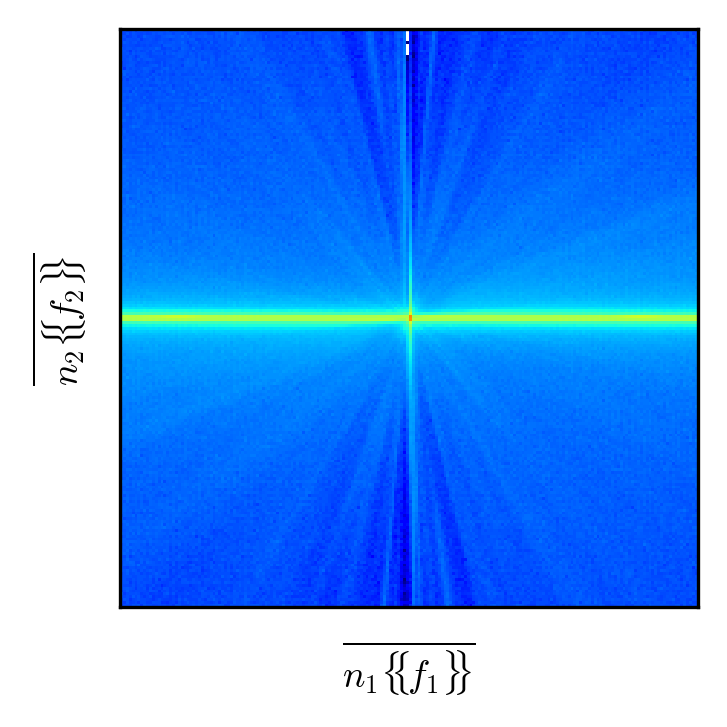}} &
            \raisebox{-0.5\height}{\includegraphics[width = .18\textwidth]{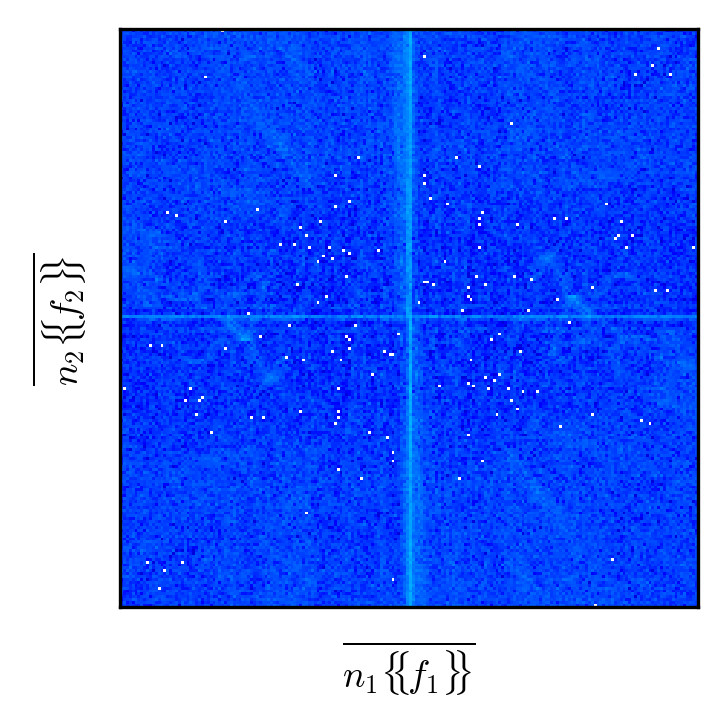}} &
            \raisebox{-0.5\height}{\includegraphics[width = .18\textwidth]{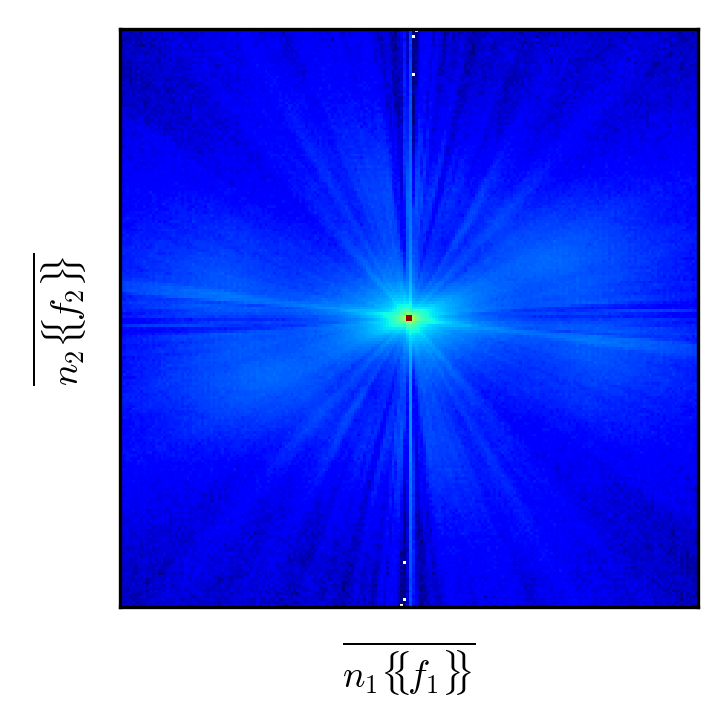}} 
        \end{tabular*}
        \caption{{\bf Data enrichment effect:} density of the three normalized input sets for five 2D Euler shock-type problems obtained from noise-free training data (used for \nDGNet approach) and corresponding corrupted training data (used in \mcDGNet approach). For all figures, the domain of interest is $[-1,1]^2$ and the color bar spans $\LRp{10^{-7}, 1}$. The maximal component is denoted by the row label, and the two axes represent the sub-maximal components.}
        \figlab{noise_corruption_effect}
\end{figure}

\subsection{Data enrichment effect by data randomization}
\seclab{P6_2D_Noise_corruption}
%We conduct a theoretical analysis of the implicit regularization benefit conferred 
We now study the data enrichment aspect of data randomization in \cref{sect:Data_rand}. Data randomization is pivotal in reinforcing long-term stability and improving generalization (see \cref{theo:data_rand} and \cref{thm:mainTheo_computable}). Nonetheless, the extent of the usefulness and scenarios in which the data randomization is beneficial for the training process is not immediately apparent. In this section, we carry out a quantitative assessment of the information enrichment effect of data randomization for five 2D Euler shock-type problems, namely, 2D Forward Facing Step (\cref{sect:forward_facing_conner}), 2D Scramjet (\cref{sect:sramjet}), 2D Airfoil (\cref{sect:airfoil}), 2D Euler Benchmarks (\cref{sect:2d_euler_benchmarks}), 2D Double Mach Reflection (\cref{sect:2d_euler_double_mach}). We observe that, in these problems, the \mcDGNet approach is either more accurate than or as accurate as the \nDGNet approach. Therefore, it is adequate to analyze the performance of the numerical flux network in the \mcDGNet approach under the presence of the data randomization technique compared to the \nDGNet approach, in which no data randomization is used. To begin with, for a given training solution snapshot, we can compute the average components $n_1 \fbxaver$, $n_2 \fbyaver$ and jump $\ubjump$ at every point on the common edge/face. As outlined in the \oDGNet framework \cref{sect:DGGNNTagent_framework}, these quantities are normalized before feeding to the neural networks by (called the triple for simplicity)
\begin{equation*}
    \begin{aligned}
        \overline{n_1 \fbxaver} = \frac{n_1 \fbxaver}{\max \LRp{\snor{n_1 \fbxaver}, \snor{n_2 \fbyaver}, \snor{\ubjump}}} \\
        \overline{n_2 \fbyaver} = \frac{n_2 \fbyaver}{\max \LRp{\snor{n_1 \fbxaver}, \snor{n_2 \fbyaver}, \snor{\ubjump}}} \\
        \overline{\ubjump} = \frac{\ubjump}{\max \LRp{\snor{n_1 \fbxaver}, \snor{n_2 \fbyaver}, \snor{\ubjump}}}.
    \end{aligned}
\end{equation*}
Here, we explicitly express quantities for 2D problems. Consequently, we achieve the constraints $\overline{n_1 \fbxaver}, \overline{n_2 \fbyaver}, \overline{\ubjump} \in \LRs{-1,1}$ in which at least one of components of the triple has an absolute value of 1, i.e., $\max \LRp{\snor{n_1 \fbxaver}, \snor{n_2 \fbyaver},  \snor{\ubjump}} = 1$. For clear visualization of 3D normalized data set in 2D planes, we categorize the normalized inputs into three sets: set 1 where $\snor{n_1 \fbxaver} = 1$, set 2 where $\snor{n_2 \fbyaver} = 1$, and set 3 where $\snor{\ubjump} = 1$. Note that the other two components are in $\LRs{-1,1}$. Subsequently, we compute the density of normalized triples for each set. We initially partition the square plane $\LRs{-1,1}^2$ into $200 \times 200$ cells and tally the number the triples that fall into each cell. In other words, we generate a bivariate histogram for the sub-maximal quantities of the triple. For ease of notation, let $N_{k, i}$ denote the number of triples belonging to cell $k$ of set $i$. $N_{k,i}$ is then again normalized to $\LRs{0,1}$, given by 
\begin{equation*}
    \overline{N}_{k,i} = \frac{N_{k,i}}{\max_{k = 1, \hdots 40000,i = 1,2,3} \LRp{N_{k,i}}}.
\end{equation*}

The density of normalized inputs induced by noise-free and noise-corrupted training data sets for various problems are shown in \cref{fig:noise_corruption_effect}. The corrupted data sets are collected from randomized samples in the first five epochs. Note that we generate new randomized samples every epoch. The noise corruption significantly enriches the training data information for the Forward Facing Step, Scramjet, Airfoil, and 2D Euler Benchmarks problems. To be more specific, the corrupted training data substantially extends the normalized training inputs, thus covering a larger proportion of planes. As a result, the flux network can adapt to a wider range of normalized inputs, leading to more accurate predictions as observed in numerical results for these problems. In contrast, for the Double Mach Reflection problem, the original noise-free data itself, after the normalization step, encompasses all possible normalized triples. As a result, the data randomization purely changes the density of normalized inputs, but no extra training information is added. This explains why the \nDGNet approach and the \mcDGNet approach achieve the same performance in the Double Mach Reflection problem. It is noteworthy that despite no extra information being gained from data randomization, the \mcDGNet approach still shows better generalization to unseen Model 2 configuration as presented in \cref{sect:2d_euler_double_mach}. This is due to the implicit regularization effect, which is always active during training in the \mcDGNet approach.

% TODO: create more noise samples instead of one epoch, it might not span the whole range of normalized inputs

% \clearpage

 \subsection{Generalization of pre-trained networks for extremely out-of-distribution scenarios}
\seclab{sramjet_generalization}

In this section, we quantitatively analyze the feasibility of employing a single pre-trained \oDGNet network from one of the following problems: 2D Forward Facing Step (\cref{sect:forward_facing_conner}), 2D Scramjet (\cref{sect:sramjet}), 2D Airfoil (\cref{sect:airfoil}), 2D Euler Benchmarks (\cref{sect:2d_euler_benchmarks}), 2D Double Mach Reflection (\cref{sect:2d_euler_double_mach}), to solve the others. To that end, we gauge the similarity between pre-trained flux networks via the profile of the normalized local linearized wave speed, which is given as

\begin{equation}
    \eqnlab{cross_solving_expected_velocity_formula}
    \overline{\lambda} = \frac{\NNflux \LRp{ \overline{n_1 \fbxaver}, \overline{n_2 \fbyaver}, \overline{\ubjump}} - \overline{n_1 \fbxaver} - \overline{n_2 \fbyaver}}{\overline{\ubjump}}.
\end{equation}

\begin{figure}[htb!]
    \centering
        \begin{tabular*}{\textwidth}{c@{\hskip -0.01cm} c@{\hskip -0.1cm} c@{\hskip -0.1cm} c@{\hskip -0.1cm} c@{\hskip -0.1cm} c@{\hskip -0.1cm}}
            \centering
            Forward & Scramjet & Airfoil & Euler-config6 & Double Mach
            \\
            \raisebox{-0.5\height}{\includegraphics[width = .20\textwidth]{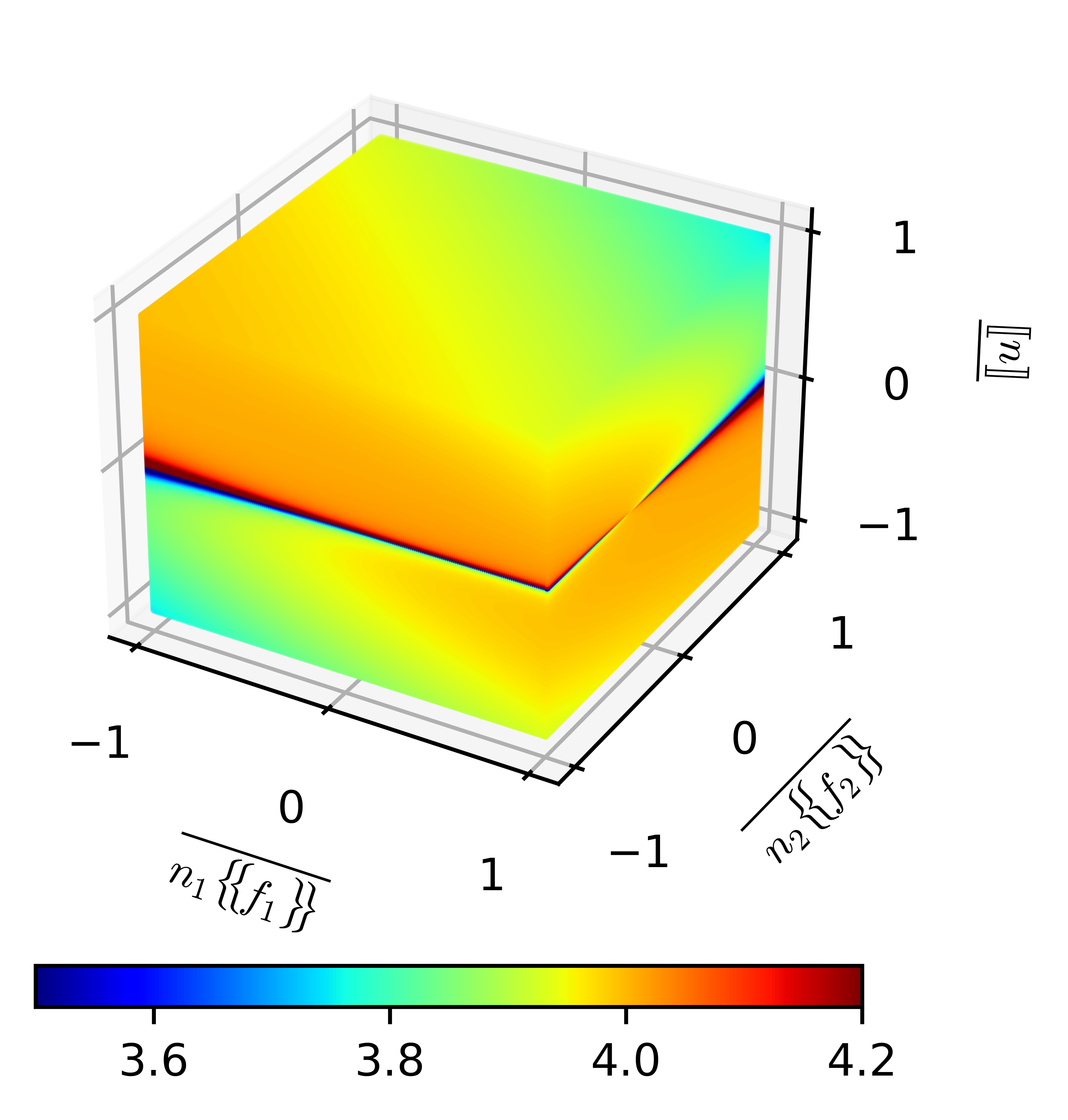}} &
            \raisebox{-0.5\height}{\includegraphics[width = .20\textwidth]{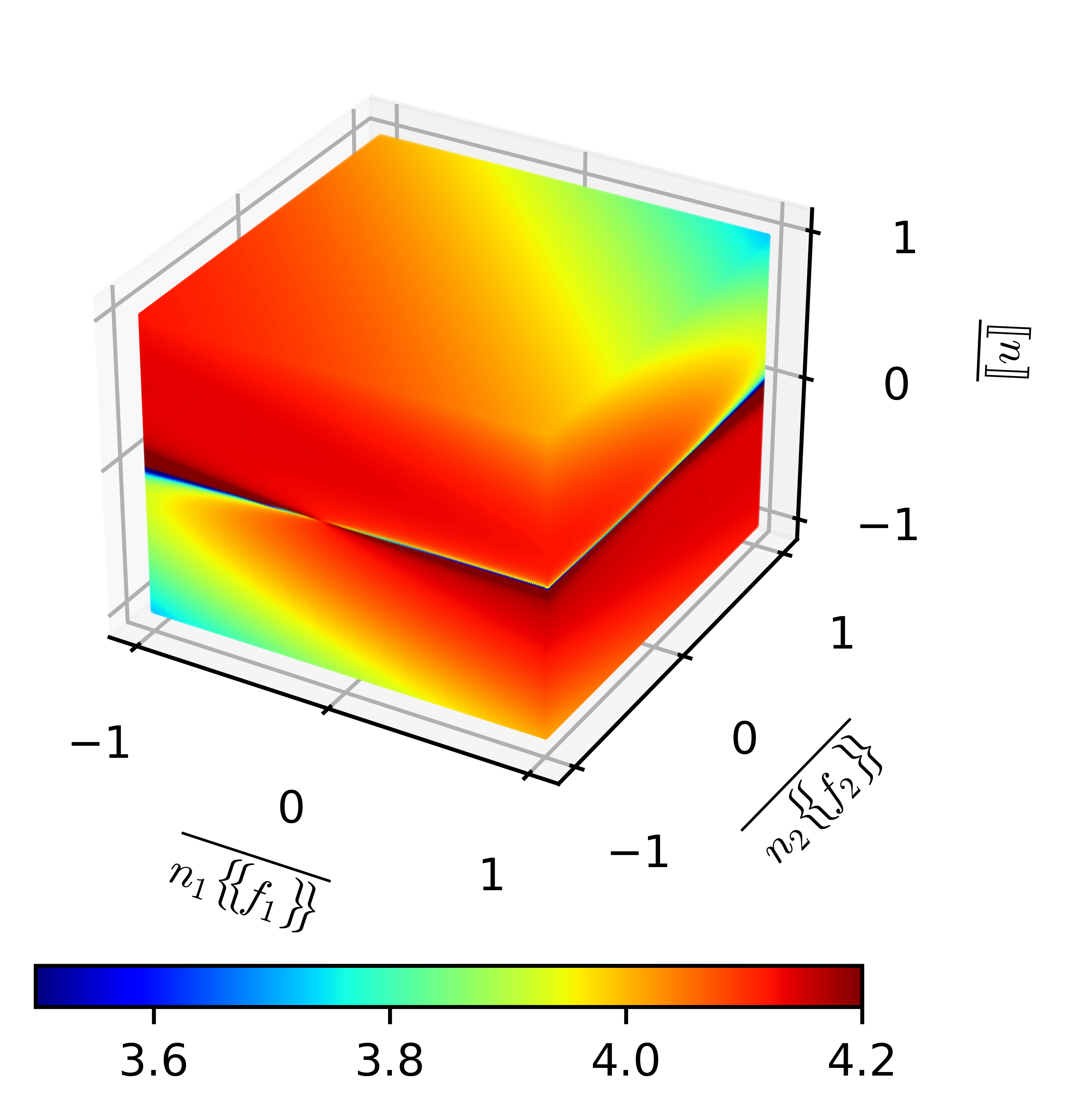}} &
            \raisebox{-0.5\height}{\includegraphics[width = .20\textwidth]{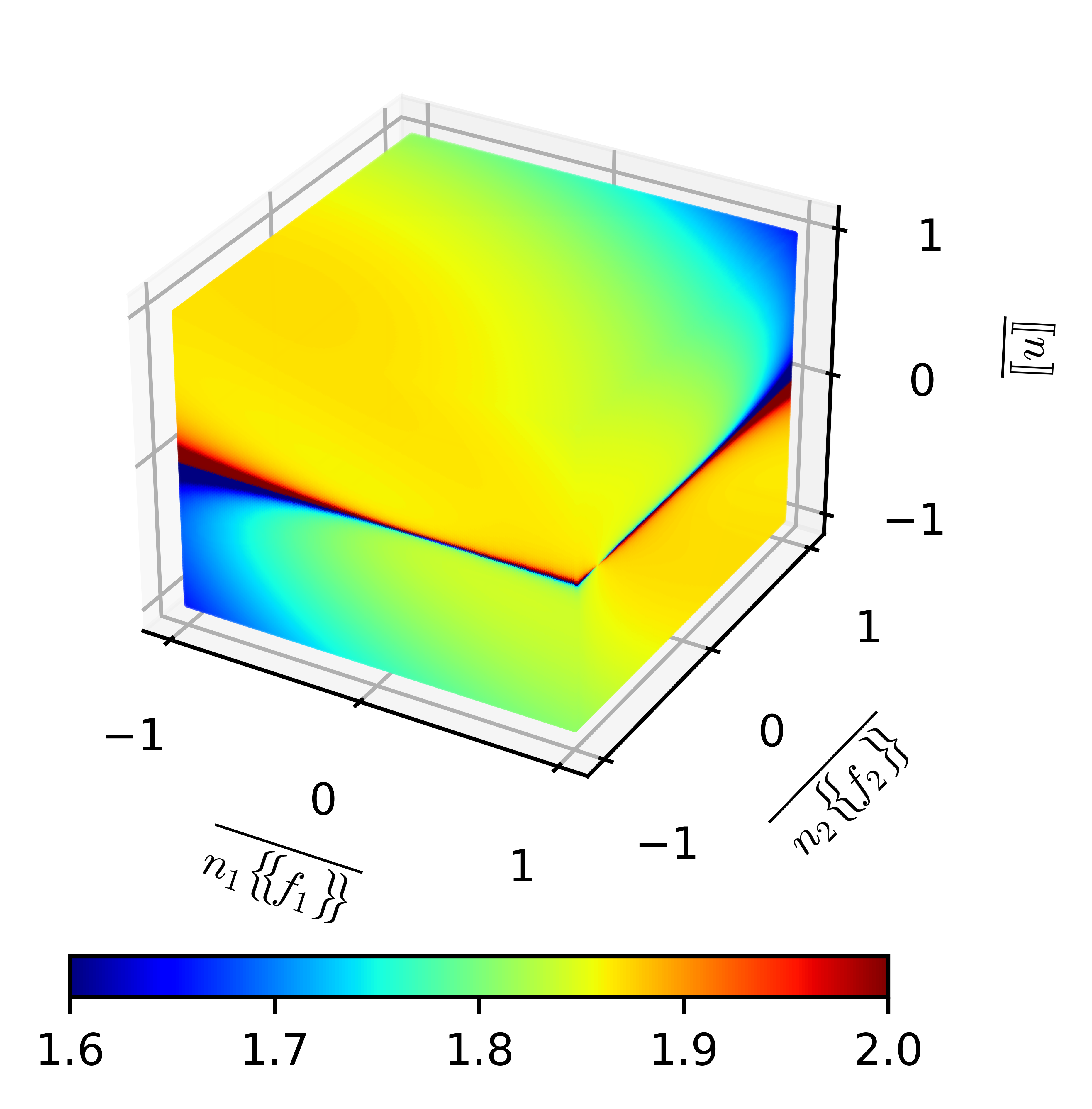}} &
            \raisebox{-0.5\height}{\includegraphics[width = .20\textwidth]{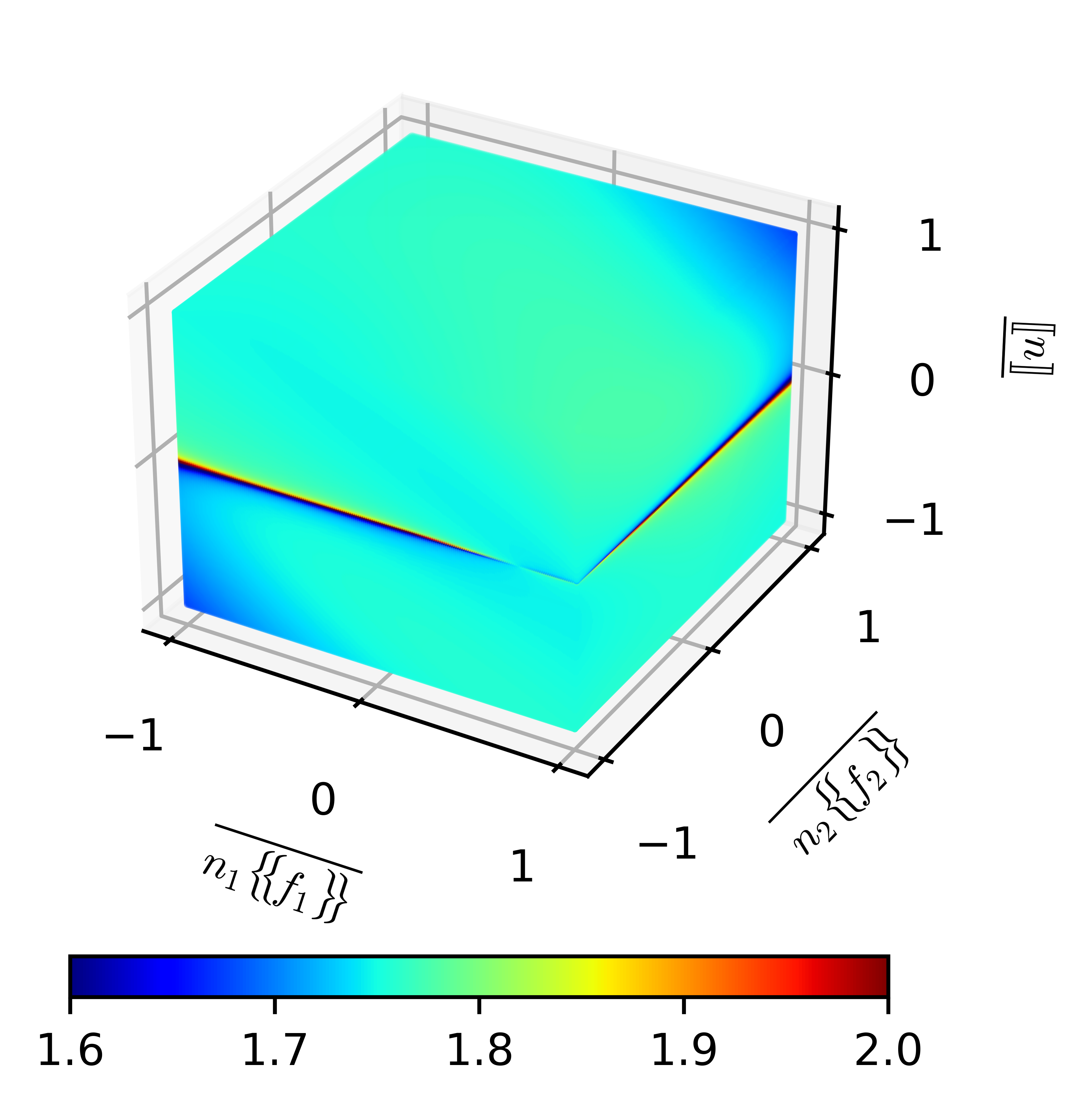}} &
            \raisebox{-0.5\height}{\includegraphics[width = .20\textwidth]{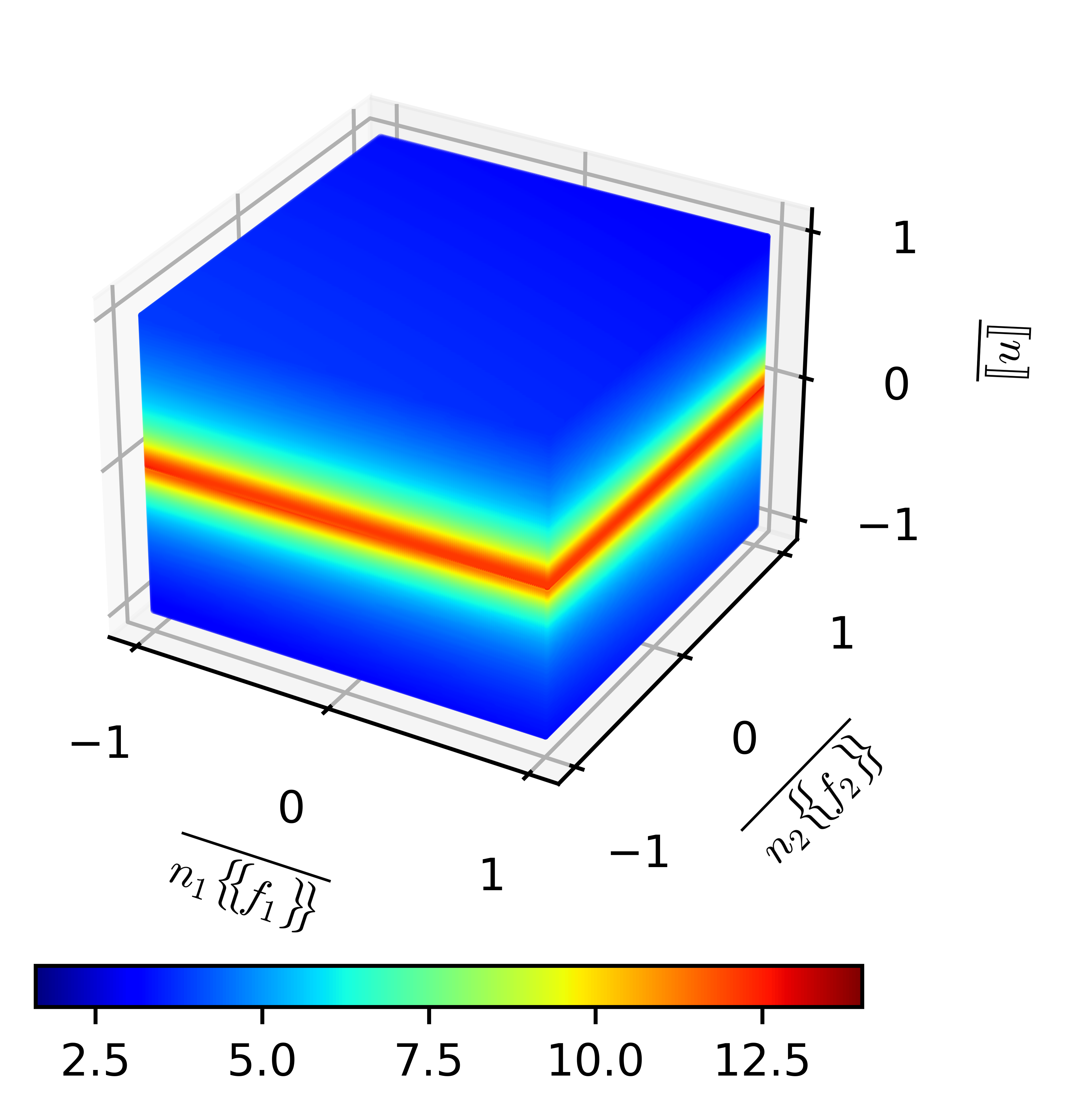}} 
            \\
            \raisebox{-0.5\height}{\includegraphics[width = .20\textwidth]{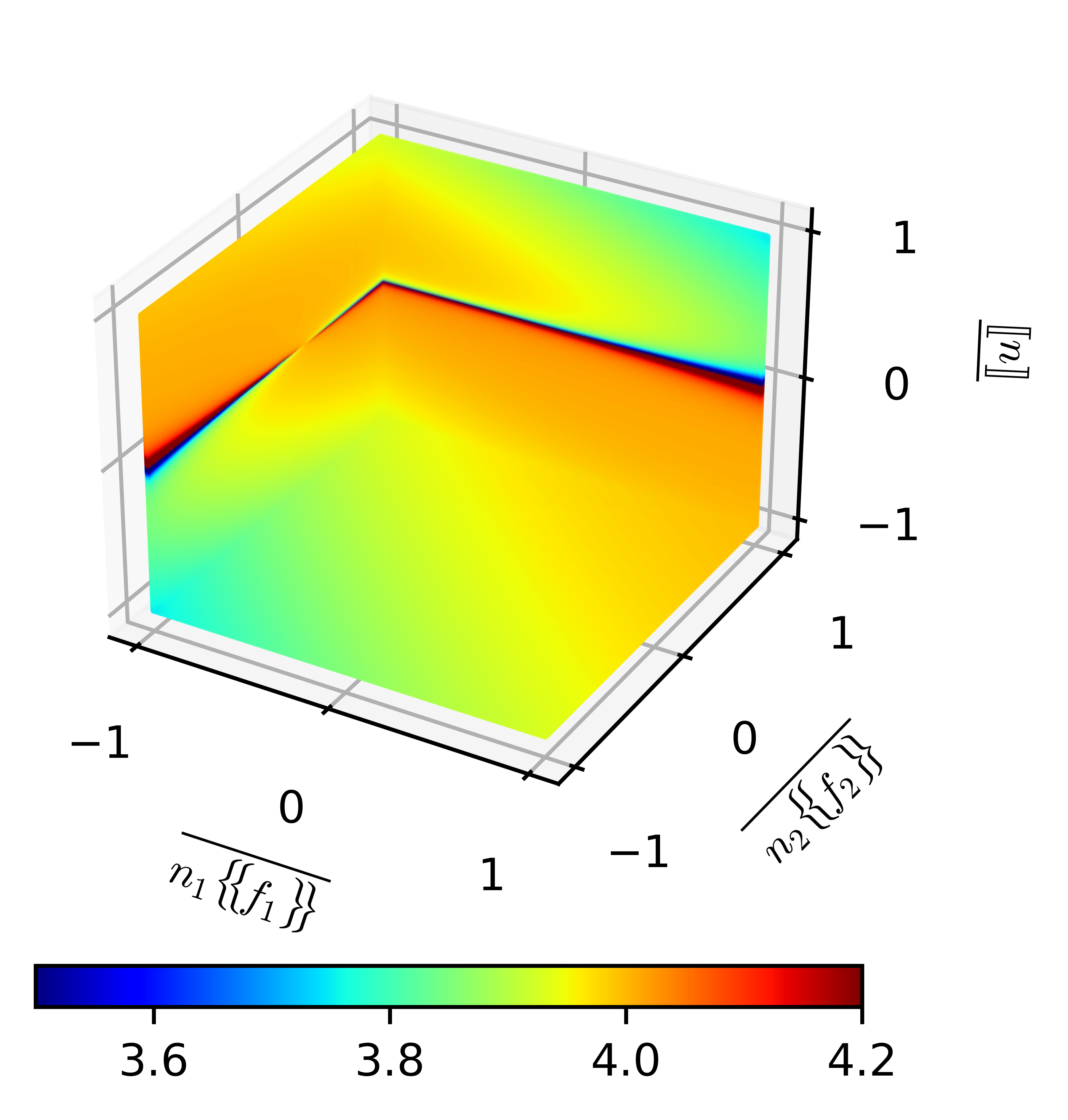}} &
            \raisebox{-0.5\height}{\includegraphics[width = .20\textwidth]{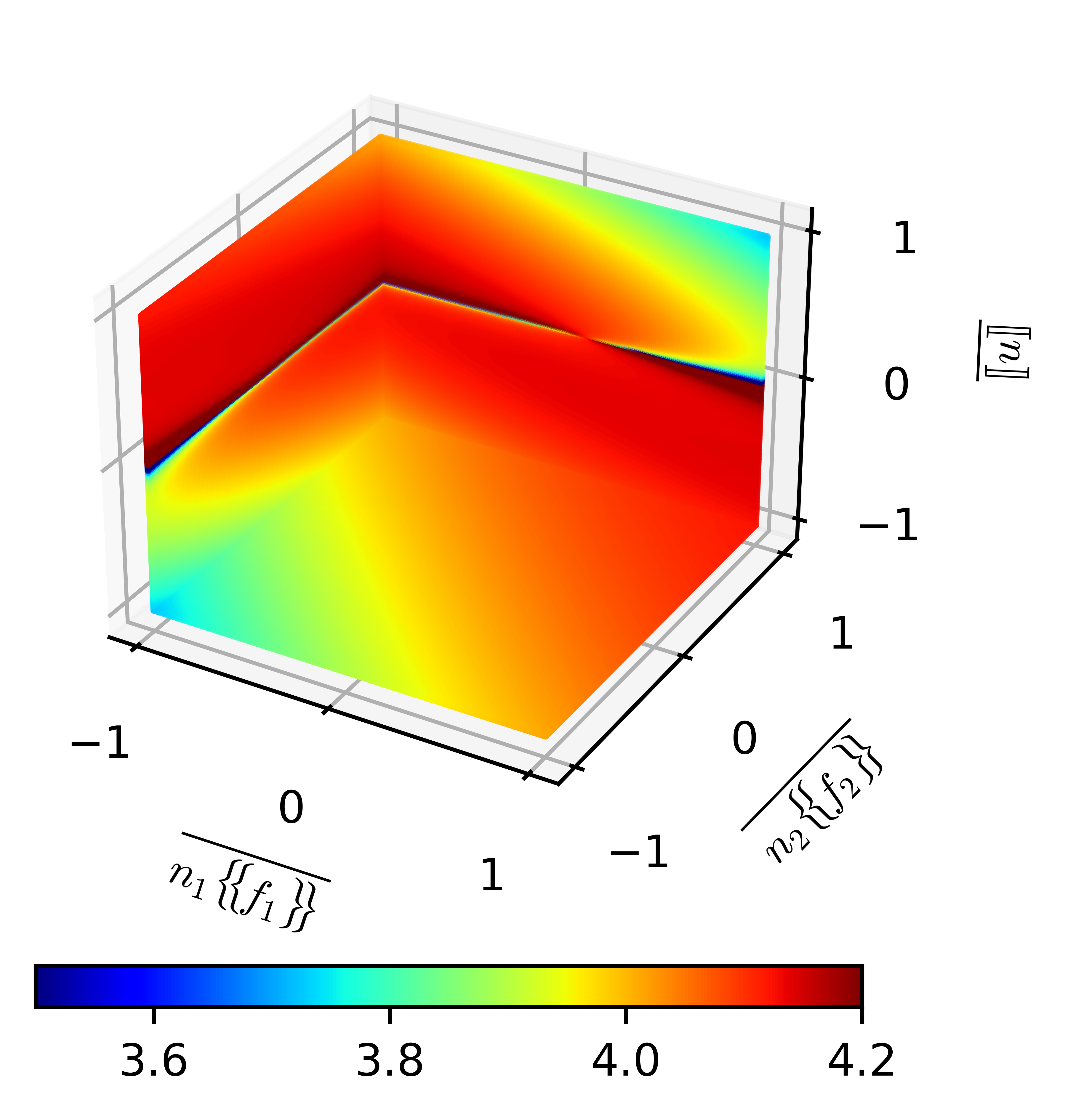}} &
            \raisebox{-0.5\height}{\includegraphics[width = .20\textwidth]{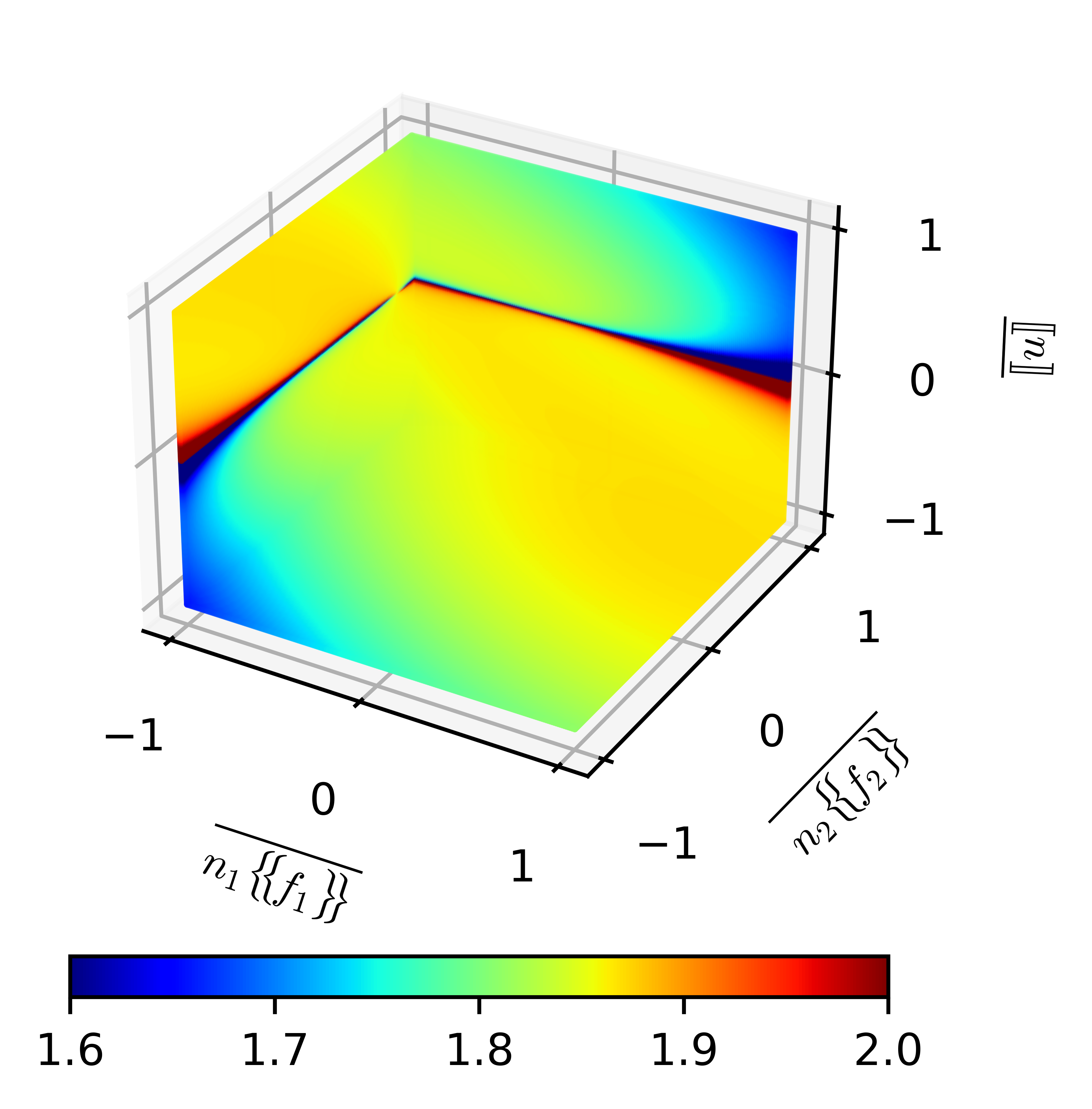}} &
            \raisebox{-0.5\height}{\includegraphics[width = .20\textwidth]{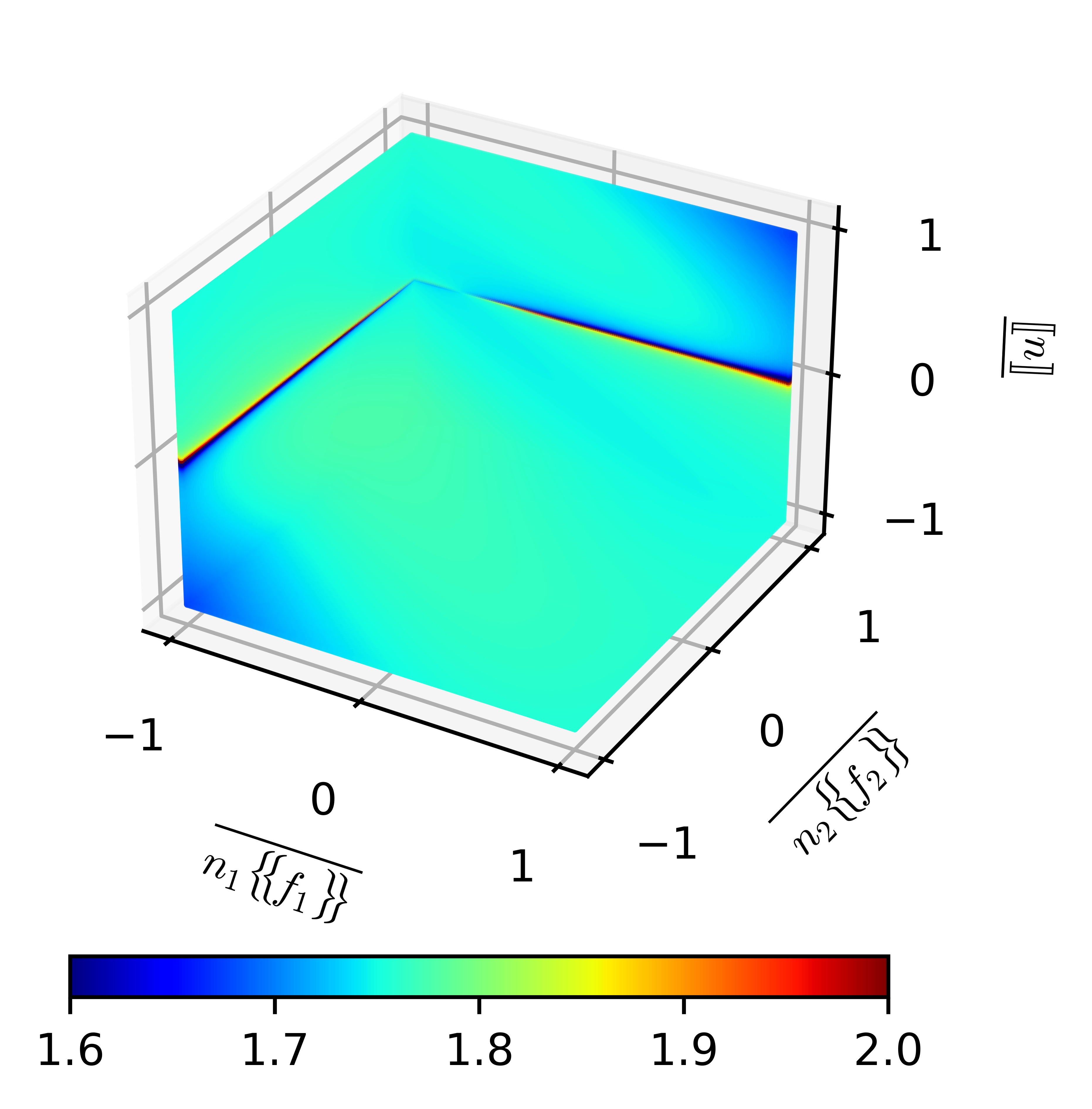}} &
            \raisebox{-0.5\height}{\includegraphics[width = .20\textwidth]{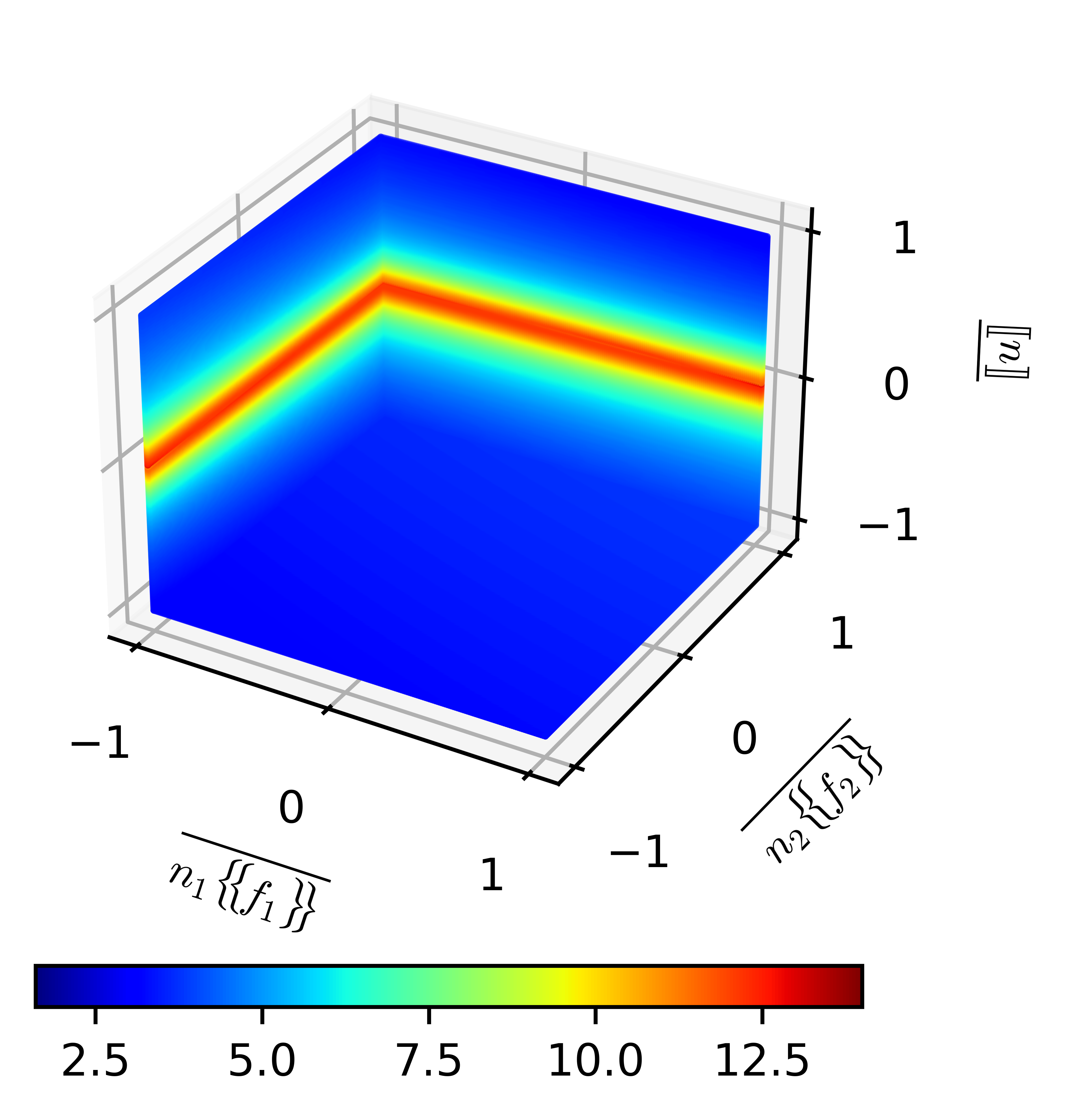}} 
        \end{tabular*}
        \caption{{\bf Pre-trained network generalization:} estimation of normalized local linearized wave speed, $\overline{\lambda}$ in \cref{eq:cross_solving_expected_velocity_formula}, obtained from pre-trained \mcDGNet network for five different problems. {\bf Top row:} plane 1 where ${n_1 \fbxaver} = 1$, plane 3 where ${\ubjump} = 1$, plane 5 where ${n_2 \fbyaver = -1}$, {\bf Bottom row:} plane 2 where $n_2 \fbyaver = 1$, plane 4 where ${n_1 \fbxaver} = -1$, plane 6 where ${\ubjump} = -1$.}
        \figlab{cross_solving_expected_velocity}
\end{figure}

\begin{table}[htb!]
    \caption{{\bf Pre-trained network generalization}: relative $L^2-$error at the time $T_\text{test}$ in Model 1 for all five shock-type problems cross-solving by pre-trained \nDGNet and \mcDGNet networks from five problems itself.}
    \tablab{Error_cross_solving}
    \begin{center}
        \begin{tabular}{l||c c c c c}
\toprule
{} & Forward & Scramjet & Airfoil & Euler-config6 & Double Mach \\
\midrule
Forward - \nDGNet        &  0.0349 &   0.0819 &  0.0105 &        0.2336$^{*}$ &      0.1336 \\
Forward - \mcDGNet       &  {\bf 0.0131} &   0.0848 &  0.0033 &        0.0993$^{*}$ &      0.0403 \\
\midrule
Scramjet - \nDGNet       &  0.0140 &   0.0812 &  0.0055 &        0.0866$^{*}$ &      0.0947 \\
Scramjet - \mcDGNet      &  0.0154 &   {\bf 0.0109} &  0.0037 &        0.0936$^{*}$ &      0.0546 \\
\midrule
Airfoil - \nDGNet        &  0.0767 &   0.1530 &  0.0031 &        0.6193$^{*}$ &      0.2888 \\
Airfoil - \mcDGNet       &  0.0557 &   0.1124 &  {\bf 0.0017} &        0.3048$^{*}$ &      0.0577 \\
\midrule
Euler-config6 - \nDGNet  &  0.0629 &   0.0618 &  0.0083 &        0.1623 &      0.1691 \\
Euler-config6 - \mcDGNet &  0.0603 &   0.0993 &  0.0018 &        {\bf 0.0229} &      0.0522 \\
\midrule
Double Mach - \nDGNet    &  0.1180$^{*}$ &   0.0855$^{*}$ &  0.0137$^{*}$ &        0.1528$^{*}$ &      {\bf 0.0168} \\
Double Mach - \mcDGNet   &  0.1400$^{*}$ &   0.0663$^{*}$ &  0.0121$^{*}$ &        0.1499$^{*}$ &      0.0171 \\
\bottomrule
\end{tabular}
\newline $^*$ solved with time step size $\frac{\dt}{20}$, otherwise NaN with $\dt$ of corresponding problem.

    \end{center}
\end{table}

We directly generate the normalized inputs to the pre-trained flux neural networks. To be more specific, we have 6 planes including plane 1 where ${n_1 \fbxaver} = 1$, plane 2 where ${n_2 \fbyaver = 1}$, plane 3 where ${\ubjump} = 1$,  plane 4 where ${n_1 \fbxaver} = -1$, plane 5 where ${n_2 \fbyaver = -1}$, and plane 6 where ${\ubjump} = -1$. In each plane, the other two components range from $\LRs{-1,1}$. Therefore, for each plane, we fix the value $1$ (plane 1,2,3) and $-1$ (for 4,5,6) for the corresponding component and generate the other $200\times 200$ pairs of two remaining components on the uniform mesh $\LRs{-1,1}^2$. The profiles of normalized local speed on the six normalized data planes obtained from the \mcDGNet numerical flux networks for different problems are shown in \cref{fig:cross_solving_expected_velocity}. The test data relative $L^2-$error in Model 1, when using pre-trained from five problems for solving others, is presented in \cref{tab:Error_cross_solving}.

We can observe that the closer the profile of the normalized local speed between two problems, the better the networks can generally be used to solve for the other. Indeed, the Forward Facing Step and Scramjet pre-trained networks are likely equivalent and can be used to solve others with a high level of accuracy. By contrast, the Double Mach Reflection pre-trained network wave speed profile is significantly larger than others and is thus unsuitable for solving other problems. The Euler Benchmark configuration 6 pre-trained network has the smallest local speed profile. Although this profile is different from the other problems as well, it still can be employed to solve others. Interestingly, we have to use a much smaller time step size when using the Double Mach Reflection pre-trained network to stably solve other problems. We notice that networks with smaller local speed profiles (Euler Benchmark configuration 6) can be used to solve problems with higher local speed profiles without such time step modification. 
%{The reason for this phenomenon is perhaps the same one as mentioned in the traditional numerical method \textcolor{red}{[Jau-Uei] add paper? and discuss a bit}}
% \jauuei

This could be due to a significant difference in the characteristic speeds between the Double Mach Reflection and the Euler benchmark problems. As a result, the characteristics of the numerical flux learned by the \oDGNet can be fundamentally distinct for these two different problems. In particular, the \oDGNet trained by the Double Mach Reflection problem could excessively penalize the jump term, hence the much smaller time step size needed to maintain the stability.

Despite the Airfoil pre-trained network having an intermediate local speed profile, it gives higher errors than all pre-trained networks for all other problems. %There is not to us  for this exceptional case at the moment of writing. 
Lastly, it is worth noting that \mcDGNet pre-trained networks have better accuracy than \nDGNet networks for most of cases: thanks to the implicit regularization feature of the \mcDGNet approach.

% \begin{table}[H]
%     \caption{{\bf Cross-solving: L2 Relative Error at the time $T_\text{test}$ using pretrained networks for different problems.} }
%     \input{Figs/2D_cross_solving/Error_cross_solving.tex}
% \end{table}

% \clearpage

\subsection{Training with HLL (Harten-Lax-van Leer) flux data} 
\seclab{forward_facing_conner_Roe_flux}

\begin{figure}[htb!]
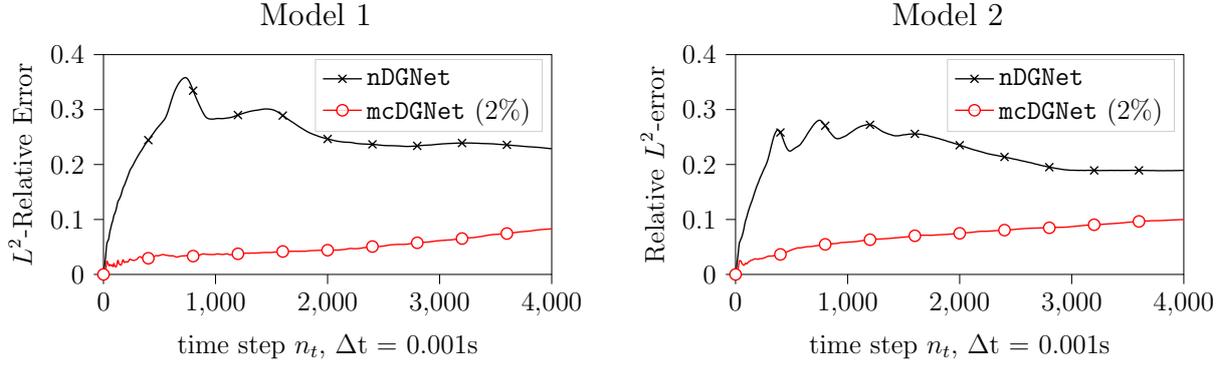

    \centering
        \begin{tabular*}{\textwidth}{c c}
            \centering
            \quad  Model 1 & \quad  Model 2
            \\
            \raisebox{-0.5\height}{\resizebox{0.48\textwidth}{!}{\input{Figs/2D_Euler_Forward_facing/Relative_error_test_data_model1_mesh_Roe_flux.tex}}} &
            \raisebox{-0.5\height}{\resizebox{0.48\textwidth}{!}{\input{Figs/2D_Euler_Forward_facing/Relative_error_test_data_model2_mesh_Roe_flux.tex}}} 
        \end{tabular*}
        \caption{{\bf 2D forward facing step - HLL flux training data:} test data relative $L^2$-error average over three conservative components $\LRp{\rho, \rho u, E}$ predictions obtained by \nDGNet and \mcDGNet approaches at different time steps for Model 1 ({\bf Left}) and Model 2 ({\bf Right}) mesh grids.}
        \figlab{Roeflux_Relative_error_test_data_model1_model2_mesh}
\end{figure}

\begin{figure}[htb!]
    \centering
        \begin{tabular*}{\textwidth}{c c@{\hskip -0.0001cm} c@{\hskip -0.002cm} c@{\hskip -0.002cm} c@{\hskip -0.002cm}}
            \centering
            & & DG & \nDGNet & \mcDGNet ($2\%$)
            \\
            \multirow{2}{*}{\rotatebox[origin=l]{90}{Model 1 \quad  }} &
            \rotatebox[origin=c]{90}{Pred} &
            \raisebox{-0.5\height}{\includegraphics[width = .31\textwidth]{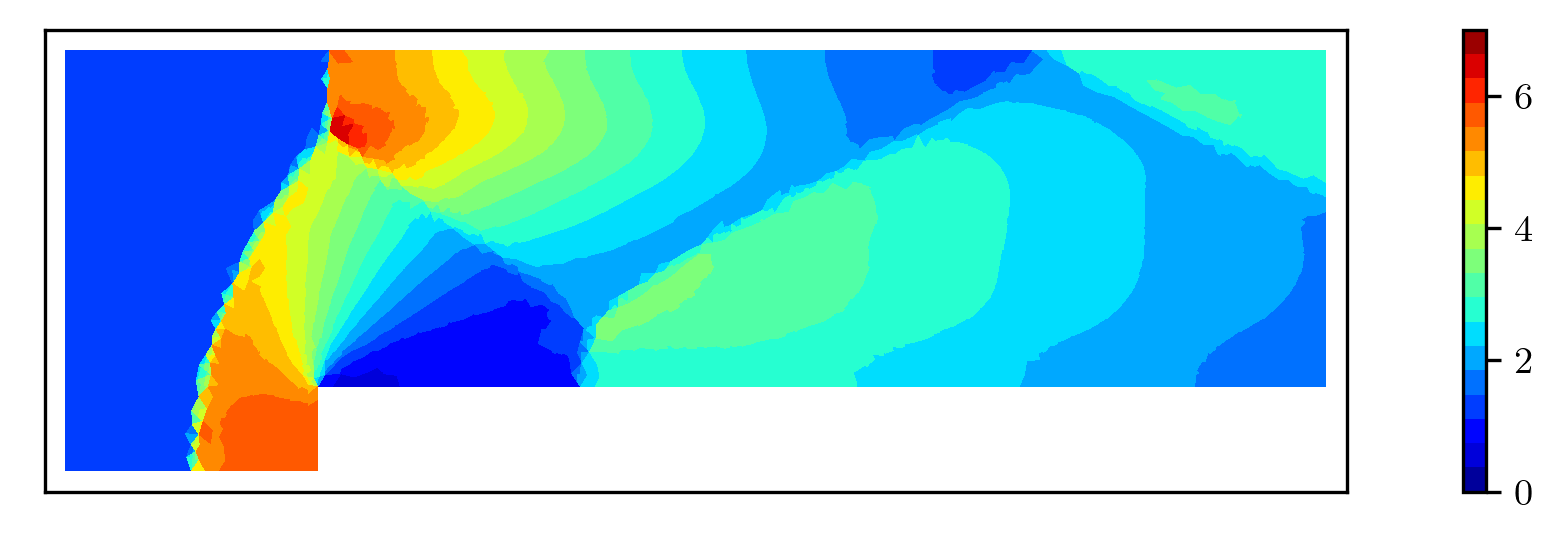}} &
            \raisebox{-0.5\height}{\includegraphics[width = .31\textwidth]{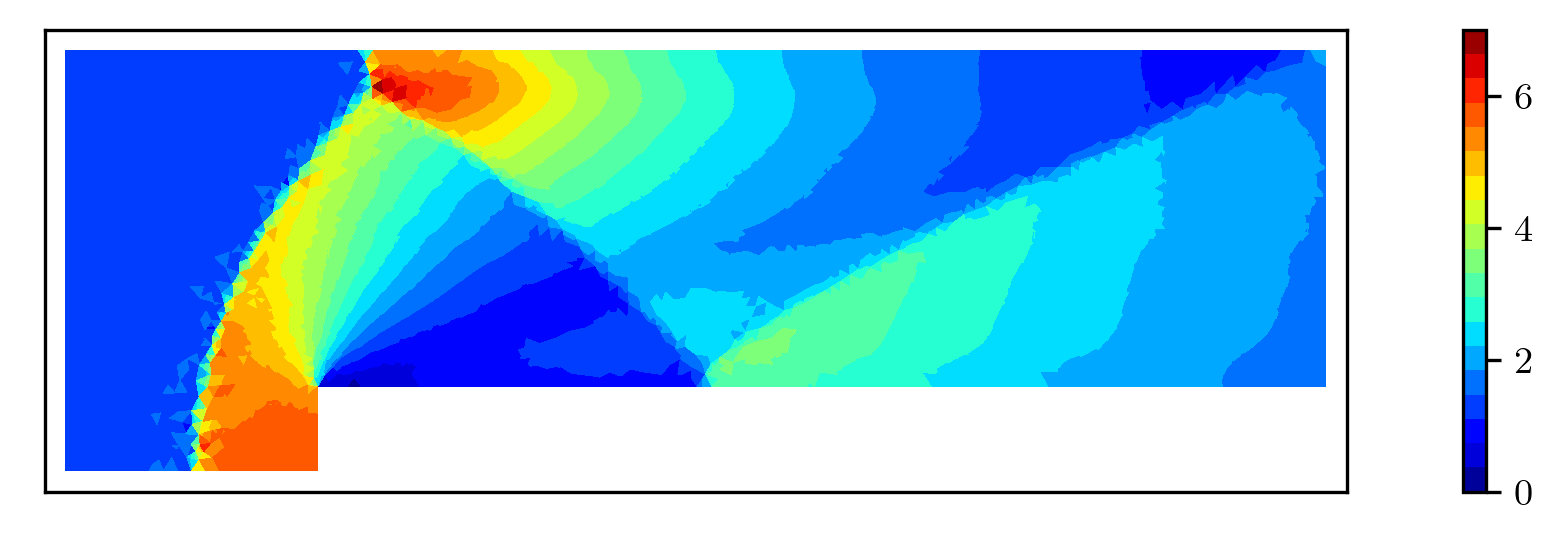}} &
            \raisebox{-0.5\height}{\includegraphics[width = .31\textwidth]{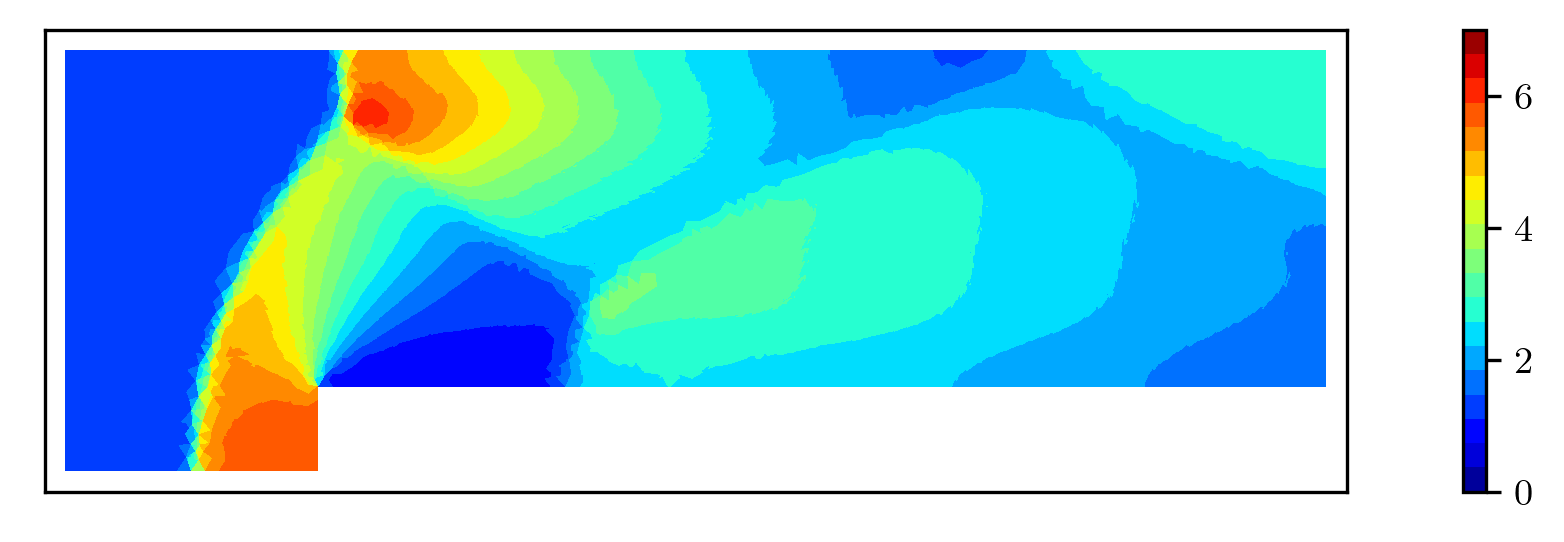}}
            \\
            &
            \rotatebox[origin=c]{90}{Error} &
            \raisebox{-0.5\height}{} &
            \raisebox{-0.5\height}{\includegraphics[width = .31\textwidth]{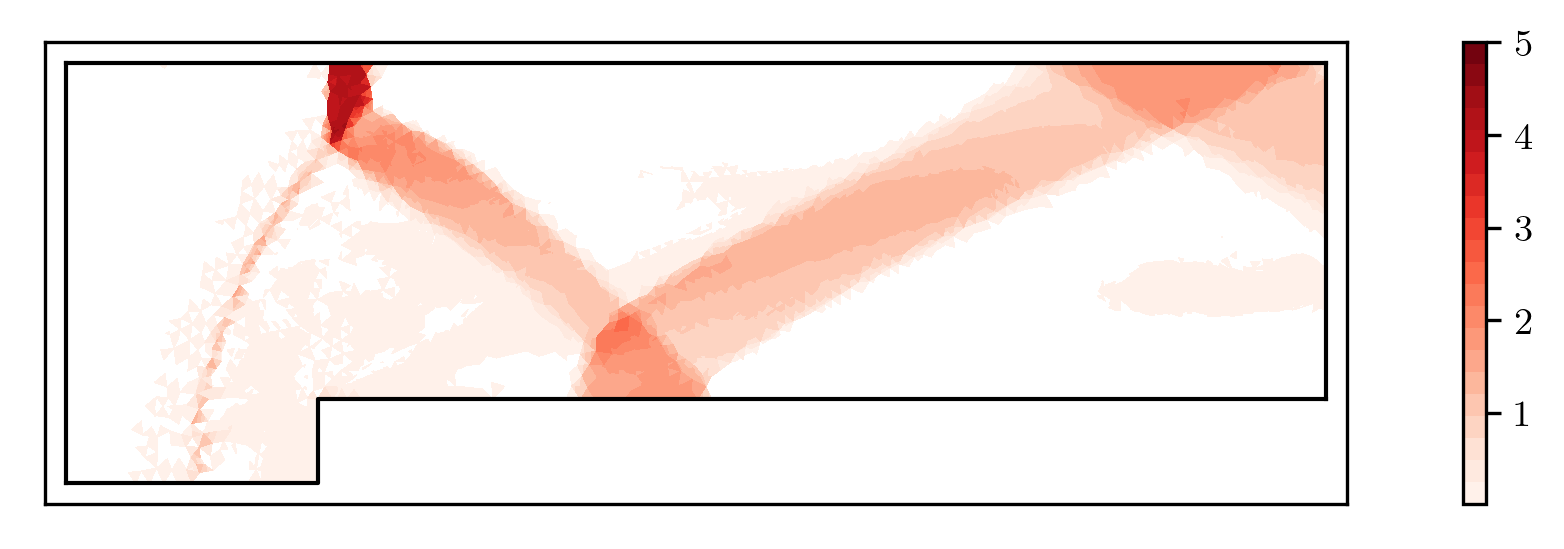}} &
            \raisebox{-0.5\height}{\includegraphics[width = .31\textwidth]{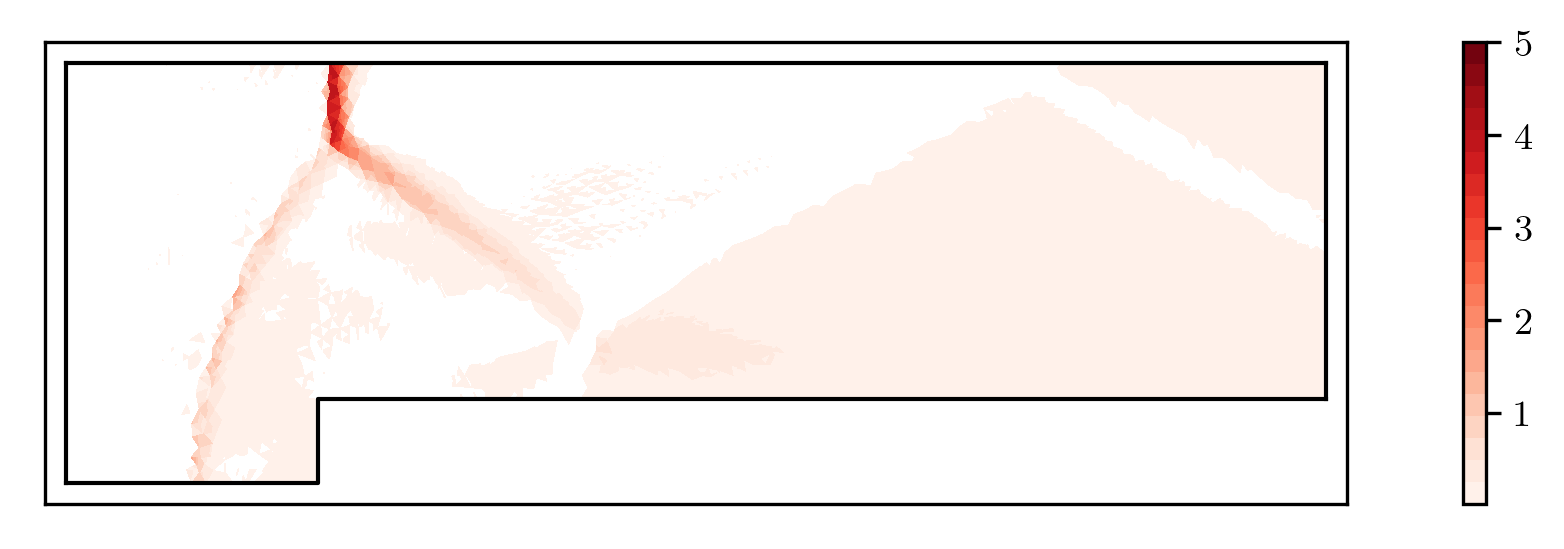}}
            \\
            \multirow{2}{*}{\rotatebox[origin=l]{90}{Model 2 \quad  }} &
            \rotatebox[origin=c]{90}{Pred} &
            \raisebox{-0.5\height}{\includegraphics[width = .31\textwidth]{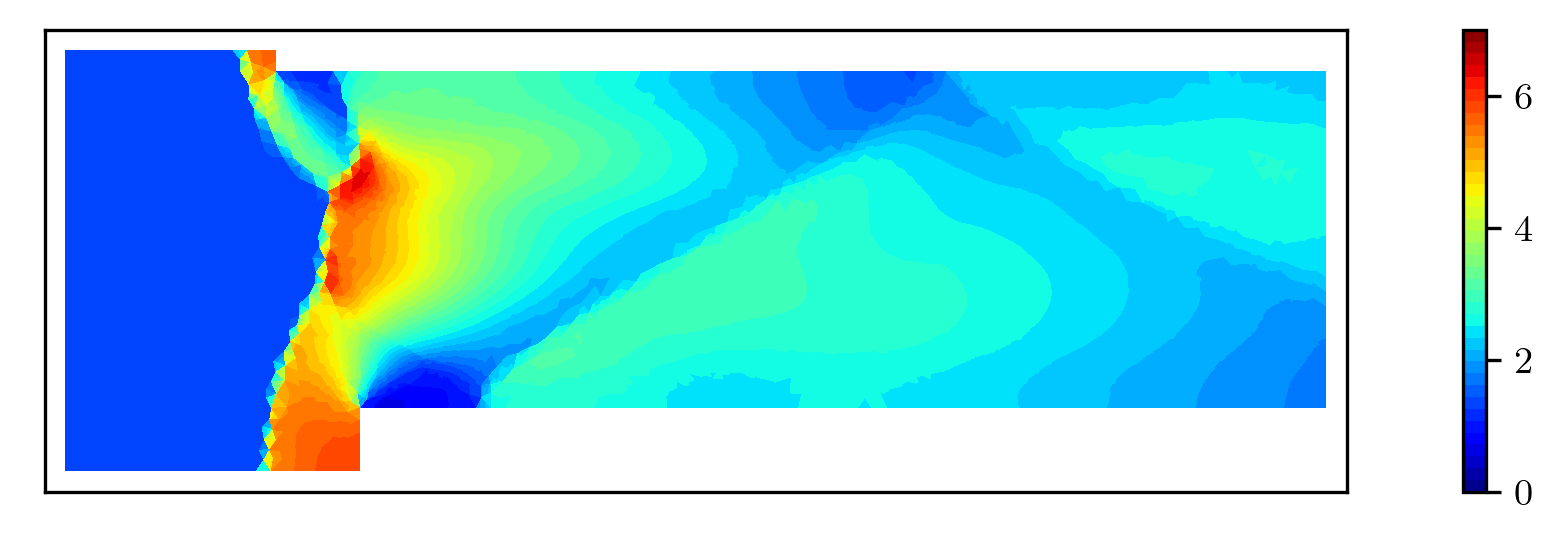}} &
            \raisebox{-0.5\height}{\includegraphics[width = .31\textwidth]{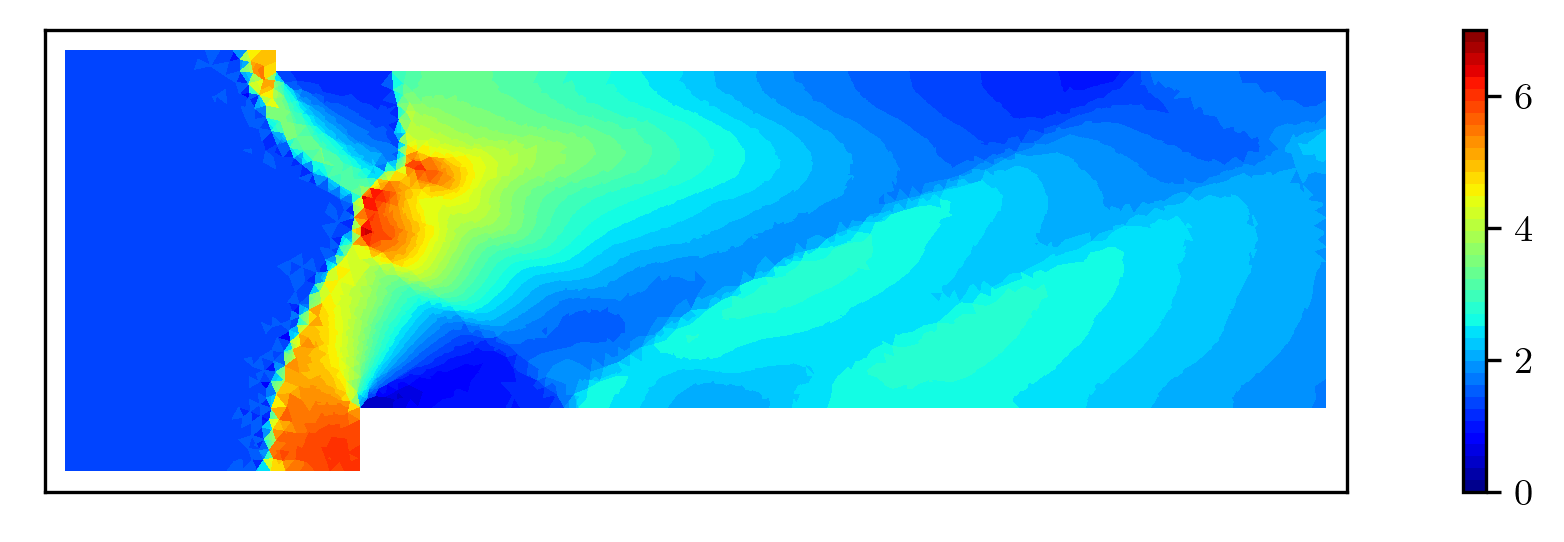}} &
            \raisebox{-0.5\height}{\includegraphics[width = .31\textwidth]{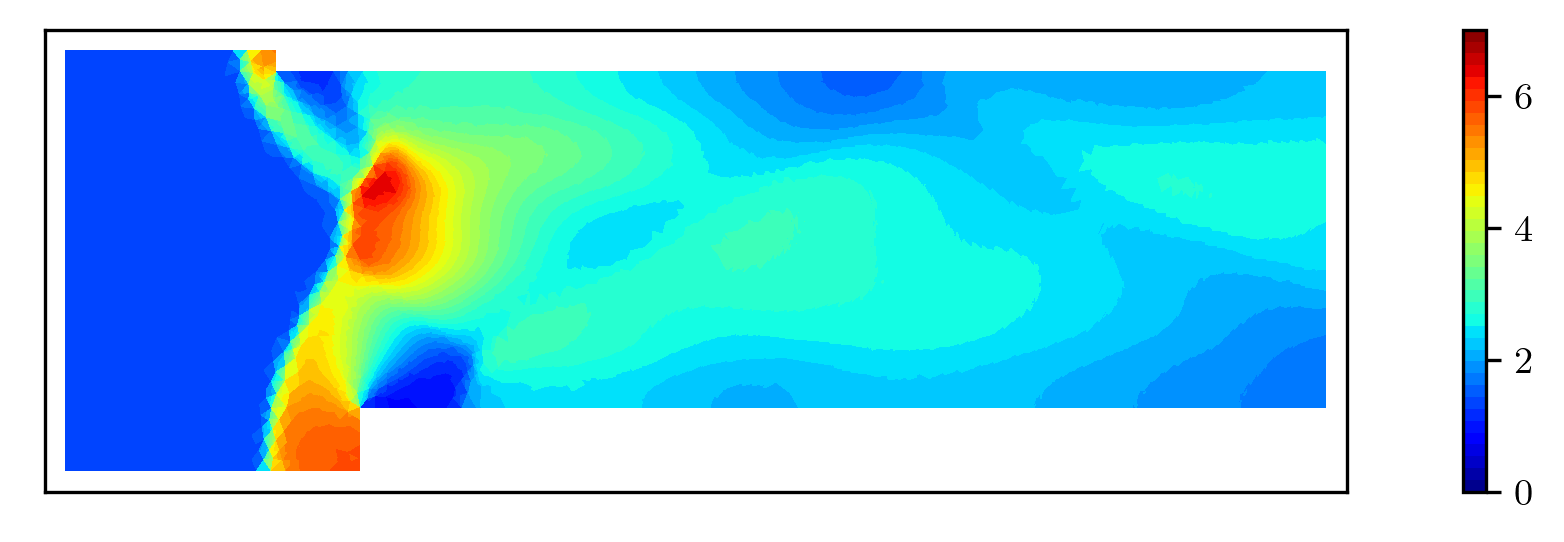}}
            \\
            &
            \rotatebox[origin=c]{90}{Error} &
            \raisebox{-0.5\height}{} &
            \raisebox{-0.5\height}{\includegraphics[width = .31\textwidth]{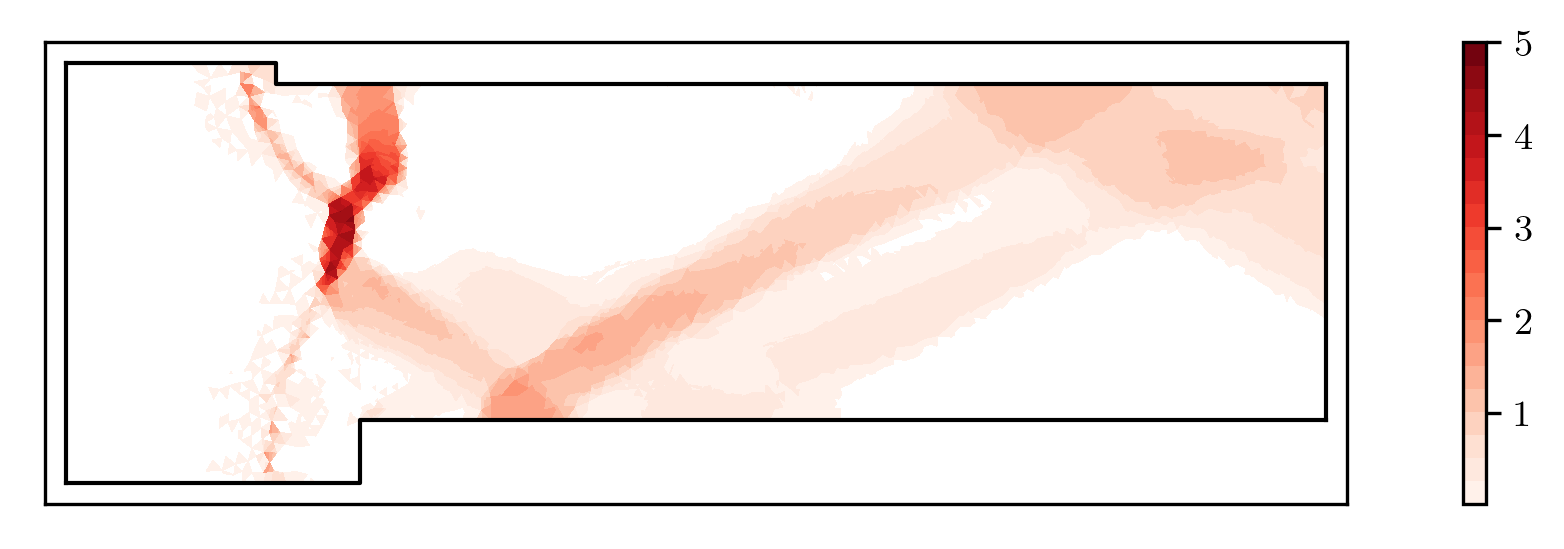}} &
            \raisebox{-0.5\height}{\includegraphics[width = .31\textwidth]{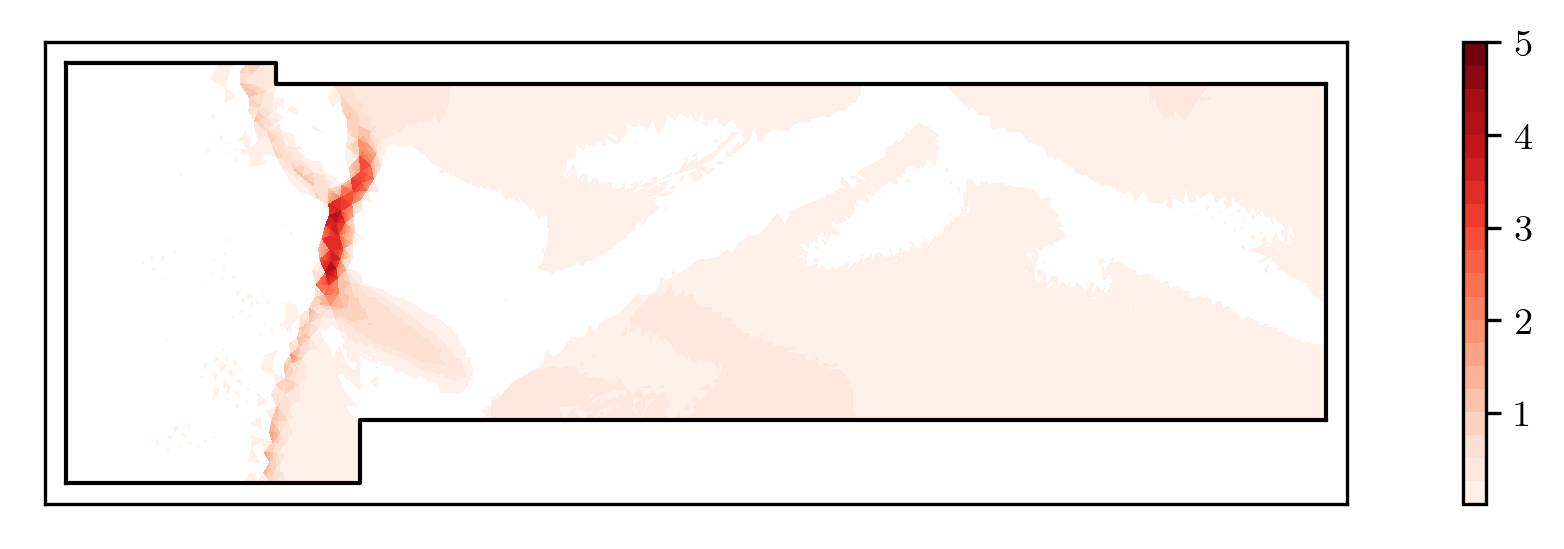}}
        \end{tabular*}
        \caption{{\bf 2D forward facing step - HLL flux training data:} predicted density field obtained by \nDGNet and \mcDGNet approaches and corresponding prediction pointwise error $\rho_{\text{DG}} - \rho_{\text{pred}}$  at time step $T_\text{test} = 4s$.} 
        \figlab{Roe_flux_prediction_solution_model1_mesh}
\end{figure}

In this problem, we implement the \oDGNet approach with the training data generated using the HLL numerical flux scheme for the Forward Facing Step problem. All the training settings are inherited from \cref{sect:forward_facing_conner}. The test data relative $L^2$-error average of three conservative components $\LRp{\rho, \rho u, E}$ between predicted \nDGNet and \mcDGNet solutions and traditional DG solutions over the test time interval $\LRs{0,4}$s is presented in \cref{fig:Roeflux_Relative_error_test_data_model1_model2_mesh}. The \mcDGNet approach gives more accurate predictions compared to the \nDGNet approach. This is due to the implicit regularization effect of data randomization. \cref{fig:Roe_flux_prediction_solution_model1_mesh}  shows the predicted density field and corresponding prediction pointwise error at the final time step $T_\text{test} = 4$s. Using the HLL flux scheme, the traditional DG method captures sharper shock compared to the case of using the Lax-Friedrichs flux scheme for both Model 1 and Model 2. The \nDGNet approach shows less accurate predictions than \mcDGNet in the vicinity of the shock on Model 1. For Model 2, the \nDGNet prediction error is significantly larger than \mcDGNet. This implies that the \nDGNet approach has worse generalization capability when solving for an unseen geometry. However, this is not the case for the \mcDGNet approach. The \mcDGNet approach is capable of predicting quite well for both Model 1 and Model 2. Interestingly, the \mcDGNet trained with HLL flux data can capture the sharper shocks compared to those \mcDGNet trained with the Lax-Freidrichs flux data, highlighting the robustness and adaptability of the \mcDGNet approach when training with data from different numerical flux schemes. Consequently, the \mcDGNet approach is a promising approach for more complex shock-type data or practical data.

% \clearpage

\subsection{Robustness to Random Neural Network Initializers}
\seclab{Random_initializers_and_noise_seeds}

\begin{figure}[htb!]
    \centering
        \begin{tabular*}{\textwidth}{c@{\hskip -0.0cm} c@{\hskip -0.0cm}}
            \centering
            \raisebox{-0.5\height}{\includegraphics[width = .48\textwidth]{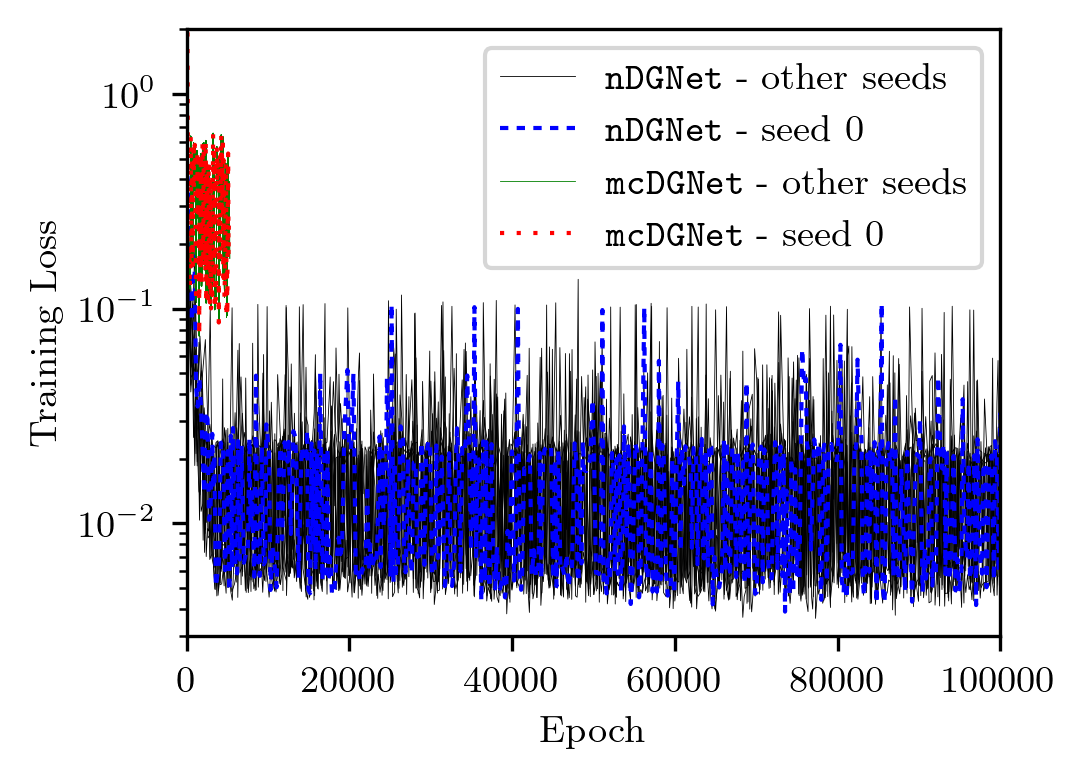}} &
            \raisebox{-0.5\height}{\includegraphics[width = .48\textwidth]{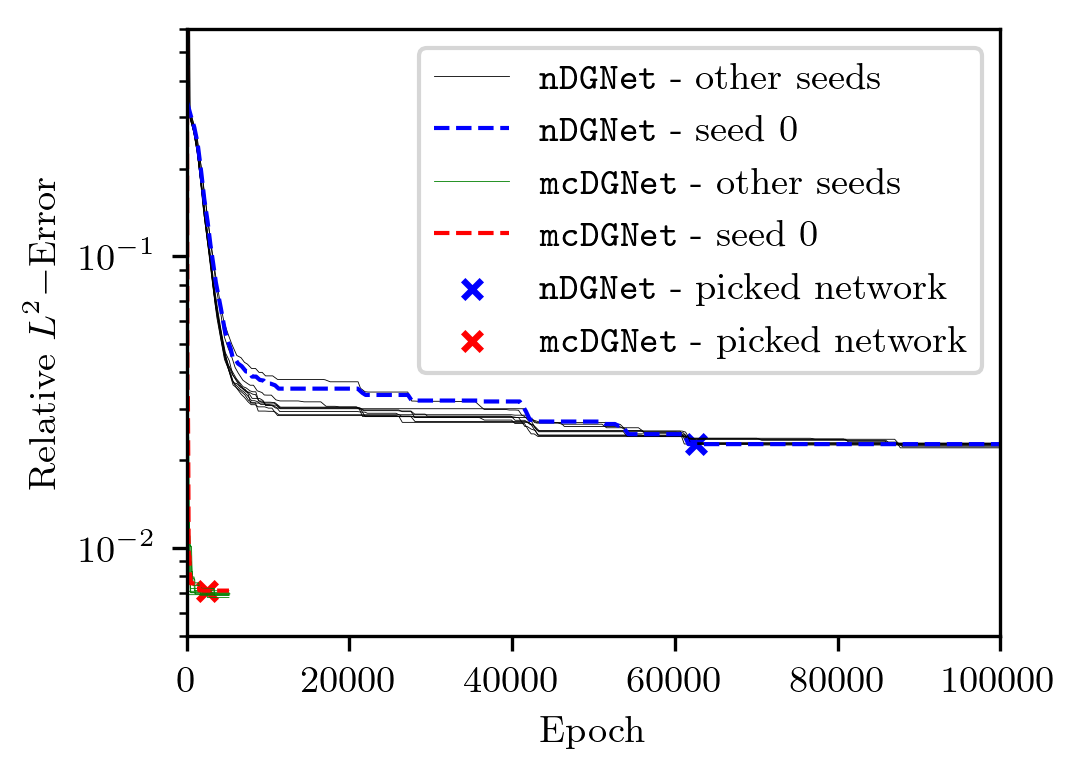}}
        \end{tabular*}
        \caption{{\bf Random initializers:} train loss ({\bf Left}) and validation loss ({\bf Right}) versus the training epoch for \nDGNet and \mcDGNet approaches over 10 random seeds.}
        \figlab{random_initializer_train_loss_validation_loss}
\end{figure}

\begin{figure}[htb!]
    \centering
    \resizebox{.8\textwidth}{!}{\input{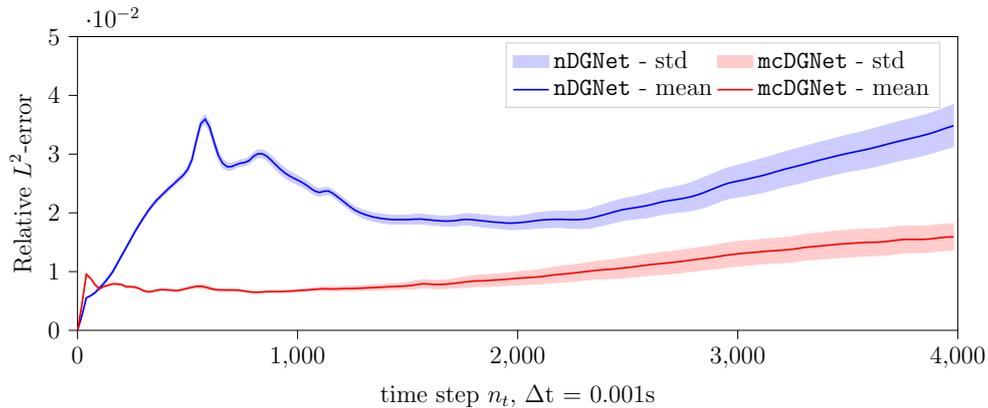}}
        \caption{{\bf Random initializers:} mean and standard deviation of test data relative $L^2$-error obtained by \nDGNet and \mcDGNet approaches over 10 instances of random neural network initializers at different time steps.}
        \figlab{mean_var_random_initializer}
\end{figure}

\begin{figure}[htb!]
    \centering
        \begin{tabular*}{\textwidth}{c@{\hskip -0.0001cm} c@{\hskip -0.002cm} c@{\hskip -0.002cm} c@{\hskip -0.002cm}}
            \centering
            & DG & \nDGNet & \mcDGNet ($2\%$)
            \\
            \rotatebox[origin=c]{90}{mean} &
            \raisebox{-0.5\height}{\includegraphics[width = .32\textwidth]{Figs/2D_Euler_Forward_facing/Model_1DG.png}} &
            \raisebox{-0.5\height}{\includegraphics[width = .32\textwidth]{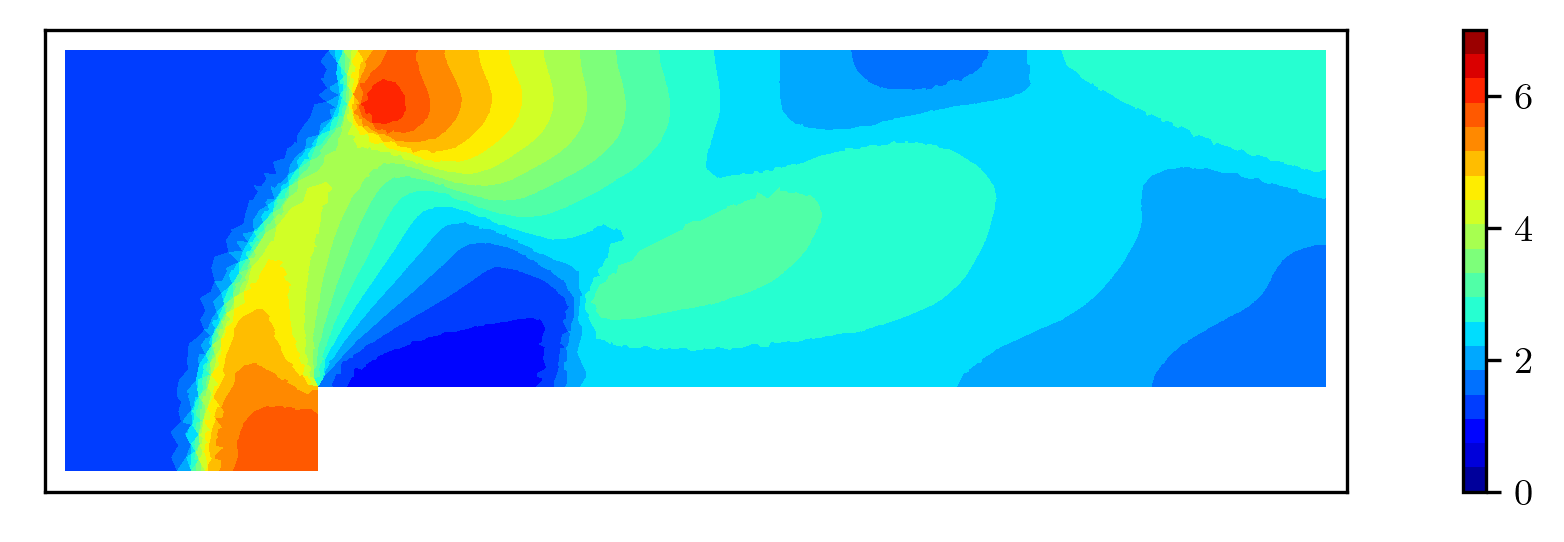}} &
            \raisebox{-0.5\height}{\includegraphics[width = .32\textwidth]{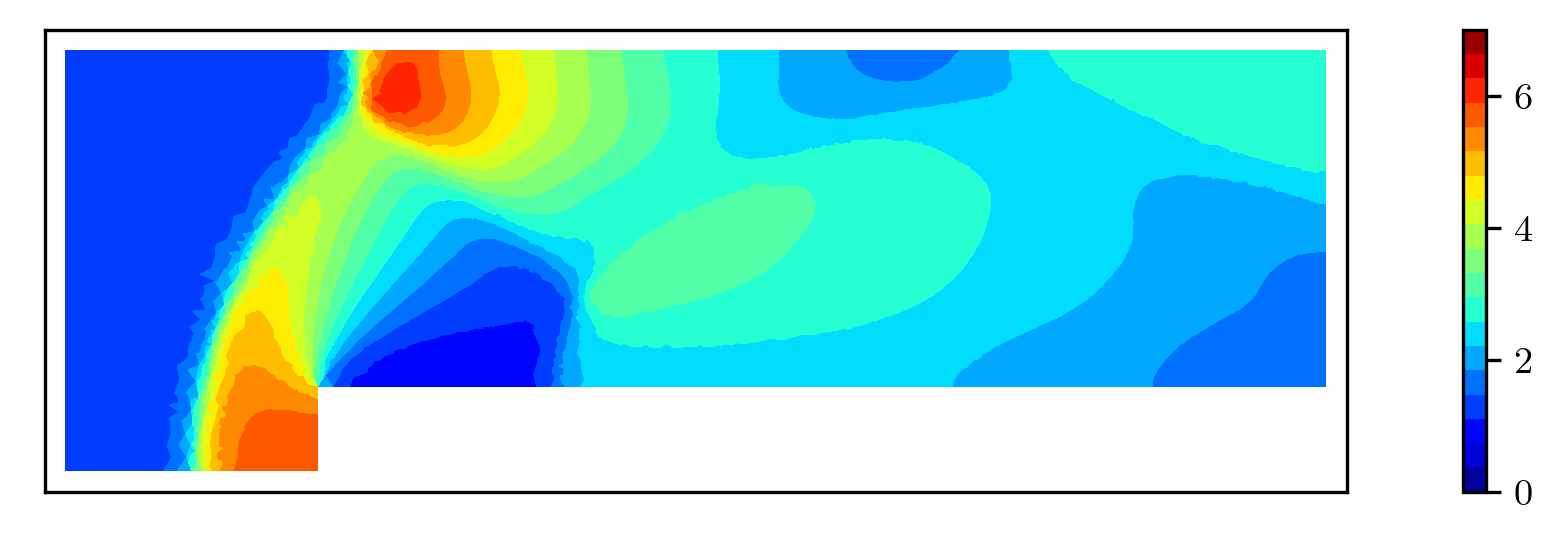}}
            \\
            \rotatebox[origin=c]{90}{std} &
            \raisebox{-0.5\height}{} &
            \raisebox{-0.5\height}{\includegraphics[width = .32\textwidth]{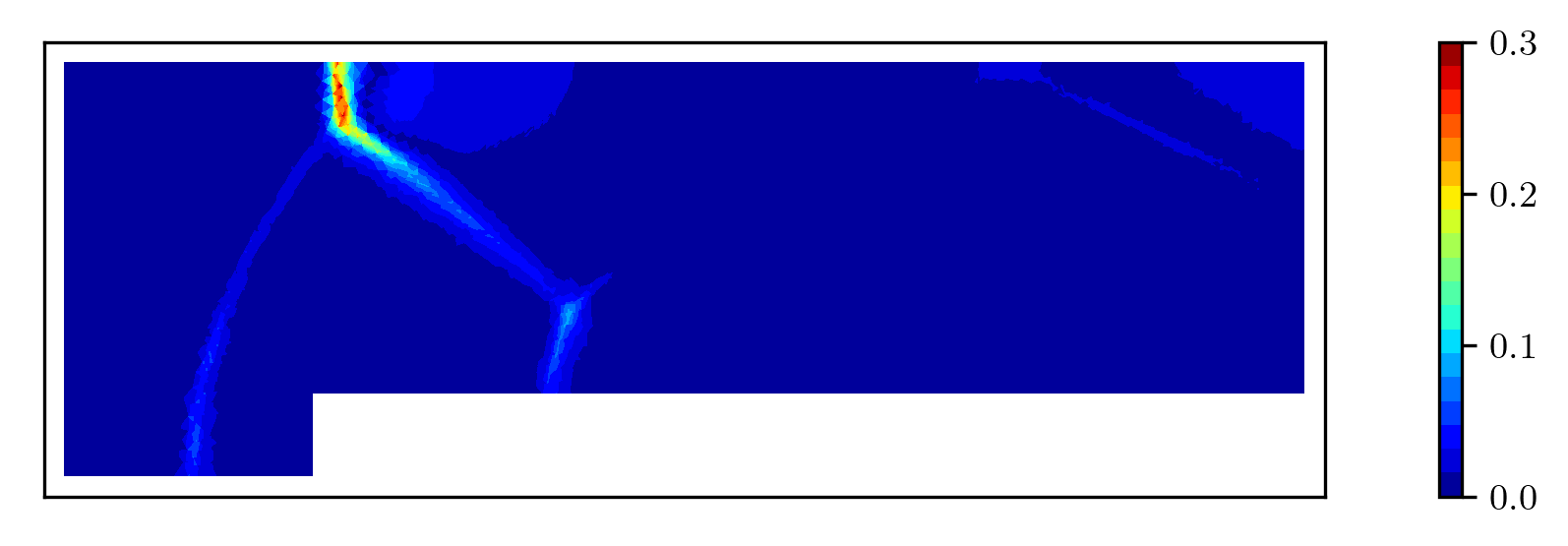}} &
            \raisebox{-0.5\height}{\includegraphics[width = .32\textwidth]{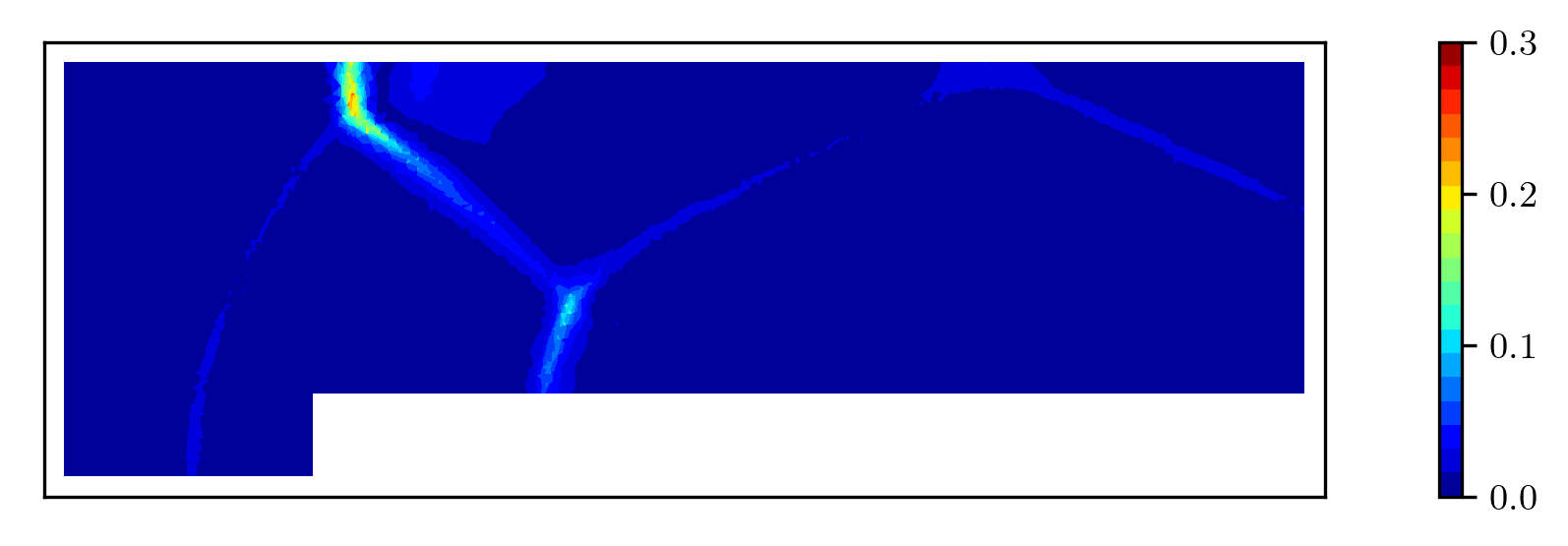}}
        \end{tabular*}
        \caption{{\bf Random initializers:} mean and standard deviation of density field obtained by \nDGNet and \mcDGNet approaches over 10 random neural network initializers at final time step $T_\text{test} = 4s$.} 
        \figlab{mean_var_predictions_random_initializer}
\end{figure}

In this section, we investigate the robustness of the \oDGNet framework with respect to random neural network initializers. We generate ten random sets of initial weights and biases of the \oDGNet networks for training on the Forward Facing Step problem. All the training settings are inherited from \cref{sect:forward_facing_conner}. The training loss and relative $L^2-$error for the validation data set are depicted in \cref{fig:random_initializer_train_loss_validation_loss}. The results show that across all random seeds, the \mcDGNet (2\%noise) approach achieves faster convergence to the best accuracy on the validation data set compared to the \nDGNet approach. In addition, we also evaluate the mean and standard deviation of the relative $L^2-$error of predictions for the test data set at test time steps, as illustrated in \cref{fig:mean_var_random_initializer}. Notably, the relative $L^2-$error mean for \nDGNet approach is consistently higher than that of \mcDGNet approach. Furthermore, the standard deviation for the \nDGNet approach increases over time more than the standard deviation for the \mcDGNet approach, implying that the \mcDGNet approach exhibits greater robustness to random initializers and is consistently more accurate than the \nDGNet approach. \cref{fig:mean_var_predictions_random_initializer} presents the mean and standard deviation of snapshot solutions from \nDGNet and \mcDGNet approaches at time $T_\text{test} = 8s$.  The \nDGNet prediction exhibits higher variability (more uncertainty) at points where intense shocks occur compared to the \mcDGNet approach. This, again, highlights the \mcDGNet approach is able to capture sharp shock structures more accurately than the \nDGNet approach.

\subsection{Hypersonic Flow through sphere cone}
\seclab{Hypersonic_Flow}

In this section, we investigate the performance of \oDGNet for the high hypersonic flow (Mach = 15) through a sphere cone, investigated in \cite{drikakis1993accuracy}. The geometry with a mesh is shown in the left subfigure in \cref{fig:Hypersonic_Flow_mesh_error} with $\rho_\infty = 0.002 \frac{kg}{m^3}, M = 15, p_0 = 170 \frac{N}{m^2}, T = 295K$. The domain is discretized to $37888$ triangular elements. The training data is generated with $\gamma \in \LRc{1.2, 1.6}$ and the validation data is generated with $\gamma = 1.4$ in the time interval $\LRs{0, 3\times 10^{-4}}$s. The time step size is $\dt = 5 \times 10^{-7}$s.

\begin{figure}[htb!]
    \centering
        \begin{tabular*}{\textwidth}{c@{\hskip -0.0cm} c@{\hskip -0.0cm}}
            \centering
            \includegraphics[width = .3\textwidth]{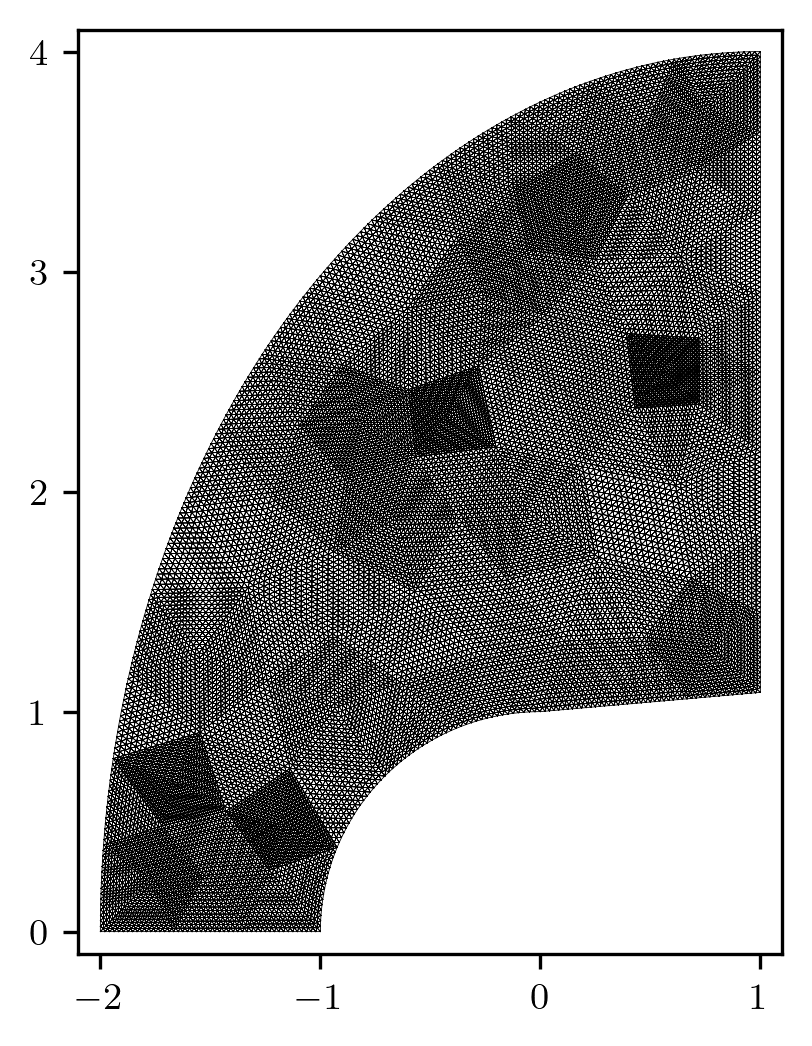} &
            \raisebox{-.0\height}{\resizebox{.7\textwidth}{!}{% This file was created with tikzplotlib v0.10.1.
\begin{tikzpicture}

\definecolor{darkgray176}{RGB}{176,176,176}
\definecolor{lightgray204}{RGB}{204,204,204}

\begin{axis}[
  width=0.7\textwidth,
  height=0.4\textwidth,
  legend cell align={left},
  legend style={
  fill opacity=0.8,
  draw opacity=1,
  text opacity=1,
  at={(0.03,0.97)},
  anchor=north west,
  draw=lightgray204
},
tick align=outside,
tick pos=left,
x grid style={darkgray176},
xlabel={time step \(\displaystyle n_t\), \(\displaystyle \Delta\)t = 5e-07s},
xmin=0, xmax=3000,
xtick={0,1000,2000,3000},
xtick style={color=black},
y grid style={darkgray176},
ylabel={Relative \(\displaystyle L^2\)-Error},
ymin=-0.0, ymax=0.2,
ytick={0, 0.1, 0.2},
ytick style={color=black},
]
\addplot [semithick, black, mark=x, mark size=2.5, mark repeat=20, mark options={solid}]
table {%
0 0
20 0.00619501434266567
40 0.00982013903558254
60 0.0120285488665104
80 0.0133296055719256
100 0.014854378066957
120 0.017359584569931
140 0.0188360996544361
160 0.0214205011725426
180 0.0223114024847746
200 0.0253668334335089
220 0.0262557491660118
240 0.0281777735799551
260 0.0301908906549215
280 0.0308875627815723
300 0.0338013395667076
320 0.0356782227754593
340 0.0368887707591057
360 0.0401219055056572
380 0.042598020285368
400 0.0448698215186596
420 0.0480800978839397
440 0.0512813031673431
460 0.054208792746067
480 0.0569460764527321
500 0.0593377165496349
520 0.0613932237029076
540 0.063091479241848
560 0.0645144060254097
580 0.0659320503473282
600 0.0672717839479446
620 0.0682600066065788
640 0.0688118636608124
660 0.0691985785961151
680 0.0698937103152275
700 0.0706328898668289
720 0.0708915442228317
740 0.0708660557866096
760 0.0709256008267403
780 0.0713970437645912
800 0.0724749490618706
820 0.0735321938991547
840 0.0741143748164177
860 0.0745159909129143
880 0.0750828012824059
900 0.0760365724563599
920 0.0775407552719116
940 0.0791831538081169
960 0.0805807635188103
980 0.0816169530153275
1000 0.0826425850391388
1020 0.0839095935225487
1040 0.0855074375867844
1060 0.0873949155211449
1080 0.0893508046865463
1100 0.0910307988524437
1120 0.0923949480056763
1140 0.0935696959495544
1160 0.0948033630847931
1180 0.0962238609790802
1200 0.0978318601846695
1220 0.0996112823486328
1240 0.101423874497414
1260 0.103104338049889
1280 0.104620359838009
1300 0.106012284755707
1320 0.107281923294067
1340 0.108633667230606
1360 0.110000908374786
1380 0.111493661999702
1400 0.113005131483078
1420 0.114599525928497
1440 0.116065889596939
1460 0.117480181157589
1480 0.118752330541611
1500 0.119964703917503
1520 0.121093437075615
1540 0.122234247624874
1560 0.123457074165344
1580 0.124654874205589
1600 0.125847667455673
1620 0.126990675926208
1640 0.128143951296806
1660 0.129189699888229
1680 0.130227208137512
1700 0.131150335073471
1720 0.132109016180038
1740 0.133057996630669
1760 0.133972689509392
1780 0.134979546070099
1800 0.135939627885818
1820 0.137014359235764
1840 0.138141706585884
1860 0.139159247279167
1880 0.140270441770554
1900 0.141322076320648
1920 0.142257243394852
1940 0.143254056572914
1960 0.144155263900757
1980 0.144944697618484
2000 0.145886570215225
2020 0.146861553192139
2040 0.14788793027401
2060 0.149084240198135
2080 0.150266319513321
2100 0.151428923010826
2120 0.152640014886856
2140 0.15384829044342
2160 0.155017271637917
2180 0.156187757849693
2200 0.157341659069061
2220 0.158453345298767
2240 0.159551307559013
2260 0.160668566823006
2280 0.161829560995102
2300 0.16300755739212
2320 0.164137840270996
2340 0.165277689695358
2360 0.166451379656792
2380 0.167552977800369
2400 0.168535828590393
2420 0.169487357139587
2440 0.170421600341797
2460 0.171289443969727
2480 0.172094196081161
2500 0.172876998782158
2520 0.173627197742462
2540 0.174336835741997
2560 0.17502461373806
2580 0.175691530108452
2600 0.176343485713005
2620 0.176955580711365
2640 0.177586764097214
2660 0.178266450762749
2680 0.178917944431305
2700 0.17950439453125
2720 0.180041655898094
2740 0.180602431297302
2760 0.181182980537415
2780 0.181729197502136
2800 0.182235866785049
2820 0.182753682136536
2840 0.183320283889771
2860 0.18388769030571
2880 0.184383377432823
2900 0.184820115566254
2920 0.185236185789108
2940 0.185658067464828
2960 0.18607372045517
2980 0.186444208025932
};
\addlegendentry{\nDGNet }
\addplot [semithick, red, mark=*, mark size=2.5, mark repeat=20, mark options={solid,fill=white}]
table {%
0 0
20 0.00738803949207067
40 0.0124597027897835
60 0.0125179868191481
80 0.00975339114665985
100 0.00958448462188244
120 0.00992880295962095
140 0.0109256077557802
160 0.0104167759418488
180 0.0116067454218864
200 0.0117220086976886
220 0.0125309517607093
240 0.0131093394011259
260 0.0134948147460818
280 0.0142381172627211
300 0.0147797055542469
320 0.0152445305138826
340 0.0156356208026409
360 0.0162688344717026
380 0.0168494507670403
400 0.0171128511428833
420 0.0177912190556526
440 0.0185832064598799
460 0.0191121324896812
480 0.0195596385747194
500 0.0201134793460369
520 0.0205648578703403
540 0.020720673725009
560 0.0206531081348658
580 0.0204699616879225
600 0.0202800668776035
620 0.0199760030955076
640 0.0195099022239447
660 0.0189430322498083
680 0.0183508098125458
700 0.0177650190889835
720 0.0171440206468105
740 0.0164604820311069
760 0.0157846417278051
780 0.015267102047801
800 0.0148890297859907
820 0.0145062357187271
840 0.014096456579864
860 0.0136463530361652
880 0.0132503313943744
900 0.0130179850384593
920 0.0128918904811144
940 0.0128587428480387
960 0.0128301335498691
980 0.0127840638160706
1000 0.0127088967710733
1020 0.0126421861350536
1040 0.0127060515806079
1060 0.0128285540267825
1080 0.0129932146519423
1100 0.0131245143711567
1120 0.0131785087287426
1140 0.0131889954209328
1160 0.0131924897432327
1180 0.0132322711870074
1200 0.0133834136649966
1220 0.0135596012696624
1240 0.0136603247374296
1260 0.0137560470029712
1280 0.0137650147080421
1300 0.0137775940820575
1320 0.0137267112731934
1340 0.0136654442176223
1360 0.0136195532977581
1380 0.0135803911834955
1400 0.0136064235121012
1420 0.0135736521333456
1440 0.013494685292244
1460 0.0134306009858847
1480 0.0134076718240976
1500 0.0133686903864145
1520 0.0133202243596315
1540 0.0132269337773323
1560 0.0131614869460464
1580 0.0130994645878673
1600 0.0130574256181717
1620 0.0129629839211702
1640 0.0128531903028488
1660 0.0126434369012713
1680 0.0125261005014181
1700 0.0123756621032953
1720 0.0123206004500389
1740 0.0122615499421954
1760 0.0121994568035007
1780 0.0121796894818544
1800 0.0120915453881025
1820 0.0120479157194495
1840 0.0119960345327854
1860 0.0119251105934381
1880 0.011888911947608
1900 0.0118249645456672
1920 0.0117705166339874
1940 0.0117421709001064
1960 0.0116502810269594
1980 0.0116232968866825
2000 0.0116380453109741
2020 0.0116233360022306
2040 0.0116746108978987
2060 0.0116638047620654
2080 0.0116202794015408
2100 0.0116330338642001
2120 0.0116052236407995
2140 0.0115651860833168
2160 0.011625949293375
2180 0.0116788847371936
2200 0.0116801289841533
2220 0.0117477253079414
2240 0.0118310768157244
2260 0.0118659753352404
2280 0.0119020286947489
2300 0.0119602642953396
2320 0.0119901252910495
2340 0.0120123866945505
2360 0.0120355486869812
2380 0.0120773315429688
2400 0.0120977256447077
2420 0.0121099753305316
2440 0.0121368877589703
2460 0.0121814412996173
2480 0.0121983839198947
2500 0.0122607303783298
2520 0.0123574640601873
2540 0.0124534722417593
2560 0.0125622898340225
2580 0.0126591864973307
2600 0.0127560347318649
2620 0.0128398006781936
2640 0.0128899728879333
2660 0.0129123479127884
2680 0.0129396831616759
2700 0.0129716657102108
2720 0.0129570513963699
2740 0.0129542704671621
2760 0.0130000561475754
2780 0.0130671225488186
2800 0.0131260510534048
2820 0.0132028739899397
2840 0.0132979741320014
2860 0.0134060410782695
2880 0.0134865362197161
2900 0.0135338259860873
2920 0.0136065986007452
2940 0.013703154399991
2960 0.0137937795370817
2980 0.013869209215045
};
\addlegendentry{\mcDGNet ($11\%$) }
\end{axis}

\end{tikzpicture}}} 
        \end{tabular*}
        \caption{{\bf 2D Hypersonic flow sphere-cone:} ({\bf Left}) Geometry and a mesh with  K = 37888 elements. ({\bf Right}) test data relative $L^2$-error average over conservative components $\LRp{\rho, \rho u, E}$ predictions obtained by \nDGNet and \mcDGNet approaches at different time steps.}
        \figlab{Hypersonic_Flow_mesh_error}
\end{figure}
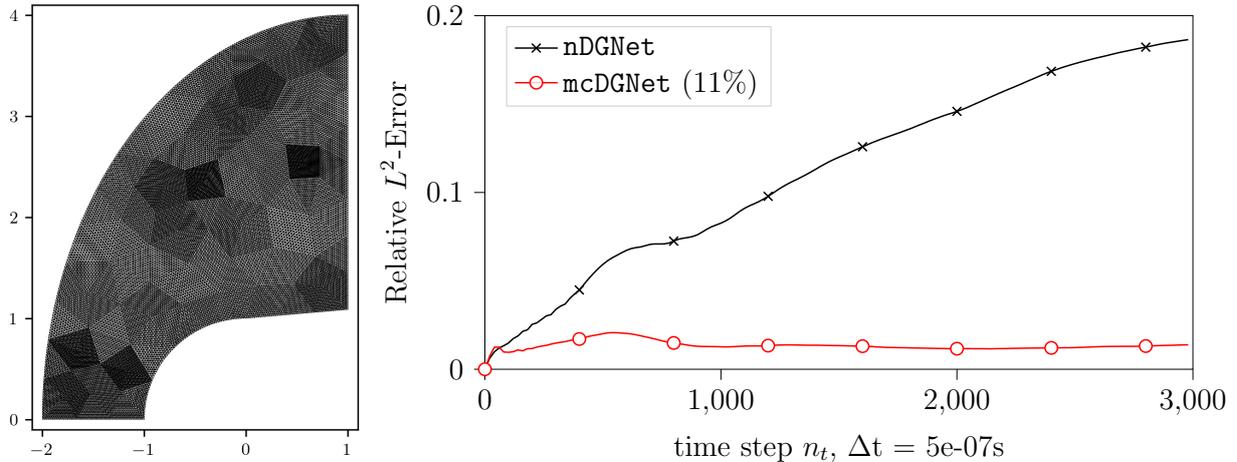

\begin{figure}[htb!]
    \centering
        \begin{tabular*}{\textwidth}{c@{\hskip -0.0001cm} c@{\hskip -0.002cm} c@{\hskip -0.002cm} c@{\hskip -0.002cm}}
            \centering
            & DG & \nDGNet  & \mcDGNet  ($11\%$)
            \\
            \rotatebox[origin=c]{90}{Pred} &
            \raisebox{-0.5\height}{\includegraphics[width = .31\textwidth]{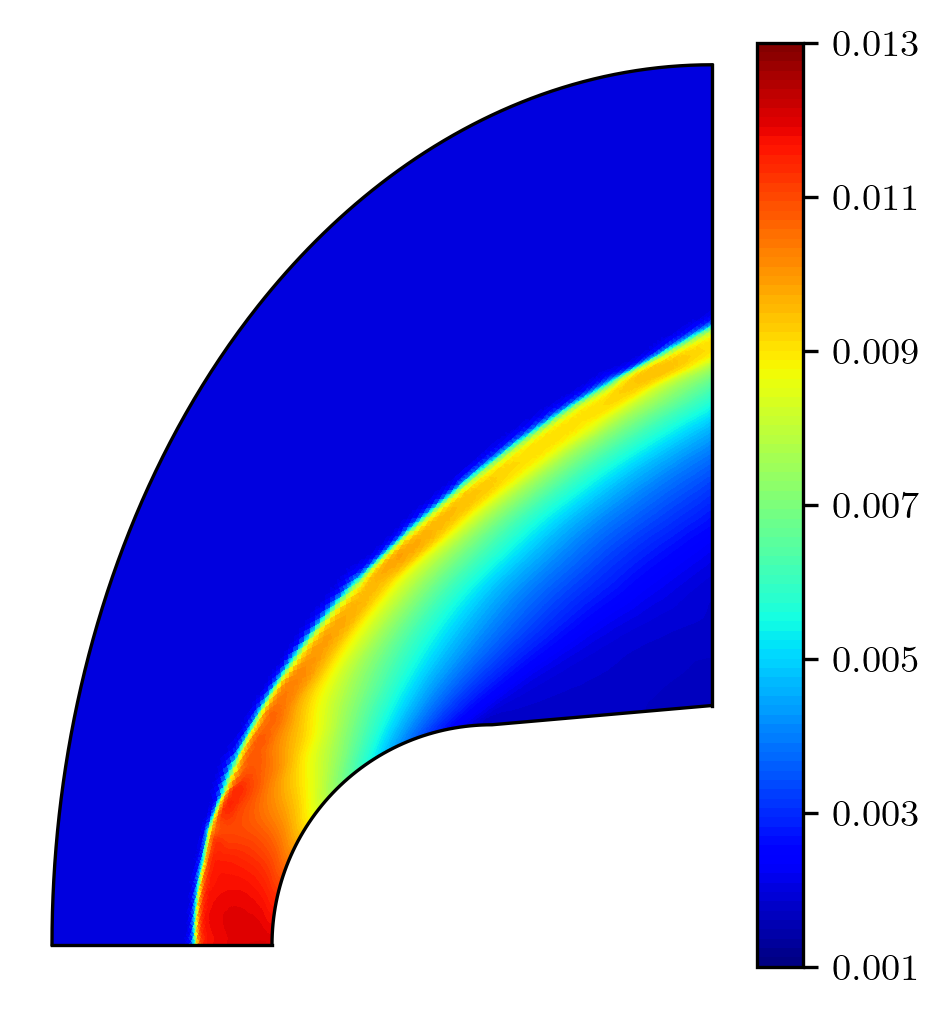}} &
            \raisebox{-0.5\height}{\includegraphics[width = .31\textwidth]{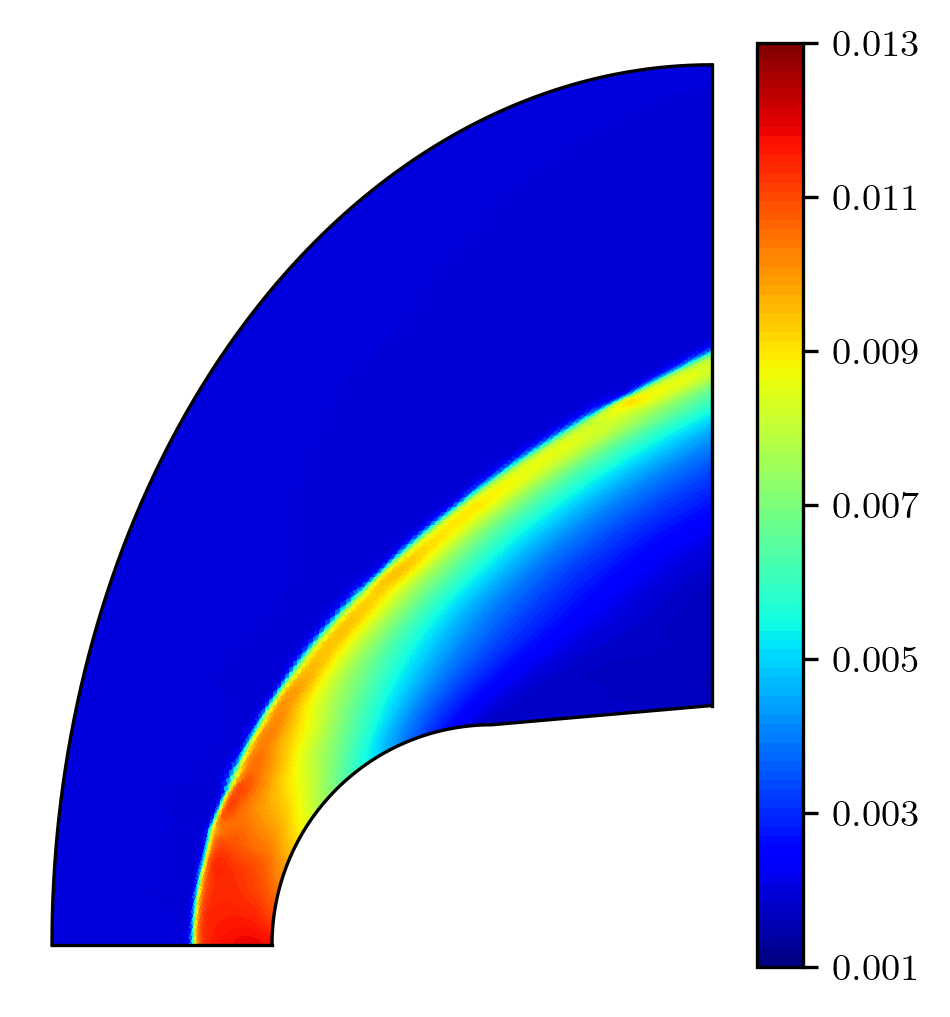}} &
            \raisebox{-0.5\height}{\includegraphics[width = .31\textwidth]{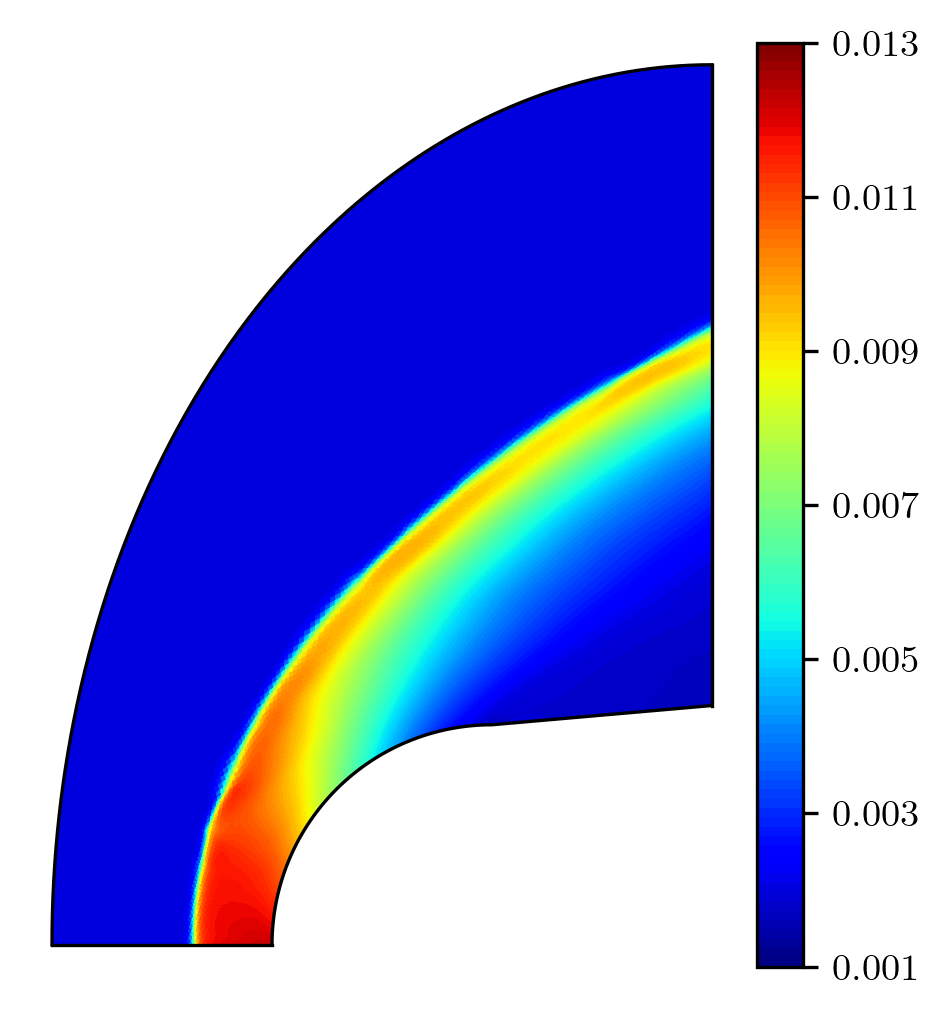}}
            \\
            \rotatebox[origin=c]{90}{Error} &
            \raisebox{-0.5\height}{} &
            \raisebox{-0.5\height}{\includegraphics[width = .31\textwidth]{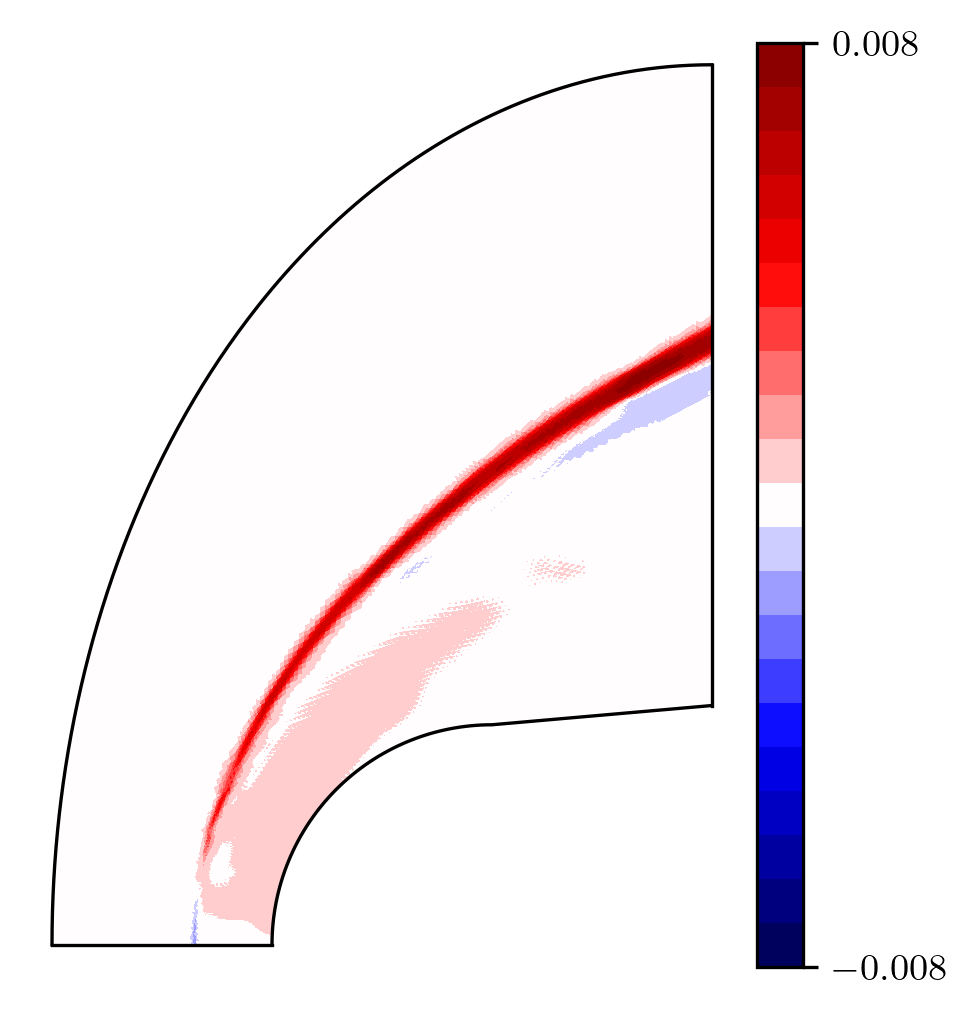}} &
            \raisebox{-0.5\height}{\includegraphics[width = .31\textwidth]{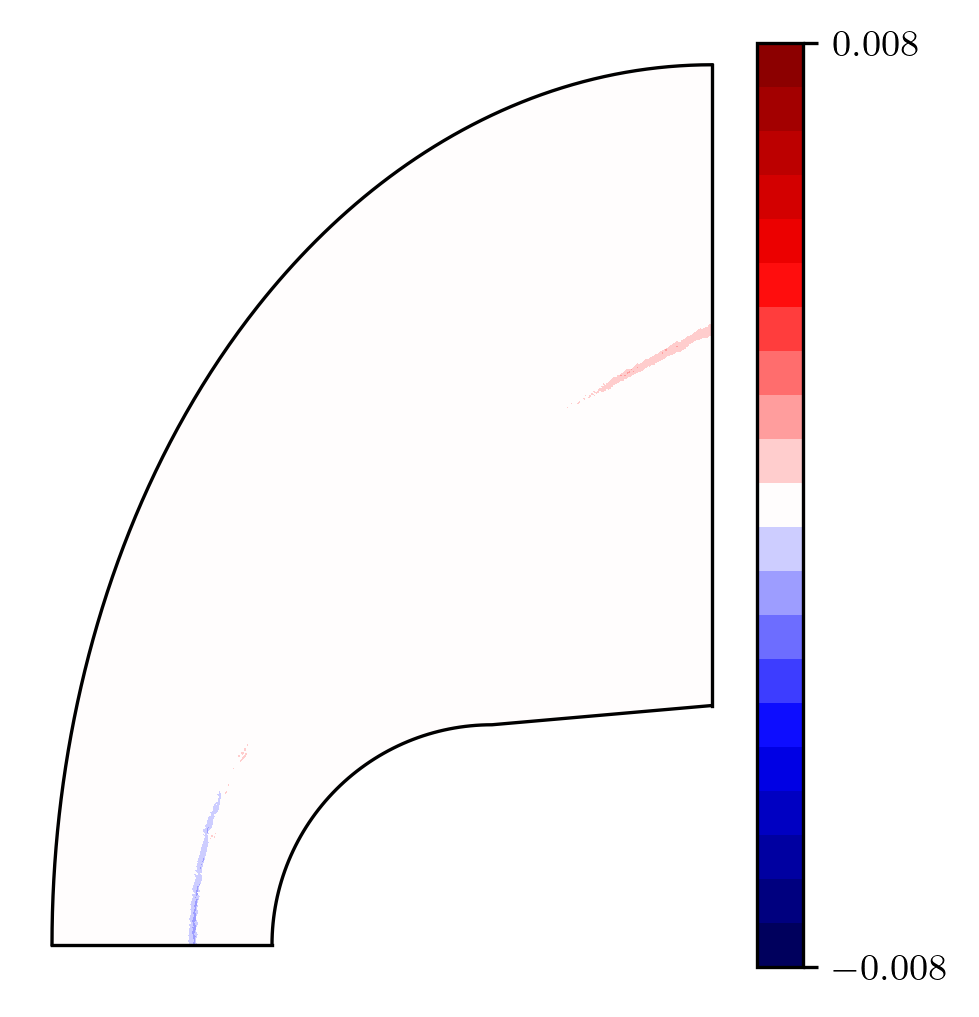}}
        \end{tabular*}
        \caption{{\bf 2D Hypersonic flow sphere-cone:} predicted density field obtained by \nDGNet and \mcDGNet approaches and corresponding prediction pointwise error $\rho_{\text{DG}} - \rho_{\text{pred}}$  at time step $T_\text{test} = 1.5 \times 10^{-3}s$.} 
        \figlab{Hypersonic_Flow_mesh_predictions}
\end{figure}

The relative $L^2$-error averaged over conservative components $(\rho, \rho u, E)$ for predictions by \nDGNet and \mcDGNet approaches at various time steps is illustrated in the right subfigure in \cref{fig:Hypersonic_Flow_mesh_error}. The results demonstrate that the \mcDGNet approach consistently outperforms the \nDGNet approach across all time steps. This superior performance can be attributed to the enhancement of \mcDGNet approach through regularization terms derived from model-constrained models and the data randomization technique. In contrast, the \nDGNet approach, being purely data-driven without any regularization terms, exhibits higher error rates.
\Cref{fig:Hypersonic_Flow_mesh_predictions} displays the predicted density fields obtained by both \nDGNet and \mcDGNet approaches, along with their corresponding pointwise prediction errors ($\rho_{\text{DG}} - \rho_{\text{pred}}$) at time step $T_\text{test} = 1.5 \times 10^{-3}s$. These results show that the \mcDGNet approach yields more accurate long-term predictions, especially in shock regions, than the \nDGNet approach.

% \clearpage

\subsection{Train, Validation and Test Computation Time}
\seclab{Train_validation_test_computation_time}

In this section, we discuss the computation cost and acceleration benefits of the proposed \oDGNet approaches. The total time for training is the sum of the cost of the training step and validation step. The validation cost is more significant than the training cost. This is because we only solve one time forward for a batch of training samples and update the network with \texttt{ADAM} optimizer, while we have to solve $\frac{T_\text{train}}{\dt}$ time steps during the validation phase from validation initial condition. Ideally, validation should be evaluated at every training epoch, but it is not a strict requirement. To decrease the overall training time, we calculate validation loss at chosen epoch intervals, for example, every 200 epochs for the Airfoil problem. This approach might lead to a trade-off where a better network might be missed if the best validation is not coincident with the designed epochs. However, based on our experiment on the Forward Facing Step problem, the best-selected network (validated every epoch) is insignificantly better than the network obtained by sparse validation every 10 epochs. The test time is the total time required to solve the equations for the test data set, either by \oDGNet networks or the traditional DG method. The speed-up is the ratio of DG test time to \oDGNet test time. \cref{tab:computational_time} summarizes the training, validation, and test computation time, along with the corresponding speed-up rate of the \oDGNet learning methods, and the GPU hardware used for the Forward Facing Step, Scramjet, Airfoil, Euler Benchmark configuration 6Double Mach Reflection, and Hypersonic Sphere Cone. The \mcDGNet converges to the selected network significantly faster than the \nDGNet method for some problems. For instance, the highest convergence rate ratio between the \mcDGNet and \nDGNet method is approximately 54 for the Scramjet problem, while the lowest one is 1.9 for the Double Mach Reflection problem. Note that in this problem the noise-free data is sufficiently informative, thus we expect similar behaviors from both \nDGNet and \mcDGNet approaches in all aspects including training, validation, and test accuracy. In comparison with the traditional DG method, we can use pre-trained \mcDGNet networks to solve shock-type problems with a speed-up rate of 5.24, 4.26, 7.61, 3.62, 3.08, and 6.32 for the Forward Facing Step, Scramjet, Airfoil, Euler Benchmark configuration 6, Double Mach Reflection, and hypersonic flow sphere-cone problems, respectively. While the speed-up is moderate since the cost of computing the Lax-Friedrichs flux scheme is not expensive, it shows that neural network approach is still beneficial as it is significantly faster than even explicit DG method with the simple Lax-Friedrichs approach.
%Indeed, in the code implementation, the numerical flux is evaluated using a well-formulated equation, rather than directly solving eigen-decomposition problems. Therefore, we anticipate a significant speed-up rate for problems with expensive flux computations. 
Also, as discussed in \cite{magiera2020constraint}, there are scenarios in particular simulations where the Riemann solver is not available. In that case, the advantages of our \oDGNet approach become clear. Indeed, we can learn surrogate models for the Riemann solver from observation data, and use surrogate models to accelerate the computation. However, it would be ideal to see an acceleration of several orders of magnitude. There are several potential ideas to gain significant speed-up for time-dependent problems. In \cite{nastorg2022dsgpsdeepstatistical}, Nastorg et al. developed an iterative procedure with a recurrent graph neural network architecture that solves the Poisson equation on an unstructured grid on a latent graph. Their approach showed a ten times speed-up over a traditional solver when GPU acceleration was available. In \cite{janny2023eaglelargescalelearningturbulent}, Janny et al. utilize a mesh transformer to cluster and pool a large graph into a low dimensional latent graph on which the solution is marched forward in time using a multi-head attention mechanism. The solution can then be upsampled to the original graph from the latent representation. This approach results in speed gains because the clustering is parallelizable and can be pre-computed in an offline manner, leaving graph pooling and time-stepping on the low dimensional latent graph as the main computational expense. We will consider encoding or pooling the mesh to a reduced dimensional latent graph to improve the computational performance of \oDGNet in future work. Finding a way to encode the graph while preserving problem shocks and network generalization capability between problems will be part of our findings.

% \usepackage{multirow}
\begin{table}[htb!]
    \caption{Training, validation, and test computation time of \nDGNet and \mcDGNet approaches and the speed improvement compared to traditional DG method for different problems.}
    \tablab{computational_time}
	\resizebox{\textwidth}{!}{
    \begin{tblr}{
  cells = {c},
  cell{1}{1} = {c=2,r=2}{},
  cell{1}{3} = {c=4}{},
  cell{1}{7} = {r=2}{},
  cell{1}{8} = {r=2}{},
  cell{1}{9} = {r=2}{},
  cell{1}{10} = {r=2}{},
  cell{3}{1} = {r=2}{},
  cell{3}{4} = {r=2}{},
  cell{3}{7} = {r=2}{},
  cell{3}{8} = {r=2}{},
  cell{3}{9} = {r=2}{},
  cell{3}{10} = {r=2}{},
  cell{5}{1} = {r=2}{},
  cell{5}{4} = {r=2}{},
  cell{5}{7} = {r=2}{},
  cell{5}{8} = {r=2}{},
  cell{5}{9} = {r=2}{},
  cell{5}{10} = {r=2}{},
  cell{7}{1} = {r=2}{},
  cell{7}{4} = {r=2}{},
  cell{7}{7} = {r=2}{},
  cell{7}{8} = {r=2}{},
  cell{7}{9} = {r=2}{},
  cell{7}{10} = {r=2}{},
  cell{9}{1} = {r=2}{},
  cell{9}{4} = {r=2}{},
  cell{9}{7} = {r=2}{},
  cell{9}{8} = {r=2}{},
  cell{9}{9} = {r=2}{},
  cell{9}{10} = {r=2}{},
  cell{11}{1} = {r=2}{},
  cell{11}{4} = {r=2}{},
  cell{11}{7} = {r=2}{},
  cell{11}{8} = {r=2}{},
  cell{11}{9} = {r=2}{},
  cell{11}{10} = {r=2}{},
  cell{13}{1} = {r=2}{},
  cell{13}{4} = {r=2}{},
  cell{13}{7} = {r=2}{},
  cell{13}{8} = {r=2}{},
  cell{13}{9} = {r=2}{},
  cell{13}{10} = {r=2}{},
  vlines,
  hline{1,3,5,7,9,11,13,15} = {-}{},
  hline{2} = {3-6}{},
  hline{4,6,8,10,12,14} = {2-3,5-6}{},
}
{Problems\\ and\\ Approaches} &                         & Training time             &                           &                  &                           & {Test\\ (sec)} & {DG\\ (sec)} & {Speed\\ Up} & GPU  \\
                              &                         & {Train \\ (sec\\ /epoch)} & {Vali-\\ dation \\ (sec)} & {Total \\ Epoch} & {Total \\ time\\ (hours)} &                &              &              &      \\
{Forward \\ Facing}           & \nDGNet  & 0.006                     & {2.07\\(10$^*$)}          & 62230            & 3.68                      & 6.45           & 33.85        & 5.24         & A100 \\
                              & \mcDGNet & 0.007                     &                           & 1560             & 0.09                      &                &              &              &      \\
Scramjet                      & \nDGNet  & 0.008                     & {1.94\\(10$^*$)}          & 151200           & 8.54                      & 5.63           & 24.02        & 4.26         & A100 \\
                              & \mcDGNet & 0.009                     &                           & 2750             & 0.16                      &                &              &              &      \\
Airfoil                       & \nDGNet  & 0.008                     & {2.68\\(200$^*$)}         & 1185800          & 7.09                      & 12.70          & 96.65        & 7.61         & A100 \\
                              & \mcDGNet & 0.011                     &                           & 472200           & 3.17                      &                &              &              &      \\
{Euler\\ Config6}             & \nDGNet  & 0.006                     & {5.44\\(500$^*$)}         & 552500           & 25.75                     & 18.82          & 68.19        & 3.62         & H100 \\
                              & \mcDGNet & 0.008                     &                           & 63000            & 3.30                      &                &              &              &      \\
{Double \\ Mach}              & \nDGNet  & 0.006                     & {8.75\\(100$^*$)}         & 172500           & 4.45                      & 37.46          & 115.45       & 3.08         & H100 \\
                              & \mcDGNet & 0.010                     &                           & 73400            & 2.35                      &                &              &              &      \\
{Sphere\\cone}                & \nDGNet  & 0005                      & {5.66\\(100$^*$)}         & 250000           & 4.27                      & 15.25          & 96.32        & 6.32         & A100 \\
                              & \mcDGNet & 0.007                     &                           & 75000            & 1.33                      &                &              &              &      
\end{tblr}
    }
    \newline $^*$validation is implemented every 10/100/200/500 epochs to reduce the total training time.
\end{table}

% TODO: adding 3D result?

% \clearpage

\section{Conclusions}
\seclab{conclusions}

In this paper, we presented the \oDGNet for solving compressible Euler equation with out-of-distribution generalization.
%, extending our previous work \cite{nguyen2022model}. 
Despite their power, neural surrogate models typically show limitations in generalizing to unseen scenarios and capturing the evolution of solution discontinuities. To address these challenges, we adopted five novel strategies: (i) leveraging time integration schemes to capture temporal correlation and exploiting neural network speed for computation time reduction; (ii) employing a model-constrained approach to ensure the learned tangent slope satisfies governing equations; (iii) utilizing a GNN-inspired architecture where edges represent Riemann solver surrogate models and nodes represent volume integration correction surrogate models, enabling discontinuity capture, aliasing error reduction, and mesh discretization generalizability; (iv) implementing an input normalization technique that allows surrogate models to generalize across different initial conditions, geometries, meshes, boundary conditions, and solution orders; and (v) incorporating a data randomization technique that not only implicitly promotes agreement between surrogate models and true numerical models up to second-order derivatives, ensuring long-term stability and prediction capacity, but also serves as a data generation engine during training, leading to enhanced generalization on unseen data. Comprehensive numerical results for 1D and 2D compressible Euler equation problems are conducted, including Sod-tube, Lax, Isentropic vortex, Forward facing step, Scramjet, Airfoil, Euler benchmarks, Double Mach Reflection, and Hypersonic Sphere Cone. Also, we showed that \oDGNet preserves convergence rates comparable to the classical DG method and exhibits robust performance across different neural network initializers.

Currently, our proof-of-concept \oDGNet can learn the DG solvers and provide accurate solutions for unseen problems with new shock structures. Though the speed-up, compared to the DG methods, is modest, our work has exhibited the attractive speed of neural network methods: they could be faster than even the explicit approach which is essentially matrix-vector products. With a proper graph auto-encoder, we expect to achieve orders of magnitude faster, and this is ongoing. 
%As discussed in \cite{magiera2020constraint}, surrogate models for Riemann solvers do not significantly enhance computational speed when well-formulated numerical fluxes are available. One of our future works is improving the speed-up factor for \oDGNet \hspace{-1ex}. 
On the other hand, applying \oDGNet to 3D large-scale problems and real-world applications such as weather forecasting is another promising future research direction that we are pursuing.

% \clearpage

\section*{Acknowledgments}
This research is partially funded by the National Science Foundation awards NSF-OAC-2212442, NSF-2108320, NSF-1808576, and NSF-CAREER-1845799; by the Department of Energy award DE-SC0018147 and DE-SC0022211. The authors would like to thank William Cole Nockolds, Wesley Lao, and Krishnanunni Chandradath Girija for fruitful discussions. The authors also acknowledge the Texas Advanced Computing Center (TACC) at The University of Texas at Austin for providing HPC, visualization, database, or grid
resources that have contributed to the research results reported within this paper. URL: http://www.tacc.utexas.edu 

\bibliographystyle{plain}
\bibliography{references, references_mcTangent}

\end{document}